\pdfoutput=1
\documentclass{article}
\PassOptionsToPackage{numbers,sort&compress}{natbib}
\usepackage[main,final]{neurips_2026}
\usepackage[utf8]{inputenc} %
\usepackage[T1]{fontenc}    %
\usepackage{url}            %
\usepackage{booktabs}       %
\usepackage{amsfonts}       %
\usepackage{amsthm}         %
\usepackage{microtype}      %
\usepackage{xcolor}         %
\usepackage{wrapfig}
\usepackage{graphicx}
\usepackage{adjustbox}
\usepackage[algo2e]{algorithm2e}
\usepackage{comment}
\usepackage{subfig}
\usepackage{enumitem}
\setlist{leftmargin=3mm}
\usepackage{multirow}
\usepackage{tcolorbox}
\tcbuselibrary{breakable,skins}
\usepackage{tikz}
\usepackage{pgfplots}\pgfplotsset{compat=1.18}
\usetikzlibrary{positioning,calc,fit,backgrounds,arrows.meta,decorations.pathreplacing,shapes.geometric}
\IfFileExists{assets/figures/splice_viz_style.tex}{%
  
\definecolor{spliceaccent}{HTML}{1F4E79}   %
\definecolor{splicefrozen}{HTML}{E9E9E9}   %
\definecolor{splicerule}{HTML}{555555}     %
\definecolor{spliceedit}{HTML}{BBD4EA}     %
\definecolor{splicewarn}{HTML}{8C2D04}     %
\definecolor{splicecard}{HTML}{FBFBFC}     %

\newcommand{\splockpath}[3]{%
  \begin{scope}[shift={(#1,#2)}, scale=#3]
    \draw[spliceaccent, line width=0.3pt] (0.3mm,0.85mm) arc (0:180:0.3mm);
    \fill[spliceaccent] (-0.5mm,-0.05mm) rectangle (0.5mm,0.85mm);
  \end{scope}%
}
\newcommand{\lockglyph}{%
  \begin{tikzpicture}[baseline=-0.4ex]
    \splockpath{0mm}{0mm}{1}
  \end{tikzpicture}%
}

\newcommand{\splchipscale}{1}
\newcommand{\splchip}[2][chipplain]{%
  \begin{tikzpicture}[baseline=-0.4ex,
      x={\splchipscale mm}, y={\splchipscale mm}]
    \foreach \i in {1,2,3}{%
      \filldraw[fill=white, draw=splicerule!70, line width=0.25pt]
        (0.45, {6.55-(\i-1)*1.10}) rectangle (2.95, {5.70-(\i-1)*1.10});}
    \foreach \i in {4,5,6}{%
      \filldraw[fill=splicefrozen, draw=spliceaccent!70, line width=0.3pt]
        (0.45, {6.55-(\i-1)*1.10}) rectangle (2.95, {5.70-(\i-1)*1.10});}
    \draw[spliceaccent, line width=0.45pt] (3.30,0.20) -- (3.30,3.25);
    \foreach \e in {#2}{%
      \ifnum\e>0\relax
        \filldraw[fill=spliceedit, draw=spliceaccent, line width=0.35pt]
          (0.45, {6.55-(\e-1)*1.10}) rectangle (2.95, {5.70-(\e-1)*1.10});
      \fi}
    \draw[#1] (0,0) rectangle (4.0,7.2);
  \end{tikzpicture}%
}

\tikzset{
  chipplain/.style ={draw=splicerule,     line width=0.40pt, rounded corners=0.6pt},
  chipelite/.style ={draw=spliceaccent,   line width=1.00pt, rounded corners=0.6pt},
  chipdiv/.style   ={draw=spliceaccent,   line width=0.60pt, rounded corners=0.6pt,
                     dash pattern=on 0.9pt off 0.7pt},
  chippruned/.style={draw=splicerule!35,  line width=0.35pt, rounded corners=0.6pt},
  blk/.style={draw=splicerule, line width=0.5pt, rounded corners=1.5pt,
    minimum height=5.4mm, text width=19mm, align=left,
    inner xsep=1.6pt, font=\scriptsize, anchor=west},
  mutblk/.style={blk, fill=white},
  frzblk/.style={blk, fill=splicefrozen, draw=spliceaccent, line width=0.75pt},
  varbox/.style={draw=splicerule!55, line width=0.4pt, rounded corners=1pt,
    fill=white, minimum height=3.2mm, minimum width=6.4mm,
    align=center, font=\tiny, inner sep=0.8pt, anchor=west},
  frzvar/.style={varbox, fill=splicefrozen, draw=spliceaccent!75},
  card/.style={draw=splicerule!45, line width=0.45pt, rounded corners=2.2pt,
    fill=splicecard},
  statecard/.style={card, draw=spliceaccent!55, line width=0.55pt},
  stg/.style={circle, fill=spliceaccent, text=white, inner sep=0pt,
    minimum size=3.0mm, font=\bfseries\tiny, align=center},
  cardttl/.style={font=\scriptsize\bfseries, text=spliceaccent,
    anchor=west, align=left, inner sep=0pt},
  lin/.style={draw=splicerule!70, line width=0.3pt, fill=white,
    minimum size=2.2mm, inner sep=0pt, font=\tiny, align=center},
  elitebadge/.style={fill=spliceaccent, draw=spliceaccent, line width=0.3pt,
    minimum size=1.5mm, inner sep=0pt},
  divbadge/.style={fill=white, draw=spliceaccent, line width=0.5pt,
    minimum size=1.5mm, inner sep=0pt,
    dash pattern=on 0.7pt off 0.5pt},
  spinebox/.style={draw=spliceaccent, line width=0.7pt, rounded corners=2.2pt,
    fill=spliceedit!25},
  spinettl/.style={font=\footnotesize\bfseries, text=spliceaccent,
    align=center, inner sep=0pt},
  spinesub/.style={font=\tiny, text=splicerule, align=center, inner sep=0pt},
  zoomlead/.style={draw=spliceaccent!70, line width=0.4pt,
    dash pattern=on 0.6pt off 0.6pt},
  blockedarr/.style={-{Bar[width=1.5mm, line width=0.55pt]},
    draw=splicewarn, line width=0.55pt},
  paneltitle/.style={font=\scriptsize\bfseries, text=spliceaccent,
    align=flush left, inner sep=0pt},
  whytxt/.style={font=\tiny, text=splicerule, align=flush left, inner sep=0pt},
  telite/.style={circle, fill=spliceaccent, draw=spliceaccent,
    line width=0.3pt, inner sep=0pt, minimum size=1.5mm},
  tdiv/.style={circle, fill=white, draw=spliceaccent, line width=0.5pt,
    inner sep=0pt, minimum size=1.7mm, dash pattern=on 0.7pt off 0.5pt},
  tedge/.style={draw=splicerule!55, line width=0.4pt},
  tleadpath/.style={draw=spliceaccent, line width=1.0pt},
  panellbl/.style={font=\bfseries\scriptsize, align=center},
  note/.style={font=\tiny\itshape, text=splicerule, align=center},
  anote/.style={font=\tiny, text=spliceaccent, align=left},
  tinylbl/.style={font=\tiny, text=splicerule, align=center, inner sep=0.5pt},
  editarr/.style={-{Stealth[length=1.3mm,width=1.1mm]},
    draw=spliceaccent, line width=0.55pt},
  flowarr/.style={-{Stealth[length=1.8mm,width=1.5mm]},
    draw=splicerule, line width=0.8pt},
  stagearr/.style={-{Stealth[length=1.6mm,width=1.4mm]},
    draw=spliceaccent, line width=0.7pt},
  backarr/.style={-{Stealth[length=1.6mm,width=1.4mm]},
    draw=spliceaccent, line width=0.7pt,
    dash pattern=on 1.4pt off 1.0pt},
  gateline/.style={draw=spliceaccent, line width=0.6pt,
    dash pattern=on 1.2pt off 0.9pt},
}

\newcommand{\pruneX}{\tikz[baseline=-0.35ex]{%
  \draw[splicewarn, line width=0.5pt] (0,0) -- (1.4mm,1.4mm)
    (0,1.4mm) -- (1.4mm,0);}}
}{}
\usepackage{listings}

\usepackage{amsmath,amssymb,amsfonts}
\usepackage{algorithm}
\usepackage{algorithmic}

\usepackage{comment}
\usepackage{bm}
\usepackage{pifont}
\newcommand{\cmark}{\ding{51}}%
\newcommand{\xmark}{\ding{55}}%
\newsavebox{\pmarkbox}
\savebox{\pmarkbox}{%
  \tikz[baseline=-0.55ex]{%
    \fill (0,0.62ex) arc (90:270:0.62ex) -- cycle;
    \draw[line width=0.55pt] (0,0) circle (0.62ex);}}
\DeclareRobustCommand{\pmark}{\usebox{\pmarkbox}}%

\def\1{\mathbf{1}}
\newcommand{\train}{\mathcal{D}}
\newcommand{\valid}{\mathcal{D_{\mathrm{valid}}}}
\newcommand{\test}{\mathcal{D_{\mathrm{test}}}}
\newcommand{\dtrain}{\mathcal{D_{\mathrm{train}}}}

\def\rvx{{\mathbf{x}}}
\def\rvy{{\mathbf{y}}}

\DeclareMathAlphabet{\mathsfit}{\encodingdefault}{\sfdefault}{m}{sl}
\SetMathAlphabet{\mathsfit}{bold}{\encodingdefault}{\sfdefault}{bx}{n}

\newcommand{\E}{\mathbb{E}}

\newcommand{\reg}{\lambda}

\DeclareMathOperator*{\argmax}{arg\,max}

\usepackage{color, colortbl}
\usepackage{bbm}
\usepackage{bm}
\usepackage{hyperref}
\usepackage[capitalize,noabbrev]{cleveref}
\lstdefinestyle{mystyle}{
    backgroundcolor=\color{backcolour},   
    commentstyle=\color{codegreen},
    keywordstyle=\color{magenta},
    numberstyle=\tiny\color{codegray},
    stringstyle=\color{codepurple},
    basicstyle=\ttfamily\footnotesize,
    breakatwhitespace=false,         
    breaklines=true,                 
    captionpos=b,                    
    keepspaces=true,                 
    numbers=left,                    
    numbersep=5pt,                  
    showspaces=false,                
    showstringspaces=false,
    showtabs=false,                  
    tabsize=2
}
\usepackage{soul}
\definecolor{LightCyan}{rgb}{0.9,1,1}
\definecolor{light-gray}{gray}{0.97}
\definecolor{codegreen}{rgb}{0,0.6,0}
\definecolor{codegray}{rgb}{0.5,0.5,0.5}
\definecolor{codepurple}{rgb}{0.58,0,0.82}
\definecolor{backcolour}{rgb}{0.95,0.95,0.92}

\crefname{assumption}{Assumption}{Assumptions}
\crefname{remark}{Remark}{Remarks}

\newcommand{\render}{\operatorname{render}}
\newcommand{\utility}{F}
\newcommand{\reward}{r}
\newcommand{\model}{M}
\newcommand{\cost}{\kappa}
\newcommand{\bw}{W}
\newcommand{\splice}{\textsc{Splice}}
\newcommand{\gateratio}{\varrho}

\providecommand{\lineage}{\mathrm{lineage}}
\providecommand{\PickDiverse}{\textsc{PickDiverse}}
\providecommand{\topk}{\mathrm{top}}
\providecommand{\Apply}{\mathrm{Apply}}
\providecommand{\Expand}{\mathrm{Expand}}

\newtheorem{definition}{Definition}
\newtheorem{assumption}{Assumption}
\newtheorem{proposition}{Proposition}
\newtheorem{theorem}{Theorem}
\newtheorem{corollary}{Corollary}
\newtheorem{conjecture}{Conjecture}
\newtheorem{lemma}{Lemma}
\newtheorem{remark}{Remark}

\tcbset{splicethmbox/.style={
  colback=gray!5,
  colframe=black!65,
  boxrule=0.9pt,
  arc=2pt,
  left=8pt, right=8pt, top=4pt, bottom=4pt,
  before skip=6pt, after skip=6pt,
  breakable,
}}
\newcommand{\splicethmboxes}{%
  \tcolorboxenvironment{definition}{splicethmbox}%
  \tcolorboxenvironment{assumption}{splicethmbox}%
  \tcolorboxenvironment{proposition}{splicethmbox}%
  \tcolorboxenvironment{theorem}{splicethmbox}%
  \tcolorboxenvironment{corollary}{splicethmbox}%
  \tcolorboxenvironment{conjecture}{splicethmbox}%
  \tcolorboxenvironment{lemma}{splicethmbox}%
  \tcolorboxenvironment{remark}{splicethmbox}%
}

\usepackage{etoc}            %
\usepackage{placeins}        %
\title{\splice{}: Structured Prompt Local Iterative Combinatorial Evolution}

\author{%
  Anish Acharya\thanks{Correspondence: \texttt{achanish@amazon.com}} \quad
  Phillip Studans\thanks{Work done while at Amazon Web Services.} \quad Amit Dhanda \quad Sina Amini Niaki\\
  \bfseries Ninad Vivek Rao \quad Vishal Amarlal Khatri \quad Brian Verkhovsky \\[2pt]
  {\normalfont Amazon Web Services}
}

\begin{document}
\etocdepthtag.toc{mtmain}
\maketitle

\begin{abstract}

Natural language optimization (NLO) aims to improve an agent by having a model
critique and rewrite its natural language policy parameters, e.g., system prompts,
tool usage instructions, reasoning directives, few-shot demonstrations. 
These artifacts are multi-section documents in which
some sections encode hard-won \emph{contracts} (safety clauses, tool schemas,
output formats) that must not drift, and others are legitimate optimization
targets. Most existing iterative NLO methods treat the artifact as a single
editable string, producing three recurrent failure modes: unsafe global rewrites, beam
collapse onto a single lineage, and unchecked bloat -- none addressed by formal
guarantees. We introduce \splice{}, a unified NLO framework that poses the optimization recipe
as \textbf{constrained 
combinatorial search over an edit
graph of block-structured, selectively-mutable states}.
Under this framework, existing
iterative algorithms
instantiate a single \emph{batched iterative search}
template or depart from it in a component we can name. Theoretically, we map each failure mode to a
matched guarantee:
\textbf{(i)}~per-block mutability preserves contract text verbatim, with a
price-of-freezing bound that is small when frozen blocks
carry small marginal gains and the submodularity ratio $\gamma$ is not small;
\textbf{(ii)}~elitist beam search with diversity slots never lets the leader
regress on validation, bounds regression under noisy re-evaluation by an
exponential tail, and keeps a competitive candidate from a different parent
whenever one exists;
\textbf{(iii)}~a ratio length regularizer comes with an explicit
accuracy--length bias bound.
Candidates are proposed from dual section-local textual gradients; under a
proposer-competence condition whose proxy we measure, edit-graph reachability
and a convergence rate bound how quickly improving children arrive.
Empirically, across seven public benchmarks under five reproduced protocols,
\splice{} wins on four and matches the strongest prior method on three. 
On a
production-style contract stress test it keeps the text of the contract clauses
that unconstrained optimizers trade away to raise the proxy metric, and
ablations attribute gains to critique (measured on Best-of-$N$) and elitist search.

\end{abstract}

\section{Introduction}
\label{sec:main.intro}
An LLM system's behavior is a policy determined by its weights and its text:
system prompts, agent instructions, tool descriptions, and the harness code
around the model. With black-box access the weights are fixed, and the text is
the only part of the policy that can be optimized. \emph{Natural language
optimization} (NLO) improves it by having a model repeatedly critique and
rewrite it; the formulation applies to any block-structured text, our
experiments optimize prompts, and we loosely call the text the \emph{prompt}.
Production prompts are multi-section documents (role framing, demonstrations,
safety constraints, tool schemas, output specifications) in one token stream,
so an edit to one section can alter behavior governed by another -- a
\textbf{regression risk}. Some sections are \emph{contracts} that must not
change, and every added token costs latency: the deployed problem is
\emph{constrained}.

Iterative prompt optimization, the most developed form of NLO, follows four
directions: \emph{generate-and-select}
rewrites~\citep{zhou2022large,yang2024large}, \emph{gradient-inspired}
refinement from textual
feedback~\citep{pryzant2023automatic,cui2024introducing,yuksekgonul2024textgrad,sharma2026modular},
\emph{compile-time} module-level
search~\citep{khattab2023dspy,opsahl2024optimizing}, and \emph{evolutionary}
population
methods~\citep{guo2023connecting,fernando2024promptbreeder,agrawal2026gepa}. They share two implicit commitments:
every part of a candidate is open to editing, and a run is a loop that
proposes, scores and keeps candidates. Left implicit, these leave three
gaps. \emph{First}, deployment requirements cannot be stated: no prior method
exposes \emph{selective mutability} as a property of the search space, and
growth is rarely priced (\citep{agrawal2026gepa} report \textsc{Mipro}v2 prompts
$\sim\!3\times$ longer at lower accuracy). \emph{Second}, representations,
proposals and selectors are confounded across implementations, so methods
cannot be compared component by component. \emph{Third}, the search dynamics
are untheorized: no monotonicity, no-collapse or length-bias guarantees. The
field largely solves the unconstrained relaxation of a constrained problem
(related work in detail: \cref{sec:rel_work}).

\begin{figure}[t]
\centering
\resizebox{0.92\textwidth}{!}{%
\begin{tikzpicture}[x=1cm, y=1cm, font=\footnotesize,
  hcard/.style={draw=spliceaccent!60, fill=spliceaccent!4, line width=1.0pt,
    rounded corners=4pt},
  hnum/.style={circle, fill=spliceaccent, text=white, inner sep=0pt,
    minimum size=3.6mm, font=\scriptsize\bfseries},
  httl/.style={anchor=west, inner sep=0pt, font=\small\bfseries,
    text=spliceaccent},
  chev/.style={-{Stealth[length=2.6mm,width=2.6mm]}, line width=1.6pt,
    draw=spliceaccent!45},
  mblk/.style={draw=splicerule!80, fill=white, line width=0.8pt,
    rounded corners=1.6pt, minimum width=34mm, minimum height=4.6mm,
    inner xsep=1.4mm, anchor=north west},
  fblk/.style={mblk, draw=spliceaccent, fill=splicefrozen, line width=1.1pt},
  snip/.style={font=\scriptsize\itshape, text=splicerule},
  gnode/.style={circle, inner sep=0pt, minimum size=1.7mm,
    fill=splicerule!30, draw=none},
  gseed/.style={circle, inner sep=0pt, minimum size=2.6mm, fill=white,
    draw=splicerule, line width=0.9pt},
  gelite/.style={circle, inner sep=0pt, minimum size=2.5mm,
    fill=spliceaccent, draw=spliceaccent},
  gbest/.style={gelite, minimum size=3.4mm, draw=white, line width=0.8pt,
    double=spliceaccent, double distance=0.6pt},
  gdiv/.style={circle, inner sep=0pt, minimum size=2.4mm, fill=white,
    draw=spliceaccent, line width=0.8pt, dash pattern=on 1pt off 0.7pt},
  ggone/.style={circle, inner sep=0pt, minimum size=2.0mm, fill=white,
    draw=splicewarn!70, line width=0.5pt, dash pattern=on 0.8pt off 0.8pt},
  gedge/.style={draw=splicerule!40, line width=0.45pt},
  stage/.style={draw=spliceaccent, fill=white, line width=1.2pt,
    rounded corners=2.5pt, text width=17mm, align=center,
    inner xsep=0.8mm, inner ysep=1.1mm},
  sttl/.style={font=\scriptsize\bfseries\scshape, text=spliceaccent},
  cyc/.style={-{Stealth[length=1.9mm,width=1.7mm]}, line width=0.8pt,
    draw=spliceaccent!80},
  chip/.style={draw=spliceaccent!85, fill=white, line width=0.9pt,
    rounded corners=5pt, inner xsep=2mm, inner ysep=1.1mm,
    font=\scriptsize},
]

\draw[hcard] (0.00, 0.00) rectangle ( 3.80,-4.55);
\draw[hcard] (4.25, 0.00) rectangle ( 8.95,-4.55);
\draw[hcard] (9.40, 0.00) rectangle (13.90,-4.55);
\draw[chev] (3.86,-2.30) -- (4.19,-2.30);
\draw[chev] (9.01,-2.30) -- (9.34,-2.30);

\node[hnum] at (0.30,-0.32) {1};
\node[httl] at (0.55,-0.32) {The Artifact};
\node[hnum] at (4.55,-0.32) {2};
\node[httl] at (4.80,-0.32) {The Feasible Set};
\node[hnum] at (9.70,-0.32) {3};
\node[httl] at (9.95,-0.32) {One \splice{} Step};

\foreach \y/\nm/\tx in {-0.72/role/{support agent},
                        -1.26/directive/{step by step},
                        -1.80/examples/{worked cases}}{%
  \node[mblk] at (0.20,\y) {};
  \node[anchor=west, inner sep=0pt, font=\scriptsize\bfseries]
    at (0.34,\y-0.23) {\nm};
  \node[snip, anchor=west, inner sep=0pt] at (1.82,\y-0.23) {\tx};
  \node[anchor=east, inner sep=0pt, text=spliceaccent, font=\scriptsize]
    at (3.52,\y-0.23) {\ding{46}};}
\foreach \y/\nm/\tx in {-2.52/{tool schema}/{lookup(id,\,$\cdot$)},
                        -3.06/privacy/{no PII}}{%
  \node[fblk] at (0.20,\y) {};
  \node[anchor=west, inner sep=0pt, font=\scriptsize\bfseries]
    at (0.34,\y-0.23) {\nm};
  \node[snip, anchor=west, inner sep=0pt] at (1.82,\y-0.23) {\tx};
  \node[anchor=east, inner sep=0pt] at (3.52,\y-0.23)
    {\tikz[baseline=-0.4ex]{\splockpath{0mm}{0mm}{1.4}}};}
\draw[spliceaccent!50, line width=0.4pt, dash pattern=on 1.5pt off 1pt]
  (0.20,-2.40) -- (3.60,-2.40);
\node[anchor=north west, inner sep=0pt, font=\scriptsize, text width=34mm,
  align=flush left] at (0.20,-3.78)
  {\textcolor{spliceaccent}{\ding{46}}\,mutable: the search edits it\\[1pt]
   \tikz[baseline=-0.4ex]{\splockpath{0mm}{0mm}{1.3}}\,frozen: a contract,
   kept verbatim};

\fill[spliceaccent!9, rounded corners=6pt] (4.62,-0.72) rectangle (7.78,-3.52);
\node[font=\small, text=spliceaccent, anchor=south east, inner sep=0pt]
  at (7.70,-3.45) {$\mathcal{X}$};
\foreach \y in {-1.17,-2.12,-3.07}{\draw[gedge] (4.95,\y) -- (7.35,\y);}
\foreach \x in {4.95,5.75,6.55,7.35}{\draw[gedge] (\x,-1.17) -- (\x,-3.07);}
\foreach \x in {4.95,5.75,6.55,7.35}{\foreach \y in {-1.17,-2.12,-3.07}{%
  \node[gnode] at (\x,\y) {};}}
\draw[splicewarn!80, line width=0.6pt, dash pattern=on 1.4pt off 1pt]
  (7.35,-2.12) -- (8.40,-2.12);
\node[ggone] at (8.48,-2.12) {};
\node[text=splicewarn, font=\scriptsize\bfseries] at (7.92,-2.12) {$\times$};
\node[text=splicewarn, font=\scriptsize, align=center, anchor=north,
  inner sep=0pt] at (8.40,-2.32) {edits\\privacy:\\not in $\mathcal{X}$};
\draw[spliceaccent!80, line width=0.8pt, dash pattern=on 1.6pt off 1.1pt]
  (4.95,-3.07) -- (4.95,-1.17);
\draw[spliceaccent, line width=1.6pt, line cap=round, line join=round,
  -{Stealth[length=2.4mm,width=2.2mm]}]
  (4.95,-3.07) -- (5.75,-3.07) -- (5.75,-2.12) -- (6.55,-2.12)
  -- (6.55,-1.17) -- (7.18,-1.17);
\node[gseed]  at (4.95,-3.07) {};
\node[gelite] at (5.75,-3.07) {};
\node[gelite] at (5.75,-2.12) {};
\node[gelite] at (6.55,-2.12) {};
\node[gelite] at (6.55,-1.17) {};
\node[gbest]  at (7.35,-1.17) {};
\node[gdiv]   at (4.95,-2.12) {};
\node[gdiv]   at (4.95,-1.17) {};
\foreach \x/\y in {6.55/-3.07, 5.75/-1.17, 7.35/-3.07}{%
  \node[fill=spliceaccent!9, inner sep=0.3pt] at (\x,\y) {\pruneX};}
\node[anchor=north, inner sep=1pt, font=\scriptsize] at (4.95,-3.18)
  {seed $x^{(0)}$};
\node[anchor=south, inner sep=1.5pt, font=\scriptsize] at (7.35,-1.05)
  {$\hat{x}$};
\node[font=\scriptsize, text=splicerule, anchor=north, inner sep=1pt]
  at (6.15,-3.58) {directive variant $\rightarrow$};
\node[font=\scriptsize, text=splicerule, rotate=90, inner sep=1pt]
  at (4.44,-2.12) {examples variant $\rightarrow$};
\node[anchor=west, font=\scriptsize, inner sep=0pt] at (4.45,-4.22)
  {\tikz[baseline=-0.5ex]\draw[spliceaccent, line width=1.4pt] (0,0) -- (0.32,0);\,leader
   \quad\tikz[baseline=-0.5ex]\draw[spliceaccent, line width=0.8pt, dash pattern=on 1.6pt off 1.1pt] (0,0) -- (0.32,0);\,diversity slot
   \quad\pruneX\,pruned};

\node[stage] (crit) at (11.65,-1.10)
  {\textcolor{spliceaccent}{\scriptsize\bfseries\scshape critique}\\[-1pt]{\scriptsize keep / repair\\[-2pt]note per block}};
\node[stage] (prop) at (12.97,-2.42)
  {\textcolor{spliceaccent}{\scriptsize\bfseries\scshape propose}\\[-1pt]{\scriptsize $k$ edits to\\[-2pt]mutable blocks}};
\node[stage] (scor) at (11.65,-3.74)
  {\textcolor{spliceaccent}{\scriptsize\bfseries\scshape score}\\[-1pt]{\scriptsize on $\valid$,\\[-2pt]length priced}};
\node[stage] (sele) at (10.33,-2.42)
  {\textcolor{spliceaccent}{\scriptsize\bfseries\scshape select}\\[-1pt]{\scriptsize elites $+$\\[-2pt]diversity slot}};
\draw[cyc] (crit.east) to[out=0, in=90]   (prop.north);
\draw[cyc] (prop.south) to[out=-90, in=0] (scor.east);
\draw[cyc] (scor.west) to[out=180, in=-90] (sele.south);
\draw[cyc] (sele.north) to[out=90, in=180] (crit.west);
\node[align=center, font=\scriptsize, text=spliceaccent] at (11.65,-2.42)
  {beam\\[-1pt]$\mathcal{C}_t$};

\node[font=\scriptsize\itshape, text=splicerule, anchor=west, inner sep=0pt]
  at (0.05,-5.07) {No proposer assumption:};
\node[chip, anchor=west] (rb1) at (2.75,-5.07)
  {\textcolor{spliceaccent}{\cmark}\ frozen text byte-identical};
\node[chip, anchor=west] (rb2) at ([xshift=2.5mm]rb1.east)
  {\textcolor{spliceaccent}{\cmark}\ leader never below the seed};
\node[chip, anchor=west] (rb3) at ([xshift=2.5mm]rb2.east)
  {\textcolor{spliceaccent}{\cmark}\ competing other-parent child kept};

\end{tikzpicture}}
\caption{\textbf{NLO as constrained combinatorial
optimization.} \textbf{(1)}~An artifact is a sequence of blocks: mutable blocks
are optimization targets, frozen blocks are contracts. \textbf{(2)}~The feasible
set $\mathcal{X}$ is an edit graph; a frozen block admits only its seed text, so
editing it is not an edge. Drawn is one \splice{} run: the leader path, a lineage
held by a diversity slot, and pruned children. \textbf{(3)}~One step critiques,
proposes $k$ edits to mutable blocks, scores children on $\valid$ with length
priced, and keeps the elites plus a child of a different parent. Frozen
\emph{text} is not frozen behavior (\cref{sec:exp.contract-main}).}
\label{fig:splice-loop}
\end{figure}
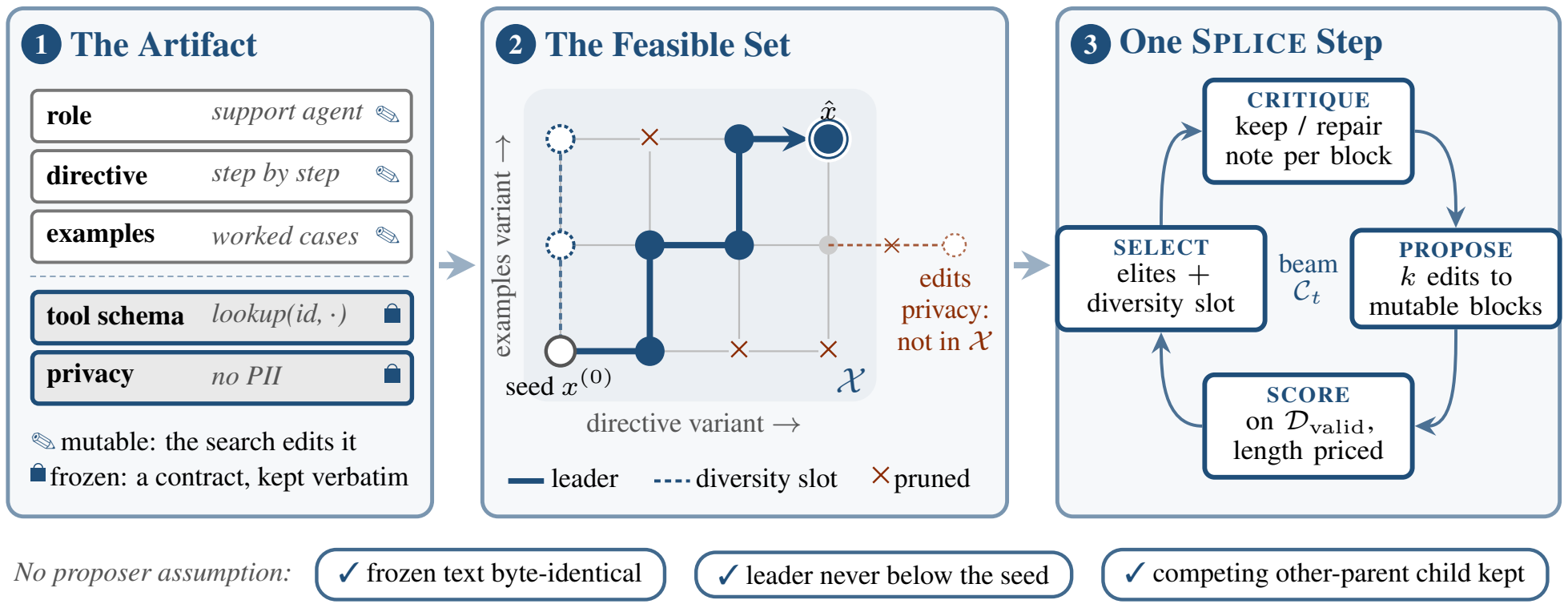

We make both commitments explicit by posing NLO as \emph{constrained
combinatorial optimization} (\cref{fig:splice-loop}). A prompt is a
block-structured state $x = (v_1,\dots,v_m,\pi)$; primitive edits connect states
into a finite edit graph; freezing a block is a \emph{feasibility constraint}
that makes every edit to it unreachable, and length is a \emph{priced penalty}.
This one move closes all three gaps: contracts and cost become part of the
problem, existing methods become instances of one batched search template that
differ in named components, and search over a finite graph under a noisy
oracle admits guarantees whose assumptions can be named. The same
propose--score--retain structure underlies self-improvement
loops~\citep{schmidhuber2003godel,zelikman2024stop,zhang2025darwin}, where
frozen blocks become invariants the loop cannot rewrite. \textbf{\splice{}}
(Structured Prompt Local Iterative Combinatorial Evolution) is this framework
together with its natural solver (\cref{alg:splice}).

{\bf \textsc{Contributions.}}
\begin{itemize}
  \item \textbf{\splice{}, a unified view of NLO as constrained combinatorial
  optimization} (\cref{sec:main.method}), separating representation, proposal
  mechanism and selector, with contracts as a feasibility constraint and length
  as a penalty (\cref{def:main-objective}). Existing
  methods~\citep{zhou2022large,pryzant2023automatic,cui2024introducing,sharma2026modular,yang2024large,opsahl2024optimizing,agrawal2026gepa,wang2024promptagent}
  either instantiate a single batched iterative search template
  (\cref{def:bips-body}) or depart from it in a component we can name: the
  scoring set, the selector, the output rule, or the expansion
  (\cref{sec:framework}); none treats a user-designated section as a constraint of the search.

  \item \textbf{The \splice{} solver, for the constrained problem}
  (\cref{sec:main.method}). Each production failure mode maps to one design
  choice: dual section-local critique edits only mutable blocks, a ratio length
  penalty prices bloat, and elitist
  selection reserves diversity slots against collapse.

  \item \textbf{A theory of search dynamics}
  (\cref{sec:theory-main}). With no assumption on the rewards or the proposer,
  the \textbf{beam leader's validation score never regresses} under cached scoring
  (\cref{thm:monotone}), the beam keeps a competitive candidate from a
  different parent whenever one exists (\cref{thm:div-preserve}), and the
  quality gate never loosens (\cref{prop:adaptive-anneal}). Assuming bounded
  rewards and i.i.d.\ batches, if candidates are instead re-scored on $n_{\mathrm{re}}$ fresh
  examples each step, the beam's best score \textbf{regresses by more than
  $\varepsilon$} with probability at most $\bw(1{+}k)\exp(-n_{\mathrm{re}}\varepsilon^2/c_\sigma)$,
  for a constant $c_\sigma\le2$ (\cref{thm:noisy-monotone,cor:min-batch}). Assuming in addition a proposer
  whose edits improve with probability at least $p^\star$ from every state
  $\varepsilon$-far from optimal (\cref{asm:coverage-body}), a full beam with no $\varepsilon$-optimal member
  \textbf{proposes an improving child in one step} with probability at least
  $1{-}(1{-}p^\star)^{\bw k}$ (\cref{thm:reachability}), and the search reaches
  $\varepsilon$-optimality at a rate we give (\cref{thm:convergence}). Finally,
  we \textbf{price the constraints}: freezing forfeits at most $\gamma^{-1}$
  times the frozen blocks' best single-block gains (\cref{prop:mut-price}), and
  length regularization biases accuracy by at most $\reg\cdot\cost(x^\star)$
  (\cref{thm:reg-bias}).

  \item \textbf{Empirical validation} (\cref{sec:main.exp}). On seven public
  benchmarks (SST-5~\citep{socher2013recursive},
  GSM8K~\citep{cobbe2021gsm8k},
  AIME-2025~\citep{balunovic2025matharena},
  ARC-Challenge~\citep{clark2018think},
  BBH~\citep{suzgun2023challenging},
  PUPA~\citep{siyan-etal-2025-papillon},
  GPQA-Diamond~\citep{rein2023gpqa}) reproducing five published
  protocols~\citep{agrawal2026gepa,juneja2025task,rishav2026contraprompt,zehle2026promptolution,zhou2025evaluating}
  without modification, and calling a margin of at most $0.5$\,pp a match,
  \splice{} wins on four and matches the strongest prior method on three (top
  margin $+5.1$\,pp on AIME-2025). On a contract stress test, unconstrained
  optimizers trade away the clauses \splice{} freezes, yet frozen text is not
  frozen behavior. Critique lifts Best-of-$N$ by $+1.2$--$1.7$\,pp, and Elitist
  Beam leads the best plain Best-of-$N$ by $+3.7$\,pp on SST-5 and $+1.3$\,pp on
  GPQA (within noise).
\end{itemize}
\begin{table}[t]
\centering
\begin{adjustbox}{max width=\textwidth}
\begin{tabular}{@{}lccccccccc@{}}
\toprule
 & {\bf Structure} & {\bf Iterative} & {\bf DualGrad} & {\bf SecLocal} & {\bf Elitism} & {\bf Diversity} & {\bf Immutable} & {\bf Priced} & {\bf Theory} \\
\midrule
APE \citep{zhou2022large}            & \xmark & \xmark & \xmark & \xmark & \xmark & \xmark & \xmark & \xmark & \xmark \\
OPRO \citep{yang2024large}           & \xmark & \cmark & \xmark & \xmark & \cmark & \xmark & \xmark & \xmark & \xmark \\
ProTeGi \citep{pryzant2023automatic} & \xmark & \cmark & \xmark & \xmark & \xmark & \xmark & \xmark & \xmark & \pmark \\
MAPO \citep{cui2024introducing}      & \xmark & \cmark & \cmark & \xmark & \xmark & \xmark & \xmark & \xmark & \xmark \\
MPO \citep{sharma2026modular}        & \cmark & \cmark & \xmark & \cmark & \xmark & \xmark & \xmark & \xmark & \xmark \\
PO2G \citep{lee2025po2g}             & \xmark & \cmark & \cmark & \xmark & \xmark & \xmark & \xmark & \xmark & \xmark \\
TextGrad \citep{yuksekgonul2024textgrad} & \xmark & \cmark & \pmark & \xmark & \xmark & \xmark & \pmark & \xmark & \xmark \\
DSPy \citep{khattab2023dspy}         & \cmark & \cmark & \xmark & \xmark & \xmark & \xmark & \xmark & \xmark & \xmark \\
MIPROv2 \citep{opsahl2024optimizing} & \cmark & \cmark & \xmark & \xmark & \xmark & \xmark & \xmark & \xmark & \xmark \\
EvoPrompt \citep{guo2023connecting}  & \xmark & \cmark & \xmark & \xmark & \cmark & \cmark & \xmark & \xmark & \xmark \\
Promptbreeder \citep{fernando2024promptbreeder} & \xmark & \cmark & \xmark & \xmark & \cmark & \cmark & \xmark & \xmark & \xmark \\
GEPA \citep{agrawal2026gepa}         & \cmark & \cmark & \xmark & \xmark & \cmark & \cmark & \xmark & \xmark & \xmark \\
\midrule
\textbf{\textsc{Splice} (This Paper)} & \cmark & \cmark & \cmark & \cmark & \cmark & \cmark & \cmark & \cmark & \cmark \\
\bottomrule
\end{tabular}
\end{adjustbox}
\caption{\footnotesize Representative prompt optimizers against the axes exposed
by our framework; \cref{tab:related-work-comparison} (\cref{sec:rel_work})
covers every optimizer we discuss or compare against. \cmark{}: the method
directly provides the property; \xmark{}: it does not; \pmark{}: partial, or
asserted but not established. ``DualGrad'': positive and negative textual
gradients. ``SecLocal'': updates at the level of named prompt sections.
``Elitism'': parents can survive a step. ``Immutable'': designated sections are
unreachable by any edit. ``Priced'': prompt length enters the objective.
``Theory'': formal guarantees about the search.}
\label{tab:related-work-main}
\end{table}

\section{Structured Prompt Local Iterative Combinatorial Evolution}
\label{sec:main.method}\label{sec:method}\label{sec:problem-setup}
We study NLO in the structured, black-box, 
sample-based regime.
A practitioner has a fixed task over an input space~$\mathcal{I}$ with unknown
distribution~$\mathcal{P}$, a black-box LLM (\cref{def:system}), a bounded
reward (\cref{def:reward}), and an initial prompt
(\cref{def:prompt-template}) they wish to improve, using an i.i.d.\ training
set $\train \sim \mathcal{P}^{N}$ to score candidates; a disjoint $\valid$
scores candidates for selection (the proposer never sees its examples), and
$\test$ is reserved for final reporting. 

\textbf{\textsc{Prompts as Block-Structured States}}

The prompt is \emph{structured} into semantically coherent
blocks that can be rewritten independently; the goal is to pick one variant
per block and an admissible ordering so as to maximise expected reward under
$\mathcal{P}$ within a finite evaluation budget.

\begin{definition}[{\bf Prompt Template}]
  \label{def:prompt-template}
  A \emph{prompt} consists of $m \ge 1$ \emph{blocks}, indexed by
  $[m] = \{1,\dots,m\}$. Each block is a semantically coherent unit whose
  textual realisation can be varied independently of the others.
\end{definition}

\begin{definition}[{\bf Block Variant Sets}]
  \label{def:variants}
  For each block $i \in [m]$, the \emph{variant set} $\mathcal{V}_i$ is a
  finite non-empty set of admissible textual variants; the initial prompt
  contributes a distinguished $v_i^{(0)} \in \mathcal{V}_i$, so that the
  initial prompt itself is always feasible.
\end{definition}

\begin{definition}[{\bf Block Mutability}]
\label{def:mutability}
Let $I_{\mathrm{mut}} \mathrel{\mathop{:}}= \{i : |\mathcal{V}_i| \ge 2\}$
and $I_{\mathrm{imm}} \mathrel{\mathop{:}}= [m] \setminus I_{\mathrm{mut}}$ be
the \emph{mutable} and \emph{immutable} block sets. A block
$i \in I_{\mathrm{imm}}$ has $\mathcal{V}_i = \{v_i^{(0)}\}$ and cannot be
edited: it encodes a \emph{contract} (safety clause, tool schema, output
format) preserved verbatim throughout optimisation.
\end{definition}

\begin{definition}[{\bf Block Orderings}]
  \label{def:orderings}
  The \emph{ordering family} $\Pi \subseteq \mathfrak{S}_m$ is a finite family
  of admissible permutations over $[m]$, with distinguished $\pi^{(0)} \in \Pi$
  corresponding to the initial layout.
\end{definition}

\begin{definition}[{\bf Prompt State and Search Space}]
\label{def:prompt-state}
A \emph{prompt state} is a tuple
$x = (v_1, \dots, v_m,\, \pi)$ with $v_i \in \mathcal{V}_i$ and $\pi \in \Pi$.
The \emph{search space} and initial state are
\begin{equation}
\label{eq:state-space}
\mathcal{X} = \Bigl(\prod_{i=1}^m \mathcal{V}_i \Bigr) \times \Pi,\qquad
|\mathcal{X}| = |\Pi| \cdot \prod_{i=1}^m |\mathcal{V}_i|,\qquad
x^{(0)} = (v_1^{(0)}, \dots, v_m^{(0)}, \pi^{(0)}).
\end{equation}
$\mathcal{X}$ is finite but combinatorially large: $|\mathcal{X}|$ quickly
exceeds any realistic evaluation budget.
\end{definition}

\begin{definition}[{\bf Render Function}]
      \label{def:render}
      The rendering function $\render : \mathcal{X} \to \Sigma^*$ maps a
      prompt state to a string over alphabet $\Sigma$ (typically Unicode) by
      concatenating $v_{\pi(1)}, \dots, v_{\pi(m)}$ in the order
      prescribed by~$\pi$ with fixed template separators. It is deterministic
      and fixed throughout optimisation.
\end{definition}

\textbf{\textsc{Combinatorial Objective}}

\begin{definition}[{\bf System Agent}]
  \label{def:system}
  The system agent is a (possibly stochastic) LLM
  $\model(\cdot\,;\,p) : \mathcal{I} \to \mathcal{Y}$ with $p \in \Sigma^*$,
  accessed strictly as a black box (no gradients, logits, or internal states).
\end{definition}

\begin{definition}[{\bf Bounded Reward}]
      \label{def:reward}
  \emph{Reward} is a bounded function $\reward : \mathcal{Y} \times \mathcal{Y}
  \to [0,1]$; canonical instantiations include $0/1$ exact match, accuracy,
  $F_1$, rubric scores, or rescaled learned-judge outputs.
\end{definition}

\begin{definition}[{\bf Population Utility}]
  \label{def:population-utility}
  The \emph{population utility} of $x \in \mathcal{X}$ is the expected reward
  under $\mathcal{P}$,
  \begin{equation}
    \label{eq:population-utility}
    \utility^{\star}(x) \;\mathrel{\mathop{:}}=\;
    \E_{(\rvx,\rvy)\sim\mathcal{P}}\!
    \bigg[ \reward\!\bigg(\model\big(\rvx;\,\render(x)\big),\, \rvy\bigg) \bigg],
  \end{equation}
  with expectation over the random example and any stochasticity in $\model$.
\end{definition}

Because $\utility^{\star}$ is inaccessible, the search observes it only
through an empirical surrogate:
\begin{definition}[{\bf Empirical Utility}]
  \label{def:empirical-utility}
  For $x \in \mathcal{X}$ and batch $B \subseteq \train$,
  \begin{equation}
    \label{eq:empirical-utility}
    \utility(x;\, B)
    \;\mathrel{\mathop{:}}=\; \frac{1}{|B|}
    \sum_{(\rvx,\rvy) \in B}
    \reward\!\bigl(\model(\rvx;\,\render(x)),\, \rvy\bigr);
  \end{equation}
  each oracle call consumes $|B|$ model evaluations, and we write
  $\utility(x) := \utility(x;\,\train)$. The gap between $\utility^{\star}$
  (what we optimise) and $\utility$ (what we observe) is the source of the
  finite-sample bounds in \cref{sec:app.theory}.
\end{definition}

\textbf{\textsc{Selective Mutability as a Feasibility Constraint}}
\label{sec:selective-mutability}\label{sec:problem.mutability}

\Cref{def:mutability} makes contract preservation a constraint on the feasible
set rather than an instruction to the proposer. Each named section of the prompt (role, objective, examples,
constraints, tool schema, output format) is tagged
\texttt{mutable}/\texttt{immutable}, and $|\mathcal{V}_i|=1$ makes every edit
to an immutable block unreachable. Freezing shrinks the
effective space by $\prod_{i\in I_{\mathrm{imm}}}|\tilde{\mathcal{V}}_i|$,
the variants the frozen blocks would otherwise admit, and guarantees the
contract's \emph{text} (safety, format, schema) by construction.

\textbf{\textsc{Length Regularisation as a Soft Penalty}}

Longer prompts add latency and cost to every call, and without a penalty
prompts keep growing, since the proposer mostly adds text. We penalise growth
rather than forbid it, using a ratio-based cost
\begin{equation}
\label{eq:ratio-cost}
\cost_{\mathrm{len}}(x) :=
\max\!\Bigl(0,\; \tfrac{\mathrm{len}(\render(x))}{\mathrm{len}(\render(x^{(0)}))} - 1\Bigr).
\end{equation}

\textbf{\textsc{The Constrained Problem}}

\begin{tcolorbox}[
  colback=gray!5,
  colframe=black!65,
  boxrule=0.9pt,
  arc=2pt,
  left=8pt, right=8pt, top=4pt, bottom=4pt,
  before skip=6pt, after skip=6pt]
\begin{definition}[{\bf Constrained Natural Language Optimization}]
\label{def:main-objective}
Given mutability tags (\cref{def:mutability}), a cost $\cost \ge 0$ with
$\cost(x^{(0)})=0$, and a weight $\reg \ge 0$, solve
\begin{equation}
\label{eq:main-objective}
x^{\star}_{\reg} \;\in\; \argmax_{x \in \mathcal{X}}\;
\utility^{\star}_{\reg}(x),
\qquad
\utility^{\star}_{\reg}(x) \;\mathrel{\mathop{:}}=\;
\utility^{\star}(x) \;-\; \reg \cdot \cost(x).
\end{equation}
\end{definition}
\end{tcolorbox}

Setting $\reg = 0$ and declaring
every block mutable recovers the unconstrained problem over the all-mutable
space $\mathcal{X}_{\mathrm{all}}$, with optimum $x^{\star}_{\mathrm{all}}$, which
most prior methods solve implicitly on a single string. 

\textbf{\textsc{Optimization is a walk over an Edit Graph}}

Problem~\eqref{eq:main-objective} is combinatorial and $\utility$ is noisy,
so any practical algorithm restricts to \emph{local} moves sampled by a
\emph{proposer} and filtered by a \emph{selector}. Prior NLO methods
(APE, OPRO, ProTeGi, MAPO, MPO) and \splice{} all fit a single template,
differing only in the state space they expose, the signal the proposer
consumes, and how the selector retains candidates between rounds.

\begin{definition}[{\bf Edit Graph}]
  \label{def:edit-graph}
  A \emph{primitive edit} is either a block replacement
  $v_i \to v'_i \in \mathcal{V}_i \setminus \{v_i\}$ or an admissible
  reordering $\pi \to \pi' \in \Pi$; each produces a state
  $x' \in \mathcal{X}$ differing from $x$ in exactly one coordinate. The
  \emph{edit graph} $\mathcal{G} = (\mathcal{X}, E)$ has radius-one
  neighbourhood $\mathcal{N}(x) = \{x' : x' \text{ is reachable from } x
  \text{ by one primitive edit}\}$.
\end{definition}
Replacement and reordering act on disjoint coordinates, so they commute,
giving $\mathrm{diam}(\mathcal{G}) \le m_{\mathrm{mut}} + \mathrm{diam}(\Pi)$
with $m_{\mathrm{mut}} = |I_{\mathrm{mut}}|$
(\cref{prop:edit-graph-diameter}).

\begin{definition}[{\bf Proposal Distribution}]
  \label{def:proposal}
  A \emph{proposal distribution} is a conditional family
  $q_\phi(u \mid x, H_t)$ over primitive edits, supported on edits that
  produce a feasible state $x' = \Apply(x, u) \in \mathcal{X}$, where
  the search history $H_t$ holds every candidate evaluated before step $t$ with its
  parent, lineage and score, and the parents' critiques at every step $s\le t$
  (step-$t$ ones computed before any step-$t$ proposal is drawn). In
  \splice{} and every baseline, $q_\phi$ is realised by an LLM conditioned on
  a textual encoding of $(x, H_t)$.
\end{definition}
Feasibility is enforced by rejecting malformed generations
(\cref{asm:proposal-distribution}).
\begin{tcolorbox}[
  colback=gray!5,
  colframe=black!65,
  boxrule=0.9pt,
  arc=2pt,
  left=8pt, right=8pt, top=4pt, bottom=4pt,
  before skip=6pt, after skip=6pt]
\begin{definition}[{\bf Batched Iterative Prompt Search}]
  \label{def:bips-body}\label{def:bips}
  Given $q_\phi$, beam width $\bw \ge 1$, per-parent fan-out $k \ge 1$, and
  selectors $\{\mathcal{S}_t\}_{t \ge 1}$, the iterates are
  \begin{equation}
    \label{eq:bips-body}
    \mathcal{C}_t = \mathcal{S}_t(\mathcal{F}_t),\quad
    \mathcal{F}_t = \mathcal{C}_{t-1} \cup
      \bigcup_{x \in \mathcal{C}_{t-1}} \{\Apply(x,u_j) : u_j \sim q_\phi(\cdot \mid x, H_t)\}_{j=1}^{k},
  \end{equation}
  starting from $\mathcal{C}_0 = \{x^{(0)}\}$, with
  $\mathcal{S}_t : 2^{\mathcal{X}} \to 2^{\mathcal{X}}$,
  $|\mathcal{S}_t(\mathcal{F})| \le \bw$; $\mathcal{F}_t$ is the \emph{frontier}.
\end{definition}
\end{tcolorbox}

Greedy hill-climbing, beam search, and elitist beam search recover as specific
selectors $\mathcal{S}_t$ on the monolithic state space;
\cref{tab:related-work-main} records where each prior method sits along
the design axes (state space, gradient signal, selector and constraints)
that together determine an algorithm (all methods: \cref{tab:related-work-comparison}).
Methods outside the template, such as GEPA and MIPROv2, depart from it
in named components (\cref{sec:framework}, \cref{fig:bips-sockets}). \splice{}'s contribution is the
particular combination of these axes, which we now describe.

\splice{} realises \cref{def:bips-body} as a loop (\cref{alg:splice},
\cref{sec:main.method-ext}). At step $t$ it re-scores the beam $\mathcal{C}_{t-1}$ on a
training batch $B_t$ and obtains section-local dual gradients from a critique
agent; builds a momentum context from the gradient history $G_{t-1}$; samples
$k$ edits per parent from $q_\phi$ \emph{restricted to mutable blocks}; scores
the children $\mathcal{N}_t$ on $\valid$ with
$\tilde\utility(x) = \utility(x;\valid) - \reg\,\cost(x)$; keeps $\bw$
members of $\mathcal{F}_t = \mathcal{C}_{t-1}\cup\mathcal{N}_t$ with an elitist
selector that reserves diversity slots; and records $(t, g_t, \Delta_t)$ in
$G_t$ for the next step's momentum. It returns $\mathrm{leader}(\mathcal{C}_T)$, where
$\mathrm{leader}(\mathcal{C}) \in \argmax_{x\in\mathcal{C}}\tilde\utility(x)$.
Selective mutability and the length penalty (above) fix what is optimised;
two independent design choices fill the loop's steps, which we take in order.

\textbf{\textsc{Critique: Dual, Section-Local Gradients with Momentum}}

\paragraph{Critique output.}
The critique agent $\model_{\mathrm{crit}}$ consumes per-example
traces on $B_t$ and returns
\begin{equation}
\label{eq:critique-output}
\mathrm{Crit}_t(x) = \bigl(\mathbf{n}(x),\, \mathbf{p}(x),\,
\{(\mathbf{n}_i,\mathbf{p}_i)\}_{i\in I_{\mathrm{mut}}},\, g(x)\bigr),
\end{equation}
with failure-patterns $\mathbf{n}$, success-patterns $\mathbf{p}$,
per-block assessments $(\mathbf{n}_i,\mathbf{p}_i)$, and a directional
summary $g(x)$ (``AWAY from $X$, TOWARD $Y$, PRESERVE $Z$'') playing
the role of a gradient. Duality keeps the proposer from chasing batch
failures while losing working content; section-localisation makes
$g(x)$ an addressable per-block update rather than a whole-prompt cue.

\paragraph{Textual momentum.}
Per-step batches make $g_t$ noisy across steps, and
natural-language gradients cannot be averaged vectorially. \splice{} instead
surfaces the gradient \emph{trajectory}
\begin{equation}
\label{eq:momentum-history}
G_t = \{(s, g_s, \Delta_s)\}_{s \le t},\quad
\Delta_s := \tilde\utility(\mathrm{leader}(\mathcal{C}_s))
          - \tilde\utility(\mathrm{leader}(\mathcal{C}_{s-1})),
\end{equation}
exposing the last $w$ entries ($w=5$) to the proposer, so it can keep
following a direction that has moved the leader. Since the leader never falls
(\cref{thm:monotone}), $\Delta_s\ge 0$: momentum records what helped.

\textbf{\textsc{Proposals and Reachability}}

The proposer is conditioned on
$\bigg(\render(x),\tilde\utility(x),\mathrm{Crit}_t(x),G_{t-1},I_{\mathrm{mut}}\bigg)$
and samples $k$ structurally-valid edits per parent at non-zero
temperature. If the proposer is competent (\cref{asm:coverage-body}), then while no beam member is
$\varepsilon$-optimal the $\bw k$ children of a step contain an improving one with probability at least
$1-(1-p^\star)^{\bw k}$ (\cref{thm:reachability}), so beam width and fan-out
trade off directly.

\textbf{\textsc{Selection: Elitism with Diversity Slots}}

\textbf{\textsc{Frontier elitism. }}
The selector ranks parents and children symmetrically on
$\mathcal{F}_t = \mathcal{C}_{t-1}\cup\mathcal{N}_t$ under the
regularised validation score $\tilde\utility(\cdot;\valid)$, computed at
first evaluation and \emph{cached} while a candidate stays in the beam
(re-rolling would introduce selection-on-noise bias). The leader score
$\ell_t := \max_{x\in\mathcal{C}_t}\tilde\utility(x;\valid)$ therefore never
falls (\cref{thm:monotone}); re-scoring instead would need large batches to
control regression (\cref{thm:noisy-monotone}).

\textbf{\textsc{Diversity slots. }}
Each candidate carries $\mathrm{id}(x)$, a parent pointer, and an
accumulated $\lineage(x)$. \splice{} reserves
$D\in\{0,\dots,\bw-1\}$ beam slots for diversity, filled by
\textsc{PickDiverse} (\cref{alg:pickdiverse}): scan the non-elite pool
in score order, accept the first candidate clearing the quality gate
$\tau_t := \gateratio\cdot\ell_t$ under progressively weaker novelty (Tier~1: new
lineage born this step; Tier~2: new lineage any generation; Tier~3:
different immediate parent), and backfill empty slots. Every ancestor set
contains the seed, so in practice Tier~3 operates: the beam keeps a
competitive different-parent candidate when one exists (\cref{thm:div-preserve}).

\textbf{\textsc{The gate, and the output rule. }}
Monotone $\ell_t$ (\cref{thm:monotone}) and $\tau_t = \gateratio\ell_t$ give a
non-decreasing gate when $\gateratio>0$: a candidate whose cached score falls below
$\tau_t$ is ineligible for diversity selection thereafter
(\cref{prop:adaptive-anneal}), which reads as annealing without a
temperature (an interpretation, not a guarantee). \splice{} returns
$\mathrm{leader}(\mathcal{C}_T)$, whose validation score is biased upward as a
maximum over the history (\cref{thm:uniform-conv}); the reported test score is
free of it.

Each design choice is a column of \cref{tab:related-work-main};
\cref{sec:theory-main} states the guarantees, and
\cref{sec:main.method-ext,sec:main.theory,sec:app.theory} give the full solver,
statements and proofs.

\section{Theoretical Guarantees}
\label{sec:theory-main}

An LLM-driven search does not admit end-to-end guarantees: the proposer is a
model we do not control, scores are noisy, and the utility need not be
submodular. We therefore state what holds at three levels of assumption: with no assumption on rewards or proposer; with
bounded rewards and i.i.d.\ batches; and, for the \emph{rate} results only,
with a competent proposer (assumptions per result: \cref{tab:theory-glossary}). Throughout, a candidate's validation score is
computed once and \emph{cached} while it stays in the beam (reachability,
convergence and the exclusion claim of \cref{prop:adaptive-anneal} assume it
cached \emph{per state}). Proofs are in \cref{sec:app.theory}.

\textbf{\textsc{No Assumption on Rewards or Proposer}}

\begin{lemma}[\bf Leader monotonicity under cached scoring]
\label{thm:monotone}
Under cached scoring and the selector
$\mathcal{S}^{\mathrm{elit}\text{-}\mathrm{div}}$, the per-step leader
score $\ell_t := \max_{x\in\mathcal{C}_t}\tilde\utility(x;\valid)$ is
non-decreasing: $\ell_t \ge \ell_{t-1}$ $\forall t\ge 1$.
\end{lemma}

It is a lemma, not a claim about search dynamics: it follows from the
definitions for any elitist selector and any proposer, and the returned prompt
therefore never scores below the seed on validation.

\begin{theorem}[\bf Diversity preservation]
\label{thm:div-preserve}
Let $1\le D\le \bw-1$ and $\gateratio\in[0,1]$. If $\mathcal{F}_t$ contains $c^{\star}$ with
$\tilde\utility(c^{\star}) \ge \gateratio\cdot\ell_t$ whose parent is not the parent of
any elite, then $\mathcal{C}_t$ contains a member scoring at least
$\gateratio\cdot\ell_t$ whose parent is not the parent of any elite.
\end{theorem}

\begin{proposition}[\bf Gate monotonicity]
\label{prop:adaptive-anneal}
Under cached scoring and $\mathcal{S}^{\mathrm{elit}\text{-}\mathrm{div}}$ with
$\gateratio>0$, the quality floor $\tau_t = \gateratio\cdot\ell_t$ is non-decreasing in $t$, and, with scores cached per state, a
candidate whose score falls below $\tau_t$ is ineligible for diversity selection
at every later step.
\end{proposition}

The diversity guarantee is \emph{conditional}: the selector retains a competitive
candidate of a different parent when one exists, and the proposer must generate
it (lineage-level variant: \cref{thm:div-preserve-ext}).

\textbf{\textsc{Bounded Rewards: Noisy No-Regression}}

Caching is a choice. If instead every frontier member were re-scored on $n_{\mathrm{re}}$
fresh i.i.d.\ examples each step and selected on those unregularized scores ($\reg=0$), regression of
the beam's best \emph{population} score $f^\star_t := \max_{x\in\mathcal{C}_t}
\utility^\star(x)$ is controlled by the batch size:

\begin{theorem}[\bf Probabilistic no-regression under re-evaluation]
\label{thm:noisy-monotone}
With rewards in $[0,1]$ and an elitist selector, for any $\varepsilon>0$,
$\Pr[f^\star_t < f^\star_{t-1} - \varepsilon] \le (|\mathcal{F}_t|-1)
\exp(-n_{\mathrm{re}}\varepsilon^2/c_\sigma)$, where $|\mathcal{F}_t| \le \bw(1+k)$, and $c_\sigma =
4\sigma^2 \le 1$ for independent batches and $c_\sigma=2$ for a shared batch.
\end{theorem}

\begin{corollary}[\bf Minimum batch size]
\label{cor:min-batch}
$\Pr[f^\star_t < f^\star_{t-1} - \varepsilon] \le \delta$ whenever
$n_{\mathrm{re}} \ge (c_\sigma/\varepsilon^2)\log(\bw(1+k)/\delta)$.
\end{corollary}

At $\bw{=}4$, $k{=}3$, $\delta{=}0.05$ and $c_\sigma{=}1$ this requires $n_{\mathrm{re}}\ge145$ at
$\varepsilon=0.2$, above most of our per-step batches, which is why \splice{}
caches scores instead of re-scoring.

\textbf{\textsc{A Competent Proposer: Rates}}

The rate results need one assumption about the LLM proposer, which we cannot
prove and instead measure through a proxy (\cref{sec:exp.pstar}):

\begin{assumption}[{\bf Proposal coverage on the $\varepsilon$-sublevel set}]
  \label{asm:coverage-body}\label{asm:proposal-coverage}
  Fix a tolerance $\varepsilon > 0$ in advance. With $u$ \emph{improving
  from} $x$ when $\tilde\utility(\Apply(x,u)) > \tilde\utility(x)$, let
  $p^\star(\varepsilon) := \min_{x:\,\max_{\mathcal{X}}\utility^\star-\utility^\star(x)\ge\varepsilon}
  \Pr_{u \sim q_\phi(\cdot \mid x, H_t)}[u\text{ is improving from }x]$,
  uniformly over histories $H_t$. We assume $p^\star(\varepsilon) > 0$.
\end{assumption}

\begin{theorem}[\bf Single-step reachability]
\label{thm:reachability}
Under \cref{asm:coverage-body}, at any step at which no beam member is
$\varepsilon$-optimal, the step's $|\mathcal{C}_{t-1}|\,k$ proposals contain one
that improves on its parent with probability at least
$1-(1-p^\star(\varepsilon))^{|\mathcal{C}_{t-1}|\,k} =: p^{\mathrm{step}}(\varepsilon)$.
\end{theorem}

\begin{theorem}[\bf Convergence rate]
\label{thm:convergence}
Under the scored form of coverage (\cref{asm:scored-coverage}) at tolerance
$\varepsilon$, let $\tilde h_0$ be the seed's gap to the scored optimum,
$K := \lceil(\tilde h_0-\varepsilon)/\Delta_{\min}\rceil$ for improvement size
$\Delta_{\min}$, and $p^{\mathrm{adv}}(\varepsilon)$ a lower bound on the
probability that the leader's $k$ proposals improve it by at least
$\Delta_{\min}$. Then $T \ge (2K + 8\ln(1/\delta))/p^{\mathrm{adv}}(\varepsilon)$
steps suffice for the leader to be $\varepsilon$-optimal on the scored objective
with probability at least $1-\delta$.
\end{theorem}

Either $\bw$ or $k$ drives the per-step failure probability to zero
exponentially. Coverage gates only these two rates, so when it fails the search
stalls rather than regresses. Terminal quality also depends on the utility's
structure: under $\gamma$-submodularity with monotone gains, block-pass greedy is exact when each
mutable block contributes one alternative (\cref{thm:greedy-approx}), and elitist
beam on the same draws stays ahead of greedy until they separate
(\cref{prop:beam-beats-greedy}).

\textbf{\textsc{The Price of the Constraints}}

\begin{proposition}[\bf Price of freezing]
\label{prop:mut-price}
At the seed ordering ($\Pi=\{\pi^{(0)}\}$), with $v_i^{(0)}\in\tilde{\mathcal{V}}_i$,
submodularity ratio $\gamma>0$ (\cref{def:gamma}) and
diminishing returns on frozen coordinates (\cref{cor:mut-price-baseline}), with $x^\star_{\mathrm{all}}$ the
optimum over the all-mutable space and $x^\star_{\mathrm{mut}}$ the unregularised
optimum over $\mathcal{X}$,
\begin{equation}
\label{eq:mut-price-body}
\utility^{\star}(x^{\star}_{\mathrm{all}}) - \utility^{\star}(x^{\star}_{\mathrm{mut}})
\;\le\; \frac{1}{\gamma}\sum_{i\in I_{\mathrm{imm}}}
\max_{v\in\tilde{\mathcal{V}}_i}\!\bigl[\utility^{\star}(x^{(0)}[i{:=}v]) - \utility^{\star}(x^{(0)})\bigr].
\end{equation}
\end{proposition}

\begin{theorem}[\bf Regularisation bias]
\label{thm:reg-bias}
For any $\reg\ge0$ and $\cost\ge0$,
$\utility^{\star}(x^{\star}_{\mathrm{mut}}) - \utility^{\star}(x^{\star}_{\reg}) \le
\reg\cdot\cost(x^{\star}_{\mathrm{mut}})$.
\end{theorem}

We write $x^\star := x^\star_{\mathrm{mut}} = x^\star_{\reg}$ at $\reg=0$, as in
every reported run. Both prices are small when the constraints are used as
intended: contract blocks carry little task reward, and $\reg$ breaks ties
toward shorter prompts. The terminal-quality results rest on $\gamma$, for which
we have no per-benchmark estimator. These bounds give the cost of freezing, not
whether freezing is needed or enough; \cref{sec:exp.contract-main} tests that.

\section{Experiments}
\label{sec:main.exp}
Our experiments address two questions.
\textbf{(Q1)} Does the batched iterative proposer--scorer template
of \cref{def:bips-body}, instantiated with elitist beam search and
lineage-aware diversity, translate the deterministic monotonicity
of \cref{thm:monotone} into measurable gains on standard
prompt-optimization benchmarks? \textbf{(Q2)} Are those gains
robust across target models, task families, and prior-work
evaluation protocols, or are they an artifact of a single setting?
To answer both we reproduce, \emph{without modification}, the
published protocols of five recent prompt-optimization papers and
compare \splice{} against their reported baselines---no re-tuning
of their pipelines, no substitution of their target models.

\begin{table}[t]
\centering
\small
\captionsetup{justification=raggedright,singlelinecheck=false}
\setlength{\tabcolsep}{5pt}
\renewcommand{\arraystretch}{1.02}
\begin{minipage}[c]{0.48\textwidth}
  \centering
  \begin{tabular}{@{}lcc@{}}
  \toprule
  \textbf{Optimizer}            & \textbf{ARC}  & \textbf{GSM8K} \\
  \midrule
  Task Description              & 79.7          & 59.4           \\
  Expert Prompt                 & 78.4          & 78.9           \\
  Llama Prompt                  & 89.7          & 79.5           \\
  CoT                           & 79.4          & 76.3           \\
  OPRO                          & 79.1          & 77.1           \\
  ProTeGi                       & 78.8          & 77.3           \\
  Evoke                         & 89.0          & 81.0           \\
  EvoPrompt                     & 89.9          & 81.4           \\
  DSPy (MIPROv2, zero-shot)     & 82.8          & 77.3           \\
  TextGrad                      & 76.5          & 81.6           \\
  UniPrompt + Beam              & 86.0          & 82.4           \\
  UniPrompt + Greedy            & 90.5          & 82.3           \\
  DSPy (BFS-RS)                 & 87.5          & 74.3           \\
  DSPy (MIPROv2, few-shot)      & 86.0          & 79.7           \\
  \midrule
  \splice{}                     & \textbf{91.0} & \textbf{83.09} \\
  \bottomrule
  \end{tabular}

  \captionof{table}{\small Optimized for model: GPT-3.5-Turbo;
datasets: ARC-Challenge, GSM8K \citep{juneja2025task}. Evoke is
\citet{hu2023evoke}.}
  \label{tab:bench-taskfacet}
\end{minipage}%
\hfill%
\begin{minipage}[c]{0.48\textwidth}
  \centering
  \begin{tabular}{@{}lcc@{}}
  \toprule
  \textbf{Optimizer}     & \textbf{AIME-25} & \textbf{PUPA}   \\
  \midrule
  Baseline               & 49.33            & 78.57           \\
  Trace (OptoPrime)      & 45.33            & 74.18           \\
  MIPROv2 (No-Demos)     & 48.67            & 91.85           \\
  MIPROv2                & 51.33            & 83.37           \\
  TextGrad               & 46.67            & 85.68           \\
  GEPA                   & 59.33            & 94.47           \\
  GEPA + Merge           & 59.33            & \textbf{96.46}  \\
  \midrule
  \splice{}              & \textbf{64.44}   & 96.24           \\
  \bottomrule
  \end{tabular}
  
  \captionof{table}{\small Optimized for model: GPT-4.1 Mini;
datasets: AIME-2025, PUPA \citep{agrawal2026gepa}. Trace (OptoPrime) is
\citet{cheng2024trace}.}
  \label{tab:bench-gepa}

  \vspace{0.5em}

  \begin{tabular}{@{}lcc@{}}
  \toprule
  \textbf{Method} & \textbf{BBH}  & \textbf{GPQA-D} \\
  \midrule
  Baseline        & 26.2          & 65.3            \\
  ContraPrompt    & 89.9          & 64.3            \\
  GEPA            & 91.9          & \textbf{68.4}   \\
  \midrule
  \splice{}       & \textbf{92.6} & \textbf{68.4}   \\
  \bottomrule
  \end{tabular}

  \captionof{table}{\small Optimized for model: Claude Haiku 4.5;
datasets: BBH, GPQA-Diamond \citep{rishav2026contraprompt}.}
  \label{tab:bench-contraprompt}
\end{minipage}%
\end{table}

\textbf{\textsc{Setup. }} We evaluate \splice{} on seven public benchmarks spanning sentiment
classification (SST-5~\citep{socher2013recursive}), mathematical
reasoning (GSM8K~\citep{cobbe2021gsm8k},
AIME-2025~\citep{balunovic2025matharena}), commonsense and multi-step
reasoning (ARC-Challenge~\citep{clark2018think},
BBH~\citep{suzgun2023challenging}), privacy-conscious delegation
(PUPA~\citep{siyan-etal-2025-papillon}), and expert-level science QA
(GPQA-Diamond~\citep{rein2023gpqa}). For every benchmark we adopt
the target model and evaluation protocol of the corresponding
published
baseline~\citep{agrawal2026gepa,juneja2025task,rishav2026contraprompt,
zehle2026promptolution,zhou2025evaluating}:
GPT-4.1~Mini~\citep{agrawal2026gepa} (AIME-2025, PUPA),
GPT-3.5-Turbo~\citep{juneja2025task} (ARC, GSM8K),
Claude~Haiku~4.5~\citep{rishav2026contraprompt} (BBH, GPQA-Diamond),
Gemma-3-27B~\citep{zehle2026promptolution} and
DeepSeek-V3.2~\citep{zhou2025evaluating} (SST-5); example optimized prompts
are in \cref{sec:app.example_prompts}, further experiments in
\cref{sec:main.exp-ext,sec:app.exp}.

\textbf{\textsc{Discussion. }}
\cref{tab:bench-taskfacet,tab:bench-gepa,tab:bench-contraprompt}
together cover six of the seven benchmarks; the seventh, SST-5,
is \cref{tab:bench-sst5} (\cref{sec:main.exp-ext}). Baselines are as published and report no seed variance, so we call a margin
of at most $0.5$\,pp a \emph{match}. On that accounting \splice{} wins on
four benchmarks (SST-5, AIME-2025, GSM8K, BBH) and matches the strongest prior
method on three (ARC-Challenge at $+0.5$\,pp, GPQA-Diamond at $0.0$, and PUPA at
$-0.22$\,pp against GEPA+Merge). The largest gains are $+5.1$\,pp on AIME-2025 over GEPA and $+2.9$\,pp on
SST-5/\texttt{gemma-3-27b-it}; GSM8K and BBH are narrow ($+0.69$, $+0.70$\,pp). Two patterns recur.
\emph{First, margins track headroom:} the lead is largest where
prior baselines sit well below the task's oracle ceiling and
tightens on near-saturated benchmarks (PUPA, BBH); GPQA-Diamond, a tie despite
headroom, is the exception.
\emph{Second, rank is stable across settings:} on every one of the five target
models we tried, \splice{} is at worst within the matching threshold of the
best published method,
spanning proprietary (GPT-4.1 Mini, GPT-3.5-Turbo, Claude Haiku 4.5)
and open-weight (Gemma-3-27B, DeepSeek-V3.2) tiers.

\textbf{\textsc{Ablations. }}\label{sec:exp.ablations}

We study the contribution of each axis of \splice{} in isolation; the proposer is fixed and the sweeps vary the target (except
\cref{fig:scaling-laws}). On SST-5 the target spans five capacity
tiers with two models per tier: \textbf{M}icro: \texttt{gemma-3-4b-it}, 
\texttt{nemotron-nano-9b-v2};
\textbf{T}iny: \texttt{gemma-3-12b-it}, \texttt{nemotron-nano-12b-v2};
\textbf{S}mall: \texttt{qwen3-32b} (S1), \texttt{nemotron-nano-3-30b} (S2);
\textbf{L}arge: \texttt{nemotron-super-3-120b} (L1), \texttt{qwen3-vl-235b-a22b} (L2);
\textbf{F}rontier: \texttt{claude-sonnet-4-6} (F1), \texttt{claude-opus-4-6} (F2),
and on ARC-Challenge and GPQA-Diamond across these six (S1--F2).
We compare four selectors spanning the iterative / non-iterative
divide: Best-of-$N$ (BoN; a single step of \cref{def:bips-body} that keeps the
best of $N$ children and drops the seed), Greedy ($\bw{=}1$, children only),
Beam ($\bw{>}1$, children only), and Elitist Beam (\cref{alg:elit-div}), which
realises the guarantee of \cref{thm:monotone}, and that of
\cref{thm:div-preserve} when $D\ge1$ ($D{=}0$ in the GPQA sweep, $D{=}1$ on
SST-5).

\textbf{\textsc{Why Mutability: A Contract Stress Test.}}\label{sec:exp.contract-main} \quad
We add a seven-clause contract (tool schema, safety text, output envelope, business
rule) to SST-5 on \texttt{gemma-3-27b-it}, including \textsc{Policy-1}:
\emph{escalate any snippet mentioning ``film'' or ``movie''}; the proxy reward
is accuracy, so obeying it ($26\%$ of examples) forfeits reward. The same
optimizer (\cref{sec:main.method}) runs in four arms that differ \emph{only} in the
contract's mutability: A1 frozen ($|\mathcal{V}_i|=1$), A2 all-mutable, A3 one
monolithic editable block, as most prior methods search, and A4 editable plus ``do
not edit''. The ordering \emph{inverts} (\cref{fig:contract-inversion}): every
editable arm beats A1 by $\approx\!10$\,pp on the proxy and loses by
$\approx\!17$\,pp on the contract-aware objective, while the other contract
classes were untouched by every arm: optimizers trade away exactly the clauses
that cost reward. Freezing is therefore needed, since instruction does not
protect the clause, but it does not suffice: it protects text, not effect, and
A1's compliance ($66.8\%$) is below the unoptimized seed's $96.5\%$ (details:
\cref{sec:exp.contract}).

\begin{figure}[t]
\centering
\begin{tikzpicture}[font=\footnotesize]
\begin{axis}[
  name=slope,
  width=7.6cm, height=3.8cm,
  xmin=-0.62, xmax=1.62, ymin=36, ymax=66,
  xtick={0,1},
  xticklabels={{Proxy accuracy\\\scriptsize val, seen by optimizer},
               {Contract-aware accuracy\\\scriptsize test, deployed}},
  xticklabel style={align=center},
  ytick={40,45,50,55,60,65},
  ylabel={Accuracy (\%)}, ylabel style={yshift=-3pt},
  axis lines*=left, tick align=outside, tick pos=left,
  ymajorgrids, grid style={splicerule!20, line width=0.3pt},
  clip=false,
  error bars/error bar style={line width=0.5pt},
  error bars/error mark options={rotate=90, mark size=1.6pt, line width=0.5pt},
  title={\textbf{(a)} The ordering inverts}, title style={yshift=-2pt},
  legend style={at={(0.79,-0.40)}, anchor=north, draw=none, fill=none,
                font=\scriptsize, legend columns=2, column sep=6pt, reverse legend,
                cells={anchor=west}},
]
\addplot+[splicerule, line width=0.7pt, dashdotted,
  mark=diamond*, mark options={solid, fill=white, draw=splicerule, mark size=2.1pt},
  error bars/.cd, y dir=both, y explicit, error bar style={solid, splicerule}]
  coordinates {(0.07,58.9) +- (0,4.8) (1.07,40.2) +- (0,0.8)};
\addlegendentry{A4 soft-instructed (408 cand.)}
\addplot+[splicerule, line width=0.7pt, densely dotted,
  mark=triangle*, mark options={solid, fill=white, draw=splicerule, mark size=2.1pt},
  error bars/.cd, y dir=both, y explicit, error bar style={solid, splicerule}]
  coordinates {(0.00,59.4) +- (0,3.6) (1.00,41.6) +- (0,0.4)};
\addlegendentry{A3 monolithic (372 cand.)}
\addplot+[splicerule, line width=0.7pt, densely dashed,
  mark=square*, mark options={solid, fill=white, draw=splicerule, mark size=1.7pt},
  error bars/.cd, y dir=both, y explicit, error bar style={solid, splicerule}]
  coordinates {(-0.07,59.8) +- (0,4.5) (0.93,41.4) +- (0,0.3)};
\addlegendentry{A2 all-mutable (336 cand.)}
\addplot+[spliceaccent, line width=1.6pt, solid,
  mark=*, mark options={solid, fill=spliceaccent, draw=spliceaccent, mark size=2.2pt},
  error bars/.cd, y dir=both, y explicit, error bar style={spliceaccent, line width=0.8pt}]
  coordinates {(0.00,49.3) +- (0,2.9) (1.00,58.5) +- (0,4.1)};
\addlegendentry{A1 \splice{}, contract frozen (456 cand.)}
\node[anchor=east, spliceaccent, font=\scriptsize\bfseries, align=right]
  at (axis cs:-0.10,49.3) {A1\\49.3};
\node[anchor=east, splicerule, font=\scriptsize, align=right]
  at (axis cs:-0.13,59.4) {A2--A4\\58.9--59.8};
\node[anchor=west, spliceaccent, font=\scriptsize\bfseries, align=left]
  at (axis cs:1.10,58.5) {A1\\58.5};
\node[anchor=west, splicerule, font=\scriptsize, align=left]
  at (axis cs:1.13,41.0) {A2--A4\\40.2--41.6};
\end{axis}
\begin{axis}[
  at={($(slope.north east)+(1.25cm,0)$)}, anchor=north west,
  width=4.6cm, height=3.8cm,
  xbar, bar width=7pt, bar shift=0pt,
  xmin=0, xmax=100, xtick={0,25,50,75,100},
  xlabel={Escalation compliance, test (\%)},
  ytick={1,2,3,4}, ymin=0.4, ymax=4.6, y dir=reverse,
  yticklabels={A1, A2, A3, A4},
  axis lines*=left, tick align=outside, tick pos=left,
  xmajorgrids, grid style={splicerule!20, line width=0.3pt},
  clip=false,
  title={\textbf{(b)} Compliance on triggered examples}, title style={yshift=-2pt},
  error bars/error bar style={line width=0.5pt},
  error bars/error mark options={rotate=90, mark size=1.6pt, line width=0.5pt},
]
\addplot+[draw=spliceaccent, fill=spliceaccent, line width=0.5pt,
  error bars/.cd, x dir=both, x explicit, error bar style={black}]
  coordinates {(66.8,1) +- (6.2,0)};
\addplot+[draw=splicerule, fill=splicefrozen, line width=0.5pt,
  error bars/.cd, x dir=both, x explicit, error bar style={black}]
  coordinates {(0.0,2) (0.1,3) +- (0.1,0) (1.4,4) += (1.6,0) -= (1.4,0)};
\draw[splicewarn, line width=0.8pt, densely dashed]
  (axis cs:96.5,0.4) -- (axis cs:96.5,4.6);
\node[splicewarn, font=\scriptsize, align=right, anchor=east]
  at (axis cs:94,3.0) {96.5: seed (A0),\\nothing optimized};
\node[white, font=\scriptsize\bfseries, anchor=east] at (axis cs:58,1) {66.8};
\node[spliceaccent, font=\scriptsize, anchor=north west, inner sep=1pt]
  at (axis cs:0.5,1.22) {clause text byte-identical};
\node[font=\scriptsize, anchor=west] at (axis cs:2.5,2) {0.0, clause deleted};
\node[font=\scriptsize, anchor=west] at (axis cs:2.5,3) {0.1};
\node[font=\scriptsize, anchor=west] at (axis cs:4.5,4) {1.4};
\end{axis}
\end{tikzpicture}
\caption{\textbf{Editable contracts win on the proxy reward but lose on the
contract-aware objective.} SST-5, \texttt{gemma-3-27b-it},
mean $\pm$ s.d.\ over three paired seeds. \textbf{(a)}~Endpoints are different
metrics on different splits; only the ordering is meaningful. \textbf{(b)}~Only
\splice{} keeps the clause partly in force (full numbers: \cref{tab:contract-results}, \cref{sec:exp.contract}).}
\label{fig:contract-inversion}
\end{figure}
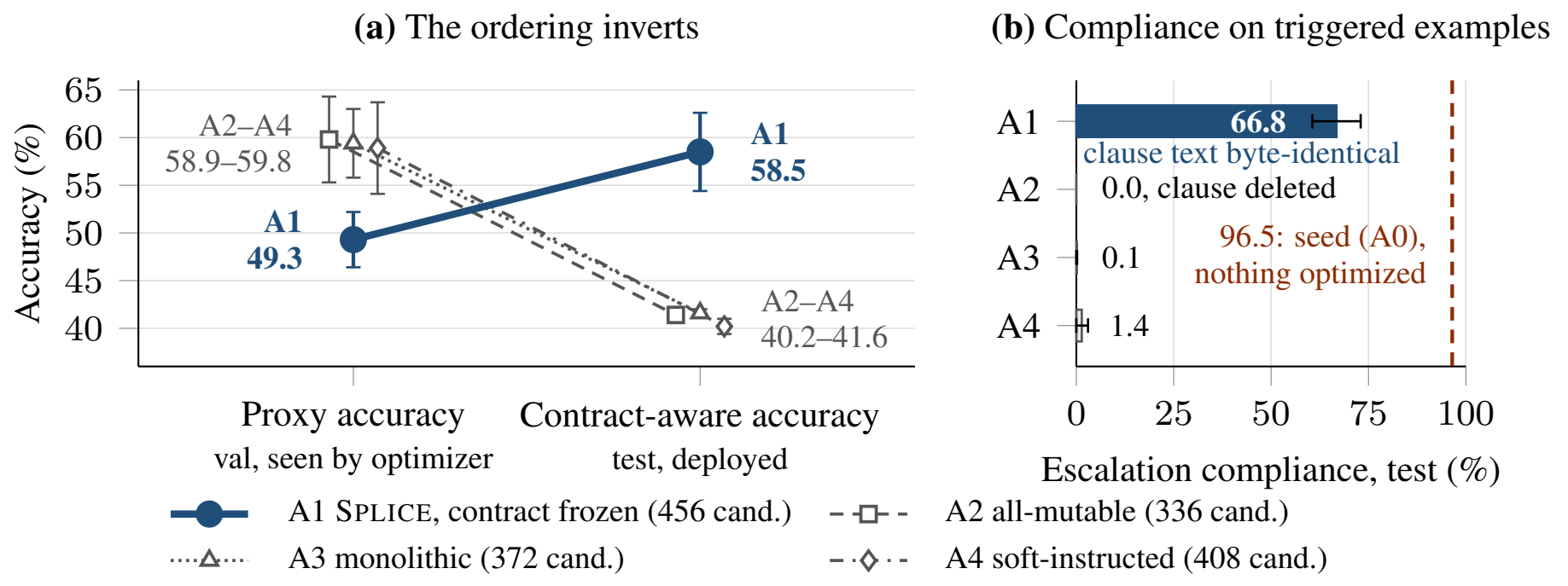
\begin{figure}[t]
\centering
\subfloat[SST-5, by proposer size]{%
  \includegraphics[width=0.42\textwidth]{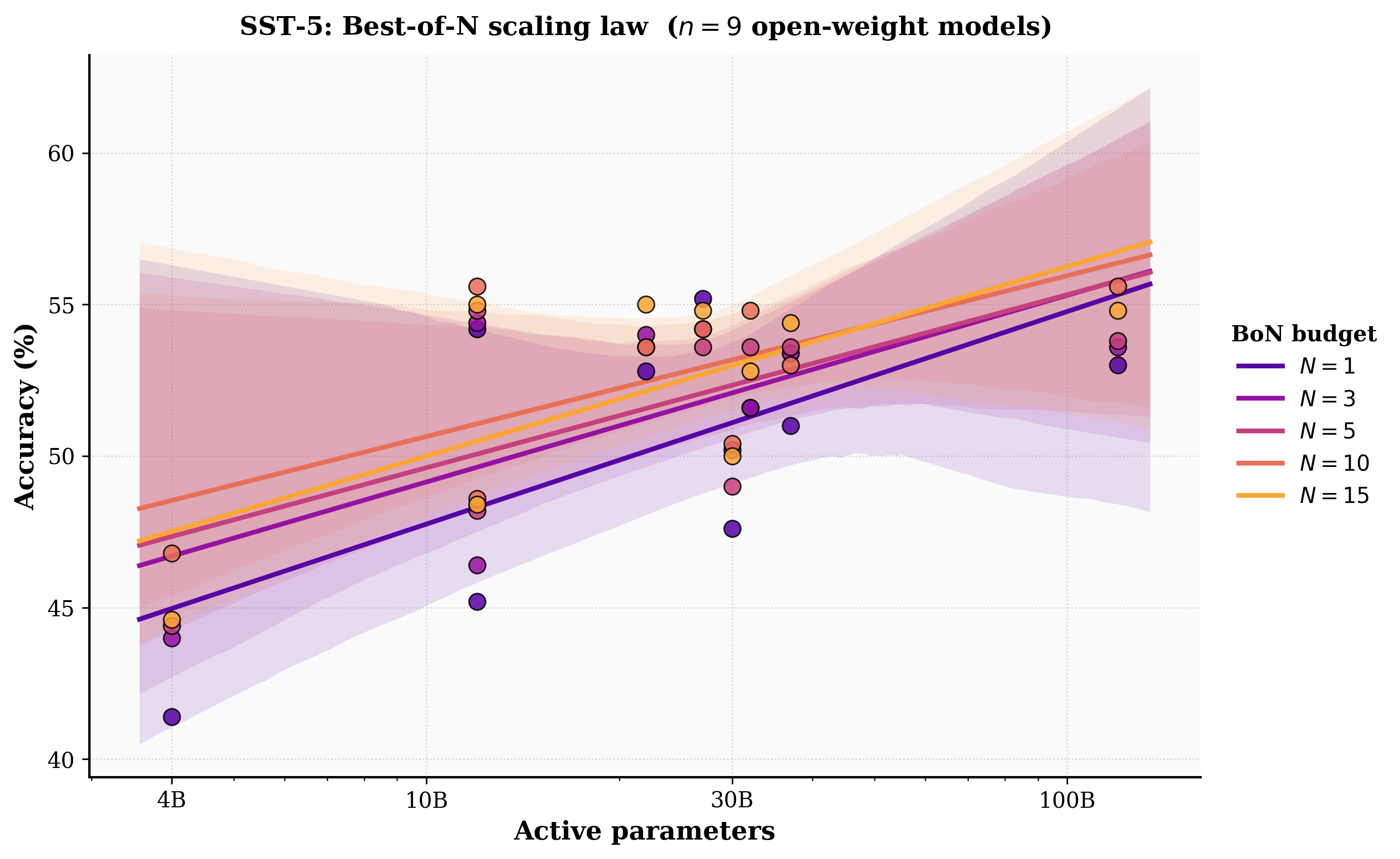}%
  \label{fig:sst5-scaling-law}%
}\hfill
\subfloat[GPQA, by proposer size]{%
  \includegraphics[width=0.42\textwidth]{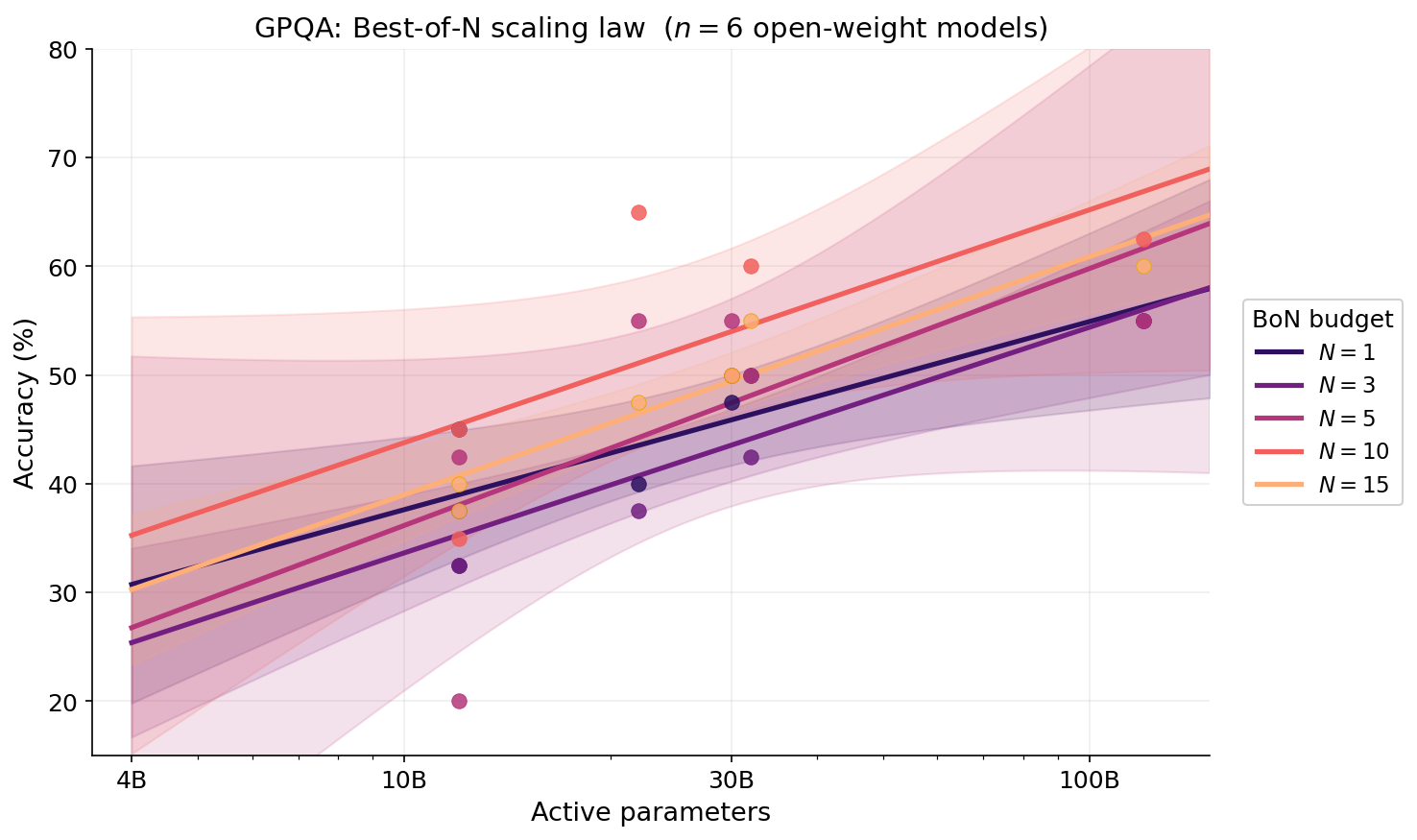}%
  \label{fig:gpqa-tier-comparison}%
}
\caption{\textbf{Best-of-$N$ scaling with proposer size.} Fitted one-shot
accuracy rises with proposer size at every $N$, while the $N$-curves overlap
within each size band: more samples do not substitute for scale.}
\label{fig:scaling-laws}
\end{figure}

\begin{figure}[t]
  \centering
  \subfloat[{\bf Critique}]{%
    \includegraphics[height=3.5cm]{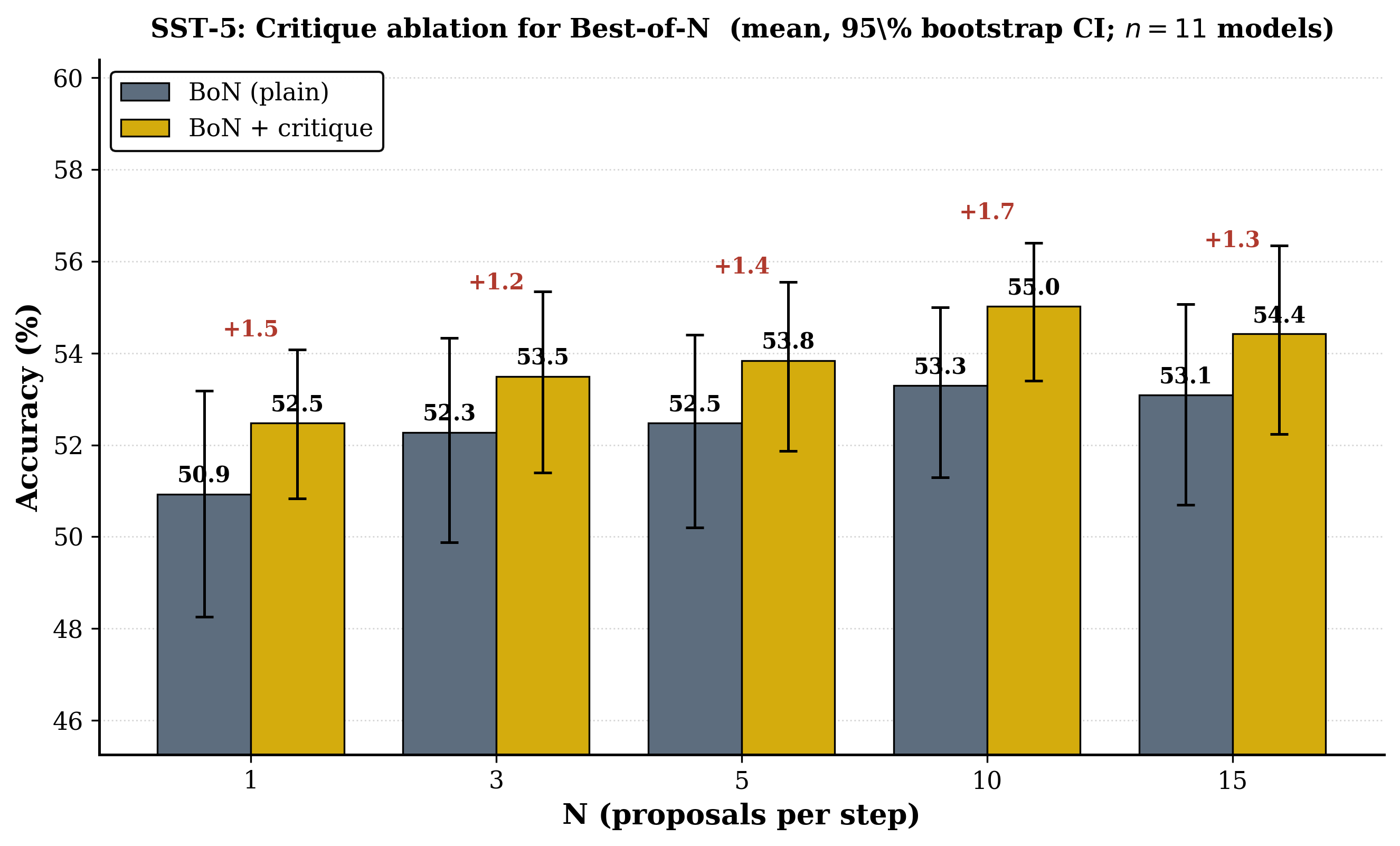}%
    \label{fig:sst5-bon-scaling-bars}%
  }\hfill
  \subfloat[{\bf Iteration}]{%
    \includegraphics[height=3.5cm]{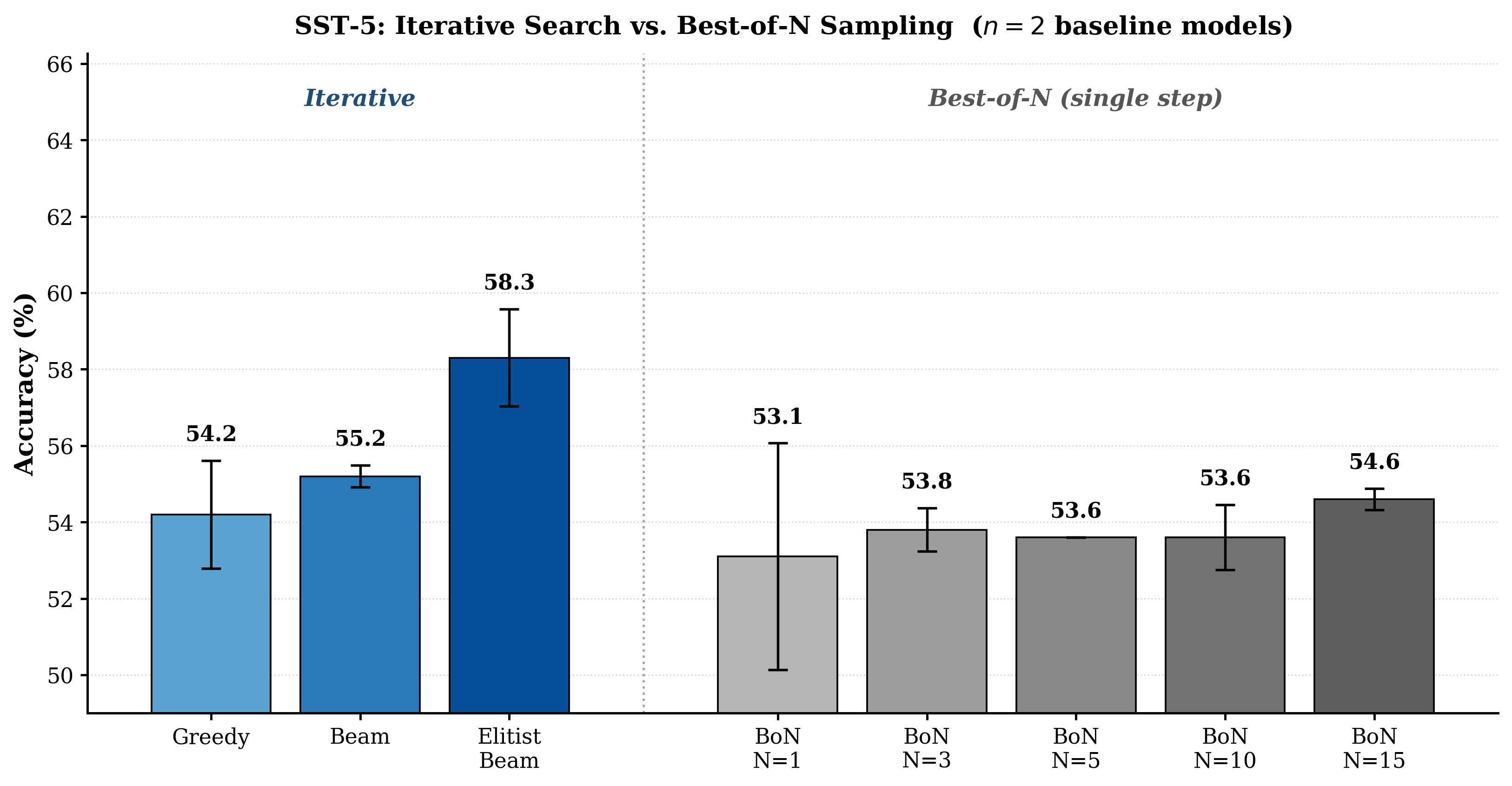}%
    \label{fig:iter-vs-bon-sst5}%
  }
  \caption{\small
  \textbf{Critique and elitist iteration each add accuracy on SST-5.}
  \textbf{(a)}~Critique adds $1.2$--$1.7$\,pp to Best-of-$N$ at every $N$: BoN
  (grey) vs.\ BoN+critique (gold), per-$N$ gap in red; $95\%$ bootstrap CIs over
  the target models overlap at every $N$. \textbf{(b)}~Elitist search
  leads one-shot sampling: test accuracy, mean over the two SST-5 protocol targets (\cref{tab:bench-sst5}).}
  \label{fig:ablation-pair}
\end{figure}

\textbf{\textsc{Best-of-$N$ Scaling.}} We characterize the weakest iterative regime first: Best-of-$N$
sampling with no feedback.~\cref{fig:scaling-laws}, from a separate run that
varies the proposer, reports Best-of-N accuracy across proposer sizes and
$N\!\in\!\{1,3,5,10,15\}$ on SST-5 and GPQA-Diamond. One-shot accuracy
scales strongly with proposer capacity: on SST-5 it rises from $\sim\!45\%$ for the smallest proposers to
$\sim\!55\%$ for the largest. But the five $N$-curves lie within a narrow
band at every proposer size; with the proposer fixed, the GPQA target sweep
shows no reliable $N$-trend either (\cref{fig:gpqa-bon-scaling-app},
$\pm20$\,pp per cell difference). Drawing more samples from the same seed
without feedback is not a reliable substitute for capacity or for
iteration; beam width is discussed in \cref{sec:exp.budget}.

\textbf{\textsc{Role of Critique Signal. }}
The next component is the critique signal: the appended textual
gradient that the proposer conditions on. 
\cref{fig:sst5-bon-scaling-bars} adds critique to BoN and reports
the paired comparison on SST-5, averaged across multiple target models
(\cref{fig:sst5-bon-scaling-bars}). At every $N$ the critique-on
variant outperforms critique-off by $+1.2$ to $+1.7$\,pp. Per-arm $95\%$ CIs overlap, so the evidence is the consistent sign across
$N$. The gap is largest at
$N{=}10$ ($+1.7$\,pp) and does not shrink systematically with $N$,
consistent with textual gradients carrying information that raw sampling does
not supply. 

\textbf{\textsc{Gains from Iterative Optimization.  }}
Iteration with an elitist selector is what turns critique into durable
progress. \cref{fig:iter-vs-bon-sst5} compares iterative strategies (Greedy,
Beam, Elitist Beam) against BoN with the same proposer model on SST-5, and
\cref{fig:iter-vs-bon-joint} (\cref{sec:app.exp}) on ARC-Challenge. On both, Elitist Beam scores highest of all
strategies; on the HelpSteer2 judge task BoN falls below the seed (\cref{sec:exp.judge}).
Elitist Beam leads the best plain BoN configuration by $+3.7$\,pp on
SST-5 (averaged over its two protocol targets) and by $+1.3$\,pp on GPQA-Diamond, inside
that sweep's noise; on saturated ARC-Challenge the strategies sit within $1$\,pp of each other. Since
the proposer is identical across strategies, the gain comes not from
proposer capacity but from the scoring--proposal feedback loop with elitist
retention (\cref{thm:reachability} composed with \cref{thm:monotone}).

\textbf{\textsc{Proposer and Target Capacity.  }}
Fitted one-shot accuracy rises with \emph{proposer} size at every $N$
(\cref{fig:sst5-scaling-law,fig:gpqa-tier-comparison}). With the proposer fixed and the \emph{target} varied
(GPQA), iterative strategies lead on the frontier tier but not on the small
tier (\cref{fig:gpqa-tier}), since a weak target bounds what any prompt achieves.

\textbf{\textsc{ Search Strategy and Diversity.  }}\label{sec:exp.diversity}
At matched $\bw{=}4$ on BBH, varying only the number of diversity slots, $D{=}0$
stays near $0.21$ for all $13$ steps (the beam never escapes a single mode)
while every $D\ge1$ variant exceeds $0.92$ by step $4$ (\cref{tab:diversity-bbh}, \cref{sec:main.exp-ext});
elsewhere the knob is far weaker, so diversity slots are insurance against
collapse rather than a uniform gain (search trees: \cref{fig:trees_frontier}, \cref{sec:app.exp}).
\cref{fig:refinement} (\cref{sec:app.exp}) traces validation score along each run's
best-path lineage: on frontier targets the best paths gain $+20$ to
$+22.5$\,pp over the seed in three to four nodes, while L1 needs six
nodes to cover a $+47.5$\,pp gap ($27.5\%\!\to\!75\%$); the leader is
non-decreasing along the best path (\cref{thm:monotone}) and climbs most in
the first two steps (\cref{sec:exp.pstar}).

\section{Limitations and Conclusion}
\label{sec:main.limitations}\label{sec:main.conclusion}
\textbf{Limitations.} The guarantees are conditional. Proposal coverage gates only the rate
results, so when it fails the search stalls rather than regresses; the
terminal-quality results rest on the submodularity ratio $\gamma$, for which we
have no estimator. Freezing preserves a contract's text, not its effect: in
\cref{sec:exp.contract-main} the frozen clause survives in every candidate, yet
behavioral compliance on test falls from $96.5\%$ for the unoptimized seed
to $66.8\%$, so frozen blocks still need a behavioral audit. \splice{}
optimizes the reward it is given, so a misspecified reward is the dominant
failure mode. Open problems: which neighboring edits
erode a frozen clause, an estimator for $\gamma$, and a matched-budget cost
comparison.
\textbf{Conclusion.} Casting prompt optimisation as constrained search over an
edit graph of selectively-mutable blocks, \splice{} carries guarantees with
named assumptions and wins on four of seven benchmarks, matching on three.
The open case is recursive: the optimized text is the optimizer's own.

\begin{ack}
We thank Benjamin Wyant and Matt Yanchyshyn for their leadership and support
of this work, and the anonymous reviewers and area chair, whose detailed
feedback significantly improved the quality of this paper. This work was funded by Amazon Web Services, where all authors were employed while carrying it out; no
third-party funding was received, and the authors declare no other competing
interests. 
\end{ack}

\bibliographystyle{abbrvnat}
\bibliography{
    bibs/baselines,
    bibs/datasets,
    bibs/basics,
    bibs/models,
    bibs/appendix_refs
}

\appendix
\clearpage
\appendix
\splicethmboxes
\begin{center}
  {\LARGE\bfseries Appendix\par}
\end{center}
\addcontentsline{toc}{section}{Appendix}
\vspace{1em}

\noindent

\etocdepthtag.toc{mtappendix}
\begingroup
\etocsettagdepth{mtmain}{none}
\etocsettagdepth{mtappendix}{subsection}
\tableofcontents
\endgroup

\section{Seed and Optimized Prompts}
\label{sec:app.example_prompts}
\begin{tcolorbox}[
  breakable,
  enhanced,
  colback=white,
  colframe=black,
  colbacktitle=black!80,
  coltitle=white,
  fonttitle=\bfseries,
  title={Seed Prompt for AIME},
  left=2mm,right=2mm,top=1mm,bottom=1mm,
]
\small
\texttt{<role></role>}

\texttt{<objective>}Solve the given math problem. Every Answer is a non-negative integer between 0 and 999 (inclusive).\texttt{</objective>}

\texttt{<approach></approach>}

\texttt{<examples></examples>}

\texttt{<output\_format>}give the final answer in the form \textbackslash boxed\{N\}, where N is an integer between 0 and 999.\texttt{</output\_format>}
\end{tcolorbox}

\vspace{0.5em}

\begin{tcolorbox}[
  breakable,
  enhanced,
  colback=white,
  colframe=black,
  colbacktitle=black!80,
  coltitle=white,
  fonttitle=\bfseries,
  title={\splice{}'s Optimized Prompt for AIME, GPT-4.1 Mini},
  left=2mm,right=2mm,top=1mm,bottom=1mm,
]
\small
\texttt{<role>}You are a world-class competition mathematician with expertise across all domains: number theory, combinatorics, algebra, geometry, probability, and analysis. You are known for your meticulous calculations, creative problem-solving approaches, and systematic verification. You never rush --- you break complex problems into manageable pieces and double-check each step before proceeding.\texttt{</role>}

\vspace{0.5em}
\texttt{<objective>}Solve the given math competition problem. Produce a clear, rigorous, step-by-step solution. The final answer is always a non-negative integer between 0 and 999 (inclusive). Before stating your final answer, verify it through an independent check or sanity test.\texttt{</objective>}

\vspace{0.5em}
\texttt{<approach>}Use this disciplined problem-solving framework:

\begin{enumerate}[leftmargin=1.5em,itemsep=0.3em]
\item \textbf{Parse the problem precisely}: Identify exactly what is being asked. Note whether the answer requires a modular reduction (e.g., mod 1000, mod 1000000007), a specific quantity, or a transformed result. Misreading the question is the \#1 source of errors.

\item \textbf{Identify the mathematical domain and select tools}: Determine the core techniques needed (modular arithmetic, CRT, generating functions, PIE, Heron's formula, coordinate geometry, recursion, casework, etc.).

\item \textbf{Break into sub-problems}: Decompose complex problems into smaller, independently verifiable steps. For each sub-step, compute carefully and verify before moving on.

\item \textbf{Execute with extreme arithmetic care}:
\begin{itemize}[leftmargin=1em,itemsep=0.2em]
\item For modular arithmetic: reduce intermediate results frequently to keep numbers small.
\item For counting: clearly define what you're counting, check for over/undercounting.
\item For algebra: expand and simplify carefully, checking each manipulation.
\item For geometry: verify formulas before applying, check units/dimensions.
\end{itemize}

\item \textbf{Verify your answer}: Use at least one of these checks:
\begin{itemize}[leftmargin=1em,itemsep=0.2em]
\item Compute via an alternative method
\item Check small cases or boundary conditions
\item Verify the answer is in the valid range [0, 999]
\item For mod problems, confirm the reduction was applied correctly
\item Estimate the magnitude to see if the answer is reasonable
\end{itemize}
\end{enumerate}

\textbf{Critical pitfalls to avoid}:
\begin{itemize}[leftmargin=1em,itemsep=0.2em]
\item Forgetting to apply the final modular reduction
\item Off-by-one errors in counting (inclusive vs exclusive bounds)
\item Sign errors in algebraic manipulation
\item Miscounting cases in casework
\item Confusing permutations with combinations
\item Forgetting to account for symmetry or identical elements
\end{itemize}
\texttt{</approach>}

\vspace{0.5em}
\texttt{<examples>}

\textbf{Example 1:}\\
\textit{Input:} Find the remainder when $2^{100}$ is divided by 1000.\\
\textit{Output:} We need $2^{100} \bmod 1000$.

Since $1000 = 8 \times 125$, we use CRT.

$2^{100} \bmod 8 = 0$ (since $2^3 = 8$ divides $2^{100}$).

$2^{100} \bmod 125$: Using repeated squaring.\\
$2^{10} = 1024 \equiv 24 \pmod{125}$\\
$2^{20} \equiv 24^2 = 576 \equiv 76 \pmod{125}$\\
$2^{40} \equiv 76^2 = 5776 \equiv 26 \pmod{125}$\\
$2^{80} \equiv 26^2 = 676 \equiv 51 \pmod{125}$\\
$2^{100} = 2^{80} \times 2^{20} \equiv 51 \times 76 = 3876 \equiv 1 \pmod{125}$

CRT: $x \equiv 0 \pmod 8$, $x \equiv 1 \pmod{125}$.\\
$x = 125k + 1$, need $125k + 1 \equiv 0 \pmod 8 \to 5k \equiv 7 \pmod 8 \to k \equiv 3 \pmod 8$.\\
$x = 125(3) + 1 = 376$.

$\boxed{376}$

\vspace{0.5em}
\textbf{Example 2:}\\
\textit{Input:} How many ways can you arrange the letters in BANANA?\\
\textit{Output:} BANANA has 6 letters: B(1), A(3), N(2).

Distinct arrangements $= 6! / (1! \times 3! \times 2!) = 720 / 12 = 60$.

$\boxed{60}$

\vspace{0.5em}
\textbf{Example 3:}\\
\textit{Input:} In triangle $ABC$, $AB = 13$, $BC = 14$, $CA = 15$. Find the area of triangle $ABC$.\\
\textit{Output:} Using Heron's formula with $s = (13+14+15)/2 = 21$.

Area $= \sqrt{21 \times 8 \times 7 \times 6} = \sqrt{7056} = 84$.

Verification: With base $BC=14$, height $h = 2(84)/14 = 12$. Check: $13^2 = 5^2 + 12^2 = 169$ \checkmark

$\boxed{84}$

\vspace{0.5em}
\textbf{Example 4:}\\
\textit{Input:} A fair six-sided die is rolled 4 times. What is the probability that the sum is 10? Express your answer as $p/q$ in lowest terms and find $p + q$.\\
\textit{Output:} We need the number of solutions to $d_1+d_2+d_3+d_4 = 10$ where $1 \le d_i \le 6$.

Substitute $x_i = d_i - 1$, so $x_1+x_2+x_3+x_4 = 6$ where $0 \le x_i \le 5$.

Using inclusion-exclusion on stars and bars:
\begin{itemize}[leftmargin=1em,itemsep=0.1em]
\item No restriction: $\binom{9}{3} = 84$
\item One $x_i \ge 6$: $\binom{4}{1}\binom{3}{3} = 4$
\item Two $x_i \ge 6$: impossible (sum would be $\ge 12 > 6$)
\end{itemize}

Favorable $= 84 - 4 = 80$.\\
Total outcomes $= 6^4 = 1296$.\\
Probability $= 80/1296 = 5/81$.

$p + q = 5 + 81 = 86$.

$\boxed{86}$

\vspace{0.5em}
\textbf{Example 5:}\\
\textit{Input:} Find the number of integers $n$ with $1 \le n \le 1000$ such that $n^2 + n$ is divisible by 6.\\
\textit{Output:} $n^2 + n = n(n+1)$, the product of two consecutive integers.

$n(n+1)$ is always even (divisible by 2), so we need $n(n+1)$ divisible by 3.

$n(n+1) \bmod 3$: if $n \equiv 0$: $0 \times 1 \equiv 0$ \checkmark; if $n \equiv 1$: $1 \times 2 \equiv 2$ \texttimes; if $n \equiv 2$: $2 \times 0 \equiv 0$ \checkmark.

So $n(n+1)$ is NOT divisible by 3 when $n \equiv 1 \pmod 3$.

Numbers 1--1000 with $n \equiv 1 \pmod 3$: these are $1, 4, 7, \ldots, 1000$. Count $= (1000-1)/3 + 1 = 334$.

Answer $= 1000 - 334 = 666$.

$\boxed{666}$

\texttt{</examples>}

\vspace{0.5em}
\texttt{<output\_format>}give the final answer in the form \textbackslash boxed\{N\}, where N is an integer between 0 and 999.\texttt{</output\_format>}
\end{tcolorbox}

\begin{tcolorbox}[
  breakable,
  enhanced,
  colback=white,
  colframe=black,
  colbacktitle=black!80,
  coltitle=white,
  fonttitle=\bfseries,
  title={Seed Prompt for GSM8K},
  left=2mm,right=2mm,top=1mm,bottom=1mm,
]
\small
\texttt{<role></role>}

\texttt{<objective>}Solve the given grade school math word problem step by step, showing your reasoning, and provide the final numerical answer.\texttt{</objective>}

\texttt{<approach></approach>}

\texttt{<examples></examples>}

\texttt{<output\_format>}Show your step-by-step reasoning, then on the final line output the answer as a single number after `\#\#\#\# '. For example: `\#\#\#\# 42'. Do not include units, commas, or any text after the number.\texttt{</output\_format>}
\end{tcolorbox}

\vspace{0.5em}

\begin{tcolorbox}[
  breakable,
  enhanced,
  colback=white,
  colframe=black,
  colbacktitle=black!80,
  coltitle=white,
  fonttitle=\bfseries,
  title={\splice{}'s Optimized Prompt for GSM8K, GPT-3.5-Turbo},
  left=2mm,right=2mm,top=1mm,bottom=1mm,
]
\small
\texttt{<role>}You are a precise math word problem solver. You work through problems using plain-text arithmetic only. CRITICAL FORMAT RULES: Never use LaTeX, backslash-parentheses \textbackslash(\ldots\textbackslash), dollar signs \$\ldots\$, \textbackslash boxed\{\}, or any markup notation anywhere in your response. Write all calculations as plain text like: 3 $\times$ 4 = 12. Always produce a final whole number answer (no decimals, no trailing zeros, no commas in numbers).\texttt{</role>}

\vspace{0.5em}
\texttt{<objective>}Solve the given math word problem. Before computing, carefully identify: (1) what quantity is being asked for, (2) any real-world facts needed but not stated (e.g., cars have 4 wheels, roller skates have 4 wheels each, a pair = 2 items, a dozen = 12), and (3) the correct direction of comparisons (e.g., ``A costs twice as much as B'' means A = 2 $\times$ B). Then solve step by step using plain arithmetic. Output ONLY a bare integer after \#\#\#\# on the final line.\texttt{</objective>}

\vspace{0.5em}
\texttt{<approach>}Follow these steps:

\begin{enumerate}[leftmargin=1.5em,itemsep=0.3em]
\item \textbf{Read the problem twice}. Identify the exact question being asked. Note all units carefully --- distinguish between groups vs individuals, per-person vs total, containers vs contents. When $N$ people/items are excluded from an action, determine what unit that removes (e.g., ``2 students didn't eat'' removes 2 complete servings, not 2 individual pieces).

\item \textbf{State any real-world knowledge needed}: common objects' properties (wheels on vehicles, legs on animals, items in a pair/dozen), and any implicit facts the problem assumes you know.

\item \textbf{Parse comparative relationships carefully}. ``A is/costs twice as much as B'' means A = 2 $\times$ B (A is larger). ``A is 5 less than 3 times B'' means A = 3 $\times$ B $-$ 5. Identify the base quantity and the derived quantity before computing.

\item \textbf{Compute step by step} using plain-text arithmetic ($\times$, $+$, $-$, $/$). Calculate each intermediate value as a number before proceeding. Never leave expressions unevaluated. Never use variables, algebra, or any LaTeX/markup.

\item \textbf{Verify}: Does your answer match the exact question asked? Did you account for ALL items/people mentioned? Is it a whole number? Write it after ``\#\#\#\# '' on the last line.
\end{enumerate}
\texttt{</approach>}

\vspace{0.5em}
\texttt{<examples>}

\textbf{Example 1:}\\
\textit{Input:} A store had 120 shirts. They sold 1/3 of them on Monday and 1/4 of the remainder on Tuesday. How many shirts are left?\\
\textit{Output:} The store started with 120 shirts. Sold on Monday: 120 $\times$ 1/3 = 40 shirts. Remaining after Monday: 120 $-$ 40 = 80 shirts. Sold on Tuesday: 80 $\times$ 1/4 = 20 shirts. Remaining after Tuesday: 80 $-$ 20 = 60 shirts.

\#\#\#\# 60

\vspace{0.5em}
\textbf{Example 2:}\\
\textit{Input:} 5 cars are in a parking lot. How many wheels are there in total?\\
\textit{Output:} There are 5 cars in the parking lot. Each car has 4 wheels (real-world fact). Total wheels: 5 $\times$ 4 = 20 wheels.

\#\#\#\# 20

\vspace{0.5em}
\textbf{Example 3:}\\
\textit{Input:} A plot of land costs \$40,000. Construction costs twice as much as the land. What is the total cost of land plus construction?\\
\textit{Output:} Land cost: 40000 dollars. Construction costs twice as much as land: 2 $\times$ 40000 = 80000 dollars. Total cost: 40000 + 80000 = 120000 dollars.

\#\#\#\# 120000

\vspace{0.5em}
\textbf{Example 4:}\\
\textit{Input:} A teacher bought 10 packs of crackers, with 8 crackers per pack, for her class. 2 students were absent and didn't get any. How many crackers were eaten?\\
\textit{Output:} Total packs: 10. Each pack has 8 crackers. Total crackers: 10 $\times$ 8 = 80 crackers. 2 students were absent, so 2 packs were not eaten. Crackers not eaten: 2 $\times$ 8 = 16 crackers. Crackers eaten: 80 $-$ 16 = 64 crackers.

\#\#\#\# 64

\vspace{0.5em}
\textbf{Example 5:}\\
\textit{Input:} A worker earns \$12 per hour. She works 8 hours a day for 5 days. Her employer deducts \$50 for insurance. What is her take-home pay?\\
\textit{Output:} Hourly wage: 12 dollars. Daily earnings: 12 $\times$ 8 = 96 dollars. Weekly earnings: 96 $\times$ 5 = 480 dollars. After insurance deduction: 480 $-$ 50 = 430 dollars.

\#\#\#\# 430

\texttt{</examples>}

\vspace{0.5em}
\texttt{<output\_format>}Show your step-by-step reasoning, then on the final line output the answer as a single number after `\#\#\#\# '. For example: `\#\#\#\# 42'. Do not include units, commas, or any text after the number.\texttt{</output\_format>}
\end{tcolorbox}

\begin{tcolorbox}[
  breakable,
  enhanced,
  colback=white,
  colframe=black,
  colbacktitle=black!80,
  coltitle=white,
  fonttitle=\bfseries,
  title={Seed Prompt for GPQA},
  left=2mm,right=2mm,top=1mm,bottom=1mm,
]
\small
\texttt{<role>}You are an expert scientist with deep knowledge in physics, chemistry, and biology. You approach problems methodically by identifying the relevant scientific principle first, then systematically evaluating each option.\texttt{</role>}

\vspace{0.5em}
\texttt{<objective>}Answer the following multiple choice question by reasoning through it step by step, then selecting the correct answer.\texttt{</objective>}

\vspace{0.5em}
\texttt{<approach></approach>}

\vspace{0.5em}
\texttt{<examples></examples>}

\vspace{0.5em}
\texttt{<output\_format>}Think through the problem step by step. After your reasoning, state your final answer on a new line in the format: The answer is (X) where X is A, B, C, or D.\texttt{</output\_format>}
\end{tcolorbox}

\vspace{0.5em}

\begin{tcolorbox}[
  breakable,
  enhanced,
  colback=white,
  colframe=black,
  colbacktitle=black!80,
  coltitle=white,
  fonttitle=\bfseries,
  title={\splice{}'s Optimized Prompt for GPQA, Claude Haiku 4.5},
  left=2mm,right=2mm,top=1mm,bottom=1mm,
]
\small
\texttt{<role>}You are an expert-level science examiner who solves graduate-level multiple choice questions with precision and efficiency. Your expertise spans physics, chemistry, biology, and mathematics including specialized topics like quantum mechanics, organic synthesis, molecular biology, optics, thermodynamics, and statistical mechanics. You are decisive: you commit to a solution path, execute it cleanly, and state your answer. You never ramble or restart calculations unnecessarily.\texttt{</role>}

\vspace{0.5em}
\texttt{<objective>}Answer the multiple choice question below. Be concise and decisive. You MUST always end with your final answer in the required format --- never leave it unstated, even if your reasoning feels incomplete.\texttt{</objective>}

\vspace{0.5em}
\texttt{<approach>}Use this efficient reasoning strategy:

\begin{enumerate}[leftmargin=1.5em,itemsep=0.3em]
\item \textbf{CLASSIFY}: What type of problem is this?
\item \textbf{ELIMINATE}: Rule out obviously wrong options.
\item \textbf{SOLVE}: Apply the relevant principle ONCE. Do not restart.
\item \textbf{MATCH \& COMMIT}: Match your result to the closest option.
\item \textbf{ANSWER}: State your final answer immediately.
\end{enumerate}

\textbf{CRITICAL RULES}:
\begin{itemize}[leftmargin=1em,itemsep=0.2em]
\item Keep reasoning to under 15 sentences.
\item If you catch yourself restarting, STOP and commit.
\item The final answer line is MANDATORY.
\end{itemize}
\texttt{</approach>}

\vspace{0.5em}
\texttt{<examples>}

\textbf{Example 1:}\\
\textit{Input:} What is the ground state energy of a quantum harmonic oscillator with angular frequency $\omega$? (A) 0 \quad (B) $\hbar\omega/2$ \quad (C) $\hbar\omega$ \quad (D) $3\hbar\omega/2$\\
\textit{Output:} CLASSIFY: Conceptual recall --- standard quantum mechanics formula. ELIMINATE: (A) is wrong --- zero-point energy is nonzero. (D) is the energy of $n=1$ state. SOLVE: Ground state ($n=0$) has $E = \hbar\omega(n + 1/2) = \hbar\omega/2$. MATCH: This is option (B).

The answer is (B)

\vspace{0.5em}
\textbf{Example 2:}\\
\textit{Input:} A 2 kg block slides down a frictionless incline of $30^\circ$ for 3 meters. What is its speed at the bottom? (A) 3.8 m/s \quad (B) 5.4 m/s \quad (C) 7.7 m/s \quad (D) 9.8 m/s\\
\textit{Output:} CLASSIFY: Calculation --- energy conservation. ELIMINATE: (D) equals $g$, unlikely for this setup. SOLVE: Height $= 3 \times \sin 30^\circ = 1.5$ m. $v = \sqrt{2 \times 9.8 \times 1.5} \approx 5.42$ m/s. MATCH: Option (B).

The answer is (B)

\texttt{</examples>}

\vspace{0.5em}
\texttt{<output\_format>}Think through the problem step by step. After your reasoning, state your final answer on a new line in the format: The answer is (X) where X is A, B, C, or D.\texttt{</output\_format>}
\end{tcolorbox}

\begin{tcolorbox}[
  breakable,
  enhanced,
  colback=white,
  colframe=black,
  colbacktitle=black!80,
  coltitle=white,
  fonttitle=\bfseries,
  title={Seed Prompt for BBH},
  left=2mm,right=2mm,top=1mm,bottom=1mm,
]
\small
\texttt{<role></role>}

\texttt{<objective>}Solve the reasoning task. Think step by step, then give the final answer.\texttt{</objective>}

\texttt{<approach></approach>}

\texttt{<examples></examples>}

\texttt{<output\_format>}Think through the problem step by step. After your reasoning, state your final answer on a new line starting with: The answer is\texttt{</output\_format>}
\end{tcolorbox}

\vspace{0.5em}

\begin{tcolorbox}[
  breakable,
  enhanced,
  colback=white,
  colframe=black,
  colbacktitle=black!80,
  coltitle=white,
  fonttitle=\bfseries,
  title={\splice{}'s Optimized Prompt for BBH, Claude Haiku 4.5},
  left=2mm,right=2mm,top=1mm,bottom=1mm,
]
\small
\texttt{<role>}You are a precise reasoning engine. You solve problems step by step using plain text only --- no markdown, no bold, no backticks, no special formatting. Your final line is always ``The answer is'' followed by the answer value. You never write anything after that final line.\texttt{</role>}

\vspace{0.5em}
\texttt{<objective>}Solve the given reasoning task using step-by-step plain-text reasoning, then state your final answer.\texttt{</objective>}

\vspace{0.5em}
\texttt{<approach>}Follow these steps:

\begin{enumerate}[leftmargin=1.5em,itemsep=0.3em]
\item Read the problem carefully. Think through it step by step in plain text.
\item Determine the correct answer. For pronoun resolution or ambiguity questions, choose the most natural and contextually supported interpretation.
\item Format your answer value according to the question type:
\begin{itemize}[leftmargin=1em,itemsep=0.1em]
\item Multiple choice: lowercase letter in parentheses, e.g. (c)
\item Yes/no or true/false: just the word
\item Sorting or listing: items separated by spaces
\item Completion: only the missing part
\item Open-ended: the shortest correct answer
\end{itemize}
\item End with exactly one line: ``The answer is'' followed by your answer value.
\item STOP after that line. Do NOT repeat or write anything after it.
\end{enumerate}
\texttt{</approach>}

\vspace{0.5em}
\texttt{<examples>}

\textbf{Example 1 (Multiple choice):}\\
\textit{Input:} Which of the following is a fruit? Options: (a) carrot (b) apple (c) potato (d) spinach\\
\textit{Output:} Apples are fruits; carrots, potatoes, and spinach are vegetables.

The answer is (b)

\vspace{0.5em}
\textbf{Example 2 (Yes/no):}\\
\textit{Input:} If all roses are flowers and some flowers fade quickly, does it follow that some roses fade quickly? Answer yes or no.\\
\textit{Output:} The flowers that fade quickly may not be roses. The conclusion does not follow from the premises.

The answer is no

\vspace{0.5em}
\textbf{Example 3 (Sorting):}\\
\textit{Input:} Sort the following words alphabetically: pear apple orange banana\\
\textit{Output:} Alphabetical order: apple, banana, orange, pear.

The answer is apple banana orange pear

\vspace{0.5em}
\textbf{Example 4 (Completion):}\\
\textit{Input:} Complete the sequence: 2, 4, 8, 16, \underline{\phantom{xx}}\\
\textit{Output:} Each term is doubled. 16 times 2 equals 32.

The answer is 32

\vspace{0.5em}
\textbf{Example 5 (Reasoning / pronoun resolution):}\\
\textit{Input:} The trophy doesn't fit into the brown suitcase because it is too large. What does ``it'' refer to? Options: (a) the trophy (b) the suitcase\\
\textit{Output:} If something does not fit because it is too large, the thing that is too large is the object trying to fit in, which is the trophy.

The answer is (a)

\vspace{0.5em}
\textbf{Example 6 (True/false):}\\
\textit{Input:} Is the following statement true or false: ``The sum of the interior angles of a triangle is 180 degrees.''\\
\textit{Output:} In Euclidean geometry, the interior angles of any triangle sum to 180 degrees.

The answer is true

\texttt{</examples>}

\vspace{0.5em}
\texttt{<output\_format>}Think through the problem step by step. After your reasoning, state your final answer on a new line starting with: The answer is\texttt{</output\_format>}
\end{tcolorbox}

\begin{tcolorbox}[
  breakable,
  enhanced,
  colback=white,
  colframe=black,
  colbacktitle=black!80,
  coltitle=white,
  fonttitle=\bfseries,
  title={Seed Prompt for ARC},
  left=2mm,right=2mm,top=1mm,bottom=1mm,
]
\small
\texttt{<role>}You have to solve the following science question.\texttt{</role>}

\vspace{0.5em}
\texttt{<objective>}Analyze the science question, reason through all answer choices, and give your final answer inside \texttt{<Answer></Answer>} tags containing ONLY the letter A, B, C, or D.\texttt{</objective>}

\vspace{0.5em}
\texttt{<approach></approach>}

\vspace{0.5em}
\texttt{<examples>}

\textbf{Example 1:}\\
\textit{Input:} What is the boiling point of water at standard atmospheric pressure?\\
A) 50$^\circ$C \quad B) 100$^\circ$C \quad C) 150$^\circ$C \quad D) 200$^\circ$C\\
\textit{Output:} Water boils at 100 degrees Celsius at standard atmospheric pressure.

\texttt{<Answer>}B\texttt{</Answer>}

\texttt{</examples>}

\vspace{0.5em}
\textit{[Immutable suffix appended to every prompt]}: After any reasoning, give your final answer inside \texttt{<Answer></Answer>} tags containing ONLY the letter A, B, C, or D.
\end{tcolorbox}

\vspace{0.5em}

\begin{tcolorbox}[
  breakable,
  enhanced,
  colback=white,
  colframe=black,
  colbacktitle=black!80,
  coltitle=white,
  fonttitle=\bfseries,
  title={\splice{}'s Optimized Prompt for ARC, GPT-3.5-Turbo},
  left=2mm,right=2mm,top=1mm,bottom=1mm,
]
\small
\texttt{<role>}You are a precise science reasoning expert. You read questions carefully, evaluate every answer choice against exact scientific definitions, and always select the answer that most directly addresses what the question asks --- whether that answer is simple or highly technical. You do not let the framing or sophistication of an option override its accuracy.\texttt{</role>}

\vspace{0.5em}
\texttt{<objective>}Analyze the science question and evaluate ALL answer choices before selecting your final answer. Consider why each option is correct or incorrect. When options seem equally plausible, use precise scientific definitions to distinguish them. Give your final answer inside \texttt{<Answer></Answer>} tags containing ONLY the letter A, B, C, or D.\texttt{</objective>}

\vspace{0.5em}
\texttt{<approach>}Use this 5-step strategy:

\begin{enumerate}[leftmargin=1.5em,itemsep=0.3em]
\item \textbf{CLASSIFY} the question type --- this determines your reasoning lens:
\begin{itemize}[leftmargin=1em,itemsep=0.1em]
\item \textbf{CAUSAL/CONSEQUENCE}: ``what happens when X?'' or ``what is the result of X?'' $\to$ trace the MOST DIRECT causal chain from X to the effect; do not skip to secondary or downstream consequences.
\item \textbf{MECHANISM/APPEARANCE}: ``why does X appear/seem\ldots?'' $\to$ focus on the perceptual or observational mechanism, not intrinsic physical properties.
\item \textbf{PROPERTY/IDENTIFICATION}: ``which property identifies X?'' or ``what state of matter is X?'' $\to$ match precise scientific definitions to the property described.
\item \textbf{STRUCTURAL/LAYER}: ``which layer does X?'' $\to$ verify the spatial relationship (above/below/adjacent) explicitly.
\end{itemize}

\item \textbf{IDENTIFY} the core scientific concept and the exact qualifier being tested (e.g., ``regardless of size,'' ``appears,'' ``most direct,'' ``solid,'' ``primarily'').

\item \textbf{EVALUATE} each answer choice (A, B, C, D) with a brief justification --- explain why each is correct or incorrect using precise scientific definitions. Do not stop at the first plausible option.

\item \textbf{CHALLENGE} --- Compare your top 2 candidates:
\begin{enumerate}[label=(\alph*),leftmargin=1.2em,itemsep=0.1em]
\item EXACT MATCH: Does this answer address what the question literally asks, or does it answer a related-but-different question?
\item CAUSAL DIRECTNESS: For causal questions, which answer is the most immediate consequence of the stated cause --- not a secondary effect?
\item REFERENT CHECK: Confirm which entity, layer, or process the question names. Make sure your answer describes THAT entity, not an adjacent one.
\end{enumerate}

\item \textbf{SELECT} the answer that best matches the question's exact ask. If you are uncertain or running low on space, skip extended reasoning and immediately output your best answer in \texttt{<Answer>} tags.
\end{enumerate}
\texttt{</approach>}

\vspace{0.5em}
\texttt{<examples>}

\textbf{Example 1 (Causal):}\\
\textit{Input:} A large forest is cut down for lumber. What is the MOST DIRECT consequence for local wildlife?\\
A) Global carbon dioxide levels increase significantly\\
B) Animals lose their habitat and food sources\\
C) Soil erosion increases over many decades\\
D) Nearby rivers become warmer due to less shade\\
\textit{Output:} Question type: CAUSAL/CONSEQUENCE --- asks for the most direct consequence of forest removal for wildlife.\\
A) CO$_2$ increase --- a real effect of deforestation, but it operates at a global scale and is an atmospheric consequence, not a direct consequence FOR local wildlife.\\
B) Animals lose habitat and food sources --- forest removal directly eliminates the shelter and food that wildlife depend on. This is the immediate, first-order effect on wildlife.\\
C) Soil erosion --- a real long-term consequence, but it operates over decades and affects the land, not wildlife directly.\\
D) Warmer rivers --- a secondary physical effect; it may eventually affect aquatic life, but it is not the most direct consequence of forest removal for wildlife generally.\\
Challenge (top 2: A vs B): A describes a global atmospheric consequence. B describes the direct, immediate effect on the animals themselves. The question asks for the most direct consequence FOR local wildlife --- B is the precise match.\\
Causal directness check: Forest removed $\to$ habitat and food gone $\to$ wildlife directly affected. B is the first-order link in the chain.

\texttt{<Answer>}B\texttt{</Answer>}

\vspace{0.5em}
\textbf{Example 2 (Mechanism):}\\
\textit{Input:} Frost forms on a cold window on a winter morning. Which process best describes how the frost formed?\\
A) Condensation, as water vapor cools below the dew point\\
B) Freezing, as liquid water droplets on the glass solidify overnight\\
C) Deposition, as water vapor converts directly to ice without passing through a liquid phase\\
D) Sublimation, as ice on nearby surfaces vaporizes and reattaches to the glass\\
\textit{Output:} Question type: MECHANISM --- asks which phase-change process produces frost.\\
A) Condensation --- gas to liquid phase change (produces dew, not frost). Only applies above 0$^\circ$C.\\
B) Freezing --- liquid to solid. Requires liquid water already present; does not describe frost forming from air.\\
C) Deposition --- direct gas to solid phase change (water vapor $\to$ ice), bypassing liquid entirely. This is exactly how frost forms on surfaces below 0$^\circ$C.\\
D) Sublimation --- solid to gas (the reverse process). Describes ice disappearing, not forming.\\
Challenge (top 2: A vs C): Both involve water vapor, but condensation produces liquid (dew, above freezing) while deposition produces solid ice (frost, below freezing). The window is below 0$^\circ$C --- deposition is the exact mechanism.\\
Exact match check: The question asks which process describes how frost forms $\to$ deposition directly answers this.

\texttt{<Answer>}C\texttt{</Answer>}

\vspace{0.5em}
\textbf{Example 3 (Appearance):}\\
\textit{Input:} Venus appears brighter than any other planet in the night sky as seen from Earth. What is the primary reason Venus appears so bright?\\
A) Venus reflects a large amount of sunlight due to its thick cloud cover\\
B) Venus is the planet closest to the Sun\\
C) Venus has a very large diameter compared to other planets\\
D) Venus emits its own light from volcanic activity\\
\textit{Output:} Question type: MECHANISM/APPEARANCE --- asks why Venus APPEARS bright from Earth, not what its intrinsic physical property is.\\
A) High reflectivity from thick clouds --- Venus has an albedo of $\sim$0.65--0.70, reflecting most incoming sunlight. High reflectivity directly causes exceptional apparent brightness as seen from Earth.\\
B) Closest to the Sun --- Mercury is even closer to the Sun. More importantly, orbital proximity to the Sun does not determine how bright a planet appears from Earth; reflectivity and distance to Earth do.\\
C) Large diameter --- Jupiter and Saturn are far larger than Venus but appear dimmer. Size alone does not determine apparent brightness.\\
D) Emits own light --- planets reflect sunlight; they do not generate visible light. Volcanic emission is negligible.\\
Challenge (top 2: A vs B): B answers ``which planet is near the Sun'' --- a related but different question. A directly answers why Venus APPEARS bright from Earth through the reflectivity mechanism.\\
Exact match check: Appearance $\to$ reflectivity mechanism $\to$ A.

\texttt{<Answer>}A\texttt{</Answer>}

\vspace{0.5em}
\textbf{Example 4 (Structural):}\\
\textit{Input:} Which layer of the Earth moves on top of the asthenosphere and is responsible for tectonic plate movement?\\
A) Inner core \quad B) Outer core \quad C) Lithosphere \quad D) Mesosphere\\
\textit{Output:} Question type: STRUCTURAL/LAYER --- asks which layer sits ON TOP OF (above) the asthenosphere.\\
A) Inner core --- Earth's deepest solid layer; does not interact with the asthenosphere.\\
B) Outer core --- liquid iron-nickel layer deep inside Earth; does not overlie the asthenosphere.\\
C) Lithosphere --- the rigid outer shell (crust + uppermost mantle) that sits directly on top of the semi-plastic asthenosphere and is divided into tectonic plates.\\
D) Mesosphere (lower mantle) --- lies BELOW the asthenosphere, not above it.\\
Challenge (top 2: C vs D): The asthenosphere is the viscous middle layer. Lithosphere is above it; mesosphere is below. The question asks for the layer moving ON TOP --- that is the lithosphere.\\
Referent check: ``moves on top of the asthenosphere'' $\to$ above $\to$ lithosphere = C.

\texttt{<Answer>}C\texttt{</Answer>}

\texttt{</examples>}

\vspace{0.5em}
\textit{[Immutable suffix appended to every prompt]}: After any reasoning, give your final answer inside \texttt{<Answer></Answer>} tags containing ONLY the letter A, B, C, or D.
\end{tcolorbox}

\section{Related Work}
\label{sec:rel_work}
\begin{table}[t]
\centering
\caption{\textbf{Every prompt optimizer we discuss or compare against, on the
axes of our framework} (\cref{def:bips}), grouped as in \cref{sec:rw-promptopt}.
\cmark: the method provides the property; \xmark: it does not; \pmark: partial,
or asserted but not established. Column definitions and per-method notes follow
the table (\cref{sec:rw-table-notes}).}
\label{tab:related-work-comparison}
\begin{adjustbox}{max width=\textwidth}
\begin{tabular}{@{}lccccccccc@{}}
\toprule
 & {\bf Structure} & {\bf Iterative} & {\bf DualGrad} & {\bf SecLocal} & {\bf Elitism} & {\bf Diversity} & {\bf Immutable} & {\bf Priced} & {\bf Theory} \\
\midrule
\multicolumn{10}{@{}l}{\emph{Generate-and-select}} \\
APE \citep{zhou2022large}                & \xmark & \xmark & \xmark & \xmark & \xmark & \xmark & \xmark & \xmark & \xmark \\
Instruction induction \citep{honovich2023instruction} & \xmark & \xmark & \xmark & \xmark & \xmark & \xmark & \xmark & \xmark & \xmark \\
AutoPrompt \citep{shin2020autoprompt} & \xmark & \cmark & \xmark & \xmark & \xmark & \xmark & \xmark & \xmark & \xmark \\
RLPrompt \citep{deng2022rlprompt}  & \xmark & \cmark & \xmark & \xmark & \xmark & \xmark & \xmark & \xmark & \xmark \\
GrIPS \citep{prasad2023grips}      & \xmark & \cmark & \xmark & \xmark & \xmark & \xmark & \xmark & \xmark & \xmark \\
\addlinespace
\multicolumn{10}{@{}l}{\emph{Iterative with textual feedback}} \\
ProTeGi \citep{pryzant2023automatic}     & \xmark & \cmark & \xmark & \xmark & \xmark & \xmark & \xmark & \xmark & \pmark \\
OPRO \citep{yang2024large}               & \xmark & \cmark & \xmark & \xmark & \cmark & \xmark & \xmark & \xmark & \xmark \\
MAPO \citep{cui2024introducing}          & \xmark & \cmark & \cmark & \xmark & \xmark & \xmark & \xmark & \xmark & \xmark \\
PO2G \citep{lee2025po2g}                 & \xmark & \cmark & \cmark & \xmark & \xmark & \xmark & \xmark & \xmark & \xmark \\
TRAS \citep{davari2025stabilizing} & \xmark & \cmark & \cmark & \xmark & \xmark & \xmark & \xmark & \xmark & \xmark \\
StraGo \citep{wu2024strago}     & \xmark & \cmark & \cmark & \xmark & \pmark & \xmark & \xmark & \xmark & \xmark \\
TextGrad \citep{yuksekgonul2024textgrad} & \xmark & \cmark & \pmark & \xmark & \xmark & \xmark & \pmark & \xmark & \xmark \\
LLM-AutoDiff \citep{yin2025llm} & \cmark & \cmark & \pmark & \xmark & \xmark & \xmark & \pmark & \xmark & \xmark \\
GRAD-SUM \citep{austin2024gradsum} & \xmark & \cmark & \pmark & \xmark & \xmark & \xmark & \xmark & \xmark & \xmark \\
Trace \citep{cheng2024trace}             & \xmark & \cmark & \xmark & \xmark & \xmark & \xmark & \xmark & \xmark & \xmark \\
Evoke \citep{hu2023evoke}                & \xmark & \cmark & \xmark & \xmark & \xmark & \xmark & \xmark & \xmark & \xmark \\
ContraPrompt \citep{rishav2026contraprompt} & \xmark & \cmark & \pmark & \xmark & \xmark & \xmark & \xmark & \xmark & \xmark \\
MAPGD \citep{han2025mapgd}      & \xmark & \cmark & \xmark & \xmark & \xmark & \xmark & \xmark & \xmark & \pmark \\
PromptAgent \citep{wang2024promptagent} & \xmark & \cmark & \xmark & \xmark & \cmark & \xmark & \xmark & \xmark & \xmark \\
MPO \citep{sharma2026modular}            & \cmark & \cmark & \xmark & \cmark & \xmark & \xmark & \xmark & \xmark & \xmark \\
SAPO \citep{kulin2026sapo}      & \cmark & \cmark & \pmark & \cmark & \cmark & \xmark & \xmark & \xmark & \xmark \\
UniPrompt \citep{juneja2025task} & \cmark & \cmark & \xmark & \cmark & \xmark & \xmark & \xmark & \xmark & \xmark \\
\addlinespace
\multicolumn{10}{@{}l}{\emph{Compile-time / module-level}} \\
DSPy \citep{khattab2023dspy}             & \cmark & \cmark & \xmark & \xmark & \xmark & \xmark & \xmark & \xmark & \xmark \\
MIPROv2 \citep{opsahl2024optimizing}     & \cmark & \cmark & \xmark & \xmark & \xmark & \xmark & \xmark & \xmark & \xmark \\
\addlinespace
\multicolumn{10}{@{}l}{\emph{Population-based}} \\
EvoPrompt \citep{guo2023connecting}      & \xmark & \cmark & \xmark & \xmark & \cmark & \cmark & \xmark & \xmark & \xmark \\
Promptbreeder \citep{fernando2024promptbreeder} & \xmark & \cmark & \xmark & \xmark & \cmark & \cmark & \xmark & \xmark & \xmark \\
GAAPO \citep{dufour2025gaapo}            & \xmark & \cmark & \xmark & \xmark & \cmark & \cmark & \xmark & \xmark & \xmark \\
GEPA \citep{agrawal2026gepa}             & \cmark & \cmark & \xmark & \xmark & \cmark & \cmark & \xmark & \xmark & \xmark \\
\quad + Merge \citep{agrawal2026gepa}    & \cmark & \cmark & \xmark & \xmark & \cmark & \cmark & \xmark & \xmark & \xmark \\
\addlinespace
\multicolumn{10}{@{}l}{\emph{Preference- and cost-aware}} \\
BPO \citep{cheng2024bpo}        & \xmark & \xmark & \xmark & \xmark & \xmark & \xmark & \xmark & \xmark & \xmark \\
CAPO\textsubscript{cost} \citep{zehle2025capo} & \xmark & \cmark & \xmark & \xmark & \cmark & \xmark & \xmark & \cmark & \pmark \\
CAPO\textsubscript{constr} \citep{dong2026capo} & \xmark & \cmark & \xmark & \xmark & \cmark & \xmark & \xmark & \cmark & \pmark \\
\midrule
\textbf{\textsc{Splice} solver (this paper)}    & \cmark & \cmark & \cmark & \cmark & \cmark & \cmark & \cmark & \cmark & \cmark \\
\bottomrule
\end{tabular}
\end{adjustbox}
\end{table}

We review NLO approaches, from prompt
optimization, its most developed form, to the self-improvement loops and
test-time search that share its structure (\cref{sec:rw-promptopt}), and then
the optimization results our analysis builds on (\cref{sec:rw-foundations}). \Cref{tab:related-work-comparison} locates each
NLO method along the axes of \cref{def:bips}.

\subsection{Natural Language Optimization}
\label{sec:rw-promptopt}

\paragraph{Discrete search.}
The text that steers a model matters: prompt quality drives few-shot performance \citep{brown2020language}, the right
directive elicits stepwise reasoning \citep{wei2022chain}, and accuracy is
sensitive even to the order of prompt parts \citep{lu2022fantastically}; we
study black-box optimization of discrete text \citep{liu2023pre}, leaving
aside soft prompts that require gradient access
\citep{lester2021power,li2021prefix}. Early methods searched tokens or phrases
by gradients \citep{shin2020autoprompt}, a learned policy
\citep{deng2022rlprompt} or edit operators \citep{prasad2023grips}, with
feedback that does not describe what failed. APE \citep{zhou2022large} and
instruction induction \citep{honovich2023instruction} let an LLM propose whole
instructions and keep the best, a single step with a greedy selector in the
terms of \cref{def:bips}.

\paragraph{Textual feedback.}
ProTeGi \citep{pryzant2023automatic} introduced the \emph{textual gradient}, a
natural-language account of minibatch errors that drives a beam over edits; its
beam keeps only new children, so a strong parent can be lost. Successors enrich
the signal: score histories \citep{yang2024large}, positive gradients and
momentum \citep{cui2024introducing,lee2025po2g}, feedback from correct
predictions to curb drift \citep{davari2025stabilizing,wu2024strago}, contrasted
reasoning traces \citep{rishav2026contraprompt}, author--reviewer loops
\citep{hu2023evoke}, aggregated feedback \citep{austin2024gradsum}, and tree
search over edit sequences \citep{wang2024promptagent}. StraGo
\citep{wu2024strago} names the failure we target, a revision that fixes current
errors but breaks passing examples, and answers it with better guidance rather
than a selection rule. Feedback and search have also been applied to whole
programs: DSPy \citep{khattab2023dspy} and MIPROv2
\citep{opsahl2024optimizing} compile per-module prompts by searching over
instruction and demonstration candidates, while TextGrad
\citep{yuksekgonul2024textgrad}, LLM-AutoDiff \citep{yin2025llm} (implemented in the
AdalFlow library) and
Trace \citep{cheng2024trace} back-propagate textual feedback through computation
graphs.
These systems optimize every prompt in a program; they structure the pipeline, whereas we structure the
interior of one artifact, so the two compose. These frameworks already hold some
text fixed. An AdalFlow prompt template is routinely assembled from several
parameters (for example a task description and few-shot demonstrations), each
marked as optimized or not; a TextGrad prompt can be split into several
variables with gradients disabled on some; and DSPy, MIPROv2 and GEPA optimize
instructions and demonstrations while the signature's field and format
scaffold stays fixed. Each amounts to section-level freezing set by the
framework or the user. None poses it as a constraint of the optimization
problem, pairs it with section-local critique, or bounds what it costs. Any of these feedback operators could serve as our proposer $q_\phi$.

\paragraph{Evolutionary and reflective search.}
EvoPrompt \citep{guo2023connecting}, Promptbreeder
\citep{fernando2024promptbreeder} and GAAPO \citep{dufour2025gaapo}
evolve prompt populations with LLM operators. GEPA
\citep{agrawal2026gepa}, the strongest published baseline on four of our seven
benchmarks (counting GEPA+Merge on PUPA), mutates by reflecting on execution traces and draws parents from a
Pareto front over per-instance scores. That front is the closest prior mechanism
to our diversity slots, but it is defined over task instances rather than
parentage, so children of one parent can fill it. (On SST-5 under the
protocol we reproduce, the published GEPA number, $42.0$, is below that
paper's unoptimized baseline, $44.6$; we report it as published.)

\paragraph{Self-improvement loops.}
The idea of a system that rewrites its own components dates to the
G\"odel machine \citep{schmidhuber2003godel}, which accepts a self-modification
only when it is provably beneficial. LLMs have made empirical versions of this
loop practical. STOP \citep{zelikman2024stop} lets a scaffolding program improve
itself, and G\"odel Agent \citep{yin2025godel} lets an agent modify its own
logic. ADAS \citep{hu2024adas} has a meta-agent write new agents in code and
keeps an archive of discoveries, the Darwin G\"odel Machine
\citep{zhang2025darwin} grows an archive of coding agents that edit their own
source, and AFlow \citep{zhang2024aflow} searches agentic workflows with Monte
Carlo tree search. FunSearch \citep{romeraparedes2024funsearch} and AlphaEvolve
\citep{novikov2025alphaevolve} pair an LLM proposer with an automatic evaluator
to evolve programs. Promptbreeder \citep{fernando2024promptbreeder} is the
self-referential case within NLO, evolving the prompts that mutate prompts.
These loops share the propose--score--retain skeleton of \cref{def:bips}, and
most depart from it in the ways \cref{sec:framework} names for GEPA and OPRO:
proposals conditioned on an archive rather than one parent, and archives that
grow without bound. Program search also anticipates selective mutability.
FunSearch evolves a single function inside a fixed program skeleton, and
AlphaEvolve evolves only the code blocks a user marks, leaving the rest fixed.
What NLO has lacked is that design posed as part of the optimization problem
at the granularity of a section inside a prompt, together with an account of
what freezing costs; self-modifying agent
loops, where the modified system is also the modifier, otherwise enforce such
limits outside the search.

\paragraph{Test-time search.}
A parallel line spends compute at inference time on each query rather than once
on the artifact. Self-consistency \citep{wang2023selfconsistency} and repeated
sampling \citep{brown2024monkeys} aggregate or verify many samples, Tree of
Thoughts \citep{yao2023tree} searches over intermediate reasoning, and
self-critique refines a single response
\citep{madaan2023selfrefine,shinn2023reflexion}. \citet{snell2024scaling} show
that allocating it adaptively between parallel samples and sequential revisions
can be more effective than scaling model parameters. NLO moves the same search off the query
path: the cost is paid once, and the improved artifact conditions every later
query. The parallel-versus-sequential question reappears in our setting as the
comparison between Best-of-$N$ sampling and iterative search under a matched
proposer (\cref{sec:main.exp}).

\paragraph{Structure and constraints.}
Recent work decomposes a prompt into sections: MPO \citep{sharma2026modular}
applies gradients to named sections, SAPO \citep{kulin2026sapo} edits
role, context, task and format segments separately, and UniPrompt
\citep{juneja2025task} edits task facets and compares greedy and beam variants,
in effect a selector ablation in a structured space. None makes a segment
unreachable or prices length. Cost-aware CAPO \citep{zehle2025capo}
penalizes relative token length in the objective, the same shape as our
$\utility - \reg\cost$, and we add the bias bound of \cref{thm:reg-bias};
BPO \citep{cheng2024bpo} rewrites prompts toward human preferences in a
single pass. Constraint-aware CAPO \citep{dong2026capo}, concurrent
with this work, enforces tool-use, brevity and compliance requirements with a
primal--dual method that \emph{prices} violations, so a contract-breaking prompt
remains feasible. Our mutability tags make it \emph{unreachable} instead, and we
reserve pricing for length, where growth is undesirable but not forbidden. The
nearest prior practice is the parameter- and scaffold-level freezing of the
program optimizers above, and the code-level freezing of program search. What
we add is to make user-designated sections a property of the feasible set,
with section-local critique and a bound on what freezing costs
(\cref{prop:mut-price}).

\subsection{Optimization Foundations}
\label{sec:rw-foundations}

\paragraph{Combinatorial and noisy search.}
Greedy maximization of monotone submodular functions achieves $(1-1/e)$
\citep{nemhauser1978analysis}, extended to approximately submodular objectives
by the submodularity ratio \citep{das2011submodular}
\citep[see also][]{mirzasoleiman2015lazier,krause2014submodular}; we use it for
block-level greedy search and the price of freezing. Choosing among near-tied
candidates on a finite batch is best-arm identification
\citep{even2006action,audibert2010best,kaufmann2016complexity}, with random
search \citep{bergstra2012random} as a baseline. Beam search
\citep{lowerre1976harpy}, simulated annealing \citep{kirkpatrick1983optimization}
and tabu search \citep{glover1989tabu} are selectors over a shared state space
and neighborhood \citep{hoos2004stochastic}, the abstraction our edit graph
adopts. Local search on tours \citep{lin1965computer,lin1973effective,helsgaun2000effective},
neural architecture search
\citep{zoph2017neural,real2019regularized,liu2019darts,elsken2019neural} and
model-based hyperparameter optimization
\citep{bergstra2011algorithms,hutter2011sequential} search structured spaces
under expensive, noisy evaluation, and evolutionary computation supplies
classical diversity mechanisms
\citep{goldberg1987genetic,holland1975adaptation,deb2002fast,lehman2011abandoning};
our diversity slots act on parentage.

\paragraph{Theory of iterative search.}
Runtime analysis of evolutionary algorithms bounds the expected time for
elitist search to reach an optimum \citep{droste2002analysis}, and our
convergence result is in the same spirit, with a proposer-competence condition, whose
average-case proxy we measure, in place of a fixed mutation operator. For prompt optimization,
existing analyses bound which candidate to evaluate or keep under noise
\citep{li2026multiobjective,lu2025fedpob}, the convergence of continuous
relaxations \citep{wang2026fozo,han2025mapgd}, or the choice of evaluation data
\citep{nian2026submodular}. None analyzes the search itself: whether the
retained candidate can regress, whether a population collapses onto one parent's children,
or what a constraint costs. These are properties of a selector on a structured
space, and \cref{sec:theory} treats them.

\paragraph{Notes on \cref{tab:related-work-comparison}.}\label{sec:rw-table-notes}
The rows cover every prompt optimizer we discuss or report a number for;
program-search systems (FunSearch, AlphaEvolve) are omitted since they optimize
code, as are the hand-written prompts in the benchmark tables (task description,
expert and CoT prompts). ``Structure'': structured or modular prompt
representation. ``DualGrad'': both positive and negative textual gradients.
``SecLocal'': gradients or updates at the level of named prompt sections.
``Elitism'': parents can survive a step. ``Diversity'': an explicit mechanism
for preserving population diversity. ``Immutable'': designated sections cannot
be modified by any reachable edit. ``Priced'': prompt length or cost enters the
objective. ``Theory'': formal guarantees about the search; \pmark{} for ProTeGi
marks a bandit selection procedure rather than a result about the search
trajectory. \pmark{} under ``Immutable'' marks a whole-variable freeze
(TextGrad's \texttt{requires\_grad=False}, AdalFlow's
\texttt{requires\_opt=False}) rather than a section inside a prompt. GEPA+Merge
adds a system-aware crossover step to GEPA and coincides with it on every axis
tabulated here. Two unrelated methods are both named CAPO:
CAPO\textsubscript{cost} \citep{zehle2025capo} is cost-aware and prunes
candidates under a token budget, while CAPO\textsubscript{constr}
\citep{dong2026capo} is concurrent constraint-aware work. Both earn ``Priced'',
and both retain candidates by truncation over parents and offspring, so both earn
``Elitism''; neither can render a section unreachable, since
CAPO\textsubscript{constr} enforces its constraints as Lagrangian penalties. SAPO
instructs its generator to preserve segments judged strong, which we do not count
as ``Immutable'' because the protection is a request to the proposer rather than
a property of the space (\cref{sec:problem.mutability}).

\section{Framework Details}
\label{sec:framework}\label{sec:app.problem-nlo}
This section supplements the framework of \cref{sec:main.method} with the two
statements the main text cites but does not state, the edit-graph diameter
(\cref{prop:edit-graph-diameter}) and the well-definedness of proposals
(\cref{asm:proposal-distribution}), and with how existing methods sit in the
batched search template as selectors (\cref{sec:algorithms-as-selectors}).
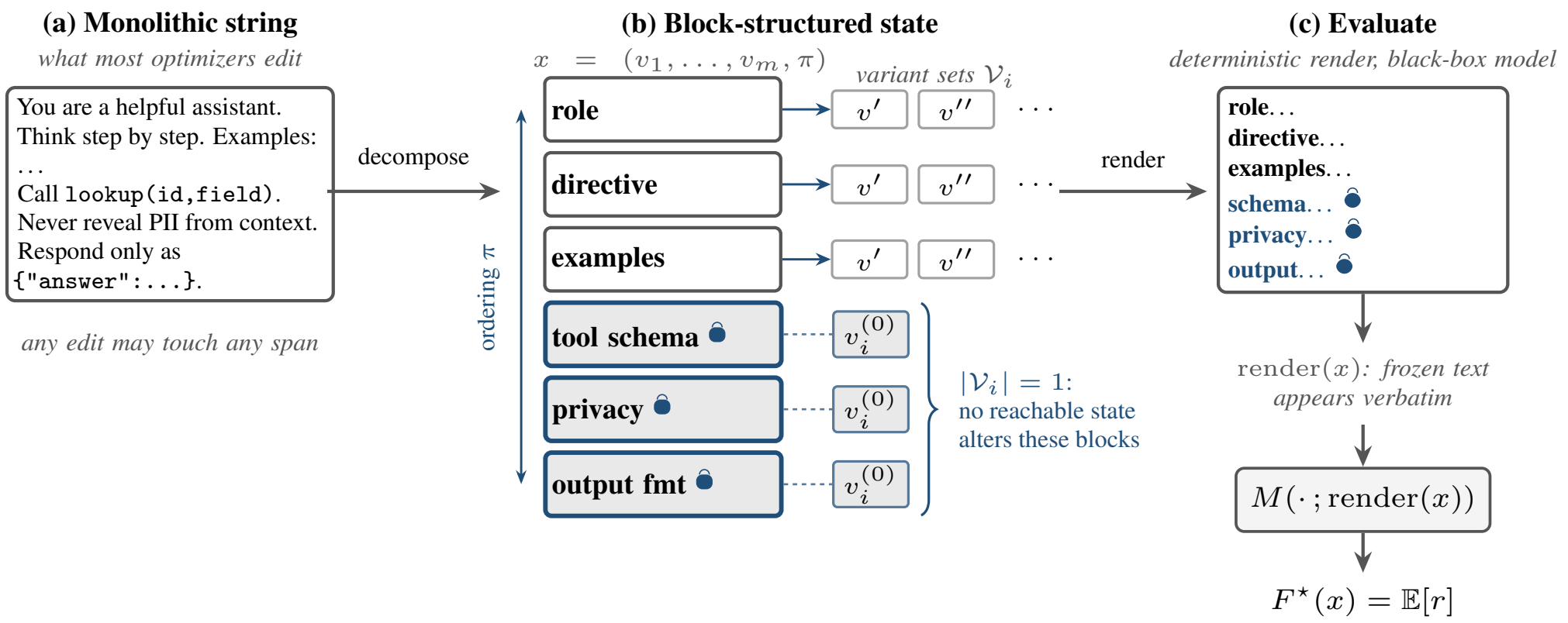
\begin{figure}[t]
\centering
\renewcommand{\lockglyph}{\begin{tikzpicture}[baseline=-0.4ex]\splockpath{0mm}{0mm}{1.4}\end{tikzpicture}}%
\begin{tikzpicture}[
  font=\small,
  x=0.985cm, y=1cm,
  blk/.style={
    draw=splicerule, line width=0.8pt, rounded corners=1.5pt,
    minimum height=5.4mm, text width=19mm, align=left,
    inner xsep=1.6pt, font=\scriptsize, anchor=west},
  mutblk/.style={blk, fill=white},
  frzblk/.style={blk, fill=splicefrozen, draw=spliceaccent, line width=1.1pt},
  varbox/.style={
    draw=splicerule!55, line width=0.6pt, rounded corners=1pt,
    fill=white, minimum height=3.2mm, minimum width=6.4mm,
    align=center, font=\tiny, inner sep=0.8pt, anchor=west},
  frzvar/.style={varbox, fill=splicefrozen, draw=spliceaccent!75},
  panellbl/.style={font=\bfseries\scriptsize, align=center},
  note/.style={font=\tiny\itshape, text=splicerule, align=center},
  anote/.style={font=\tiny, text=spliceaccent, align=left},
  editarr/.style={-{Stealth[length=1.3mm,width=1.1mm]},
    draw=spliceaccent, line width=0.55pt},
  flowarr/.style={-{Stealth[length=1.8mm,width=1.5mm]},
    draw=splicerule, line width=0.8pt},
]

\node[panellbl] at (1.32, 0) {(a) Monolithic string};
\node[note, text width=30mm] at (1.32, -0.30) {what most optimizers edit};

\node[draw=splicerule, line width=0.8pt, rounded corners=2pt, fill=white,
      text width=26mm, align=flush left, font=\tiny, inner sep=2.6pt,
      anchor=north] at (1.32, -0.52) (str)
  {You are a helpful assistant.\\
   Think step by step. Examples: \dots\\
   Call \texttt{lookup(id,field)}.\\
   Never reveal PII from context.\\
   Respond only as \texttt{\{"answer":\dots\}}.};

\node[note, text width=31mm, anchor=north, below=1.4mm of str]
  {any edit may touch any span};

\node[panellbl] at (6.6, 0) {(b) Block-structured state};
\node[note, text width=40mm] at (5.72, -0.30)
  {$x=(v_1,\dots,v_m,\pi)$};

\node[mutblk] at (4.55, -0.72) (b1) {\textbf{role}};
\node[mutblk] at (4.55, -1.36) (b2) {\textbf{directive}};
\node[mutblk] at (4.55, -2.00) (b3) {\textbf{examples}};
\node[frzblk] at (4.55, -2.64) (b4) {\textbf{tool schema}~\lockglyph};
\node[frzblk] at (4.55, -3.28) (b5) {\textbf{privacy}~\lockglyph};
\node[frzblk] at (4.55, -3.92) (b6) {\textbf{output fmt}~\lockglyph};

\foreach \b in {b1,b2,b3}{
  \node[varbox] at ($(\b.east)+(0.42,0)$)      (\b x) {$v'$};
  \node[varbox] at ($(\b x.east)+(0.08,0)$)    (\b y) {$v''$};
  \node[font=\tiny, anchor=west] at ($(\b y.east)+(0.06,0)$) (\b z) {$\cdots$};
  \draw[editarr] (\b.east) -- (\b x.west);
}
\node[note, text width=24mm, text=splicerule] at (7.95, -0.44)
  {variant sets $\mathcal{V}_i$};

\foreach \b in {b4,b5,b6}{
  \node[frzvar] at ($(\b.east)+(0.42,0)$) (\b x) {$v^{(0)}_i$};
  \draw[draw=spliceaccent!75, line width=0.5pt,
        dash pattern=on 1.1pt off 1.1pt] (\b.east) -- (\b x.west);
}

\draw[editarr, {Stealth[length=1.2mm]}-{Stealth[length=1.2mm]}]
  (4.36,-0.72) -- (4.36,-3.92);
\node[anote, rotate=90, anchor=south, text width=0mm] at (4.28,-2.32) {};
\node[font=\tiny, text=spliceaccent, rotate=90, anchor=south]
  at (4.30,-2.32) {ordering $\pi$};

\draw[decorate, decoration={brace, amplitude=1.3mm},
      draw=spliceaccent, line width=0.6pt]
  ($(b4x.north east)+(0.10,0.06)$) -- ($(b6x.south east)+(0.10,-0.06)$);
\node[anote, text width=25mm, anchor=west, align=flush left]
  at ($(b5x.east)+(0.30,0)$)
  {$|\mathcal{V}_i| = 1$:\\ no reachable state\\ alters these blocks};

\node[panellbl] at (11.65, 0) {(c) Evaluate};
\node[note, text width=38mm] at (11.65, -0.30) {deterministic render, black-box model};

\node[draw=splicerule, line width=0.8pt, rounded corners=2pt, fill=white,
      text width=23mm, align=left, font=\tiny, inner sep=2.4pt,
      anchor=north] at (11.65,-0.52) (rendered)
  {\textbf{role}\dots\\[0.5pt]
   \textbf{directive}\dots\\[0.5pt]
   \textbf{examples}\dots\\[0.5pt]
   \textcolor{spliceaccent}{\textbf{schema}\dots~\lockglyph}\\[0.5pt]
   \textcolor{spliceaccent}{\textbf{privacy}\dots~\lockglyph}\\[0.5pt]
   \textcolor{spliceaccent}{\textbf{output}\dots~\lockglyph}};

\node[note, text width=34mm, below=4.2mm of rendered] (renderlbl)
  {$\render(x)$: frozen text\\ appears verbatim};

\node[draw=splicerule, line width=0.7pt, rounded corners=2pt,
      fill=splicefrozen!50, minimum width=20mm, minimum height=5.4mm,
      font=\scriptsize, below=3.6mm of renderlbl] (model)
  {$\model(\cdot\,;\render(x))$};

\node[font=\scriptsize, below=3.4mm of model] (rew)
  {$\utility^{\star}(x)=\E\!\left[\reward\right]$};

\draw[flowarr] (rendered.south) -- (renderlbl.north);
\draw[flowarr] (renderlbl.south) -- (model.north);
\draw[flowarr] (model.south) -- (rew.north);

\draw[flowarr] (2.68,-1.42) -- (4.18,-1.42);
\node[font=\tiny, align=center, anchor=south] at (3.43,-1.36)
  {decompose};

\draw[flowarr] (9.02,-1.42) -- (10.30,-1.42);
\node[font=\tiny, align=center, anchor=south] at (9.66,-1.34)
  {render};

\end{tikzpicture}

\caption{\textbf{A prompt as a block-structured state.}
\textbf{(a)} Most prior optimizers edit a deployed prompt as one
string, so any proposed rewrite may touch any span, including
the tool schema that downstream parsers consume and the privacy clause
that survived review. \textbf{(b)} We represent the same prompt as a
state $x=(v_1,\dots,v_m,\pi)$: one variant per semantically coherent
block, plus an admissible ordering (\cref{def:prompt-state}). Mutable
blocks (white) range over variant sets $\mathcal{V}_i$ that the
proposer explores. Immutable blocks (shaded, locked) satisfy
$|\mathcal{V}_i|=1$, so no state in the search space differs from the
seed on those coordinates (\cref{def:mutability}); the space itself preserves
the contract \emph{text}; whether the model still obeys it is measured in
\cref{sec:exp.contract-main}. \textbf{(c)} A deterministic render maps the state to a
string, which the black-box model consumes to produce reward
(\cref{def:render,def:population-utility}). Because $\render$ is
deterministic, frozen blocks appear in the rendered prompt verbatim.
Most prior work optimizes the object in (a); the constrained problem
of \cref{def:main-objective} is posable only over the object in (b).}
\label{fig:state-space}
\end{figure}

\begin{figure}[t]
\centering
\begin{tikzpicture}[x=1cm, y=1cm, font=\footnotesize,
  hcard/.style={draw=spliceaccent!60, fill=spliceaccent!4, line width=1.0pt,
    rounded corners=4pt},
  hnum/.style={circle, fill=spliceaccent, text=white, inner sep=0pt,
    minimum size=3.4mm, font=\scriptsize\bfseries},
  httl/.style={anchor=west, inner sep=0pt, font=\small\bfseries,
    text=spliceaccent},
  hsub/.style={anchor=west, inner sep=0pt, font=\scriptsize\itshape,
    text=splicerule},
  stage/.style={draw=spliceaccent, fill=white, line width=1.2pt,
    rounded corners=2.5pt},
  sttl/.style={font=\footnotesize\bfseries\scshape, text=spliceaccent,
    inner sep=0pt},
  ssub/.style={font=\scriptsize, inner sep=0pt},
  chev/.style={-{Stealth[length=1.8mm,width=2.2mm]}, line width=1.3pt,
    draw=spliceaccent!45},
  cyc/.style={-{Stealth[length=1.9mm,width=1.7mm]}, line width=0.8pt,
    draw=spliceaccent!80},
  cell/.style={anchor=west, font=\scriptsize, align=flush left,
    text width=26.0mm, inner sep=0pt},
  mname/.style={anchor=west, font=\scriptsize\bfseries, align=flush left,
    text width=17mm, inner sep=0pt},
  nonamed/.style={font=\scriptsize\itshape, text=splicerule!70,
    inner sep=0pt},
  sharedbox/.style={draw=splicerule!70, fill=white, line width=0.9pt,
    rounded corners=2pt},
  selbox/.style={draw=spliceaccent, fill=white, line width=0.9pt,
    rounded corners=1.6pt},
  selsplice/.style={draw=spliceaccent, fill=spliceedit!60, line width=1.4pt,
    rounded corners=1.6pt},
  widened/.style={draw=spliceaccent, fill=spliceedit!70, line width=1.0pt,
    rounded corners=1.6pt},
  rowrule/.style={draw=spliceaccent!18, line width=0.4pt},
]

\node[httl] at (0.15,-0.55) {Template};
\node[anchor=west, inner sep=0pt, font=\scriptsize, text=splicerule]
  at (0.15,-0.86) {\cref{def:bips}};

\foreach \l/\r/\n in {2.05/4.80/1, 5.05/7.80/2, 8.05/10.80/3, 11.05/13.80/4}{%
  \draw[stage] (\l,-0.30) rectangle (\r,-1.10);
}
\foreach \l in {4.80, 7.80, 10.80}{%
  \draw[chev] (\l+0.02,-0.70) -- (\l+0.23,-0.70);}

\node[sttl, anchor=base] at (3.425,-0.66) {\tikz[baseline=-0.62ex]\node[hnum]{1};\ propose $q_\phi$};
\node[ssub] at (3.425,-0.90) {$\mathcal{F}_t=\mathcal{C}_{t-1}\cup\Expand$};
\node[sttl, anchor=base] at (6.425,-0.66) {\tikz[baseline=-0.62ex]\node[hnum]{2};\ score $\Psi_t$};
\node[ssub] at (6.425,-0.90) {$\mathcal{X}\to\mathbb{R}$, on a named split};
\node[sttl, anchor=base] at (9.425,-0.66) {\tikz[baseline=-0.62ex]\node[hnum]{3};\ select $\mathcal{S}_t$};
\node[ssub] at (9.425,-0.90)
  {$\mathcal{C}_t\subseteq\mathcal{F}_t$, $|\mathcal{C}_t|\le\bw$};
\node[sttl, anchor=base] at (12.425,-0.66) {\tikz[baseline=-0.62ex]\node[hnum]{4};\ return $\mathcal{O}$};
\node[ssub] at (12.425,-0.90)
  {$\hat x=\mathcal{O}(\mathcal{C}_{0:T};\Psi_{1:T})$};

\draw[cyc] (9.425,-0.30) -- (9.425,-0.08) -- (3.425,-0.08) -- (3.425,-0.30);
\node[fill=white, inner xsep=3pt, inner ysep=0.5pt, font=\scriptsize,
  text=spliceaccent] at (6.425,-0.08)
  {from $\mathcal{C}_0=\{x^{(0)}\}$, steps $t=1,\dots,T$};

\draw[hcard] (0.00,-1.30) rectangle (13.90,-4.18);
\node[httl] at (0.15,-1.53) {Instances};
\node[hsub] at (1.62,-1.53)
  {matched proposer, so apart from the budget they differ only in $\mathcal{S}_t$
   (\cref{sec:algorithms-as-selectors})};

\fill[spliceaccent!11, rounded corners=2.5pt] (7.98,-1.72) rectangle (10.87,-4.10);

\foreach \l/\r in {2.05/4.80, 5.05/7.80, 11.05/13.80}{%
  \draw[sharedbox] (\l,-1.76) rectangle (\r,-4.06);}
\node[cell] at (2.10,-2.91) {same state space, edit graph and proposal model};
\node[cell] at (5.10,-2.91) {$\Psi_t = \tilde\utility(\cdot\,;\valid)$};
\node[cell] at (11.10,-2.91)
  {final leader, $\argmax_{\mathcal{C}_T}\Psi_T$\\[1.6mm]
   the slot also admits \emph{best retained},
   $\argmax_{\cup_t \mathcal{C}_t}\Psi$};

\foreach \y in {-2.18, -2.60, -3.02, -3.44}{%
  \draw[rowrule] (0.12,\y) -- (1.90,\y);}

\foreach \yt/\yb/\nm/\sel in {%
  -1.76/-2.18/{Best-of-$N$}/{best child, seed dropped},
  -2.18/-2.60/{Greedy}/{$\topk_{1}(\mathcal{F}\setminus\mathcal{C}_{t-1},\Psi)$},
  -2.60/-3.02/{Beam (ProTeGi)}/{$\topk_{\bw}(\mathcal{F}\setminus\mathcal{C}_{t-1},\Psi)$},
  -3.02/-3.44/{Elitist Beam}/{$\topk_{\bw}(\mathcal{F},\Psi)$}}{%
  \draw[selbox] (8.08,\yt-0.03) rectangle (10.77,\yb+0.03);
  \node[mname] at (0.15,{(\yt+\yb)/2}) {\nm};
  \node[cell]  at (8.14,{(\yt+\yb)/2}) {\sel};}

\draw[selsplice] (8.08,-3.46) rectangle (10.77,-4.10);
\node[mname, text=spliceaccent] at (0.15,-3.78) {\splice{}};
\node[cell] at (8.14,-3.78)
  {$\topk_{\bw-D}(\mathcal{F},\Psi)$ $+$ $D$ diversity slots};

\draw[hcard] (0.00,-4.34) rectangle (13.90,-7.30);
\node[httl] at (0.15,-4.57) {Departures};
\node[hsub] at (1.85,-4.57)
  {each embeds into \cref{def:bips} only after widening the tinted slot};
\draw[widened] (11.30,-4.65) rectangle (11.62,-4.49);
\node[anchor=west, inner sep=0pt, font=\scriptsize] at (11.70,-4.57)
  {widened slot};

\foreach \y in {-5.50, -5.94, -6.54}{%
  \draw[rowrule] (0.12,\y) -- (13.78,\y);}

\foreach \l/\yt/\yb in {%
  2.05/-4.78/-5.50, 5.05/-4.78/-5.50, 8.05/-4.78/-5.50,
  2.05/-5.50/-5.94,
  2.05/-5.94/-6.54,
  8.05/-6.54/-7.20}{%
  \draw[widened] (\l,\yt-0.03) rectangle ({\l+2.75},\yb+0.03);}

\node[mname] at (0.15,-5.14) {GEPA\\($+$Merge)};
\node[cell] at (2.10,-5.14) {two parents, outside $\mathcal{N}(x)$};
\node[cell] at (5.10,-5.14) {per-instance Pareto preorder};
\node[cell] at (8.10,-5.14) {unbounded archive};
\node[nonamed] at (12.425,-5.14) {not named};

\node[mname] at (0.15,-5.72) {MIPROv2};
\node[cell] at (2.10,-5.72) {no parent, no local move};
\foreach \x in {6.425, 9.425, 12.425}{\node[nonamed] at (\x,-5.72) {not named};}

\node[mname] at (0.15,-6.24) {OPRO};
\node[cell] at (2.10,-6.24) {meta-prompt over history};
\node[nonamed] at (6.425,-6.24) {not named};
\node[cell] at (8.10,-6.24) {parents can survive a step};
\node[nonamed] at (12.425,-6.24) {not named};

\node[mname] at (0.15,-6.87) {PromptAgent};
\foreach \x in {3.425, 6.425, 12.425}{\node[nonamed] at (\x,-6.87) {not named};}
\node[cell] at (8.10,-6.87) {tree with visit-count bonus};

\end{tikzpicture}

\caption{\textbf{One template, four slots, and what each method plugs in.}
\emph{Top:} batched iterative search (\cref{def:bips}). From
$\mathcal{C}_0=\{x^{(0)}\}$, each step expands the beam into a frontier with a
proposal model $q_\phi$, scores it with a scoring functional $\Psi_t$, keeps
at most $\bw$ candidates with a selector $\mathcal{S}_t$, and after $T$ steps
returns one artifact through an output rule $\mathcal{O}$.
\emph{Middle:} the selectors compared in \cref{sec:main.exp} are instances that
share every slot but the selector.
\emph{Bottom:} four published methods depart from the template, and the tinted
cells are the slots each one widens. ``Not named'' marks a slot the paper does not
specify for that method, not a claim that it matches the template.}
\label{fig:bips-sockets}
\end{figure}
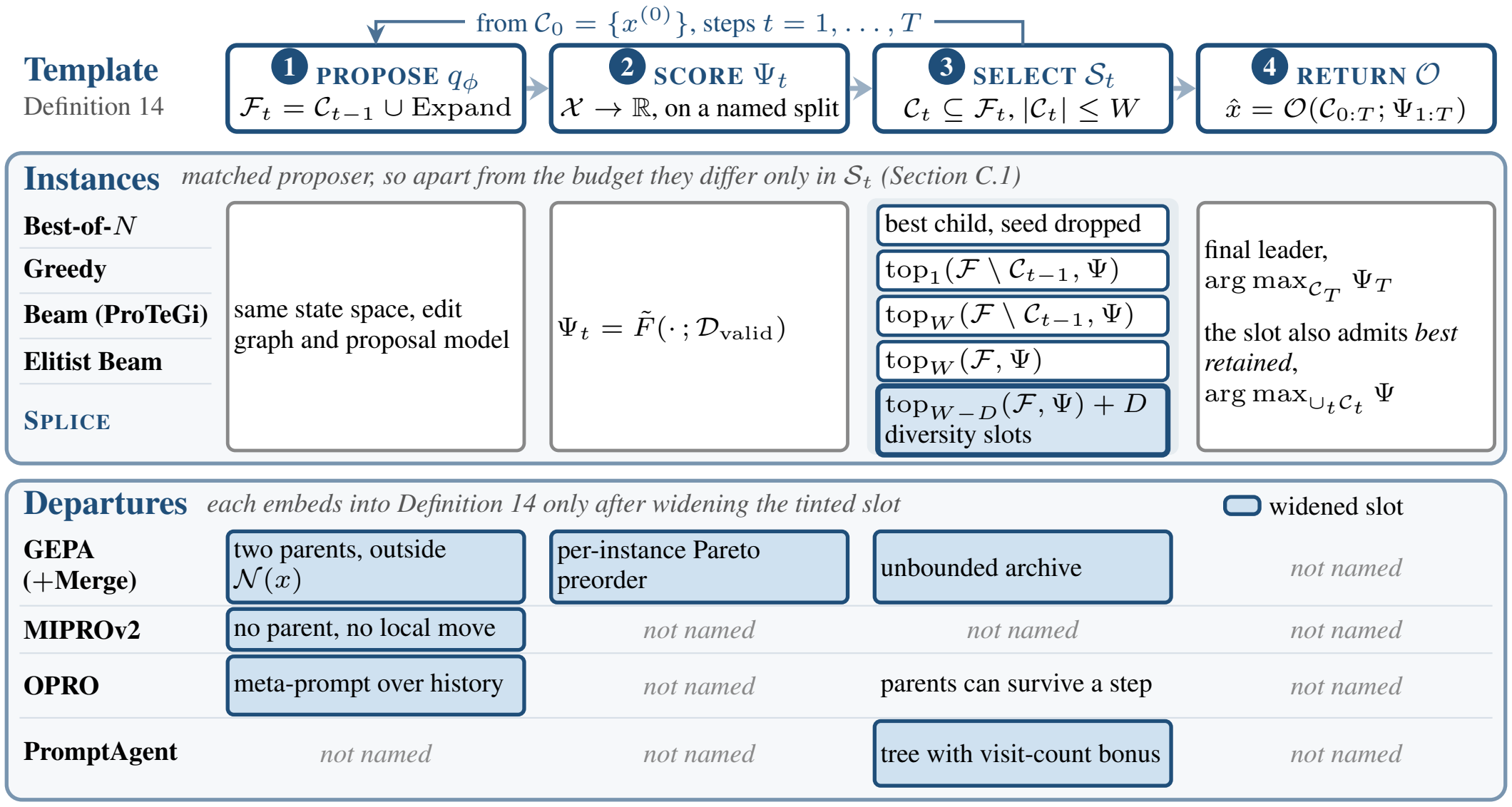
\begin{proposition}[Diameter of the edit graph]
\label{prop:edit-graph-diameter}
Let $\mathrm{diam}(\Pi)$ denote the diameter of the permutation subgraph
induced by the chosen ordering moves. Then
\begin{equation}
\label{eq:edit-graph-diameter}
\mathrm{diam}(\mathcal{G}) \;\le\; \sum_{i=1}^m \1\{|\mathcal{V}_i| \ge 2\}
\;+\; \mathrm{diam}(\Pi).
\end{equation}
\end{proposition}

\begin{assumption}[Well-defined proposal distribution]
\label{asm:proposal-distribution}
For every $x \in \mathcal{X}$ and history $H_t$, $q_\phi(\cdot \mid x, H_t)$
is a probability distribution over edits whose support satisfies
$\Apply(x, \mathrm{supp}\,q_\phi(\cdot \mid x, H_t)) \subseteq \mathcal{N}(x)
\cup \{x\}$: no proposal can leave $\mathcal{X}$ or write to an immutable
block.
\end{assumption}

\subsection{Algorithms as Selectors}
\label{sec:algorithms-as-selectors}

The selectors compared in \cref{sec:main.exp} are instances of
\cref{def:bips} that differ in the retention rule. With
$\Psi_t=\tilde\utility(\cdot\,;\valid)$ and the final-leader output rule:
\emph{Best-of-$N$} is a single step ($T{=}1$, $\bw{=}1$, $k{=}N$) that keeps the
best child and drops the seed; \emph{greedy} is beam search with $\bw{=}1$,
$\topk_{1}(\mathcal{F}\setminus\mathcal{C}_{t-1},\Psi)$: the best child replaces
the parent even when it scores lower; \emph{beam search} in the style of ProTeGi keeps
the top-$\bw$ \emph{children},
$\mathcal{S}^{\mathrm{beam}} := \topk_{\bw}(\mathcal{F}\setminus\mathcal{C}_{t-1},\Psi)$,
dropping parents each step; \emph{elitist beam search} keeps the top-$\bw$ of
the \emph{full} frontier, $\topk_{\bw}(\mathcal{F},\Psi)$; and
the \splice{} solver extends the elitist rule with a diversity reservation read from
candidate annotations (\cref{sec:main.method}).

The published baselines of \cref{sec:main.exp} are not re-run inside the
template; their numbers come from the protocols we reproduce. Several of them
are also not instances of \cref{def:bips}, and it is more useful to name the
difference than to stretch the template over it. GEPA scores candidates by a
Pareto preorder over per-instance results and keeps an unbounded archive;
GEPA+Merge additionally recombines two parents, a move outside
$\mathcal{N}(x)$; MIPROv2 searches instruction--demonstration combinations
without a parent or a local move; OPRO conditions each proposal on a
meta-prompt over the top-scoring history rather than on one parent; and
PromptAgent retains an unbounded tree with a visit-count bonus. OPRO is the
closest case: it fits \cref{def:bips} if the parent is taken as nominal and the
meta-prompt is passed through $H_t$, which \cref{def:proposal} allows. Each method embeds into \cref{def:bips}
only after widening a named component (a vector-valued scoring functional, a
multi-parent expansion, a parent-free proposal, an unbounded pool, a nominal
parent), and we do not
claim that the template's component-by-component comparison transfers to them.

Casting the methods this way exposes the axes along which they differ (beam
size $\bw$, parent survival, diversity, scoring set and output rule), and
\cref{tab:related-work-comparison} records several of them (elitism,
diversity, structure) for each prior method.
The axes internal to the \splice{} solver are ablated in \cref{sec:main.exp};
the external baselines are compared as whole systems.

\section{Solver Details}
\label{sec:main.method-ext}
This section collects the pseudocode and illustrations of the \splice{} loop
described in \cref{sec:main.method}, and the details of proposal generation.
\begin{algorithm}[t]
\SetAlgoLined
\SetKwInOut{Input}{input}
\SetKwInOut{Output}{output}
\Input{initial state $x^{(0)} \in \mathcal{X}$; proposal model
$q_\phi$; critique agent $\model_{\mathrm{crit}}$; beam width $\bw$;
diversity slots $D$; quality-gate ratio $\gateratio \in [0,1]$; proposals per
candidate $k$; horizon $T$; momentum window $w$; minimum improvement
$\Delta_{\min}$; patience $T_{\mathrm{pat}}$; training data $\dtrain$; validation data
$\valid$; regularizers $\{R_j\}$ with weights $\{\lambda_j\}$.}

{\bf Initialize:} $\mathcal{C}_0 \gets \{x^{(0)}\}$; score $\tilde\utility(x^{(0)}) \gets \utility(x^{(0)};\,\valid)$.\par
Gradient history $G_0 \gets \emptyset$.\par

\For{$t = 1, \dots, T$}{
  \tcp{\small \emph{i. Re-run the beam on a training batch (critique evidence only).}}
  Sample $B_t \subseteq \dtrain$; compute $\utility(x;\, B_t)$ and per-example outcomes $\forall x \in \mathcal{C}_{t-1}$.\par

  \tcp{\small \emph{ii. Dual, section-local critique, one per parent.}}
  $\mathrm{Crit}_t(x) \gets \model_{\mathrm{crit}}\bigl(x,\; B_t\text{-outcomes of } x\bigr)\quad \forall x \in \mathcal{C}_{t-1}$\par

  \tcp{\small \emph{iii. Momentum context from the gradient trajectory and cached scores.}}
  $\mathrm{Mom}_t \gets \textsc{BuildMomentum}\bigl(G_{t-1},\, \{\tilde\utility(\mathrm{leader}(\mathcal{C}_s))\}_{s<t},\, \{\tilde\utility(x)\}_{x\in\mathcal{C}_{t-1}};\; w, \Delta_{\min}\bigr)$\par

  \tcp{\small \emph{iv. Propose $k$ candidates per beam member (edit only mutable blocks).}}
  $\mathcal{N}_t \gets \bigcup_{x \in \mathcal{C}_{t-1}} \bigl\{ \Apply(x,u_j) :
    u_j \overset{\text{i.i.d.}}{\sim} q_\phi(\cdot \mid x,\, \tilde\utility(x),\, \mathrm{Crit}_t(x),\, \mathrm{Mom}_t),\,
    j=1,\dots,k \bigr\}$\par

  Remove from $\mathcal{N}_t$ any duplicate of a member of $\mathcal{C}_{t-1}$ or of an earlier element of $\mathcal{N}_t$.\par

  \tcp{\small \emph{v. Evaluate new candidates on validation, once (cached scoring).}}
  \For{$x \in \mathcal{N}_t$}{
    $\tilde{\utility}(x) \gets \utility(x;\,\valid) - \sum_j \lambda_j R_j(x)$\par
  }

  \tcp{\small \emph{vi. Elitist selection with diversity slots (\cref{alg:elit-div}).}}
  $\mathcal{C}_t \gets \mathcal{S}_t^{\mathrm{elit}\text{-}\mathrm{div}}\bigl(
     \mathcal{C}_{t-1} \cup \mathcal{N}_t,\; \tilde{\utility};\;
     \bw, D, \gateratio, t\bigr)$\par

  \tcp{\small \emph{vii. Record the leader change and the critique behind it.}}
  $x^{\mathrm L}_t \gets \mathrm{leader}(\mathcal{C}_t)$\par
  $\Delta_t \gets \tilde{\utility}(x^{\mathrm L}_t) - \tilde{\utility}(\mathrm{leader}(\mathcal{C}_{t-1}))$\par
  $G_t \gets G_{t-1} \cup \{(t,\, \mathbf{g}_t,\, \Delta_t)\}$, where $\mathbf{g}_t$ is the summary of $\mathrm{Crit}_t(\mathrm{parent}(x^{\mathrm L}_t))$, or a no-change marker if $x^{\mathrm L}_t \in \mathcal{C}_{t-1}$\par

  \tcp{\small \emph{viii. Early stopping.}}
  \If{the leader has gained less than $\Delta_{\min}$ over the last $T_{\mathrm{pat}}$ steps}{
    \textbf{break}\par
  }
}
\Return $\hat{x} \gets \mathrm{leader}(\mathcal{C}_t)$ at the last completed step
\caption{The \splice{} solver.}
\label{alg:splice}
\end{algorithm}

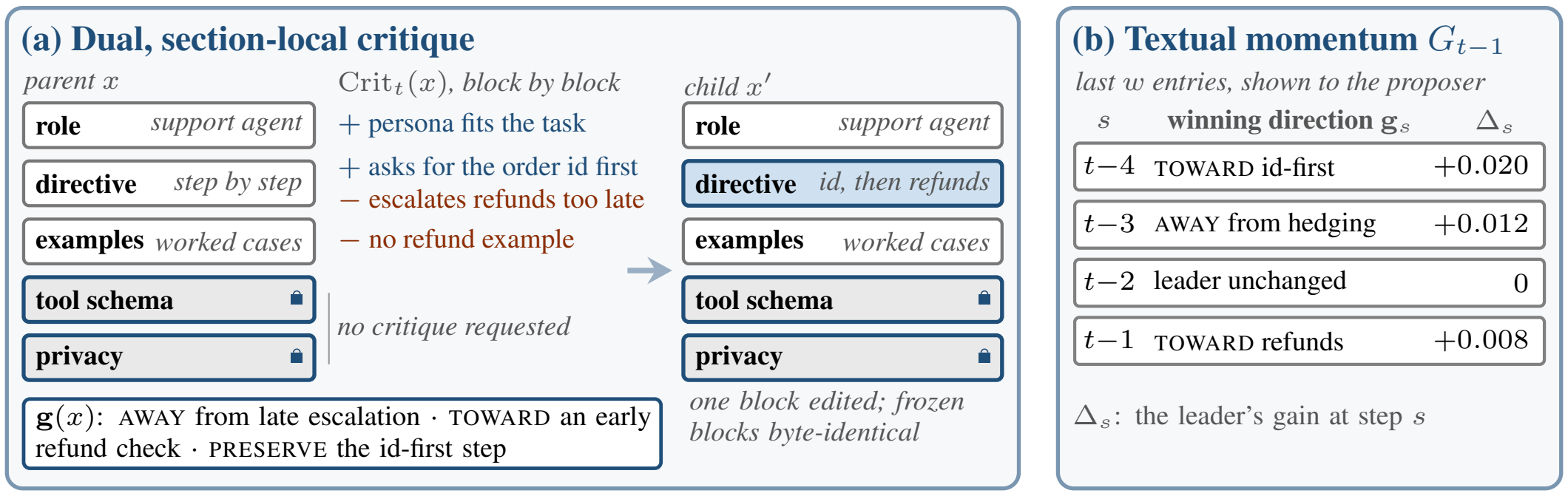
\begin{figure}[t]
\centering
\begin{tikzpicture}[x=1cm, y=1cm, font=\scriptsize,
  hcard/.style={draw=spliceaccent!60, fill=spliceaccent!4, line width=1.0pt,
    rounded corners=4pt},
  httl/.style={anchor=west, inner sep=0pt, font=\small\bfseries,
    text=spliceaccent},
  sub/.style={anchor=west, inner sep=0pt, font=\scriptsize\itshape,
    text=splicerule},
  mb/.style={draw=splicerule!80, fill=white, line width=0.8pt,
    rounded corners=1.4pt, minimum width=24mm, minimum height=4.0mm,
    inner xsep=1.2mm, anchor=west, align=left},
  fb/.style={mb, fill=splicefrozen, draw=spliceaccent, line width=0.6pt},
  mbx/.style={draw=splicerule!80, fill=white, line width=0.8pt, rounded corners=1.4pt},
  fbx/.style={fill=splicefrozen, draw=spliceaccent, line width=1.0pt, rounded corners=1.4pt},
  ebx/.style={fill=spliceedit!70, draw=spliceaccent, line width=1.0pt, rounded corners=1.4pt},
  eb/.style={mb, fill=spliceedit!70, draw=spliceaccent, line width=0.6pt},
  keep/.style={anchor=west, inner sep=0pt, text=spliceaccent},
  fix/.style={anchor=west, inner sep=0pt, text=splicewarn},
  chev/.style={-{Stealth[length=2.2mm,width=2.4mm]}, line width=1.4pt,
    draw=spliceaccent!45},
  tape/.style={draw=splicerule!75, fill=white, line width=0.8pt,
    rounded corners=1.2pt},
]

\draw[hcard] (0.00, 0.00) rectangle ( 9.05,-4.30);
\draw[hcard] (9.40, 0.00) rectangle (13.90,-4.30);

\node[httl] at (0.12,-0.30) {(a) Dual, section-local critique};
\node[sub] at (0.15,-0.68) {parent $x$};
\node[sub] at (2.95,-0.68) {$\mathrm{Crit}_t(x)$, block by block};
\node[sub] at (6.05,-0.68) {child $x'$};

\foreach \y/\b/\t/\s in {-1.05/role/{support agent}/mbx, -1.57/directive/{step by step}/mbx,
                         -2.09/examples/{worked cases}/mbx,
                         -2.61/{tool schema}/{\lockglyph}/fbx, -3.13/privacy/{\lockglyph}/fbx}{%
  \draw[\s] (0.15,\y+0.20) rectangle (2.75,\y-0.20);
  \node[anchor=west, inner sep=0pt, font=\scriptsize\bfseries] at (0.25,\y) {\b};
  \node[anchor=east, inner sep=0pt, font=\scriptsize\itshape, text=splicerule] at (2.65,\y) {\t};}

\node[keep] at (2.95,-1.05) {$+$ persona fits the task};
\node[keep] at (2.95,-1.43) {$+$ asks for the order id first};
\node[fix]  at (2.95,-1.72) {$-$ escalates refunds too late};
\node[fix]  at (2.95,-2.09) {$-$ no refund example};
\node[anchor=west, inner sep=0pt, text=splicerule, font=\scriptsize\itshape]
  at (2.95,-2.87) {no critique requested};
\draw[splicerule!60, line width=0.4pt] (2.88,-2.55) -- (2.88,-3.19);

\draw[draw=spliceaccent, fill=white, line width=0.9pt, rounded corners=1.6pt]
  (0.15,-3.50) rectangle (5.85,-4.12);
\node[anchor=west, inner sep=0pt, text width=55mm, align=left] at (0.25,-3.81)
  {$\mathbf{g}(x)$: \textsc{away} from late escalation $\cdot$ \textsc{toward} an early
   refund check $\cdot$ \textsc{preserve} the id-first step};

\draw[chev] (5.55,-2.35) -- (5.95,-2.35);
\foreach \y/\b/\t/\s in {-1.05/role/{support agent}/mbx, -1.57/directive/{id, then refunds}/ebx,
                         -2.09/examples/{worked cases}/mbx,
                         -2.61/{tool schema}/{\lockglyph}/fbx, -3.13/privacy/{\lockglyph}/fbx}{%
  \draw[\s] (6.05,\y+0.20) rectangle (8.90,\y-0.20);
  \node[anchor=west, inner sep=0pt, font=\scriptsize\bfseries] at (6.15,\y) {\b};
  \node[anchor=east, inner sep=0pt, font=\scriptsize\itshape, text=splicerule] at (8.80,\y) {\t};}
\node[anchor=north west, inner sep=0pt, text width=25mm, align=left,
  text=splicerule, font=\scriptsize\itshape] at (6.10,-3.42)
  {one block edited; frozen blocks byte-identical};

\node[httl] at (9.52,-0.30) {(b) Textual momentum $G_{t-1}$};
\node[sub] at (9.55,-0.68) {last $w$ entries, shown to the proposer};
\node[font=\scriptsize\bfseries, text=splicerule, anchor=west] at (9.62,-1.02) {$s$};
\node[font=\scriptsize\bfseries, text=splicerule, anchor=west] at (10.25,-1.02) {winning direction $\mathbf{g}_s$};
\node[font=\scriptsize\bfseries, text=splicerule, anchor=east] at (13.62,-1.02) {$\Delta_s$};

\foreach \i/\s/\g/\d in {%
  0/{t{-}4}/{\textsc{toward} id-first}/{+0.020},
  1/{t{-}3}/{\textsc{away} from hedging}/{+0.012},
  2/{t{-}2}/{leader unchanged}/{0},
  3/{t{-}1}/{\textsc{toward} refunds}/{+0.008}}{%
  \pgfmathsetmacro{\yy}{-1.42-0.52*\i}
  \draw[tape] (9.55,{\yy+0.21}) rectangle (13.75,{\yy-0.21});
  \node[anchor=west, inner sep=0pt] at (9.62,\yy) {$\s$};
  \node[anchor=west, inner sep=0pt] at (10.25,\yy) {\g};
  \node[anchor=east, inner sep=0pt] at (13.62,\yy) {$\d$};}

\node[anchor=north west, inner sep=0pt, text width=42mm, align=left,
  text=splicerule] at (9.55,-3.55)
  {$\Delta_s$: the leader's gain at step $s$};

\end{tikzpicture}
\caption{\textbf{The critique and the momentum context} (illustrative content, on
the running example of \cref{fig:splice-loop}). \textbf{(a)} The critique agent
reads a parent's traces on the training batch and returns, for each mutable
block, what to keep ($\mathbf{p}_i$) and what to repair ($\mathbf{n}_i$), plus a
directional summary $\mathbf{g}(x)$ \eqref{eq:critique-output}. Frozen blocks receive no
critique. The proposer edits the block the critique points at, here the
directive, and leaves the frozen blocks byte-identical. \textbf{(b)} The
proposer also sees the momentum context, the last $w$ entries of the gradient
trajectory
\eqref{eq:momentum-history}: the direction behind each new leader and the
leader's gain $\Delta_s$, so it can keep following a direction that worked.}
\label{fig:critique}
\end{figure}
\subsection{Proposals}
\label{sec:method.propose}

The proposer is conditioned on the parent, its cached validation score, its
critique, and the momentum context,
\[
\bigl(\render(x),\;\tilde\utility(x),\;\mathrm{Crit}_t(x),\;\mathrm{Mom}_t\bigr),
\]
where $\mathrm{Mom}_t$ summarizes $G_{t-1}$ together with recent leader scores and the
current beam's scores. All of it is fixed before any step-$t$ proposal is
drawn and is a function of the history $H_t$ of \cref{def:proposal},
which is what makes a parent's $k$ proposals i.i.d.\ given $H_t$.
The proposer sees validation \emph{scores} but never a validation example, and
samples $k$ structurally valid edits per parent at non-zero temperature, each
confined to the mutable blocks. A proposal may also reorder the blocks, an
ordering move in $\Pi$. \cref{prop:mut-price} is stated at the seed ordering,
$\Pi = \{\pi^{(0)}\}$, so it prices freezing for search confined to that
ordering and says nothing about runs that reorder.

\cref{sec:theory} gives the probability that a step's draws contain an improving
one (\cref{thm:reachability}) and what is known about terminal quality
(\cref{thm:greedy-approx,rem:matroid}).

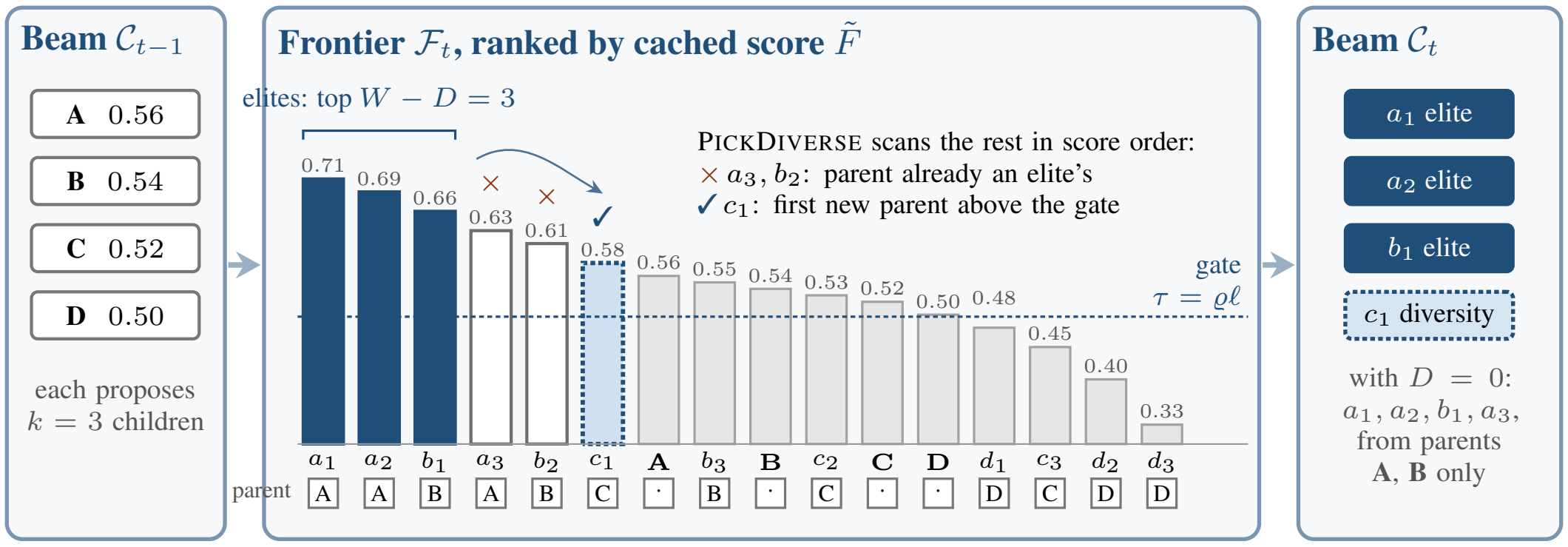
\begin{figure}[t]
\centering
\begin{tikzpicture}[x=1cm, y=1cm, font=\footnotesize,
  hcard/.style={draw=spliceaccent!60, fill=spliceaccent!4, line width=1.0pt,
    rounded corners=4pt},
  httl/.style={anchor=west, inner sep=0pt, font=\small\bfseries,
    text=spliceaccent},
  pchip/.style={draw=splicerule!80, fill=white, line width=0.8pt,
    rounded corners=1.6pt, minimum width=15mm, minimum height=4.2mm,
    inner sep=0pt, font=\scriptsize},
  echip/.style={pchip, fill=spliceaccent, draw=spliceaccent, text=white},
  dchip/.style={pchip, fill=spliceedit!60, draw=spliceaccent, line width=1.0pt,
    dash pattern=on 1.2pt off 0.8pt},
  bele/.style={fill=spliceaccent, draw=spliceaccent, line width=0.8pt},
  bdiv/.style={fill=spliceedit!70, draw=spliceaccent, line width=1.1pt,
    dash pattern=on 1.2pt off 0.8pt},
  bskip/.style={fill=white, draw=splicerule!85, line width=0.8pt},
  brest/.style={fill=splicerule!15, draw=splicerule!55, line width=0.6pt},
  par/.style={draw=splicerule!80, line width=0.6pt, fill=white,
    minimum size=2.6mm, inner sep=0pt, font=\tiny, align=center},
  nm/.style={font=\tiny, inner sep=0pt},
  chev/.style={-{Stealth[length=2.2mm,width=2.4mm]}, line width=1.4pt,
    draw=spliceaccent!45},
]

\draw[hcard] (0.00, 0.00) rectangle ( 1.95,-4.75);
\draw[hcard] (2.30, 0.00) rectangle (11.20,-4.75);
\draw[hcard] (11.55,0.00) rectangle (13.90,-4.75);
\draw[chev] (1.98,-2.30) -- (2.27,-2.30);
\draw[chev] (11.23,-2.30) -- (11.52,-2.30);

\node[httl] at (0.12,-0.30) {Beam $\mathcal{C}_{t-1}$};
\node[httl] at (2.42,-0.30) {Frontier $\mathcal{F}_t$, ranked by cached score $\tilde\utility$};
\node[httl] at (11.67,-0.30) {Beam $\mathcal{C}_t$};

\foreach \y/\n/\s in {-0.95/A/0.56, -1.55/B/0.54, -2.15/C/0.52, -2.75/D/0.50}{%
  \node[pchip] at (0.97,\y) {\textbf{\n}\;\; $\s$};}
\node[font=\scriptsize, text=splicerule, align=center, text width=17mm]
  at (0.97,-3.55) {each proposes\\ $k = 3$ children};

\def\base{-3.90}
\foreach \i/\n/\s/\p/\r in {%
   0/{a_1}/0.71/A/0,  1/{a_2}/0.69/A/0,  2/{b_1}/0.66/B/0,
   3/{a_3}/0.63/A/2,  4/{b_2}/0.61/B/2,  5/{c_1}/0.58/C/1,
   6/{\mathbf{A}}/0.56/{\textperiodcentered}/3,  7/{b_3}/0.55/B/3,
   8/{\mathbf{B}}/0.54/{\textperiodcentered}/3,  9/{c_2}/0.53/C/3,
  10/{\mathbf{C}}/0.52/{\textperiodcentered}/3, 11/{\mathbf{D}}/0.50/{\textperiodcentered}/3,
  12/{d_1}/0.48/D/3, 13/{c_3}/0.45/C/3, 14/{d_2}/0.40/D/3, 15/{d_3}/0.33/D/3}{%
  \pgfmathsetmacro{\xl}{2.65+0.50*\i}
  \pgfmathsetmacro{\xc}{\xl+0.18}
  \pgfmathsetmacro{\yt}{\base+(\s-0.30)/0.45*2.6}
  \ifcase\r\relax
    \draw[bele]  (\xl,\base) rectangle ({\xl+0.36},\yt);
  \or
    \draw[bdiv]  (\xl,\base) rectangle ({\xl+0.36},\yt);
  \or
    \draw[bskip] (\xl,\base) rectangle ({\xl+0.36},\yt);
  \or
    \draw[brest] (\xl,\base) rectangle ({\xl+0.36},\yt);
  \fi
  \node[nm] at (\xc,{\base-0.16}) {$\n$};
  \node[par] at (\xc,{\base-0.44}) {\p};
  \pgfmathsetmacro{\ly}{\yt+0.12+0.15*(\i==12)}
  \node[nm, text=splicerule] at (\xc,\ly) {\scalebox{0.85}{$\s$}};}
\draw[draw=splicerule!60, line width=0.4pt] (2.60,\base) -- (11.10,\base);
\node[nm, text=splicerule, anchor=east] at (2.55,{\base-0.44}) {parent};

\draw[draw=spliceaccent, line width=0.6pt]
  (2.65,-1.18) -- (2.65,-1.10) -- (4.01,-1.10) -- (4.01,-1.18);
\node[font=\scriptsize, text=spliceaccent, anchor=south] at (3.33,-1.06)
  {elites: top $\bw - D = 3$};

\pgfmathsetmacro{\ygate}{\base+(0.497-0.30)/0.45*2.6}
\draw[gateline] (2.60,\ygate) -- (11.10,\ygate);
\node[font=\scriptsize, text=spliceaccent, anchor=south east, inner sep=1.5pt,
  align=right] at (11.10,\ygate) {gate\\ $\tau = \gateratio\ell$};

\pgfmathsetmacro{\ya}{\base+(0.63-0.30)/0.45*2.6+0.42}
\pgfmathsetmacro{\yb}{\base+(0.61-0.30)/0.45*2.6+0.42}
\pgfmathsetmacro{\yc}{\base+(0.58-0.30)/0.45*2.6+0.42}
\node[text=splicewarn, font=\scriptsize\bfseries] at (4.33,\ya) {$\times$};
\node[text=splicewarn, font=\scriptsize\bfseries] at (4.83,\yb) {$\times$};
\node[text=spliceaccent, font=\scriptsize] at (5.33,\yc) {\cmark};
\node[font=\scriptsize, align=left, anchor=west, text width=52mm]
  at (6.05,-1.50)
  {\textsc{PickDiverse} scans the rest in score order:\\
   \textcolor{splicewarn}{$\times$}\,$a_3, b_2$: parent already an elite's\\
   \textcolor{spliceaccent}{\cmark}\,$c_1$: first new parent above the gate};
\draw[-{Stealth[length=1.4mm,width=1.2mm]}, draw=spliceaccent!80,
  line width=0.5pt] (4.20,{\ya+0.22}) to[out=30, in=150] (5.33,{\yc+0.20});

\foreach \y/\n in {-0.95/{a_1}, -1.55/{a_2}, -2.15/{b_1}}{%
  \node[echip] at (12.72,\y) {$\n$ \scriptsize elite};}
\node[dchip] at (12.72,-2.75) {$c_1$ \scriptsize diversity};
\node[font=\scriptsize, text=splicerule, align=center, text width=21mm]
  at (12.72,-3.75)
  {with $D = 0$: $a_1, a_2, b_1, a_3$,\\ from parents \textbf{A}, \textbf{B} only};

\end{tikzpicture}
\caption{\textbf{One selection step, with numbers} ($\bw = 4$, $D = 1$,
$k = 3$, $\gateratio = 0.7$). The four beam members each propose three children, and
parents and children are ranked together on scores cached at first
evaluation (\cref{alg:elit-div}). The top $\bw - D = 3$ fill the elite slots;
all are children of \textbf{A} or \textbf{B}. The leader score is $\ell = 0.71$, so the quality
gate is $\tau = \gateratio\ell \approx 0.50$ (dashed line). Carried beam members
(\textbf{A}--\textbf{D}) have parents from an earlier step, shown as $\cdot$. \textsc{PickDiverse}
(\cref{alg:pickdiverse}) scans the remaining candidates in score order and
skips $a_3$ and $b_2$, whose parents are already elites' parents; $c_1$ is the
first candidate above the gate with a new parent and takes the diversity slot.
This is Tier~3: Tiers~1--2 admit nothing here, because the seed is not in this
frontier and every other candidate descends from it. Had no candidate above $\tau$ had a new
parent, the slot would be backfilled with the best remaining candidate above
$\tau$. With $D = 0$ the next beam would come from two parents only.}
\label{fig:select-example}
\end{figure}
\begin{algorithm}
\SetAlgoLined
\SetKwInOut{Input}{input}
\SetKwInOut{Output}{output}
\Input{frontier $\mathcal{F} = \mathcal{C}_{t-1} \cup \mathcal{N}_t$;
  regularized score $\tilde\utility$; beam width $\bw$; diversity slots $D$;
  quality-gate ratio $\gateratio \in [0,1]$; current step $t$.}
\Output{next beam $\mathcal{C}_t \subseteq \mathcal{F}$, $|\mathcal{C}_t| \le \bw$.}
\tcp{\small \emph{Degenerate cases.}}
\If{$D \le 0$ \textbf{or} $|\mathcal{F}| \le \bw$ \textbf{or} $\mathcal{N}_t = \emptyset$}{
  \Return $\topk_{\bw}(\mathcal{F};\, \tilde\utility)$\par
}
$D \gets \min(D,\, \bw - 1)$ \tcp*{\small guarantee at least one elite slot}
\tcp{\small \emph{Elite slots by score.}}
$\mathcal{E} \gets \topk_{\bw-D}(\mathcal{F};\, \tilde\utility)$\par
$\ell \gets \max_{x \in \mathcal{E}} \tilde\utility(x)$\par
$\tau \gets \gateratio \cdot \ell$ \tcp*{\small quality floor relative to the leader}
$\Lambda \gets \bigcup_{x \in \mathcal{E}} \bigl(\lineage(x) \setminus \{\mathrm{id}(x)\}\bigr)$\par
$P \gets \{\mathrm{parent}(x) : x \in \mathcal{E}\}$\par
$\mathcal{Q} \gets (\mathcal{F} \setminus \mathcal{E})$ sorted by $\tilde\utility$ desc.\par
$\mathcal{K} \gets \emptyset$\par
\tcp{\small \emph{Fill $D$ diversity slots.}}
\For{$j = 1, \dots, D$}{
  $c \gets \PickDiverse(\mathcal{Q},\, \Lambda,\, P,\, t,\, \tau)$\; \tcp{\cref{alg:pickdiverse}}
  \If{$c$ is \textnormal{none}}{\textbf{break}\;}
  $\mathcal{K} \gets \mathcal{K} \cup \{c\}$;\quad $\mathcal{Q} \gets \mathcal{Q} \setminus \{c\}$\par
  $\Lambda \gets \Lambda \cup (\lineage(c) \setminus \{\mathrm{id}(c)\})$\par
  $P \gets P \cup \{\mathrm{parent}(c)\}$\par
}
\tcp{\small \emph{Backfill any unfilled diversity slots above the floor.}}
$\mathcal{K} \gets \mathcal{K} \cup \topk_{D - |\mathcal{K}|}\!\bigl(\{c \in \mathcal{Q} : 
\tilde\utility(c) \ge \tau\};\, \tilde\utility\bigr)$\par
\Return $\mathcal{E} \cup \mathcal{K}$ sorted by $\tilde\utility$ desc.\par
\caption{$\mathcal{S}_t^{\mathrm{elit}\text{-}\mathrm{div}}$: elitist selection with diversity slots.}
\label{alg:elit-div}
\end{algorithm}
\begin{algorithm}%
\SetAlgoLined
\SetKwInOut{Input}{input}
\SetKwInOut{Output}{output}
\Input{sorted-desc candidates $\mathcal{Q}$; used-ancestor set $\Lambda$;
  used-parent set $P$; current step $t$; quality floor $\tau$;
  $\mathrm{gen}(c)$ is the step at which $c$ was proposed.}
\Output{a diverse candidate $c \in \mathcal{Q}$, or \textsc{none}.}
\tcp{\small \emph{T1: brand-new lineage, born this step.}}
\For{$c \in \mathcal{Q}$}{
  \If{$\tilde\utility(c) < \tau$}{\textbf{break}\;}
  \If{$(\lineage(c) \setminus \{\mathrm{id}(c)\}) \cap \Lambda = \emptyset$ \textbf{and} $\mathrm{gen}(c) = t$}{\Return $c$\;}
}
\tcp{\small \emph{T2: brand-new lineage, any generation.}}
\For{$c \in \mathcal{Q}$}{
  \If{$\tilde\utility(c) < \tau$}{\textbf{break}\;}
  \If{$(\lineage(c) \setminus \{\mathrm{id}(c)\}) \cap \Lambda = \emptyset$}{\Return $c$\;}
}
\tcp{\small \emph{T3: at least a different immediate parent.}}
\For{$c \in \mathcal{Q}$}{
  \If{$\tilde\utility(c) < \tau$}{\textbf{break}\;}
  \If{$\mathrm{parent}(c) \notin P$}{\Return $c$\;}
}
\Return \textsc{none}\par
\caption{$\PickDiverse$ (\texttt{mode=lineage}): one diversity pick.}
\label{alg:pickdiverse}
\end{algorithm}
\begin{figure}[t]
\centering
\begin{tikzpicture}[x=1cm, y=1cm, font=\scriptsize,
  hcard/.style={draw=spliceaccent!60, fill=spliceaccent!4, line width=1.0pt,
    rounded corners=4pt},
  httl/.style={anchor=west, inner sep=0pt, font=\small\bfseries,
    text=spliceaccent},
  cell/.style={anchor=center, inner sep=0pt},
  win/.style={draw=spliceaccent, fill=spliceedit!60, line width=1.0pt,
    rounded corners=1.4pt},
]

\draw[hcard] (0.00, 0.00) rectangle ( 5.40,-3.40);
\draw[hcard] (5.75, 0.00) rectangle (13.90,-3.40);

\node[httl] at (0.12,-0.30) {(a) The cost};
\draw[draw=splicerule!60, line width=0.4pt] (0.75,-2.85) -- (5.05,-2.85);
\draw[draw=splicerule!60, line width=0.4pt] (0.75,-2.85) -- (0.75,-0.85);
\node[anchor=north, inner sep=1pt] at (2.90,-3.06)
  {length ratio $\mathrm{len}(x)/\mathrm{len}(x^{(0)})$};
\node[rotate=90, anchor=south, inner sep=1pt] at (0.62,-1.85) {$\cost_{\mathrm{len}}$};
\foreach \r/\xx in {0.5/0.75, 1/2.15, 1.5/3.55, 2/4.95}{%
  \draw[splicerule!60] (\xx,-2.85) -- (\xx,-2.92);
  \node[anchor=north, inner sep=0.5pt, font=\tiny, text=splicerule] at (\xx,-2.92) {$\r$};}
\draw[draw=spliceaccent, line width=1.1pt] (0.75,-2.85) -- (2.15,-2.85) -- (4.95,-0.95);
\fill[white, draw=splicerule, line width=0.6pt] (2.15,-2.85) circle (0.06);
\node[anchor=south, inner sep=2pt] at (2.15,-2.85) {seed};
\node[anchor=west, inner sep=0pt, text=spliceaccent, align=left] at (1.00,-1.15)
  {$\cost_{\mathrm{len}} = \max(0,\, \text{ratio} - 1)$};
\node[anchor=west, inner sep=0pt, text=splicerule, font=\scriptsize\itshape]
  at (0.85,-2.35) {shortening is free};
\node[anchor=west, inner sep=0pt, text=splicerule, font=\scriptsize\itshape,
  rotate=34, anchor=south west] at (2.95,-2.70) {growth is charged};

\node[httl] at (5.87,-0.30) {(b) What $\reg$ changes};
\node[cell, font=\scriptsize\bfseries, text=splicerule] at (7.10,-0.85) {candidate};
\node[cell, font=\scriptsize\bfseries, text=splicerule] at (8.55,-0.85) {$\utility(x;\valid)$};
\node[cell, font=\scriptsize\bfseries, text=splicerule] at (9.75,-0.85) {ratio};
\node[cell, font=\scriptsize\bfseries, text=splicerule] at (11.05,-0.85) {$\tilde\utility$, $\reg = 0$};
\node[cell, font=\scriptsize\bfseries, text=splicerule] at (12.85,-0.85) {$\tilde\utility$, $\reg = 0.05$};
\draw[splicerule!40, line width=0.4pt] (6.05,-1.07) -- (13.70,-1.07);

\draw[win] (10.50,-1.78) rectangle (11.60,-2.20);
\draw[win] (12.30,-1.22) rectangle (13.40,-1.64);
\foreach \y/\c/\f/\r/\a/\b in {%
  -1.43/{parent $x$}/{0.80}/{1.00}/{0.800}/{0.800},
  -1.99/{child $x'$}/{0.81}/{1.30}/{0.810}/{0.795}}{%
  \node[cell] at (7.10,\y) {\c};
  \node[cell] at (8.55,\y) {$\f$};
  \node[cell] at (9.75,\y) {$\r$};
  \node[cell] at (11.05,\y) {$\a$};
  \node[cell] at (12.85,\y) {$\b$};}
\node[anchor=north west, inner sep=0pt, text width=76mm, align=left]
  at (6.05,-2.45)
  {The child gains $0.01$ but is $30\%$ longer. At $\reg = 0.05$ the growth
   costs $0.05 \times 0.30 = 0.015$, more than the gain, so the selector keeps
   the parent. Whatever $\reg$ is, the accuracy given up is at most
   $\reg\cdot\cost(x^\star)$ (\cref{thm:reg-bias}).};

\end{tikzpicture}
\caption{\textbf{Pricing growth} (illustrative numbers). \textbf{(a)} The ratio
cost of \eqref{eq:ratio-cost} is zero at or below the seed's length and grows
linearly beyond it, so it has the same meaning on every task. \textbf{(b)} A
child that buys a small gain with a long addition wins at $\reg = 0$ and loses
at $\reg = 0.05$; the tinted cells mark the candidate the selector prefers at each $\reg$.}
\label{fig:length}
\end{figure}

\section{Theory: Supporting Results and Proofs}
\label{sec:main.theory}\label{sec:app.theory}
This section extends \cref{sec:theory-main}: the notation, the assumptions the
main text does not state, the results it only points to, and proofs of every
result. \Cref{tab:theory-glossary} gives the assumptions behind each result.
\label{sec:theory}

The analysis has two goals. First, to place the \splice{} solver on firm footing:
each component of \cref{sec:main.method} other than the critique (selective
mutability, elitist selection, diversity slots, and length regularization)
is paired with a formal guarantee that makes its
contribution precise rather than anecdotal. Second, to name what the
experiments of \cref{sec:main.exp} can check: some results predict a quantity
an ablation measures, and others are not tested in this paper. Proofs follow each
group of results.

\begin{table}[t]
\centering
\small
\caption{\small \textbf{The results of \cref{sec:theory} at a glance.} Each row
gives a result, what it says, and what it rests on. Proposal coverage, in its
population form (Asm.~\ref{asm:proposal-coverage}) and its scored form
(Asm.~\ref{asm:scored-coverage}), is the assumption about the LLM proposer that
we cannot verify; only the three rate results in the bottom block use it. The
middle block rests on $\gamma$-submodularity (Asm.~\ref{asm:gamma-submod})
instead, for which we have no per-benchmark estimator.}
\label{tab:theory-glossary}
\renewcommand{\arraystretch}{1.12}
\begin{tabular}{@{}l>{\raggedright\arraybackslash}p{7.3cm}>{\raggedright\arraybackslash}p{3.7cm}c@{}}
\toprule
\textbf{Result} & \textbf{What it says} & \textbf{Rests on} & \textbf{Cov.} \\
\midrule
\multicolumn{4}{@{}l}{\emph{Hold whether or not the proposer is competent}} \\
\mbox{Prop.~\ref{prop:edit-graph-diameter}} & \emph{Diameter.} Any two states are at most $m_{\mathrm{mut}}+\mathrm{diam}(\Pi)$ edits apart & connected ordering moves & \xmark \\
\mbox{Lem.~\ref{thm:monotone}} & \emph{Leader.} The leader's score never decreases & cached scoring, parent retention & \xmark \\
\mbox{Prop.~\ref{prop:beam-beats-greedy}} & \emph{Beam vs greedy.} The elitist leader scores at least greedy's state until the beam drops that state & coupled draws, per-state cache & \xmark \\
\mbox{Thm.~\ref{thm:div-preserve}} & \emph{Diversity.} If a competitive child of another parent exists, the beam keeps one & a qualifying candidate in $\mathcal{F}_t$ & \xmark \\
\mbox{Prop.~\ref{prop:adaptive-anneal}} & \emph{Gate.} The quality gate $\tau_t=\gateratio\ell_t$ never falls & Lem.~\ref{thm:monotone}, per-state cache & \xmark \\
\mbox{Thm.~\ref{thm:noisy-monotone}} & \emph{Noisy no-regression.} Under re-scoring, an $\varepsilon$-regression has probability at most $\bw(1{+}k)e^{-n_{\mathrm{re}}\varepsilon^2/c_\sigma}$ & Asm.~\ref{asm:bounded-reward}, fresh batches, elitist selector & \xmark \\
\mbox{Cor.~\ref{cor:min-batch}} & \emph{Batch size.} Batch size $n_{\mathrm{re}} \ge (c_\sigma/\varepsilon^2)\log(\bw(1{+}k)/\delta)$ suffices for $\delta$-monotonicity & as Thm.~\ref{thm:noisy-monotone} & \xmark \\
\mbox{Thm.~\ref{thm:uniform-conv}} & \emph{Uniform convergence.} Validation scores over the search history are within $\varepsilon$ of population scores & Asm.~\ref{asm:bounded-reward}; (ii) single-edit proposals & \xmark \\
\mbox{Thm.~\ref{thm:reg-bias}} & \emph{Length bias.} Regularizing with $\reg$ costs at most $\reg\cdot\cost(x^\star)$ in utility & $\cost \ge 0$ & \xmark \\
\midrule
\multicolumn{4}{@{}l}{\emph{Rest on $\gamma$-submodularity, not coverage}} \\
\mbox{Prop.~\ref{thm:greedy-approx}} & \emph{Greedy exactness.} With one alternative per block, block-pass greedy reaches the optimum & $\gamma > 0$, monotone $g$, one alternative per block & \xmark \\
\mbox{Prop.~\ref{prop:mut-price}} & \emph{Price of freezing.} At the seed ordering, freezing costs at most $\gamma^{-1}$ times the frozen blocks' best single-block gains over their counterfactual variants & fixed ordering, $\gamma$ on $E^{\mathrm{all}}$, diminishing returns & \xmark \\
\mbox{Conj.~\ref{conj:beam-improvement}} & \emph{Beam width.} At each $\gamma$, worst-case leader quality is non-decreasing in $\bw$ & $\gamma$-submodular instances & \xmark \\
\midrule
\multicolumn{4}{@{}l}{\emph{Rate results: need proposal coverage}} \\
\mbox{Thm.~\ref{thm:reachability}} & \emph{Reachability.} While no beam member is $\varepsilon$-optimal, a step finds an improving child with probability at least $1-(1-p^\star)^{|\mathcal{C}_{t-1}|k}$ & Asm.~\ref{asm:proposal-coverage}, \ref{asm:proposal-distribution}, per-state cache & \cmark \\
\mbox{Thm.~\ref{thm:convergence}} & \emph{Convergence.} Horizon $T \ge (2K + 8\ln(1/\delta))/p^{\mathrm{adv}}$ steps reach $\varepsilon$-optimality on the scored objective $\tilde\utility$ with probability $1-\delta$ & Asm.~\ref{asm:scored-coverage}, Lem.~\ref{thm:monotone}, per-state cache & \cmark \\
\mbox{Cor.~\ref{cor:convergence-population}} & \emph{Population form.} The same for population utility, up to $2\varepsilon_n$ plus the penalty at $x^\star$ & as Thm.~\ref{thm:convergence}, plus Thm.~\ref{thm:uniform-conv} & \cmark \\
\bottomrule
\end{tabular}
\end{table}

\paragraph{Two optima, and which one every statement here targets.}
We write $x^\star_{\mathrm{all}}$ for the maximizer of
$\utility^\star$ over the \emph{all-mutable} space
$\mathcal{X}_{\mathrm{all}}$. That is the right reference for pricing the
constraints (\cref{prop:mut-price}) and the wrong one for analyzing the
search, which never leaves $\mathcal{X}$. Throughout this section, therefore,
\begin{equation}
\label{eq:xstar-convention}
x^\star \;\in\; \argmax_{x \in \mathcal{X}} \utility^\star(x),
\end{equation}
the optimum over the constrained space with contract blocks frozen (the
$x^\star_{\mathrm{mut}}$ of \cref{prop:mut-price}). The distinction is not
cosmetic. Measured against $x^\star_{\mathrm{all}}$, a state that is optimal
within $\mathcal{X}$ may still be $\varepsilon$-suboptimal with no improving
neighbor available, which would make \cref{asm:proposal-coverage} false and
the target of \cref{thm:convergence} unreachable at any horizon;
\cref{sec:exp.contract} measures that residual gap at $10.5$\,pp on a
deliberately expensive contract, so the difference is real rather than
hypothetical. Every $\varepsilon$-optimality claim below is against the best
\emph{feasible} prompt, and the price of feasibility itself is
\cref{prop:mut-price}'s separate subject.

\paragraph{What the concentration results need from the metric.}
Relating a batch score to the population score takes two properties of the
metric.
\begin{assumption}[Bounded, decomposable metric]
\label{asm:bounded-reward}
The per-example score lies in $[0,1]$ (\cref{def:reward}), and the score of a
batch is the mean of its per-example scores \eqref{eq:empirical-utility}.
\end{assumption}

Boundedness means, by Hoeffding's lemma, that each centered score is
sub-Gaussian with variance proxy $\sigma^2 \le 1/4$ (\cref{lem:hoeffding}).
Decomposability is what lets a batch of $n$ examples estimate the population
score to within $O(n^{-1/2})$. Accuracy and exact match satisfy both.
Set-level metrics, such as macro-$F_1$ and the mean Spearman $\rho$ of the
judge task (\cref{sec:exp.judge}), are not decomposable, because they are not
averages of per-example terms; $\rho$ also ranges over $[-1,1]$ rather than
$[0,1]$. Only the concentration results use \cref{asm:bounded-reward}, since
they are the results that relate a batch score to the population score
(\cref{thm:uniform-conv,thm:noisy-monotone,cor:min-batch,cor:convergence-population}),
and for set-level metrics they do not apply as stated. The remaining results
(monotonicity, diversity, reachability, convergence on the scored objective,
and the prices of regularization and mutability) do not relate a batch score
to the population score, so they do not use \cref{asm:bounded-reward} and
hold for any scoring function.

\paragraph{Cached scores.}
We write $\hat\utility_n(x)$ for $\utility(x;S)$ on a sample $S$ of $n$
examples, so that $\tilde\utility(x) = \hat\utility_n(x) - \sum_j \lambda_j R_j(x)$
with $S = \valid$.
\splice{} scores a candidate on $\valid$ when it is first evaluated and caches
that score while the candidate stays in the beam; a state that leaves the beam
and is later re-proposed verbatim is evaluated again (\cref{sec:main.method-ext}).
We call this the \emph{cached-scoring convention}, and \cref{thm:monotone}
needs nothing more. The definitions and results that treat
$\tilde\utility(\cdot) := \tilde\utility(\cdot\,;\valid)$ as a function on
$\mathcal{X}$
(\cref{def:improving,def:delta-improving,asm:scored-coverage,thm:reachability,prop:beam-beats-greedy,thm:convergence,cor:convergence-population,prop:adaptive-anneal})
assume in addition that scores are \emph{cached per state}: every evaluation
of a state returns the same score, as with a deterministic system agent.
\cref{thm:uniform-conv}(ii) covers the general case, where re-evaluations are
priced by a factor $T+1$. The exception is the re-evaluation regime of
\cref{thm:noisy-monotone,cor:min-batch}, the alternative design in which every
frontier member is re-scored on fresh examples at each step.

\subsection{Probabilistic Preliminaries}
\label{sec:app.theory:prelim}

We record the standard facts used below, so that every proof reduces to an
explicit application of one of them.

\begin{lemma}[Hoeffding's inequality]
\label{lem:hoeffding}
Let $Z_1,\dots,Z_n$ be independent with $Z_j \in [0,1]$ and
$\bar Z = \tfrac1n\sum_j Z_j$. Then for any $\varepsilon>0$,
$\Pr\bigl[|\bar Z - \E \bar Z| > \varepsilon\bigr] \le 2\exp(-2n\varepsilon^2)$.
Moreover $\bar Z - \E\bar Z$ is sub-Gaussian with variance proxy
$\sigma^2/n$ where $\sigma^2 \le 1/4$.
\end{lemma}

\begin{lemma}[Difference of independent sub-Gaussians]
\label{lem:subg-diff}
If $U, V$ are independent, mean-zero, sub-Gaussian with variance proxies
$s^2$ each, then $U - V$ is mean-zero sub-Gaussian with proxy $2s^2$, so
$\Pr[U - V > \varepsilon] \le \exp\bigl(-\varepsilon^2/(4s^2)\bigr)$.
\end{lemma}

\begin{proof}
The moment generating function of a sub-Gaussian with proxy $s^2$ satisfies
$\E e^{\theta U} \le e^{\theta^2 s^2/2}$. By independence
$\E e^{\theta(U-V)} \le e^{\theta^2 s^2/2}e^{\theta^2 s^2/2} =
e^{\theta^2 (2s^2)/2}$, which is the proxy-$2s^2$ bound; the tail bound is
the Chernoff optimization $\theta = \varepsilon/(2s^2)$.
\end{proof}

\begin{lemma}[Lower-tail Chernoff bound, explicit form]
\label{lem:chernoff}
Let $N_T$ be a sum of $T$ Bernoulli indicators, each with conditional success
probability at least $p \in (0,1]$ given the past. Fix an integer
$K \ge 1$ and $\delta \in (0,1)$. If
\begin{equation}
\label{eq:chernoff-T}
T \;\ge\; \frac{2K + 8\ln(1/\delta)}{p},
\end{equation}
then $\Pr[N_T < K] \le \delta$.
\end{lemma}

\begin{proof}
Because each indicator has conditional success probability at least $p$
given the past, $N_T$ stochastically dominates $\mathrm{Bin}(T,p)$: couple
each indicator to a $\mathrm{Bernoulli}(p)$ variable that it dominates
pointwise, and sum. It therefore suffices to bound the lower tail of
$Y \sim \mathrm{Bin}(T,p)$ with mean $\mu = Tp$. The multiplicative Chernoff
bound gives, for $0 < K < \mu$,
\[
\Pr[Y < K] \;\le\; \exp\Bigl(-\tfrac{(\mu-K)^2}{2\mu}\Bigr).
\]
Write $\beta = \ln(1/\delta)$. By \eqref{eq:chernoff-T}, $\mu \ge 2K + 8\beta$. Then
$K \le \mu/2$, so $\mu - K \ge \mu/2$ and
$\tfrac{(\mu-K)^2}{2\mu} \ge \tfrac{(\mu/2)^2}{2\mu} = \tfrac{\mu}{8} \ge
\tfrac{2K+8\beta}{8} \ge \beta$. Hence $\Pr[Y<K] \le e^{-\beta} = \delta$.
\end{proof}

\Cref{lem:chernoff} gives \cref{thm:convergence} an explicit requirement,
\eqref{eq:chernoff-T}, in place of an unspecified universal constant; it is
additive rather than multiplicative in $\ln(1/\delta)$.

\subsection{Edit-Graph Reachability}
\label{sec:theory:graph}

The diameter bound \cref{prop:edit-graph-diameter} is a worst-case
statement about the geometry of $\mathcal{X}$: any two states are connected
by at most $m_{\mathrm{mut}}+\mathrm{diam}(\Pi)$ primitive edits, and
selective mutability reduces this by excluding immutable blocks from the
replacement phase. But \splice{} does not enumerate neighborhoods; it
samples from $q_\phi$, and progress at step $t$ depends on whether an
\emph{improving} neighbor is among the $\bw k$ samples drawn.

\begin{definition}[Improving proposal]
\label{def:improving}
For a state $x$ and history $H_t$, a proposal $u \sim q_\phi(\cdot \mid
x, H_t)$ is \emph{improving from $x$} if
$\tilde\utility(\mathrm{Apply}(x,u)) > \tilde\utility(x)$.
\end{definition}

\par\noindent\emph{\Cref{asm:proposal-coverage} is stated in the main text (p.~\pageref{asm:proposal-coverage}).}\par

\cref{asm:proposal-coverage} is a \emph{minimum-competence} condition:
whenever a sufficient improvement is available, one is proposed with
non-vanishing probability. It is not a property we prove about the LLM.
Three clarifications matter, because the form of the statement carries most of
what a reader should take from it.

\emph{It is required at one tolerance, and only away from the target.} Asking
for a single $p^\star$ bounded away from zero at \emph{every} non-optimal state
would not be plausible: near the optimum the remaining gap is small, most
single-coordinate edits are neutral or worse, and the improving probability
decays toward zero. The assumption avoids that only because $\varepsilon$ is
fixed in advance. Were it required for every $\varepsilon > 0$, it would
reinstate the implausible version: any non-optimal state lies in the
sublevel set at $\varepsilon$ equal to its own gap. With $\varepsilon$ fixed,
coverage is invoked only on states at least $\varepsilon$ from the target,
which is all \cref{thm:reachability} uses, since it is stated for that
$\varepsilon$; the convergence rate rests on the scored analogue,
\cref{asm:scored-coverage}, in the same form. The decay then shows up as a comparison \emph{across}
instantiations: a practitioner who demands a closer solution assumes coverage
at a smaller $\varepsilon$, which is a weaker landscape and a larger step
requirement in \cref{thm:convergence}. That is the graceful degradation
familiar from stochastic local search, and \cref{sec:exp.pstar} shows the
improving rate falling as the leader climbs.

\emph{It constrains the landscape, the sampler and the scoring noise jointly.}
Three things must hold for $p^\star(\varepsilon) > 0$. The state must have an
improving single-coordinate neighbor; at a radius-one local maximum that is
still $\varepsilon$-suboptimal it does not, and block interactions (a single edit
fails while a coordinated pair improves) are exactly the mechanism producing such states. The proposer must
put positive probability on such a neighbor, which
\cref{asm:proposal-distribution} does not give, since it bounds the support of
$q_\phi$ from above rather than below. And because improvement is judged on the
scored objective, the neighbor's validation score must exceed its parent's,
which depends on the realized validation sample as well as on the population
gap. The sublevel set is defined by population utility because the
reachability argument reasons about which parents are truly suboptimal;
the event inside is judged on the score because that is what the selector sees.
In particular, $p^\star(\varepsilon) > 0$ requires every state with no
score-improving neighbor, a maximizer of the validation score over $\mathcal{X}$
included, to be $\varepsilon$-optimal in population, so the assumption also
asserts that validation overfitting at the score maxima is below
$\varepsilon$.

\emph{Coverage is needed at every parent, not just one.} The reachability
bound multiplies failure probabilities across all $|\mathcal{C}_{t-1}|\,k$
draws, and each factor is at most $1-p^\star(\varepsilon)$ only if the parent it
was drawn from is itself $\varepsilon$-suboptimal. That is why
\cref{lem:beam-subopt} asks it of every parent. If some parent is already
$\varepsilon$-optimal the hypothesis fails and the theorem says nothing, though
the beam then holds an $\varepsilon$-optimal candidate, which may not be the
leader.
\cref{sec:exp.pstar} estimates the improving rate directly from the per-step
proposal acceptance rate.

\paragraph{Which results depend on coverage.}
Coverage, in its population form (\cref{asm:proposal-coverage}) and its scored
form (\cref{asm:scored-coverage}, stated with the convergence rate), is the
assumption in this section we cannot verify, so it matters exactly where it is
used: the population form in reachability and the scored form in the
convergence results, all of which bound a \emph{rate}. \cref{tab:theory-glossary}
makes the dependency structure explicit. The main consequence is that
everything that holds by construction is coverage-free: the returned prompt
never scores below the seed on the validation split, the beam keeps a
competitive distinct-parent candidate whenever one is proposed, and the text of
frozen blocks is byte-identical to the seed's, whether or not the proposer is
competent. When coverage fails, \splice{} makes no progress and returns its
monotone best-so-far. \emph{Coverage failure costs speed, not these
guarantees.} None of them says that the target obeys a frozen clause; that is
behavior, which \cref{sec:exp.contract} measures.

Coverage is also not a convenience specific to \splice{}. It is a condition
on the problem class: if no improving edit can be proposed from a reachable
non-optimal state, no proposer--critic loop can progress from it, so the
limitation belongs to the paradigm rather than to our analysis of it. Prior
work does carry formal guarantees: ProTeGi's candidate selection is a
bandit procedure, CAPO\textsubscript{cost}'s survival step uses racing with a multiple-testing
correction, and constrained variants prove compliance results under their own
assumptions. These guarantees concern selecting among candidates, not the
trajectory of the search that produced them, so they do not need a condition
on what the proposer can generate. We name that condition and measure it.

\par\noindent\emph{\Cref{thm:reachability} is stated in the main text (p.~\pageref{thm:reachability}).}\par

The bound is on the draws, before deduplication. An improving draw whose child
duplicates a member of $\mathcal{C}_{t-1}$, or an earlier draw of the step, is
removed from $\mathcal{N}_t$; the state it reached is then already in the beam
or in $\mathcal{N}_t$, so it adds no new candidate.

The exponent counts the draws made, $|\mathcal{C}_{t-1}|\,k$, and not
the budget $\bw k$; the two coincide only once the beam is full. They do
coincide for the whole run after the first step whenever $1+k \ge \bw$, as at
$\bw{=}4$ with $k{=}3$ (GPQA and the seeded SST-5 sweeps) or $k{=}5$ (BBH),
since the step-$1$ frontier holds $1+k \ge \bw$ candidates and elitist
selection retains $\bw$ of them, so $|\mathcal{C}_{t-1}| = \bw$ for every
$t \ge 2$ whenever the frontier supplies $\bw$ distinct candidates and, when
$\gateratio > 0$, at least $D$ non-elite frontier members clear the quality
gate (automatic for $\gateratio = 0$, as on GPQA and BBH). At $t=1$
the exponent is $k$, because $\mathcal{C}_0 = \{x^{(0)}\}$ is a single parent.
We state the theorem in the exact form rather than the budgeted one because
substituting $\bw k$ at $t=1$ would overstate the first step's guarantee, and
the first step is where the run has the least to fall back on.

\cref{thm:reachability} quantifies the joint effect of beam width and
proposals-per-candidate: increasing either drives the failure probability to
zero exponentially. It is a statement about \emph{some} parent improving, and
it is the wrong quantity for predicting whether the search makes progress. At
$\bw k = 12$ a per-proposal rate of $0.25$ gives $p^{\mathrm{step}} = 0.97$ and
even $0.08$ gives $0.63$, so by this measure fan-out looks almost freely
substitutable for proposer quality. Progress of the \emph{leader}, which is what
\cref{thm:convergence} counts, runs on the leader's own $k$ draws, and there the
same rates give $0.58$ and $0.22$ at $k = 3$. So the operating condition is
$p^\star(\varepsilon)$ above $1/k$, not $1/\bw k$, and beam
width buys reachability without buying the proven rate. The coverage rate is
also not a property of the proposer alone: \cref{sec:exp.pstar} measures it
with the proposer fixed and finds it rises with the capacity of the target
model, so where a given proposer falls relative to that condition depends on
what it is prompting.

\paragraph{Proofs.}
\label{sec:app.theory:graph}

\begin{proof}[Proof of \cref{prop:edit-graph-diameter}]
Fix $x = (v, \pi)$ and $x' = (v', \pi')$ in $\mathcal{X}$, where $v, v'$ are
the block-variant assignments and $\pi, \pi'$ the orderings. We exhibit a path
in the edit graph of the stated length.

A block-replacement move alters one coordinate $v_i$ and leaves $\pi$ fixed;
an ordering move alters $\pi$ and leaves every $v_i$ fixed. The two families
therefore act on disjoint coordinates of the pair $(v,\pi)$ and commute as
operators on $\mathcal{X}$, so we may schedule all replacements before all
ordering moves without changing the endpoint.

\emph{Replacement phase.} For each $i$ with $v_i \ne v'_i$, one
block-replacement move sends $v_i \mapsto v'_i$; this is admissible because
$v'_i \in \mathcal{V}_i$ by $x' \in \mathcal{X}$. Immutable blocks satisfy
$v_i = v_i^{(0)} = v'_i$ by \cref{def:mutability} and so contribute no move.
A block with $|\mathcal{V}_i| = 1$ likewise cannot differ. Hence the number of
replacement moves is at most
$\sum_{i \in I_{\mathrm{mut}}} \1\{|\mathcal{V}_i| \ge 2\}
\le m_{\mathrm{mut}}$.

\emph{Ordering phase.} The reachable orderings form the connected set $\Pi$
under the admissible ordering moves, so $\pi$ and $\pi'$ are joined by at most
$\mathrm{diam}(\Pi)$ such moves.

Concatenating the two phases gives a path of length at most
$m_{\mathrm{mut}} + \mathrm{diam}(\Pi)$, and since $x,x'$ were arbitrary this
bounds the diameter, which is \eqref{eq:edit-graph-diameter}. Selective
mutability enters only through the replacement phase: freezing block $i$
removes it from $I_{\mathrm{mut}}$ and hence deletes one unit from the bound.
\end{proof}

\label{sec:app.theory:reach}

The exponent in \cref{thm:reachability} counts every draw made in
the expansion, $|\mathcal{C}_{t-1}|\,k$, and each draw only carries the coverage
guarantee if the parent it was drawn from is itself $\varepsilon$-suboptimal. We
isolate that condition first.

\begin{lemma}[When every parent carries the coverage guarantee]
\label{lem:beam-subopt}
Suppose $\max_{x \in \mathcal{C}_{t-1}} \utility^\star(x) \le
\utility^\star(x^\star) - \varepsilon$, i.e.\ the beam contains no
$\varepsilon$-optimal candidate. Then every $x \in \mathcal{C}_{t-1}$ lies in
the $\varepsilon$-sublevel set of \cref{asm:proposal-coverage}, and hence every draw
$u \sim q_\phi(\cdot\mid x, H_t)$ from every parent is improving from its
parent with probability at least $p^\star(\varepsilon)$.
\end{lemma}

\begin{proof}
Immediate: if the maximum population utility over the beam is at most
$\utility^\star(x^\star) - \varepsilon$ then each individual member satisfies
$\utility^\star(x^\star) - \utility^\star(x) \ge \varepsilon$, which is
precisely the constraint indexing the minimum in \cref{asm:proposal-coverage}.
\end{proof}

\begin{remark}
\label{rem:leader-not-enough}
The hypothesis cannot be weakened to ``the beam \emph{leader} is
$\varepsilon$-suboptimal''. The leader is the maximizer of the \emph{empirical}
score $\tilde\utility(\cdot;\valid)$, not of $\utility^\star$, so under
scoring noise a beam whose leader is $\varepsilon$-suboptimal may still
contain an $\varepsilon$-optimal member. \Cref{lem:beam-subopt} is stated in
population terms for exactly this reason. If the hypothesis fails, the beam
already contains an $\varepsilon$-optimal candidate, but that does not meet the
conclusion of \cref{thm:convergence}, which concerns the leader's score: the
$\varepsilon$-optimal member need not be the leader.
\end{remark}

\begin{proof}[Proof of \cref{thm:reachability}]
Condition on $\mathcal{C}_{t-1}$ and the history $H_t$, and assume the
hypothesis of \cref{lem:beam-subopt}. Index the draws of the expansion by
$(x,j)$ with $x \in \mathcal{C}_{t-1}$ and $j \in [k]$, so there are
$|\mathcal{C}_{t-1}| \cdot k \le \bw k$ of them.

\emph{Independence.} Within a parent, the $k$ draws are i.i.d.\ from
$q_\phi(\cdot \mid x, H_t)$ by \eqref{eq:bips-body}. Across parents,
each parent's expansion is a separate invocation of the proposer with its own
randomness, and all of them condition on the same fixed $(\mathcal{C}_{t-1},
H_t)$; hence the full collection $\{u_{x,j}\}$ is mutually independent
given $(\mathcal{C}_{t-1}, H_t)$.

\emph{Failure probability.} Let $A_{x,j}$ be the event that $u_{x,j}$ is
improving from $x$ in the sense of \cref{def:improving}. By
\cref{lem:beam-subopt}, $\Pr[A_{x,j}] \ge p^\star(\varepsilon)$ for every
$(x,j)$. The event that the expansion contains no improving child is
$\bigcap_{x,j} A_{x,j}^{c}$, and by independence
\[
\Pr\Bigl[\bigcap_{x,j} A_{x,j}^{c}\Bigr]
= \prod_{x,j}\Pr[A_{x,j}^{c}]
\le \bigl(1 - p^\star(\varepsilon)\bigr)^{|\mathcal{C}_{t-1}|k}.
\]
Taking complements gives \cref{thm:reachability}. Feasibility of every
draw (no proposal leaves $\mathcal{X}$ or writes to an immutable
block) is \cref{asm:proposal-distribution}, so every improving draw reaches
a state of $\mathcal{X}$. The event is on the draws $\{u_{x,j}\}$, not on
$\mathcal{N}_t$ after deduplication: a draw whose child duplicates a member of
$\mathcal{C}_{t-1}$ or an earlier draw is removed, and the state it reached is
already in the beam or in $\mathcal{N}_t$.

One tempting step is unavailable, although the budgeted form reads like a
harmless simplification. Bounding
$(1-p^\star)^{|\mathcal{C}_{t-1}|k}$ above by
$(1-p^\star)^{\bw k}$ does not follow from $|\mathcal{C}_{t-1}| \le \bw$:
because $0 \le 1-p^\star \le 1$, raising it to the \emph{larger} exponent makes
it \emph{smaller}, so $(1-p^\star)^{\bw k} \le
(1-p^\star)^{|\mathcal{C}_{t-1}|k}$ and the substitution runs the wrong way.
The budgeted form is therefore a valid bound only when the beam is full. It
fails at $t=1$, where $|\mathcal{C}_0| = 1$, and the exponent is $k$.
\end{proof}

\begin{remark}[What reachability does and does not give]
\label{rem:reach-vs-advance}
\Cref{thm:reachability} bounds the probability that \emph{some} child beats
\emph{its own parent}. That is weaker than the leader advancing: a child
improving on a non-leader parent need not exceed $\ell_{t-1}$. The
distinction is what separates $p^{\mathrm{step}}$ from the leader-advance rate
$p^{\mathrm{adv}}$ of \cref{thm:convergence}.
\end{remark}

\subsection{Approximate Submodularity and Greedy Approximation}
\label{sec:theory:submod}

\cref{thm:reachability} bounds the probability that some proposal improves on
its parent, which need not move the leader, but is silent about \emph{how much} progress is possible overall. A
terminal-quality bound needs structural information about $\utility^\star$.
We import the standard framework: approximate submodularity of the utility
as a set function over block-variant commits. Fixing $\pi=\pi^{(0)}$, let
\begin{equation}
\label{eq:ground-sets}
E \;=\; \bigl\{(i, v) : i \in I_{\mathrm{mut}},\; v \in \mathcal{V}_i \setminus \{v_i^{(0)}\}\bigr\},
\qquad
E^{\mathrm{all}} \;=\; \bigl\{(i, v) : i \in [m],\; v \in \tilde{\mathcal{V}}_i \setminus \{v_i^{(0)}\}\bigr\},
\end{equation}
where $\tilde{\mathcal{V}}_i$ denotes the \emph{counterfactual variant set} of
block $i$ (the variants block $i$ would admit had it not been frozen), with
$v_i^{(0)} \in \tilde{\mathcal{V}}_i$ for every $i$ and
$\tilde{\mathcal{V}}_i = \mathcal{V}_i$ for every $i \in I_{\mathrm{mut}}$, so
that $\mathcal{X}_{\mathrm{all}} = (\prod_{i} \tilde{\mathcal{V}}_i) \times \Pi$. So
$E$ collects the alternative variants across mutable blocks, each tagged with
its block so that equal text in two blocks gives two elements, and
$E^{\mathrm{all}} \supseteq E$ additionally carries the alternatives that
freezing removed. Call $A\subseteq E^{\mathrm{all}}$ \emph{feasible} if it has
at most one entry per block, and for feasible $A$ let $x_A$ substitute each
$(i,v)\in A$ into $x^{(0)}$. The gain function, defined on feasible sets, is
\begin{equation}
\label{eq:gain-function}
g(A) \;:=\; \utility^\star(x_A) - \utility^\star(x^{(0)}), \qquad g(\emptyset) = 0,
\end{equation}
well defined on every feasible subset of $E^{\mathrm{all}}$ because \eqref{eq:population-utility}
depends on a state only through $\render(x)$ and so extends verbatim from
$\mathcal{X}$ to the all-mutable space $\mathcal{X}_{\mathrm{all}}$ of
\eqref{eq:main-objective}. Prompt optimization over mutable blocks is
constrained maximization of $g$ on $E$; the larger ground set enters only in
\cref{prop:mut-price}, to express what the restriction to $E$ costs.

Two points about $\tilde{\mathcal{V}}_i$, since it is the one object here not
already supplied by \cref{sec:main.method}. First, it is what makes the price
of freezing expressible at all. By \cref{def:mutability} an immutable block has
$\mathcal{V}_i = \{v_i^{(0)}\}$, so any quantity written as a maximum over
$\mathcal{V}_i$ for $i \in I_{\mathrm{imm}}$ is identically zero, and a bound
in that form could only ever assert that freezing is free. The counterfactual
set is the smallest addition that lets the question be posed. Second, it is
\emph{not observable}: recovering $\tilde{\mathcal{V}}_i$ would mean running
the search with the contract released, which is precisely what deployment
forbids. $E^{\mathrm{all}}$ therefore buys a meaningful statement, not a
measurable one (\cref{rem:mut-price-condition}). Search itself never touches
$E^{\mathrm{all}} \setminus E$: every algorithm in this paper optimizes $g$
over $E$ alone.

We note the shape of the constraint on $E$ now, because it matters for what
the approximation guarantee below covers: the feasible sets are those with at
most one entry per block, which form a \emph{partition matroid} with one part
per mutable block and capacity one, not a cardinality constraint.

\begin{definition}[Submodularity ratio, \citealp{das2011submodular}]
\label{def:gamma}
The \emph{submodularity ratio} of $g$ on a ground set $E'$ is
\begin{equation}
\label{eq:gamma}
\gamma(g; E') \;:=\; \min_{\substack{L, S \subseteq E',\; L \cap S = \emptyset \\ L \cup S \text{ feasible},\; g(L \cup S) > g(L)}}
  \frac{\sum_{e \in S} [g(L \cup \{e\}) - g(L)]}{g(L \cup S) - g(L)} \;\le\; 1,
\end{equation}
the minimum running over the pairs on which $g$ is defined. It is the ratio of
\citet{das2011submodular} restricted to the partition matroid; when every
subset is feasible the two coincide. $\gamma = 1$ recovers standard
submodularity; $\gamma = 0$ indicates fully complementary interactions.
\end{definition}

\begin{assumption}[$\gamma$-submodular utility]
\label{asm:gamma-submod}
The gain function $g$ of \eqref{eq:gain-function} has submodularity ratio
$\gamma(g; E^{\mathrm{all}}) \ge \gamma$ for some $\gamma \in (0, 1]$.
\end{assumption}

We state \cref{asm:gamma-submod} on the larger ground set because
\cref{prop:mut-price} compares across the frozen/unfrozen boundary and so needs
$\gamma$ to control marginals on frozen coordinates too. Nothing is lost for
the results that live on $E$: since $E \subseteq E^{\mathrm{all}}$, the minimum
in \eqref{eq:gamma} runs over at least as many pairs on the larger set, so
$\gamma(g; E^{\mathrm{all}}) \le \gamma(g; E)$ and the assumption implies
$\gamma(g; E) \ge \gamma$. \cref{thm:greedy-approx} and
\cref{conj:beam-improvement} may therefore be read with $\gamma$ on $E$. The
cost of the stronger form is that $\gamma$ becomes a property of a landscape
that includes unobservable coordinates. We report no estimate of $\gamma$ on
either ground set: estimating it per benchmark would take block-subset
ablations we have not run, and it remains open (\cref{sec:main.limitations}).

Beyond its scope, \cref{asm:gamma-submod} is a structural assumption on the
task, not the algorithm: block interactions are bounded. It does not assert
monotonicity, that no block-level commit lowers utility. That would rarely hold
on the whole ground set when $\mathcal{V}_i$ holds every admissible text
(\cref{def:variants}), and only \cref{thm:greedy-approx} asks for it, on $E$
and in the single-alternative regime.

The shape of the constraint decides what is provable, and the two regimes it
admits behave very differently. When each mutable block has a single
alternative to its seed text, every subset of $E$ is feasible and the partition
constraint does no work. This regime is an idealization: by
\cref{def:variants}, $\mathcal{V}_i$ holds every variant the proposer could
produce, so no run of \splice{} is in it. We analyze it because it isolates what
$\gamma > 0$ buys, for an idealized greedy that sees the exact gain function.
\emph{Block-pass greedy} starts from $A_0 = \emptyset$; at each pass it
evaluates $g(A_j \cup \{e\})$ for every remaining $e \in E$, commits the
element with the largest positive marginal gain, and stops when no marginal is
positive. In this regime the guarantee is not an approximation factor but an
exact one.

\begin{proposition}[Block-pass greedy is exact in the single-alternative regime]
\label{thm:greedy-approx}
Suppose each mutable block contributes exactly one alternative, so that
$E = \{(i, \hat v_i) : i \in I_{\mathrm{mut}}\}$ and every subset of $E$ is
feasible, and that $g$ is monotone non-decreasing on $E$ ($g(A) \le g(A')$
whenever $A \subseteq A' \subseteq E$). Under \cref{asm:gamma-submod},
block-pass greedy returns a commit set $A_{\mathrm{gr}}$ with
\begin{equation}
\label{eq:greedy-bound}
g(A_{\mathrm{gr}}) \;=\; \max_{A \subseteq E} g(A),
\end{equation}
that is, $x_{A_{\mathrm{gr}}}$ attains the best utility reachable by block replacement at
the baseline ordering.
\end{proposition}

Two things follow, and the second is a limitation. First, in this regime a positive submodularity ratio is doing
more work than an approximation guarantee would suggest: $\gamma > 0$ forbids
exactly the configuration that traps greedy, a set of blocks that is valuable
jointly while each of its members is worthless alone. Second, the regime is
narrow. As soon as some block offers two or more alternatives the feasible sets
form a non-trivial \emph{partition matroid}, greedy must also choose \emph{which}
variant to commit per block, and no terminal-quality guarantee we are aware of
covers deterministic block-pass greedy there: the best available result for
monotone $\gamma$-weakly-submodular maximization under a matroid constraint is
$(1+1/\gamma)^{-2}$ for a \emph{randomized} greedy variant
\citep{chen2017weakly}, and whether deterministic greedy attains any
non-trivial ratio beyond cardinality constraints was open in that work.
\cref{rem:matroid} records what this does and does not justify. We therefore
claim no approximation factor in the multi-alternative regime rather than
import one that does not apply.

Wider beams maintain multiple partial configurations and can commit to
coordinated multi-block edits greedy would reject. Since
\cref{thm:greedy-approx} leaves nothing to improve when a single alternative
per block is on offer, the interesting claim lives in the multi-alternative
regime:

\begin{conjecture}[Beam improvement]
\label{conj:beam-improvement}
In the multi-alternative regime, fix $\gamma \in (0,1]$ and a proposer that puts
positive probability on every feasible single-block edit, score with the exact
gain $g$, and run elitist beam search of width $\bw$ until no proposal can
improve a beam member. Let $\alpha_\gamma(\bw)$ be the worst case, over
instances whose submodularity ratio is at least $\gamma$, of the expected ratio
between the returned leader's gain and $\max_{A} g(A)$ over feasible $A$. Then
$\alpha_\gamma$ is non-decreasing in $\bw$, and, over the instances where some optimal commit set requires two or
more blocks to move together, strictly increasing for $\bw \le m_{\mathrm{mut}}$.
\end{conjecture}

We state \cref{conj:beam-improvement} as a conjecture, and deliberately in
qualitative form. A proof would require a coupling between the beam's $\bw$
lineages that elitist selection makes dependent, so the
product-of-failure-probabilities argument does not apply; and we do not
conjecture a specific functional form for $\alpha_\gamma$, because no
deterministic-greedy baseline exists in this regime
(\cref{thm:greedy-approx}) for such a form to interpolate from. What \emph{is}
provable is a pathwise comparison:

\begin{proposition}[Elitist beam stays ahead of greedy until they separate]
\label{prop:beam-beats-greedy}
Run elitist beam search, with selector $\mathcal{S}^{\mathrm{elit}\text{-}\mathrm{div}}$
for any $D \le \bw - 1$ (which includes \splice{}), and greedy on the same
proposal draws, with scores cached per state and shared by both runs: whenever greedy's state $x^{\mathrm{g}}_{t}$ lies in
$\mathcal{C}_t$, both procedures see the same $k$ children of it. Let
$\nu := \min\{t : x^{\mathrm{g}}_t \notin \mathcal{C}_t\}$, with $\nu = \infty$
if no such step exists. Then
\begin{equation}
\label{eq:beam-greedy}
\ell_t \;\ge\; \tilde\utility\bigl(x^{\mathrm{g}}_{\min(t,\nu)}\bigr)
\qquad\text{for every } t \le T .
\end{equation}
For $\bw = 1$ the two procedures coincide under a common tie-breaking rule, so
$\nu = \infty$ and $\ell_t = \tilde\utility(x^{\mathrm{g}}_t)$ for every $t$.
\end{proposition}

Here greedy is hill climbing with retention: it moves to its best child only
if that child scores at least as high as its current state. The Greedy arm of
\cref{sec:main.exp} is the children-only beam with $\bw{=}1$ of
\cref{sec:algorithms-as-selectors}, which can regress, so
\cref{prop:beam-beats-greedy} does not describe it.

The comparison stops at $\nu$ because greedy then proposes from a state the
beam no longer holds, and the beam never scores those children;
\cref{rem:beam-greedy-after} gives a three-step instance with $\bw = 2$ in which
greedy finishes above the beam leader. The domination is also on the
\emph{scored} objective $\tilde\utility$ that both selectors act on, since a
score ordering need not agree with the population ordering, and it needs the
coupling: two runs with independent draws cannot be compared path by path.

\cref{prop:beam-beats-greedy} says that an elitist beam is never behind the
greedy path while it holds that path, and is at least level with greedy at the
step it drops it. It
does not say the beam finishes ahead of greedy; \cref{conj:beam-improvement} is
the empirically motivated claim that the benefit \emph{scales} with $\bw$.

\paragraph{Proofs.}
\label{sec:app.theory:submod}

The feasible sets of \eqref{eq:gain-function} are the subsets of
$E = \{(i,v) : i \in I_{\mathrm{mut}},\, v \in \mathcal{V}_i \setminus \{v_i^{(0)}\}\}$
containing at most one element per block. These form a partition matroid with
one part per mutable block and capacity one, which is \emph{not} a cardinality
constraint. The distinction collapses in one regime and bites in the other, and
the two need separating before either can be argued. When every mutable block
offers exactly one alternative, each part contributes one element, every subset
is feasible, and the rank budget $m_{\mathrm{mut}}$ is non-binding; otherwise
the matroid constraint is non-trivial. \Cref{thm:greedy-approx} covers the first case and
\cref{rem:matroid} states what is and is not known about the second.

\begin{remark}[Why no reduction to a cardinality constraint is needed]
\label{rem:no-reduction}
One route to a terminal-quality bound fixes a designated alternative
$\hat v_i$ per mutable block, observes that the resulting ground set
$\hat E = \{(i,\hat v_i)\}$ makes the partition constraint vacuous and the
cardinality budget $|\hat E| = m_{\mathrm{mut}}$ non-binding, and invokes
\citet[Thm.~3]{das2011submodular} on the reduction. That route is available but
pointless, because the same observation that permits it makes its conclusion strictly weaker than the truth. Once the
feasible family is the full power set, the $(1-e^{-\gamma})$ factor is not
tight: \cref{thm:greedy-approx} shows greedy is \emph{exactly} optimal there, by
a three-line argument that uses only $\gamma > 0$ and monotonicity and needs no
external approximation theorem. The reduction also does not extend: it
compares greedy against the best subset of $\hat E$ rather than against the
optimum over $E$, and those differ as soon as a block offers a second
alternative, which is the regime \cref{rem:matroid} addresses.
\end{remark}

\begin{proof}[Proof of \cref{thm:greedy-approx}]
By hypothesis $E = \{(i,\hat v_i) : i \in I_{\mathrm{mut}}\}$ carries exactly one
element per part, so every subset of $E$ is feasible and the cardinality budget
$m_{\mathrm{mut}} = |E|$ is non-binding. Let $A_{\mathrm{gr}}$ be greedy's output. Greedy
halts only when no remaining element has positive marginal gain, so
$g(A_{\mathrm{gr}} \cup \{e\}) - g(A_{\mathrm{gr}}) \le 0$ for every $e \in E \setminus A_{\mathrm{gr}}$; by the
monotonicity hypothesis of \cref{thm:greedy-approx} each such marginal is also
non-negative, hence
\begin{equation}
\label{eq:app-greedy-zero-marginals}
g(A_{\mathrm{gr}} \cup \{e\}) - g(A_{\mathrm{gr}}) \;=\; 0
\qquad \text{for all } e \in E \setminus A_{\mathrm{gr}}.
\end{equation}

Suppose for contradiction that $g(E) > g(A_{\mathrm{gr}})$. Then $L := A_{\mathrm{gr}}$ and
$S := E \setminus A_{\mathrm{gr}}$ are disjoint with $g(L \cup S) = g(E) > g(L)$, so
$(L,S)$ is admissible in the minimum defining $\gamma(g;E)$ in
\eqref{eq:gamma}, and that definition gives
\[
\gamma\,\bigl[g(E) - g(A_{\mathrm{gr}})\bigr]
\;\le\; \sum_{e \in E \setminus A_{\mathrm{gr}}}\bigl[g(A_{\mathrm{gr}} \cup \{e\}) - g(A_{\mathrm{gr}})\bigr]
\;=\; 0
\]
by \eqref{eq:app-greedy-zero-marginals}. Since $\gamma > 0$ this forces
$g(E) \le g(A_{\mathrm{gr}})$, contradicting the supposition. Hence $g(A_{\mathrm{gr}}) = g(E)$, and
$g(E) \ge g(A)$ for every $A \subseteq E$ by monotonicity, so
$g(A_{\mathrm{gr}}) = \max_{A \subseteq E} g(A)$; with
$g(A) = \utility^\star(x_A) - \utility^\star(x^{(0)})$ from
\eqref{eq:gain-function} this is \eqref{eq:greedy-bound}.

The hypothesis $\gamma > 0$ is needed here for the conclusion to hold, not
only to bound its loss: it rules out a set of blocks that is jointly valuable while each
member is individually worthless, which is the only configuration that could
strand greedy at a zero-marginal set. The conclusion fails at
$\gamma = 0$. Take two blocks with
$g(\emptyset) = g(\{a\}) = g(\{b\}) = 0$ and $g(\{a,b\}) = 1$: this is monotone
with $\gamma(g;E) = 0$, greedy halts immediately and returns $0$, and the
optimum is $1$.
\end{proof}

\begin{remark}[The multi-alternative regime is a partition matroid, and is open]
\label{rem:matroid}
\Cref{thm:greedy-approx} assumes one alternative per mutable block, an
idealization no run of \splice{} satisfies, since $\mathcal{V}_i$ holds every
variant the proposer could produce (\cref{def:variants}). The assumption is a
real restriction, and dropping it changes the problem class rather than the
constant: when some $|\mathcal{V}_i| > 2$ the feasible sets form a partition
matroid of rank $m_{\mathrm{mut}}$ that is no longer the full power set of the
ground set, and block-pass greedy must additionally decide \emph{which} variant
to commit in each block, by an inner maximization that no
cardinality-constrained surrogate records.

We know of no terminal-quality guarantee covering deterministic greedy in that
setting. The strongest available result for maximizing a monotone
$\gamma$-weakly-submodular function under a matroid constraint is
$(1+1/\gamma)^{-2}$, achieved by a \emph{randomized} greedy variant
\citep{chen2017weakly}; that work reports it as the first non-trivial guarantee
beyond cardinality constraints and states that the deterministic greedy
algorithm's ratio for such constraints was unknown. Two consequences for how
this paper should be read. First, the relevant factor is \emph{not}
$\gamma/(1+\gamma)$, and the familiar $1/2$ for matroid-constrained
\emph{submodular} greedy does not transfer to the weakly submodular case by
analogy: $(1+1/\gamma)^{-2}$ is $1/4$ at $\gamma = 1$. Second, \splice{}'s
selector is deterministic, so even that result does not apply to it unmodified.
We therefore state no approximation factor in the multi-alternative regime and
treat the terminal quality of deterministic block-pass greedy under a partition
matroid as open. Settling it, or showing that randomization is necessary, would
strengthen \cref{sec:theory} at the point it is currently weakest.
\end{remark}

\begin{proof}[Proof of \cref{prop:beam-beats-greedy}]
The claim is a statement about two algorithms run on shared randomness, so we
first fix the coupling it is made under. At every step, the $k$ draws greedy
makes from its current state $x^{\mathrm{g}}_{t}$ are the same $k$ draws the
beam makes from $x^{\mathrm{g}}_{t}$ whenever $x^{\mathrm{g}}_{t} \in
\mathcal{C}_t$. Scores are the cached validation scores
$\tilde\utility(\cdot;\valid)$ of \cref{sec:main.method}, cached per state and
shared by the coupled runs, so a given state carries the same score in both. Since
$\mathcal{C}_0 = \{x^{(0)}\} = \{x^{\mathrm{g}}_0\}$, $\nu \ge 1$.

\emph{Steps $t < \nu$.} By definition of $\nu$, $x^{\mathrm{g}}_t \in
\mathcal{C}_t$, so $\ell_t = \max_{x\in\mathcal{C}_t}\tilde\utility(x) \ge
\tilde\utility(x^{\mathrm{g}}_t)$.

\emph{Step $t = \nu < \infty$.} Since $x^{\mathrm{g}}_{\nu-1} \in
\mathcal{C}_{\nu-1}$, the coupling puts greedy's $k$ children of
$x^{\mathrm{g}}_{\nu-1}$ into the beam's expansion, or into
$\mathcal{C}_{\nu-1}$ if deduplication removed them as duplicates of beam
members. Greedy's next state is either one of these children or
$x^{\mathrm{g}}_{\nu-1}$ itself, so in every case $x^{\mathrm{g}}_\nu \in
\mathcal{F}_\nu = \mathcal{C}_{\nu-1} \cup \mathcal{N}_\nu$. Elitist selection
retains the top $\bw - D \ge 1$ members of $\mathcal{F}_\nu$ by score, and
$x^{\mathrm{g}}_\nu$ is not retained, so every retained elite scores at least
$\tilde\utility(x^{\mathrm{g}}_\nu)$; in particular $\ell_\nu \ge
\tilde\utility(x^{\mathrm{g}}_\nu)$.

\emph{Steps $t > \nu$.} By \cref{thm:monotone}, $\ell_t \ge \ell_\nu \ge
\tilde\utility(x^{\mathrm{g}}_\nu) = \tilde\utility(x^{\mathrm{g}}_{\min(t,\nu)})$.

For $\bw = 1$, $D = 0$ and $\mathcal{C}_t$ is the single highest-scoring member
of $\{x\} \cup \{\text{children of } x\}$ for the current state $x$, which is
greedy's rule; with a common tie-breaking rule the two runs visit the same
states, so $x^{\mathrm{g}}_t \in \mathcal{C}_t$ for all $t$ and $\nu = \infty$.
\end{proof}

\begin{remark}[The comparison cannot extend past $\nu$]
\label{rem:beam-greedy-after}
After displacement greedy draws from a state the beam no longer holds, so its
children are never scored by the beam and elitism cannot retain them. Take
$\bw = 2$, $D = 0$, $k = 2$ and cached scores. At $t=1$ the seed ($0.50$) has
children scoring $0.60$ and $0.40$; greedy moves to the $0.60$ child and the
beam keeps it and the seed. At $t=2$ the seed draws children scoring $0.90$ and
$0.85$, and the $0.60$ state draws $0.61$ and $0.30$ (shared with greedy);
greedy moves to $0.61$, and the beam keeps $\{0.90, 0.85\}$, so $\nu = 2$. At
$t = 3$ every child of the beam scores $0.10$, while greedy's own draw from the
$0.61$ state scores $0.95$. Then $\ell_3 = 0.90 < 0.95$, while
\eqref{eq:beam-greedy} requires only $\ell_3 \ge 0.61$. Each step has positive
probability under a proposer that conditions on the history, as
\cref{def:proposal} allows.
\end{remark}

\begin{remark}[Scored versus population comparison]
\label{rem:beam-greedy-scored}
The argument is carried out on the cached score $\tilde\utility(\cdot;\valid)$,
which is the quantity both selectors act on, and it is exact there. The
corresponding population statement $\utility^\star(\mathrm{leader}(\mathcal{C}_t)) \ge
\utility^\star(x^{\mathrm{g}}_{\min(t,\nu)})$ does not follow pathwise, because a score
ordering need not agree with the population ordering; it holds up to the
uniform deviation of \cref{thm:uniform-conv}, i.e.\ with
$2\varepsilon$ slack on the event of that theorem. We state the proposition
on the scored objective for this reason.
\end{remark}

\begin{remark}[Why \cref{conj:beam-improvement} resists this argument]
\label{rem:conj}
The proof above compares the beam with greedy only up to the displacement step
$\nu$, and says nothing about how the beam's advantage grows with $\bw$. Showing
that $\alpha_\gamma$ grows with $\bw$ would require the $\bw$ lineages to
contribute independent progress, and the natural
product-of-failure-probabilities argument does not apply: elitist selection
couples the lineages, since which candidates survive depends on the scores of
all of them jointly, and the reserved diversity slots induce further
dependence through the shared quality gate $\tau_t = \gateratio\ell_t$. We therefore
leave \cref{conj:beam-improvement} as a conjecture and rely on
\cref{prop:beam-beats-greedy} wherever the comparison is needed.
\end{remark}

\subsection{Elitist Monotonicity and Noisy No-Regression}
\label{sec:theory:elit}

This subsection characterizes what elitist selection buys over the
children-only beam of ProTeGi. Throughout we use \splice{}'s
\emph{cached-scoring convention} stated at the start of this section: each
state $x$ carries the regularized validation score $\tilde\utility(x;\valid)$
realized when $x$ was first evaluated, not recomputed while it stays in the
beam. This avoids the
selection-on-noise bias that re-rolling would introduce, and its
consequence for monotonicity is direct.

\par\noindent\emph{\Cref{thm:monotone} is stated in the main text (p.~\pageref{thm:monotone}).}\par

We state \cref{thm:monotone} as a lemma rather than a theorem because it is a
consequence of the definitions, not a claim about search dynamics. The proof is two lines: $\mathcal{C}_{t-1} \subseteq
\mathcal{F}_t$ and scores are cached, so a maximum over a superset of a set
whose values have not changed cannot fall. It therefore holds for
\emph{any} elitist selector over \emph{any} proposal distribution, including
an adversarial one. No property of prompt space is used and no experiment
can falsify it.

It is nonetheless necessary: cached scoring is a design decision with a real
cost, and this lemma justifies it. Later arguments use $\ell_t$'s monotonicity as a
hypothesis. \emph{The output rule} (\cref{sec:main.method}) returns
$\mathrm{leader}(\mathcal{C}_T)$, and \cref{thm:monotone} is what makes this the
best-scoring candidate of the run, so the returned prompt never scores below
the seed; without monotonicity the final leader could fall below a candidate the
run had already seen.
\emph{Gate monotonicity} (\cref{prop:adaptive-anneal}): the diversity
gate is $\tau_t = \gateratio\,\ell_t$, so monotone $\ell_t$ is what makes $\tau_t$
non-decreasing and a candidate that falls below it \emph{permanently}
ineligible; if $\ell_t$
could fall the gate reopens and the argument collapses. And \emph{the
convergence rate} (\cref{thm:convergence}) analyzes the leader score as a
monotone accumulator of $\Delta_{\min}$ increments, which is what makes
``improving steps required'' a well-defined finite quantity. And \emph{the
comparison with greedy} (\cref{prop:beam-beats-greedy}) uses it after the
displacement step. One result we do \emph{not}
claim as a dependency: \cref{thm:div-preserve-ext} evaluates its gate at a
single step and never compares $\ell_t$ to $\ell_{t-1}$, so it holds
verbatim under re-evaluation.

The contrast with children-only beam search is still sharp: under
$\mathcal{S}^{\mathrm{beam}}$ (\cref{sec:algorithms-as-selectors}) the previous leader is never retained, so a
step in which all $\bw k$ children score below $\ell_{t-1}$ forces
$\ell_t<\ell_{t-1}$. But \cref{thm:monotone} buys this on the \emph{score
estimate}, and the substantive monotonicity statement, the one that
constrains behavior rather than fixing a definition, is the next result,
which bounds regression of the beam's best \emph{true} score under noise.

\par\noindent\emph{\Cref{thm:noisy-monotone} is stated in the main text (p.~\pageref{thm:noisy-monotone}).}\par

\par\noindent\emph{\Cref{cor:min-batch} is stated in the main text (p.~\pageref{cor:min-batch}).}\par

For $\bw=4$, $k=3$, $\delta=0.05$ and independent batches ($c_\sigma=1$ at
$\sigma^2=1/4$), \cref{cor:min-batch} reads
$n_{\mathrm{re}} \ge \varepsilon^{-2}\ln(320) \approx 5.77\,\varepsilon^{-2}$: $n_{\mathrm{re}} \ge 145$ at
$\varepsilon = 0.2$, and $n_{\mathrm{re}} \ge 5.8\times10^{4}$ at $\varepsilon = 0.01$. A
shared batch doubles both. Read against our per-step training batches
($|B_t| \in \{10, 64, 216\}$), only BBH's
reaches the $\varepsilon = 0.2$ requirement, and a batch of $64$ buys
$\varepsilon \approx 0.30$. So a per-step no-regression guarantee for the
re-evaluation regime is coarse at every batch size we use, which is the
quantitative reason \splice{} does not select on re-evaluated scores. We have
not swept the batch size, so the corollary is a statement about the
alternative design rather than one we test.

\paragraph{The two scoring conventions are a trade-off.}
It would be convenient to read \cref{thm:monotone} as the guarantee and
\cref{thm:noisy-monotone} as a fallback for an ablation, but the two are not
\emph{guaranteed} versus \emph{realistic} but two designs with opposite
failure modes. \emph{Cached scoring} gives deterministic non-regression on
the score \emph{estimate}, but the estimate can be stale: a candidate that
drew a lucky batch keeps its inflated score for the rest of the run, so the
leader may be a noise artifact. The guarantee is exact; the quantity
guaranteed is imperfect. \emph{Re-evaluation} tracks the true objective more
closely, but re-rolls selection-on-noise at every step, so the leader can
regress in truth, and \cref{thm:noisy-monotone} is what bounds that
regression, with \cref{cor:min-batch} naming the batch size that makes it
small. Read together: \emph{exact monotonicity on a possibly-stale estimate,
or probabilistic monotonicity on the true score at a stated batch-size
price.}

The distinction matters when reading the experiments. \splice{} re-runs its
beam on a fresh training batch every step to produce critiques, so a log of
\emph{those} scores can move down from one step to the next without
contradicting \cref{thm:monotone}: the lemma is about the cached selection
score, not about fresh-batch re-evaluations. Score-progression traces should
therefore be read as monotone in the cached score only, not as evidence of a
monotone true-objective curve.

The rate counts progress in steps of a fixed size $\Delta_{\min} > 0$.

\begin{definition}[$\Delta_{\min}$-improving proposal]
\label{def:delta-improving}
A proposal $u \sim q_\phi(\cdot \mid x, H_t)$ is
\emph{$\Delta_{\min}$-improving from $x$} if
$\tilde\utility(\mathrm{Apply}(x,u)) \ge \tilde\utility(x) + \Delta_{\min}$.
\end{definition}

Selection keeps any improvement, so smaller gains occur and are retained; they
never lower the leader, and the definition only decides which proposals the
rate counts. With a $0/1$ metric and $\reg = 0$, every positive gain on a
validation split of $n$ examples is at least $1/n$, so $\Delta_{\min} = 1/n$
makes every improving proposal $\Delta_{\min}$-improving. Setting
$\Delta_{\min}$ to \splice{}'s \texttt{min\_improvement} parameter (default
$0.005$) matches the threshold the algorithm uses for early stopping, but the
analysis does not require that choice.

Before stating the rate we fix the quantity it is stated in, since the
argument accumulates progress in the scored objective and the conclusion a
reader wants is about the population one. Every ingredient of the proof lives in the \emph{scored}
objective $\tilde\utility$ of \cref{sec:main.method}:
\cref{def:delta-improving} measures improvement in $\tilde\utility$, and
\cref{thm:monotone} makes $\ell_t = \max_{\mathcal{C}_t}\tilde\utility$
non-decreasing. We therefore state the rate on $\tilde\utility$ and pay the
transfer to $\utility^\star$ explicitly in \cref{cor:convergence-population},
rather than assert a population conclusion the accumulation does not deliver.
Write $\tilde x^\star \in \argmax_{x\in\mathcal{X}}\tilde\utility(x)$.

\begin{assumption}[Scored coverage]
\label{asm:scored-coverage}
Fix a tolerance $\varepsilon \ge \Delta_{\min}$, and let
\begin{equation}
\label{eq:pstar-scored}
\tilde p^\star(\varepsilon) \;:=\;
\inf_{H_t}\;
\min_{\substack{x \,:\; \tilde\utility(\tilde x^\star) - \tilde\utility(x) \,\ge\, \varepsilon}}\;
\Pr_{u \sim q_\phi(\cdot \mid x, H_t)}\bigl[u \text{ is } \Delta_{\min}\text{-improving from } x\bigr],
\end{equation}
the minimum over states whose score is still $\varepsilon$ below the best,
taken uniformly over histories $H_t$, for the realized validation split. We assume
$\tilde p^\star(\varepsilon) > 0$ at this one $\varepsilon$.
\end{assumption}

This is \cref{asm:proposal-coverage}'s minimum-competence condition read on
the sublevel set of the objective the algorithm can see, with an improvement
counted only when it reaches $\Delta_{\min}$. The two conditions range over
different sets of states, since validation noise moves states into and out of
the sublevel set, and neither implies the other. \cref{thm:reachability} needs
the population form, because it reasons about the population suboptimality of
parents; the rate below needs this one, because the leader is chosen by score.

\par\noindent\emph{\Cref{thm:convergence} is stated in the main text (p.~\pageref{thm:convergence}).}\par

\begin{corollary}[Population form, with the transfer priced]
\label{cor:convergence-population}
In the setting of \cref{thm:convergence}, under
\cref{asm:bounded-reward,asm:proposal-distribution} and with a validation
split of $n = |\valid|$ i.i.d.\ examples, let $\varepsilon_n$ satisfy
$2(|\mathcal{B}_T|+1)\exp(-2n\varepsilon_n^2) \le \delta/2$, with
$\mathcal{B}_T$ the data-independent ball of \cref{thm:uniform-conv}. The
quantities $\tilde p^\star$, $K$ and $p^{\mathrm{adv}}$ depend on the realized
validation split, and so does the horizon $T_{\mathrm{req}} := (2K +
8\ln(2/\delta))/p^{\mathrm{adv}}(\varepsilon)$. Then with probability at least
$1-\delta$, either \cref{asm:scored-coverage} fails at $\varepsilon$, or
$T < T_{\mathrm{req}}$, or
\begin{equation}
\label{eq:convergence-population}
\utility^\star(\mathrm{leader}(\mathcal{C}_T)) \;\ge\;
\utility^\star(x^\star) \;-\; \varepsilon \;-\; 2\varepsilon_n
\;-\; \textstyle\sum_j \lambda_j R_j(x^\star).
\end{equation}
\end{corollary}

Three of the four terms on the right of \eqref{eq:convergence-population} are
the price of reporting a population guarantee for a procedure that
optimizes a regularized finite-sample score, and each is separately
controllable: $\varepsilon$ by the horizon, $\varepsilon_n$ by the validation
size, $\sum_j \lambda_j R_j(x^\star)$ by the regularization weights (for a
single penalty, the term $\reg\,\cost(x^\star)$ that \cref{thm:reg-bias}
isolates). The first two pull against each other: a longer horizon
drives the $\varepsilon$ requirement down but enlarges $\mathcal{B}_T$, and
$\varepsilon_n$ grows as $\sqrt{T\log(\Delta_{\mathcal{G}}+1)/n}$, so at a fixed validation
size the population guarantee does not improve indefinitely with search. What no choice of $T$ removes is the gap between
$x^\star$ and $x^\star_{\mathrm{all}}$, which is why
\eqref{eq:xstar-convention} fixes the target at the best feasible prompt and
\cref{prop:mut-price} prices the rest.

One scope note. \splice{} stops early once the leader has gained less than
$\Delta_{\min}$ (\texttt{min\_improvement}) over the last $T_{\mathrm{pat}}$
steps (five or ten in our runs), so a run can terminate before the horizon
\cref{thm:convergence} asks for. The bound therefore describes the
unstopped process; early stopping trades the guarantee for budget exactly when
$p^{\mathrm{adv}}$ is small enough to make the horizon impractical, and
\cref{thm:monotone} is what makes that trade safe: the run exits with its
monotone best-so-far rather than with a regression.

Two features of \cref{thm:convergence} matter. The exponent is $k$,
\emph{not} $\bw k$: \cref{thm:reachability} guarantees a child that beats
\emph{its own parent}, whereas advancing the leader needs a child that beats
\emph{the leader}, and only the leader's own $k$ draws are guaranteed to do
that. The rate is therefore conservative, since a child of another lineage can
also overtake the incumbent. And the dependence on $\delta$ is
additive with explicit constants rather than an unspecified multiplicative
$C\log(1/\delta)$, which is both sharper and checkable.

The bound has three empirical handles: the numerator scales with the
initial gap, the denominator grows with $k$ alone, and $\Delta_{\min}$
enters linearly. The $k$-dependence is the one the rate predicts. Because coverage enters
through $\tilde p^\star(\varepsilon)$, the $\varepsilon$-dependence is explicit on
both sides: tightening the target (smaller $\varepsilon$) enlarges the
numerator's $(\tilde h_0-\varepsilon)$ term and, because it enlarges the
sublevel set over which $\tilde p^\star(\varepsilon)$ is a minimum, can only
lower $p^{\mathrm{adv}}(\varepsilon)$, so $T$ grows on both counts. For small $\tilde p^\star$,
$p^{\mathrm{adv}} \approx k\, \tilde p^\star$, giving $T \propto 1/(k\,
\tilde p^\star)$, linear in $1/\tilde p^\star$ with per-parent fan-out, not beam width, as
the practitioner's compensating lever. When $p^{\mathrm{adv}}$ becomes very
small the bound stays true and reports that the remaining gap is no longer
reachable, and early stopping then ends the run with the leader it has.

\paragraph{Proofs.}
\label{sec:app.theory:elit}

\begin{proof}[Proof of \cref{thm:monotone}]
Let $x^{\mathrm{L}}_{t-1} \in \argmax_{x \in \mathcal{C}_{t-1}}
\tilde\utility(x;\valid)$, so $\tilde\utility(x^{\mathrm{L}}_{t-1}) = \ell_{t-1}$.
Three facts combine.

\emph{(i) Retention in the frontier.} By \eqref{eq:bips-body} the frontier is
$\mathcal{F}_t = \mathcal{C}_{t-1} \cup \mathcal{N}_t \supseteq
\mathcal{C}_{t-1}$, so $x^{\mathrm{L}}_{t-1} \in \mathcal{F}_t$. This is the step
that fails for children-only beam search, where
$\mathcal{F}_t = \mathcal{N}_t$.

\emph{(ii) Score stability.} Under the cached-scoring convention,
$\tilde\utility(x^{\mathrm{L}}_{t-1};\valid)$ is the value realized when
$x^{\mathrm{L}}_{t-1}$ was first evaluated and is not recomputed on retention. Hence
its score in round $t$ equals $\ell_{t-1}$ exactly.

\emph{(iii) Elitist selection.} $\mathcal{S}^{\mathrm{elit}\text{-}\mathrm{div}}$
fills $\mathcal{E}_t$ with the top $\bw - D$ elements of $\mathcal{F}_t$ by
score. Either $x^{\mathrm{L}}_{t-1} \in \mathcal{E}_t$, in which case
$\ell_t \ge \tilde\utility(x^{\mathrm{L}}_{t-1}) = \ell_{t-1}$; or it is excluded,
which by the definition of ``top $\bw-D$ by score'' requires at least
$\bw - D$ elements of $\mathcal{F}_t$ with score at least
$\tilde\utility(x^{\mathrm{L}}_{t-1})$, all of which are in $\mathcal{E}_t \subseteq
\mathcal{C}_t$, so again $\ell_t \ge \ell_{t-1}$.

In both cases $\ell_t \ge \ell_{t-1}$, and $t$ was arbitrary.
\end{proof}

\begin{remark}
\label{rem:monotone-scope}
The argument uses only parent retention~(i) and score
stability~(ii), and no property of $q_\phi$ or of the landscape, which is why
\cref{thm:monotone} holds for any elitist selector over any proposal
distribution, including an adversarial one, and why it is stated as a lemma in
the body. It bounds the trajectory of the score \emph{estimate}; the
corresponding statement about the true objective is
\cref{thm:noisy-monotone}.
\end{remark}

\begin{proof}[Proof of \cref{thm:noisy-monotone}]
Here every member of $\mathcal{F}_t$ is re-evaluated at step $t$ on $n_{\mathrm{re}}$ fresh
examples and selection acts on $\hat\utility_{n_{\mathrm{re}}}$, so the cached-scoring
argument above is unavailable. Write $f^\star_t = \max_{x \in
\mathcal{C}_t}\utility^\star(x)$ and let $\bar x_{t-1} \in
\argmax_{x\in\mathcal{C}_{t-1}}\utility^\star(x)$. Condition on everything up
to the drawing of the step-$t$ evaluation examples. The frontier
$\mathcal{F}_t$ is then fixed (its parents were chosen at step $t-1$ and its
children by the proposer), and $\bar x_{t-1} \in \mathcal{F}_t$ is fixed
with it; only the step-$t$ estimates remain random.

Suppose the regression event $\{f^\star_t < f^\star_{t-1} - \varepsilon\}$
occurs. Then $\bar x_{t-1}$ was not retained. The selector is elitist, so the
top-scoring member $x'$ of $\mathcal{F}_t$ is retained, and since
$\bar x_{t-1} \in \mathcal{F}_t$ was not, $x' \ne \bar x_{t-1}$ and
$\hat\utility_{n_{\mathrm{re}}}(x') \ge \hat\utility_{n_{\mathrm{re}}}(\bar x_{t-1})$; and every retained candidate has $\utility^\star(x') \le
f^\star_t < f^\star_{t-1} - \varepsilon$. Hence there is
$x' \in \mathcal{F}_t \setminus \{\bar x_{t-1}\}$ with
\begin{equation}
\label{eq:app-regress-gap}
\utility^\star(\bar x_{t-1}) - \utility^\star(x') \;>\; \varepsilon
\qquad\text{and}\qquad
\hat\utility_{n_{\mathrm{re}}}(x') \;\ge\; \hat\utility_{n_{\mathrm{re}}}(\bar x_{t-1}).
\end{equation}
Writing $\eta(x) := \hat\utility_{n_{\mathrm{re}}}(x) - \utility^\star(x)$, this gives
$\eta(x') - \eta(\bar x_{t-1}) > \varepsilon$ for that $x'$.

Fix any $x' \in \mathcal{F}_t \setminus \{\bar x_{t-1}\}$ with population gap
above $\varepsilon$. If the two candidates are scored on independent batches,
each $\eta$ is mean-zero sub-Gaussian with proxy $\sigma^2/n_{\mathrm{re}}$
(\cref{lem:hoeffding}), their difference has proxy $2\sigma^2/n_{\mathrm{re}}$
(\cref{lem:subg-diff}), and
$\Pr[\eta(x') - \eta(\bar x_{t-1}) > \varepsilon] \le
\exp(-n_{\mathrm{re}}\varepsilon^2/4\sigma^2)$. If they share a batch, the difference is a
mean of $n_{\mathrm{re}}$ i.i.d.\ per-example differences in $[-1,1]$, and Hoeffding's
inequality gives $\exp(-2n_{\mathrm{re}}\varepsilon^2/2^2) = \exp(-n_{\mathrm{re}}\varepsilon^2/2)$. Both
are $\exp(-n_{\mathrm{re}}\varepsilon^2/c_\sigma)$ for the stated $c_\sigma$.

Because $\mathcal{F}_t$ is fixed given the conditioning, a union bound over its
at most $|\mathcal{F}_t| - 1$ members other than $\bar x_{t-1}$ is valid and
gives \cref{thm:noisy-monotone}. Taking the union over the retained beam
$\mathcal{C}_t$ instead, as a smaller prefactor would suggest, would not be:
$\mathcal{C}_t$ is chosen by the very estimates being bounded
(\cref{rem:adaptive-union}). Finally $|\mathcal{F}_t| \le
|\mathcal{C}_{t-1}| + k|\mathcal{C}_{t-1}| \le \bw(1+k)$, and integrating over
the conditioning preserves the bound.
\end{proof}

\begin{proof}[Proof of \cref{cor:min-batch}]
Impose $\bw(1+k)\exp(-n_{\mathrm{re}}\varepsilon^2/c_\sigma) \le \delta$, which implies the
conclusion of the corollary through \cref{thm:noisy-monotone}, since $|\mathcal{F}_t| - 1 < \bw(1+k)$. Taking
logarithms gives $n_{\mathrm{re}} \ge (c_\sigma/\varepsilon^2)\ln(\bw(1+k)/\delta)$, which is
\cref{cor:min-batch}. With $\bw=4$, $k=3$, $\delta=0.05$, $c_\sigma=1$: $n_{\mathrm{re}} \ge
\varepsilon^{-2}\ln 320$, i.e.\ $145$ at $\varepsilon=0.2$ and
$5.77\times10^{4}$ at $\varepsilon=0.01$.
\end{proof}

\begin{proof}[Proof of \cref{thm:convergence}]
Let $\tilde h_0 = \tilde\utility(\tilde x^\star) - \tilde\utility(x^{(0)})$ and
$K := \lceil (\tilde h_0 - \varepsilon)/\Delta_{\min}\rceil$. Every quantity in
this proof is the scored objective $\tilde\utility$; the transfer to
$\utility^\star$ is \cref{cor:convergence-population}. If $K \le 0$ then
$\tilde h_0 \le \varepsilon$, so the target already holds at $t = 0$ and, by
\cref{thm:monotone}, at every $T$; we assume $K \ge 1$ below, as
\cref{lem:chernoff} requires.

\emph{Step 1: a bounded number of advances suffices.} Call step $t$ an
\emph{advance} if $\ell_t \ge \ell_{t-1} + \Delta_{\min}$. By
\cref{thm:monotone} the leader score is non-decreasing, so it never gives
back progress and the advances accumulate. Each advance raises the leader by
at least $\Delta_{\min}$, so after $K$ advances the leader has gained at least
$K\Delta_{\min} \ge \tilde h_0 - \varepsilon$, leaving
$\tilde\utility(\tilde x^\star) - \ell_t \le \varepsilon$; that is, $K$
advances suffice for $\varepsilon$-optimality in $\tilde\utility$. Note that
$K$ is a ratio of two quantities in the same units, which is what makes it the
right count; dividing a population gap by a scored increment would not be.

\emph{Step 2: each step advances with probability at least
$p^{\mathrm{adv}}$.} Let
$t_\varepsilon := \min\{t : \tilde\utility(\tilde x^\star) - \ell_t \le \varepsilon\}$ be
the first $\varepsilon$-optimal step, and fix $t \le t_\varepsilon$. Let
$x^{\mathrm{L}}_{t-1}$ denote the current leader; since $t \le t_\varepsilon$ it
satisfies $\tilde\utility(\tilde x^\star) - \tilde\utility(x^{\mathrm{L}}_{t-1})
\ge \varepsilon$, so it lies in the sublevel set of \eqref{eq:pstar-scored}. The
leader contributes $k$ of the expansion's draws. Each is
$\Delta_{\min}$-improving from $x^{\mathrm{L}}_{t-1}$ (\cref{def:delta-improving})
with probability at least $\tilde p^\star(\varepsilon)$
by \eqref{eq:pstar-scored}, and by
\eqref{eq:bips-body} they are i.i.d.\ given the history, so
\[
\Pr\bigl[\text{some child of the leader beats it by at least } \Delta_{\min}\bigr]
\;\ge\; 1 - \bigl(1-\tilde p^\star(\varepsilon)\bigr)^{k} \;=:\; p^{\mathrm{adv}}(\varepsilon).
\]
Any such child lies in $\mathcal{F}_t$ and scores at least
$\ell_{t-1} + \Delta_{\min}$, so by the elitist argument of \cref{thm:monotone}
the new leader satisfies
$\ell_t \ge \ell_{t-1} + \Delta_{\min}$: the step is an advance. The bound
holds conditionally on the entire past, since $\tilde p^\star(\varepsilon)$ is a
uniform lower bound over states in the sublevel set and over histories.

\emph{Step 3: Chernoff, with the indicators patched after the target is met.}
Let $I_t$ indicate that step $t$ is an advance. Step 2 bounds its conditional
success probability only for $t \le t_\varepsilon$, so set
$I'_t := I_t\,\1\{t \le t_\varepsilon\} + \1\{t > t_\varepsilon\}$. The event
$\{t \le t_\varepsilon\}$ is fixed by the past at step $t$, since it says the target was
not met by step $t-1$, so every $I'_t$ has conditional success probability at
least $p^{\mathrm{adv}}(\varepsilon)$ given the past, and \cref{lem:chernoff}
with $p = p^{\mathrm{adv}}(\varepsilon)$ applies to $N'_T := \sum_{t \le T} I'_t$:
\[
T \;\ge\; \frac{2K + 8\ln(1/\delta)}{p^{\mathrm{adv}}(\varepsilon)}
\quad\Longrightarrow\quad \Pr[N'_T < K] \;\le\; \delta .
\]
On $\{t_\varepsilon > T\}$ every $I'_t$ equals $I_t$, so $N'_T$ is the number of
advances, which is below $K$ by Step 1. Hence $\Pr[t_\varepsilon > T] \le
\Pr[N'_T < K] \le \delta$. On $\{t_\varepsilon \le T\}$ the target is met at $t_\varepsilon$ and
kept to $T$ by \cref{thm:monotone}, since $\ell_T \ge \ell_{t_\varepsilon}$. So
$\tilde\utility(\mathrm{leader}(\mathcal{C}_T)) \ge \tilde\utility(\tilde x^\star) -
\varepsilon$ with probability at least $1-\delta$, which is
\cref{thm:convergence}.
\end{proof}

\begin{proof}[Proof of \cref{cor:convergence-population}]
Write $\hat x := \mathrm{leader}(\mathcal{C}_T)$ and $n := |\valid|$. Apply \cref{thm:uniform-conv}(i) to the set
$\mathcal{U} := \mathcal{B}_T \cup \{x^\star\}$. It is fixed independently of
$\valid$: $\mathcal{B}_T$ is determined by the edit graph and the seed, and
$x^\star$ by $\utility^\star$ and $\mathcal{X}$. In the setting of
\cref{thm:convergence} scores are cached per state, so each member of
$\mathcal{U}$ carries one score and the factor $T+1$ of
\cref{thm:uniform-conv}(ii) does not arise. The set contains $\hat x$, since
$\hat x \in \mathcal{H}_T \subseteq \mathcal{B}_T$ by
\cref{thm:uniform-conv}(ii). Using the history itself as the union set would
not be valid, for the reason \cref{rem:adaptive-union} gives. By hypothesis
$2(|\mathcal{B}_T|+1)\exp(-2n\varepsilon_n^2) \le \delta/2$, so with
probability at least $1-\delta/2$,
\begin{equation}
\label{eq:app-conv-unif}
|\hat\utility_n(x) - \utility^\star(x)| \;\le\; \varepsilon_n
\qquad\text{for every } x \in \mathcal{B}_T \cup \{x^\star\}.
\end{equation}
Conditional on the validation split and the scores, \cref{thm:convergence} at
confidence $\delta/2$, which is what the $8\ln(2/\delta)$ term provides,
applies whenever \cref{asm:scored-coverage} holds and $T \ge T_{\mathrm{req}}$.
So the event that both hold and the conclusion of \cref{thm:convergence} fails
has probability at most $\delta/2$, and with the failure of
\eqref{eq:app-conv-unif} the total is at most $\delta$. The hypothesis and
$T_{\mathrm{req}}$ depend on the split, which is why the corollary is stated as
this joint bound rather than conditionally on the hypothesis. Outside the two
failure events, either the hypothesis fails, or $T < T_{\mathrm{req}}$, or three
inequalities compose. First, since every $R_j \ge 0$ and $\hat x \in \mathcal{B}_T$,
\[
\utility^\star(\hat x) \;\ge\; \hat\utility_n(\hat x) - \varepsilon_n
\;=\; \tilde\utility(\hat x) + \textstyle\sum_j\lambda_jR_j(\hat x) - \varepsilon_n
\;\ge\; \tilde\utility(\hat x) - \varepsilon_n .
\]
Second, \cref{thm:convergence} gives $\tilde\utility(\hat x) \ge
\tilde\utility(\tilde x^\star) - \varepsilon$, and $\tilde x^\star$ maximizes
$\tilde\utility$ over $\mathcal{X} \ni x^\star$, so
$\tilde\utility(\hat x) \ge \tilde\utility(x^\star) - \varepsilon$. Third,
\[
\tilde\utility(x^\star) \;=\; \hat\utility_n(x^\star) - \textstyle\sum_j\lambda_jR_j(x^\star)
\;\ge\; \utility^\star(x^\star) - \varepsilon_n - \textstyle\sum_j\lambda_jR_j(x^\star)
\]
by \eqref{eq:app-conv-unif}. Chaining the three gives
\eqref{eq:convergence-population}.
\end{proof}

\begin{remark}[Which term is irreducible]
\label{rem:conv-transfer}
Of the three penalties in \eqref{eq:convergence-population}, $\varepsilon$
shrinks with the horizon, $\varepsilon_n$ with the validation size, and
$\sum_j\lambda_jR_j(x^\star)$ with the regularization weights, so each can be driven down
by a choice the practitioner controls. The quantity that no horizon reduces is
the one \eqref{eq:convergence-population} does \emph{not} contain: the gap
between $x^\star$ and $x^\star_{\mathrm{all}}$. Stating the corollary against
$x^\star$ rather than $x^\star_{\mathrm{all}}$ is therefore not a weakening for
convenience: against $x^\star_{\mathrm{all}}$ the conclusion would be
unreachable at every $T$ whenever freezing costs more than $\varepsilon$, and
\eqref{eq:pstar-scored}'s sublevel set would contain states with no improving
neighbor, making the hypothesis false as well. \Cref{prop:mut-price} is where
that gap is addressed, and it is addressed as a price rather than as a rate.
\end{remark}

\begin{remark}[Why the rate carries $k$ and not $\bw k$]
\label{rem:adv-rate}
Step 2 uses only the $k$ draws taken from the leader, because those are the
ones whose improvement is guaranteed to move $\ell_t$: a child improving on a
non-leader parent need not exceed $\ell_{t-1}$
(\cref{rem:reach-vs-advance}). The resulting rate
$p^{\mathrm{adv}} = 1-(1-\tilde p^\star)^{k}$ is therefore conservative:
children of other lineages can also exceed the incumbent leader, and nothing in
the bound credits them. It should not be replaced by the reachability rate
$p^{\mathrm{step}} = 1-(1-p^\star)^{|\mathcal{C}_{t-1}|k}$ of
\cref{thm:reachability}, which bounds a different event (some child beats its
own parent) under a different coverage condition; using it would overstate
what reachability supports, so we keep both quantities and label which is which, in the theory
and in \cref{sec:exp.pstar}.
\end{remark}

\subsection{Diversity Preservation and the Quality Gate}
\label{sec:theory:div}

Monotonicity describes the leader score but says nothing about the beam's
\emph{lineage structure}. Under pure elitism ($D=0$) the leader climbs while
all beam members may become descendants of one early winner (beam
collapse).

\begin{definition}[Beam collapse]
\label{def:collapse}
Let $\mathrm{anc}(x) := \lineage(x) \setminus \{\mathrm{id}(x)\}$ denote the
proper ancestors of $x$, and
$\mathrm{anc}^\circ(x) := \mathrm{anc}(x) \setminus \{\mathrm{id}(x^{(0)})\}$
those other than the seed. A beam $\mathcal{C}$ with at least two members is
\emph{collapsed} if every pair of distinct members shares an ancestor other
than the seed, that is, if no $x \ne x'$ in $\mathcal{C}$ have
$\mathrm{anc}^\circ(x) \cap \mathrm{anc}^\circ(x') = \emptyset$.
\end{definition}

The seed has to be excluded, because every candidate after the first step
descends from it. That is also the subtlety in \cref{alg:pickdiverse}: its
Tier~1 and Tier~2 tests check disjointness of $\mathrm{anc}$, which
\emph{includes} the seed, so a candidate other than the seed can pass them only
while the seed is the sole elite. The selector we ran therefore enforces diversity
through its third tier (a distinct immediate parent) and through the
backfill of \cref{alg:elit-div}, and the guarantee below is stated for that
mechanism. A selector that applied the same tests to $\mathrm{anc}^\circ$ would
enforce the stronger, lineage-level property; we state what it would guarantee
but did not run it.

\begin{theorem}[Diversity preservation]
\label{thm:div-preserve-ext}
Part~(i) is \cref{thm:div-preserve}. Let $1 \le D \le \bw - 1$, $\gateratio\in[0,1]$, and let $\mathcal{E}_t$ be the elite slots of
\cref{alg:elit-div}.
\begin{enumerate}[label=(\roman*),nosep,leftmargin=*]
\item \emph{(As run.)} If $\mathcal{F}_t$ contains $c^{\star}$ with
$\tilde\utility(c^{\star}) \ge \gateratio\cdot\ell_t$ whose parent is not the parent of
any elite, then $\mathcal{C}_t$ contains a member scoring at least $\gateratio\cdot\ell_t$
whose parent is not the parent of any elite.
\item \emph{(Lineage-level variant.)} If the Tier~1--2 tests of
\cref{alg:pickdiverse} use $\mathrm{anc}^\circ$ in place of $\mathrm{anc}$, and
$\mathcal{F}_t$ contains $c^{\star}$ with $\tilde\utility(c^{\star}) \ge
\gateratio\cdot\ell_t$ and $\mathrm{anc}^\circ(c^{\star})\cap\mathrm{anc}^\circ(e) =
\emptyset$ for every elite $e$, then $\mathcal{C}_t$ is not collapsed whenever
it has at least two members.
\end{enumerate}
\end{theorem}

The guarantee is \emph{conditional}: the selector retains a distinct-parent
(or, in the variant, lineage-disjoint) competitive candidate \emph{when one
exists}; generating one is the proposer's job. Part (i) is also weak by
design. Two members with different parents may still share a recent
ancestor, so it rules out the beam narrowing to one parent's children, not to
one lineage. That is the property the $D$-sweep of \cref{sec:main.exp}
exercises, and its effect there is large; whether the lineage-level variant
would add to it is untested.

\par\noindent\emph{\Cref{prop:adaptive-anneal} is stated in the main text (p.~\pageref{prop:adaptive-anneal}).}\par

\cref{prop:adaptive-anneal} is what an annealing reading of the gate rests on,
and it is all that is proved: the gate rises with the leader, so a candidate
that falls below it is never readmitted. The reading itself is that early on
$\ell_t$ is low and many candidates qualify for the diversity slots, while late
the gate admits only competitive ones, a shift from exploration to exploitation
with no temperature parameter. That shift is not a theorem, because eligibility
compares a candidate with the leader, $\tilde\utility(c) \ge \gateratio\,\ell_t$, and
whether fewer candidates qualify later depends on how new children score
relative to a rising leader, which the selector does not control; children whose
scores rise with the leader stay eligible. The exclusion is per candidate, not
per lineage: a lineage whose current members fall below the gate can re-enter
through a new, higher-scoring child. With $\gateratio=0$ the gate is inactive and the
proposition is empty; \cref{sec:exp.setup} records which runs set $\gateratio>0$.
We do not report a lineage-count trajectory or a $\gateratio$-sweep, so the schedule is
neither proved nor measured.

\paragraph{Proofs.}
\label{sec:app.theory:div}

\begin{proof}[Proof of \cref{thm:div-preserve-ext}]
Write $\mathcal{E}_t$ for the elites, $P = \{\mathrm{parent}(e) : e \in
\mathcal{E}_t\}$ for their parents, and $\mathcal{Q}_t = \mathcal{F}_t
\setminus \mathcal{E}_t$ for the pool \textsc{PickDiverse}
(\cref{alg:pickdiverse}) scans in descending score order. Since $D \le \bw-1$,
$\mathcal{E}_t \ne \emptyset$, and since the elites are the top of the
frontier, $\ell_t = \max_{\mathcal{E}_t}\tilde\utility$, so the gate is
$\tau_t = \gateratio\ell_t$. If \cref{alg:elit-div} takes its degenerate branch
($|\mathcal{F}_t| \le \bw$ or $\mathcal{N}_t = \emptyset$) it returns the top
$\min(\bw, |\mathcal{F}_t|)$ of the frontier; in the first case that is all of
$\mathcal{F}_t$, which contains $c^\star$ and an elite and so the claim holds
directly, and we assume the main branch below.

\emph{Part (i).} $c^\star \notin \mathcal{E}_t$, since an elite's parent lies in
$P$ by definition. So $c^\star \in \mathcal{Q}_t$, and because the scan is in
descending score order and $\tilde\utility(c^\star) \ge \tau_t$, no
\textbf{break} occurs before $c^\star$ is reached. If Tier~1 or Tier~2 returns
some $c'$, then $\mathrm{anc}(c') \cap \Lambda = \emptyset$. Every non-seed
elite $e$ has $\mathrm{parent}(e) \in \mathrm{anc}(e) \subseteq \Lambda$. So if
$c'$ is not the seed, $\mathrm{parent}(c') \in \mathrm{anc}(c')$ differs from
the parent of every non-seed elite, and also from the seed's undefined parent,
which lies in $P$ only when the seed is an elite (for instance its sole elite).
If $c'$ is the seed, the seed is not an elite and $P$ holds no undefined parent.
Either way $\mathrm{parent}(c') \notin P$. Otherwise Tier~3 reaches
$c^\star$, which satisfies $\mathrm{parent}(c^\star) \notin P$, so it returns
$c^\star$ or an earlier $c''$ with $\mathrm{parent}(c'') \notin P$. In every
case the first diversity pick has a parent outside $P$ and, since the scan stops
at the gate, a score of at least $\tau_t = \gateratio\ell_t$.

\emph{Part (ii).} If $c^\star \in \mathcal{E}_t$, the hypothesis applied to
$e = c^\star$ gives $\mathrm{anc}^\circ(c^\star) = \emptyset$, so $c^\star$ has
disjoint non-seed ancestry from every other member of $\mathcal{C}_t$, and the
beam is not collapsed as soon as it has a second member, which the statement
assumes. (The selector returns a single candidate only when every non-elite
candidate falls below the gate and $\bw - D = 1$.) Otherwise
$c^\star \in \mathcal{Q}_t$, it clears the gate, and it passes Tier~2 on
$\mathrm{anc}^\circ$, so the scan returns $c^\star$ or an earlier pick passing
Tier~1 or Tier~2, either of which has $\mathrm{anc}^\circ$ disjoint from every
elite's. That pick and any elite form the required pair, so $\mathcal{C}_t$ is
not collapsed in the sense of \cref{def:collapse}.
\end{proof}

\begin{remark}[The guarantee is conditional, and on the proposer]
\label{rem:div-conditional}
Both hypotheses are about the \emph{frontier}, not the selector: the theorem
says that a competitive diverse candidate is retained \emph{when one exists}.
If every proposal descends from the current leader, no such $c^\star$ exists
and the statement is vacuous; generating diverse candidates is the
proposer's responsibility. Two further points about what ran. The backfill
step of \cref{alg:elit-div} fills any slot \textsc{PickDiverse} leaves empty
with the best remaining candidate above the gate, so a $D$-slot arm is not
``$\bw-D$ elites plus $D$ diversity picks'' but at most that many diversity
picks with the rest chosen by score. And the theorem gives one diverse pair,
not $D+1$: a pairwise-diversity claim for $D\ge2$ would need a hypothesis on
the frontier at every slot, which we do not state.
\end{remark}

\begin{proof}[Proof of \cref{prop:adaptive-anneal}]
By \cref{thm:monotone}, $\ell_t \ge \ell_{t-1}$; since $\gateratio > 0$, multiplying
through gives $\tau_t = \gateratio\ell_t \ge \gateratio\ell_{t-1} = \tau_{t-1}$, so the gate
sequence is non-decreasing.

For the exclusion claim, suppose $\tilde\utility(x) < \tau_t$ for some
candidate $x$ and step $t$. With scores cached per state $\tilde\utility(x)$ is a fixed
number, unchanged at every later step, while $\tau_s \ge \tau_t$ for all
$s \ge t$ by the first part. Hence $\tilde\utility(x) < \tau_s$, and $x$ fails
the gate of \cref{alg:pickdiverse} at every $s \ge t$. The claim is about
states only: new children of the same lineage are new states with new scores
and may clear the gate. Without the per-state cache the claim holds per
evaluation: a state re-proposed after leaving the beam is scored afresh and
may clear the gate. Caching is essential: under re-evaluation
$\tilde\utility(x)$ is re-drawn each step and a candidate below the gate can
re-enter, which is the regime \cref{thm:noisy-monotone} governs.
\end{proof}

\subsection{Generalization, Regularization, and the Price of Freezing}
\label{sec:theory:gen-reg}

Three results close the analysis: a uniform-convergence bound relating
empirical and population scores across the search
(\cref{thm:uniform-conv}), the bias of length regularization
(\cref{thm:reg-bias}), and the approximation penalty of restricting search
to mutable blocks (\cref{prop:mut-price}).

\begin{theorem}[Uniform convergence over the search history]
\label{thm:uniform-conv}
Let $\mathcal{H}_T := \bigcup_{t=0}^{T} (\mathcal{C}_t \cup \mathcal{N}_t)$ be
the evaluated history, let
$\mathcal{B}_T := \{x \in \mathcal{X} : d_{\mathcal{G}}(x^{(0)}, x) \le T\}$ be
the radius-$T$ ball of the edit graph around the seed, and let $\Delta_{\mathcal{G}}$ be the
maximum degree of $\mathcal{G}$, so that
$|\mathcal{B}_T| \le \min\{|\mathcal{X}|,\,(\Delta_{\mathcal{G}}+1)^T\}$. Under
\cref{asm:bounded-reward} and a validation split $\valid$ of $n$ i.i.d.\ examples:
\begin{enumerate}[label=(\roman*),nosep,leftmargin=*]
\item for any set $\mathcal{U} \subseteq \mathcal{X}$ fixed independently of
$\valid$,
\begin{equation}
\label{eq:uniform-conv}
\Pr\!\left[\,\sup_{x \in \mathcal{U}} |\hat{\utility}_n(x) - \utility^\star(x)| > \varepsilon\,\right]
\;\le\; 2\,|\mathcal{U}|\,\exp(-2 n \varepsilon^2);
\end{equation}
\item under \cref{asm:proposal-distribution}, $\mathcal{H}_T \subseteq
\mathcal{B}_T$ on every trajectory. Let $\mathcal{L}_T$ be the set of all
evaluations the run makes up to step $T$, each a state $x$ with the score
$\hat\utility_n$ it received. Deduplication allows at most one evaluation of a
state per step, so no state is evaluated more than $T+1$ times, and
\begin{equation}
\label{eq:uniform-conv-adaptive}
\Pr\!\left[\,\sup_{(x,\hat\utility_n) \in \mathcal{L}_T} |\hat{\utility}_n - \utility^\star(x)| > \varepsilon\,\right]
\;\le\; 2\,(T+1)\,|\mathcal{B}_T|\,\exp(-2 n \varepsilon^2).
\end{equation}
If every evaluation of a state returns the same score (a deterministic system
agent, or scores cached per state), the factor $T+1$ can be dropped.
\end{enumerate}
\end{theorem}

The two parts differ in exactly the way that matters here. The history
$\mathcal{H}_T$ is \emph{not} fixed independently of $\valid$: validation scores
decide which parents survive, and the proposer conditions on those scores
(\cref{sec:main.method}), so which candidates exist at all depends on the selection
split. A union bound over a set chosen with the data does not control the
deviation on that set. If many candidates of equal population utility are
scored and only the luckiest are expanded, the supremum over what was kept is
the maximum of many deviations, not of a few. So part (i) applied to
$\mathcal{H}_T$ as if it were fixed (which gives a half-width scaling as
$\sqrt{\log|\mathcal{H}_T|}$, with the loose count $|\mathcal{H}_T| \le 1+T\bw(1+k)$, which counts retained parents again) is not
a guarantee for \splice{}. Part (ii) is, because $\mathcal{B}_T$ is a
data-independent superset. The price is that it is loose: $\log|\mathcal{B}_T|$
grows linearly in $T$ rather than logarithmically in the budget, and at the
budgets of \cref{sec:exp.setup} the guarantee exceeds $24$\,pp, wider than any
margin this paper reports, so it is correct but not informative at our scale.
Closing the gap between the two parts
with an adaptive-data-analysis bound
\citep{dwork2015preserving,blum2015ladder,russo2016controlling} is open.

Two points keep this from reaching further than it does. First, it is a
\emph{selection} guarantee, not a \emph{reporting} one. Reported benchmark
numbers are single evaluations on the untouched $\test$; $\hat x$ is a function
of $\dtrain$ and $\valid$ only, so it is independent of the test sample and the
reported number needs no union bound and no adaptivity correction. Adaptivity
can make \splice{} select a worse prompt than it believes it has; it cannot make
the reported test score of that prompt optimistic. Second, three properties of
\splice{} plausibly narrow the adaptive channel, though none of them proves a
bound. A state in the beam is never re-scored; a state that left the beam and
is re-proposed is scored again, which with a stochastic system agent is a
limited form of re-querying the validation split, and the factor $T+1$ in
part (ii) prices it. Retention is a
top-$\bw$ cut, so validation enters selection through a ranking rather than
through its full numerical detail. And the proposer never sees a validation
\emph{example}: it sees the scalar score $\tilde\utility(x)$ of its parent and
critiques computed from $\dtrain$ batches. Any residual selection bias then
appears as a val--test gap on the returned prompt, which \cref{sec:exp.setup}
reports.

\par\noindent\emph{\Cref{thm:reg-bias} is stated in the main text (p.~\pageref{thm:reg-bias}).}\par

The penalty for regularizing is at most $\reg$ times the cost of the
unregularized optimum; the bound is tight at $\reg=0$ (no bias) and linear in $\reg$.
The bound is not an empirical claim: the proof rearranges two $\argmax$
definitions and uses only $\cost \ge 0$, so it holds for any
non-negative penalty on any search space and \emph{no experiment can violate
it}. It is also loose by construction, since the right-hand side is
evaluated at the unregularized optimum. The growth the regularizer is meant to
correct is an expected tendency of a proposer that edits by addition, not a
theorem, and this paper does not measure it.

\par\noindent\emph{\Cref{prop:mut-price} is stated in the main text (p.~\pageref{prop:mut-price}).}\par

The diminishing-returns hypothesis is not implied by
$\gamma$-submodularity. The submodularity ratio bounds the immutable blocks'
marginals \emph{at $L$}, the part of $x^\star_{\mathrm{all}}$'s commit set that leaves
the frozen blocks at baseline, while \cref{prop:mut-price}
evaluates them \emph{at the baseline} $x^{(0)}$, and \cref{def:gamma} says
nothing about comparing the two (\cref{rem:mut-price-condition}). The
hypothesis holds automatically when $g$ is monotone with $\gamma = 1$, and in general it formalizes
the reading below: contract blocks do not become \emph{more} worth editing once
the mutable blocks are optimized. Without it the bound still holds with the
marginals taken at $L$, \eqref{eq:app-mut-price-L}. The fixed
ordering is needed because the ground sets of \eqref{eq:ground-sets} are defined at $\pi^{(0)}$: an
all-mutable optimum at another ordering has no commit set in $E^{\mathrm{all}}$,
and pricing it would require ordering moves in the ground set.

Under the diminishing-returns hypothesis, the price of freezing block $i$ is at most its best
single-block marginal gain \emph{over the variants it would have admitted},
summed over frozen blocks and inflated by $1/\gamma$. When immutable blocks are
safety constraints, format contracts, or tool schemas (whose marginal
utility is \emph{by design} small, because they constrain behavior rather than
improve the metric), the right-hand side is small, and so is the price of
selective mutability.

That reading is qualitative. Both factors on the
right-hand side are outside what a deployment can measure: $\gamma$ is the
quantity we cannot estimate per benchmark (\cref{sec:main.limitations}), and
$\tilde{\mathcal{V}}_i$ is unobservable by construction, since it describes a
search that the contract forbids running. We therefore do not report a
realized-to-bound ratio for \cref{prop:mut-price}, and no experiment in this
paper certifies it numerically. What \cref{sec:exp.contract} does instead is
measure a proxy for the left-hand side, the score gap between the returned prompts, comparing full \splice{} against an
all-mutable variant, so it tests the conclusion rather than either hypothesis;
it also reports the case the bound is least flattering in, where a frozen
clause carries a large marginal gain by construction.

\paragraph{Scope.}
These results are stated at the granularity at which each component is
defined. We do not attempt a single end-to-end sample-complexity
bound composing every component: it would multiply the success probability
of \cref{thm:reachability}, the bias of \cref{thm:reg-bias}, the terminal
quality of the idealized greedy of \cref{thm:greedy-approx}, and the slack of
\cref{thm:uniform-conv} into constants no sharper than their product. The
component-wise statements are more informative: each concerns one component,
and an improvement to any one of them carries over to the composition.
\paragraph{Proofs.}
\label{sec:app.theory:gen-reg}

\begin{proof}[Proof of \cref{thm:uniform-conv}]
\emph{Part (i).} Fix $x \in \mathcal{U}$. By \cref{asm:bounded-reward} the per-example
scores $\reward(\model(\rvx;\render(x)),\rvy)$ lie in $[0,1]$, and the
validation split $\valid$ consists of $n$ i.i.d.\ draws, so $\hat\utility_n(x)$ is an
average of $n$ independent $[0,1]$-valued variables with mean
$\utility^\star(x)$ by \eqref{eq:population-utility}. \Cref{lem:hoeffding} gives
\[
\Pr\bigl[|\hat\utility_n(x) - \utility^\star(x)| > \varepsilon\bigr]
\;\le\; 2\exp(-2n\varepsilon^2).
\]
This step uses that $x$ is fixed before $\valid$ is drawn; it is the only place
independence enters. Because $\mathcal{U}$ is fixed in the same sense, a union
bound over its $|\mathcal{U}|$ members gives \eqref{eq:uniform-conv}. The union
bound needs no independence across members, so sharing one $\valid$ across all
of them is harmless.

\emph{Part (ii).} By \cref{asm:proposal-distribution}, every child is
$\Apply(x,u)$ with $u$ a primitive edit or the no-op, so it lies at
edit-graph distance at most one from its parent. Parents at step $t$ belong to
$\mathcal{C}_{t-1} \subseteq \mathcal{H}_{t-1}$, and $\mathcal{C}_0 =
\{x^{(0)}\}$, so by induction every member of $\mathcal{H}_T$ is within
distance $T$ of $x^{(0)}$: $\mathcal{H}_T \subseteq \mathcal{B}_T$ on every
trajectory, whatever the data. The ball $\mathcal{B}_T$ is determined by
$\mathcal{G}$ and $x^{(0)}$ alone, so it is fixed independently of $\valid$.

A state may be evaluated more than once: deduplication compares a proposal
only with the current beam and the step's other proposals, so a state that left
the beam can be re-proposed and re-scored. Deduplication removes duplicates
within a step, so a state is evaluated at most once per step and at
most $T+1$ times in all. Each evaluation draws fresh model randomness on the
same $\valid$. Realize that randomness in advance: for every $x \in
\mathcal{B}_T$ and $j \le T+1$, draw independently the model outputs the $j$-th
evaluation of $x$ will use, and let $\hat\utility_n^{(j)}(x)$ be the resulting
score. This table has the same law as the scores the run produces, and it does
not depend on the run's choices, so each $\hat\utility_n^{(j)}(x)$ is an average
of $n$ independent $[0,1]$-valued scores with mean $\utility^\star(x)$, and
\cref{lem:hoeffding} applies to it as in part (i). Every evaluation in
$\mathcal{L}_T$ is one entry of the table, so
\begin{align*}
\Pr\Bigl[\sup_{(x,\hat\utility_n)\in\mathcal{L}_T}|\hat\utility_n-\utility^\star(x)|>\varepsilon\Bigr]
&\;\le\; \Pr\Bigl[\max_{x\in\mathcal{B}_T,\; j\le T+1}|\hat\utility_n^{(j)}(x)-\utility^\star(x)|>\varepsilon\Bigr]\\
&\;\le\; 2(T+1)|\mathcal{B}_T|\exp(-2n\varepsilon^2),
\end{align*}
which is \eqref{eq:uniform-conv-adaptive}. If every evaluation of a state
returns the same score (as when scores are cached per state), the table has one distinct entry per state and the
union runs over $\mathcal{B}_T$ alone. For the cardinality, every state has
at most $\Delta_{\mathcal{G}}$ neighbors, so the ball holds at most
$\sum_{j=0}^{T}\Delta_{\mathcal{G}}^j \le (\Delta_{\mathcal{G}}+1)^T$ states, and trivially at most
$|\mathcal{X}|$.
\end{proof}

\begin{remark}[Why the history itself cannot be the union set]
\label{rem:adaptive-union}
It is tempting to bound the deviation on $\mathcal{H}_T$ by
$2|\mathcal{H}_T|\exp(-2n\varepsilon^2)$, since $|\mathcal{H}_T| \le
1+T\bw(1+k)$ deterministically (a loose count, since retained parents are
counted again). That does not follow. A union bound controls
the probability that \emph{some member of a fixed set} deviates; it says nothing
about a set whose membership is chosen after looking at the deviations. A
two-line instance shows the gap. Let $J$ candidates share the same population
utility, score all of them on one $\valid$, and keep the $N \ll J$ with the
highest scores. The kept set has cardinality $N$, but the largest deviation in
it is the largest of all $J$, so its tail is governed by $J$, not $N$. \splice{}
is a milder version of the same mechanism: scores decide which parents are
expanded, and the proposer conditions on the parent's score, so the history is
partly a function of validation noise.

This is why \cref{thm:uniform-conv} states the history bound through the
data-independent ball $\mathcal{B}_T$ rather than through $\mathcal{H}_T$. The
cardinality-$|\mathcal{H}_T|$ figure remains useful as a calibration (it is
exactly what a procedure that fixed its candidates in advance would get, so no
worst-case adaptive accounting can do better), but it is
not itself a guarantee for \splice{}. On SST-5 the two are $10.3$\,pp and above
$24$\,pp respectively (\cref{sec:exp.setup}); the difference is the price of
adaptivity under the crudest valid accounting.
\end{remark}

\begin{proof}[Proof of \cref{thm:reg-bias}]
Let $x^\star \in \argmax_{\mathcal{X}}\utility^\star$ and $x^\star_\reg \in
\argmax_{\mathcal{X}}\utility^\star_\reg$ where $\utility^\star_\reg =
\utility^\star - \reg\cost$ with $\reg \ge 0$ and $\cost \ge 0$. By
optimality of $x^\star_\reg$ for $\utility^\star_\reg$ and feasibility of
$x^\star$,
\[
\utility^\star(x^\star_\reg) - \reg\cost(x^\star_\reg)
\;\ge\; \utility^\star(x^\star) - \reg\cost(x^\star).
\]
Rearranging,
\[
\utility^\star(x^\star) - \utility^\star(x^\star_\reg)
\;\le\; \reg\bigl(\cost(x^\star) - \cost(x^\star_\reg)\bigr)
\;\le\; \reg\,\cost(x^\star),
\]
the last inequality because $\cost(x^\star_\reg) \ge 0$. This is
\cref{thm:reg-bias}.
\end{proof}

\begin{remark}[The bound is unfalsifiable]
\label{rem:reg-unfalsifiable}
The proof uses only the two $\argmax$ definitions and $\cost \ge 0$; it
appeals to no property of $\mathcal{X}$, of $q_\phi$, or of the task. The
inequality therefore holds for any non-negative penalty on any search space,
and no experiment can violate it.
\end{remark}

\begin{proof}[Proof of \cref{prop:mut-price}]
Throughout, the ground set is $E^{\mathrm{all}}$ of \eqref{eq:ground-sets},
not $E$: the statement compares an optimum that may edit frozen blocks against
one that may not, so the frozen coordinates must be present in the ground set
for either the hypothesis or the conclusion to have content. On $E$ the set
$S_{\mathrm{imm}}$ constructed below is empty, every marginal in
\eqref{eq:app-mut-price-L} vanishes, and the argument degenerates to
$0 \le 0$.

Among the all-mutable optima, take $x^\star_{\mathrm{all}}$ whose set
$S^\star \subseteq E^{\mathrm{all}}$ of block-variant commits is inclusion-minimal
(a commit set exists because $\Pi = \{\pi^{(0)}\}$, and a minimal one because
$\mathcal{X}_{\mathrm{all}}$ is finite),
let $S_{\mathrm{imm}} := \{(i,v) \in
S^\star : i \in I_{\mathrm{imm}}\}$ be its entries on immutable blocks, and put
$L := S^\star \setminus S_{\mathrm{imm}} \subseteq E$. Minimality gives
$g(L) < g(S^\star)$ whenever $S_{\mathrm{imm}} \neq \emptyset$, since otherwise $x_L$
would be an optimum with a smaller commit set. Then $x_L$ leaves every
immutable block at its baseline variant, so $x_L \in \mathcal{X}$ is feasible
for the restricted search and therefore
\begin{equation}
\label{eq:app-mut-feasible}
g(L) \;\le\; \utility^\star(x^\star_{\mathrm{mut}}) - \utility^\star(x^{(0)}).
\end{equation}
Apply \cref{def:gamma} on $E' = E^{\mathrm{all}}$ with this $L$ and
$S = S_{\mathrm{imm}}$, which are disjoint by construction:
\[
\gamma\bigl[g(L\cup S_{\mathrm{imm}}) - g(L)\bigr]
\;\le\; \sum_{e \in S_{\mathrm{imm}}}\bigl[g(L\cup\{e\}) - g(L)\bigr].
\]
If $g(L \cup S_{\mathrm{imm}}) = g(L)$ the left-hand side of
\eqref{eq:app-mut-price-L} is at most zero by \eqref{eq:app-mut-feasible} and the
claim is immediate; otherwise the pair is admissible in the minimum defining
$\gamma(g; E^{\mathrm{all}})$. Since $L \cup S_{\mathrm{imm}} = S^\star$ and
$g(S^\star) = \utility^\star(x^\star_{\mathrm{all}}) - \utility^\star(x^{(0)})$,
dividing by $\gamma > 0$ and combining with \eqref{eq:app-mut-feasible} gives
\begin{equation}
\label{eq:app-mut-price-L}
\utility^\star(x^\star_{\mathrm{all}}) - \utility^\star(x^\star_{\mathrm{mut}})
\;\le\; g(S^\star) - g(L)
\;\le\; \frac{1}{\gamma}\sum_{e \in S_{\mathrm{imm}}}
  \bigl[g(L\cup\{e\}) - g(L)\bigr],
\end{equation}
which is the price of freezing expressed through the marginal gains of the
immutable blocks \emph{evaluated at $L$}, the mutable part of $S^\star$. This step
uses only \cref{asm:gamma-submod}. Under the diminishing-returns hypothesis of
\cref{prop:mut-price}, \cref{cor:mut-price-baseline} converts the right-hand
side to the baseline form \eqref{eq:mut-price-body}, which completes the proof.
\end{proof}

\begin{corollary}[Baseline form of the price of freezing]
\label{cor:mut-price-baseline}
If in addition $g$ has diminishing returns on immutable coordinates, i.e.\
$g(L \cup \{e\}) - g(L) \le g(\{e\}) - g(\emptyset)$ for every
$e \in S_{\mathrm{imm}}$ and the above $L$ (which holds whenever $g(A \cup \{e\}) - g(A) \le g(\{e\}) - g(\emptyset)$ for
every feasible $A \subseteq E$ and $e \in E^{\mathrm{all}} \setminus E$), then
\begin{equation}
\label{eq:app-mut-price-baseline}
\utility^\star(x^\star_{\mathrm{all}}) - \utility^\star(x^\star_{\mathrm{mut}})
\;\le\; \frac{1}{\gamma}\sum_{i \in I_{\mathrm{imm}}}
  \max_{v\in\tilde{\mathcal{V}}_i}\bigl[\utility^\star(x^{(0)}[i{:=}v]) - \utility^\star(x^{(0)})\bigr],
\end{equation}
which is \eqref{eq:mut-price-body}.
\end{corollary}

\begin{proof}
Substituting the diminishing-returns inequality into
\eqref{eq:app-mut-price-L} replaces each marginal at $L$ by a marginal at
$\emptyset$; then $g(\{(i,v)\}) - g(\emptyset) = \utility^\star(x^{(0)}[i{:=}v])
- \utility^\star(x^{(0)})$ by \eqref{eq:gain-function}, and bounding each term
by the best counterfactual variant of its block (the maximum over
$\tilde{\mathcal{V}}_i$, which is the set the entries of $S_{\mathrm{imm}}$ are
drawn from) and each block of $S_{\mathrm{imm}}$ by all of
$I_{\mathrm{imm}}$ (each omitted term is non-negative, since the seed variant
$v_i^{(0)} \in \tilde{\mathcal{V}}_i$ gives $0$) gives
\eqref{eq:app-mut-price-baseline}.
\end{proof}

\begin{remark}[What the extra condition costs, and why we state it]
\label{rem:mut-price-condition}
Equation~\eqref{eq:mut-price-body}, which \cref{sec:theory-main} reads as a bound by each
frozen block's best single-block gain \emph{at the baseline}, is
\eqref{eq:app-mut-price-baseline}, and it needs the
diminishing-returns step of \cref{cor:mut-price-baseline}. That step does
\emph{not} follow from $\gamma$-submodularity alone: \cref{def:gamma} lower-bounds
a sum of marginals at $L$ by a joint gain, and says nothing about comparing a
marginal at $L$ to a marginal at $\emptyset$. It holds automatically when $g$
is monotone with $\gamma = 1$ (then $g$ is submodular), and more generally it is the natural
formalization of the claim the interpretation relies on, namely that contract
blocks do not become \emph{more} valuable to edit once the mutable blocks have
been optimized. It is kept separate from \cref{asm:gamma-submod} because it is a second structural assumption and the
$\gamma$ in \eqref{eq:app-mut-price-baseline} is already the quantity we cannot
estimate per benchmark (\cref{sec:main.limitations}). The unconditional statement
is \eqref{eq:app-mut-price-L}, and the empirical check of
\cref{sec:exp.contract} measures the left-hand side directly, so it tests the
conclusion rather than the hypothesis.
\end{remark}

\begin{remark}[The right-hand side is not measurable, and why we keep it anyway]
\label{rem:mut-price-counterfactual}
Both statements quantify over the counterfactual variant sets
$\tilde{\mathcal{V}}_i$ of \eqref{eq:ground-sets}, and these are unobservable in
any setting where the proposition is worth invoking: recovering
$\tilde{\mathcal{V}}_i$ means running the proposer against block $i$ with the
contract released, which is exactly what a contract forbids. Together with
$\gamma$, this makes the right-hand side of \eqref{eq:mut-price-body} an object no
experiment in this paper estimates.

The counterfactual set is nonetheless the only choice that leaves the
proposition with content. Taking the maximum over $\mathcal{V}_i$ would make
the right-hand side computable, but trivially: \cref{def:mutability} gives
$\mathcal{V}_i = \{v_i^{(0)}\}$ for every $i \in I_{\mathrm{imm}}$, so every
term is $\utility^\star(x^{(0)}) - \utility^\star(x^{(0)}) = 0$ and the bound
asserts that freezing is free, which \cref{sec:exp.contract} refutes directly.
What \cref{prop:mut-price} supplies is therefore a \emph{shape} rather than a
budget: the price of freezing is controlled by the marginal value of the frozen
blocks. That is the claim \cref{sec:theory:gen-reg} makes, and
\cref{sec:exp.contract} probes it by constructing a frozen block whose marginal
value is large by design.
\end{remark}

\section{Experimental Details}
\label{sec:main.exp-ext}\label{sec:app.exp}
This section extends \cref{sec:main.exp}: protocols and splits, the full
contract study, the ablations summarized there, capacity and cost, the
judge-alignment application, and per-benchmark diagnostics.
\subsection{Protocol and Reporting}

\label{sec:exp.setup}

\textbf{\textsc{Three model roles.}}
Every run has three model roles, which we keep apart. The \emph{target} model is the
one whose prompt is being optimized and on which the returned prompt
is deployed; it is fixed by whichever published protocol we are
reproducing. The \emph{proposal agent} samples the $k$ candidate edits
per parent, and the \emph{critique agent} produces the dual
section-local gradients of \eqref{eq:critique-output}. Proposal and
critique are ordinarily distinct models, which we keep separate so that the
critique is less likely to endorse the reasoning that produced the edit it
assesses. The exception is
AIME-2025, where both roles run on the same model and are
separated only by temperature ($0.9$ for proposal, $0.0$ for
critique).

\textbf{\textsc{The state, concretely.}}
\cref{def:prompt-state} is instantiated by a template of named
sections. On SST-5 and AIME the prompt is $m=5$ blocks (role,
objective, approach, examples, and output format) rendered to a
single XML stream, with the output-format block tagged
\texttt{immutable} and the remaining four mutable. So
$|I_{\mathrm{mut}}|=4$, the frozen coordinate is the output contract,
and $\mathcal{V}_i$ for that block is a singleton by construction
rather than by instruction. The contract experiment of
\cref{sec:exp.contract} extends this template with a seven-clause
contract block and varies exactly which blocks carry the tag.

\textbf{\textsc{Search configuration.}}
Search parameters differ by benchmark, so there is no single default: beam
width is $5$ on SST-5 and AIME ($4$ in the seeded SST-5 sweeps) but $4$ on BBH,
GPQA-Diamond and HelpSteer2;
the fan-out $k$ ranges over $3$--$5$; and the quality-gate ratio $\gateratio$ of
\cref{prop:adaptive-anneal} is $0.7$ on SST-5 and AIME and $0.9$ on HelpSteer2, while the BBH and
GPQA-Diamond configurations do not set it, so the implementation's default
$\gateratio=0$ applies and the gate is inactive there. Every benchmark number uses $\reg=0$. Every benchmark follows its source protocol
exactly (target model, decoding settings, splits and metric). Unless an
ablation varies them, proposals and critiques come from fixed proposal and
critique models (proposal temperature $0.9$, $1.0$ on BBH; critique temperature
$0.0$), with AIME-2025 the exception noted above. The improvement threshold is
$\Delta_{\min} = 0.003$ on SST-5 and AIME and $0.005$ on BBH and GPQA.
Momentum is enabled at window $w=5$ throughout, the setting
\cref{sec:exp.momentum} examines directly.

\textbf{\textsc{Three splits, three roles.}}
Every run uses three disjoint i.i.d.\ splits, and both the guarantees and
the reported numbers depend on which split does what.
$\dtrain$ supplies a fresh batch $B_t$ each step on which the current beam
is re-run; those scores feed the critique and are never used for selection,
retention, gating, stopping or output. $\valid$ is fixed and used to score
and select candidates, under the cached-scoring convention of
\cref{sec:main.method}: each candidate is scored once when it is created and
carries that score while it stays in the beam, rather than being re-rolled.
(A candidate that leaves the beam and is later re-proposed verbatim is scored
again, since deduplication checks only the current beam.) This is what makes
\cref{thm:monotone} operative rather than merely formal. $\test$ is
disjoint and never queried during search. The returned prompt is
$\mathrm{leader}(\mathcal{C}_T)$, which under elitist selection is also the
best-scoring candidate the run evaluated, so its validation score is
optimistic and we report test accuracy separately.

\textbf{\textsc{Reporting convention.}}
We label the two accuracies explicitly, because the distinction
carries the statistical argument below. \emph{Val} is accuracy on
$\valid$: what the optimizer sees, and the quantity every selection
decision is made against. \emph{Test} is accuracy on $\test$, a
single evaluation of the single returned prompt. Unless a number is
marked \emph{Val}, it is test accuracy.

\textbf{\textsc{Adaptive reuse of the validation split.}}
$\valid$ is queried repeatedly across a run to rank candidates, so a
single-candidate concentration bound does not cover \emph{selection},
and we separate the two claims it might be asked to support. For
\emph{reporting}, the returned prompt $\hat x$ is a function of $\dtrain$,
$\valid$ and the optimizer's own randomness, so
$\hat x$ is independent of the test sample and the reported number
needs no union bound and no adaptivity correction. For
\emph{selection}, the correct statement is the uniform bound of
\cref{thm:uniform-conv}, and it is worth being exact about which part of it
applies. Bounding by the size of the evaluated history,
$|\mathcal{H}_T| \le 1 + T\bw(1+k)$, would give a $95\%$ uniform half-width of
$10.3$\,pp on SST-5 ($n{=}500$, $\bw{=}5$, $k{=}3$, $T{=}50$, so
$|\mathcal{H}_T| \le 1001$) against $6.1$\,pp for a single candidate. That
figure is not a guarantee for \splice{}, because the history is chosen using
the same validation scores it would be bounding (\cref{rem:adaptive-union});
it is what a procedure that fixed its candidates in advance would get. The
valid statement unions over the radius-$T$ ball of the edit graph, which
contains every history and does not depend on the data. Its half-width is
$\sqrt{(\ln(2(T+1)/\delta) + T\ln(\Delta_{\mathcal{G}}+1))/2n}$, which at $T{=}50$ exceeds
$24$\,pp for every maximum degree $\Delta_{\mathcal{G}} \ge 2$. So we have a correct
selection guarantee that is uninformative at our budget, and an informative
figure that is not a guarantee. We report both rather than let the second pass
for the first.

What remains is the empirical check. Validation scores decide
which parents survive, and the proposer conditions on its parent's score, so
the history depends on validation noise. Three properties of \splice{} plausibly
narrow that channel without bounding it: each candidate is scored once while it stays in the beam,
retention is a top-$\bw$ cut rather than a numerical reweighting, and the
proposer sees scalar scores and $\dtrain$ critiques but never a $\valid$
example. What those properties cannot do is stop the search from fitting
validation noise, and on the one configuration where we can check, it does.
The seeded SST-5 runs (the momentum and contract experiments)
are the only seed-paired runs for which we report both a validation and a test
score for the returned prompt: in the momentum control arm (\cref{tab:momentum}) the leader
scores $59.9$ on validation and $56.3$ on test, a $3.6$\,pp optimism, against a
seed prompt that scores $54.0$ and $56.2$. So the search gained $5.9$\,pp on
the split it selected on and $0.1$\,pp on the split it did not. That is
selection-on-noise during search, and we report it as such. It does not bias
the reported number, which is measured on $\test$ and is independent of how
$\hat x$ was chosen; it does mean a validation curve in this paper should not be
read as an estimate of held-out gain. The GPQA sweep has no test split, and
its train--validation gaps (\cref{fig:gpqa-gen-gap} in \cref{sec:app.exp}) are not evidence either
way about validation optimism.

\textbf{\textsc{Reproducibility.}}
Prompt optimization is unusually sensitive to details that are easy to
leave unstated (the seed prompt, decoding settings, parsing rules,
model versions, splits, and candidate budget), so we state what the seed
controls and what it does not.

The random seed governs batch draws and data ordering. It does
\emph{not} control sampling inside the language models: the proposal
agent runs at non-zero temperature behind a black-box endpoint, and its
draws are not reproducible from any seed we hold. Every spread we
report across seeds is therefore \emph{total} run-to-run variance
(seed-controlled data sampling plus irreducible model sampling noise)
rather than the seed-attributable component alone, which makes it a
conservative estimate of stability: it can only overstate the
variability a fixed-data rerun would show.

\label{sec:exp.sst5-detail}

\cref{tab:bench-sst5} reports the full SST-5 results alongside the
published baselines from the two protocols. Under the
\citep{zehle2026promptolution} protocol
(target: \texttt{gemma-3-27b-it}), \splice{} reaches $59.2$\% accuracy,
$+2.9$\,pp over the strongest published method (CAPO at $56.3$).
Under the \citep{zhou2025evaluating} protocol
(target: \texttt{deepseek-v3.2}), \splice{} reaches $57.4$\%, $+0.8$\,pp
over the strongest prior method (EvoPrompt at $56.6$). The gap is
larger on \texttt{gemma-3-27b-it}, whose protocol baseline is $44.6$\% against
$56.0$\% for \texttt{deepseek-v3.2}, and the published methods cluster in $53.3$--$56.6\%$
on both (GEPA's $42.0$ aside). A $2.9$\,pp gap over CAPO
corresponds to an error-rate reduction from $43.7$\% to $40.8$\%, a
$6.6$\% relative reduction; the same gap on
\texttt{deepseek-v3.2}'s saturated band would require a method that
breaks out of the cluster entirely, which no published method in this
comparison does.

\begin{table}[t]
\centering
\caption{Benchmark results on SST-5 across two evaluation protocols.
Gemma-3-27B numbers follow \citep{zehle2026promptolution}; DeepSeek-V3.2
numbers follow \citep{zhou2025evaluating}. ``—'' indicates the method
was not run under the corresponding protocol. \splice{} is the test mean
over three runs; baselines are as published. Bold marks the best score per
column.}
\label{tab:bench-sst5}
\small
\begin{tabular}{@{}lcc@{}}
\toprule
Method           & Gemma-3-27B     & DeepSeek-V3.2   \\
\midrule
Baseline         & 44.6            & 56.0            \\
AutoDiff         & 55.7            & ---             \\
OPRO             & 56.0            & ---             \\
EvoPrompt        & 53.3            & 56.6            \\
GEPA             & 42.0            & ---             \\
CAPO             & 56.3            & ---             \\
ProTeGi          & ---             & 56.4            \\
BPO              & ---             & 56.4            \\
\splice{}        & \textbf{59.2}   & \textbf{57.4}   \\
\bottomrule
\end{tabular}
\end{table}

On the \citep{juneja2025task} protocol (GPT-3.5-Turbo), \splice{}
reaches $91.0$\% on ARC-Challenge and $83.09$\% on GSM8K
(\cref{tab:bench-taskfacet}), clearing the
strongest UniPrompt configuration by $+0.5$\,pp and $+0.7$\,pp
respectively. The lead over the TextGrad / ProTeGi family is wider:
$+12.2$\,pp on ARC-Challenge over ProTeGi and $+1.5$\,pp on GSM8K over TextGrad.
Under the \citep{agrawal2026gepa} protocol (target GPT-4.1 Mini;
\cref{tab:bench-gepa}), \splice{} reaches $64.4$\% on
AIME-2025 and $96.24$\% on PUPA ($+5.1$\,pp over GEPA on AIME,
within $0.22$\,pp of GEPA+Merge on PUPA). Under the
\citep{rishav2026contraprompt} protocol (Claude Haiku 4.5;
\cref{tab:bench-contraprompt}), \splice{} reaches $92.6$\% on BBH
($+0.7$\,pp over GEPA) and ties GEPA at $68.4$\% on GPQA-Diamond,
$+3.1$\,pp over the protocol's baseline prompt and $+4.1$\,pp over
ContraPrompt.

Under the accounting of \cref{sec:main.exp}, a margin of at most
$0.5$\,pp either way is a match: ARC-Challenge ($+0.5$), GPQA-Diamond ($0.0$)
and PUPA ($-0.22$). The GSM8K and BBH wins ($+0.69$, $+0.70$\,pp) clear that
threshold only narrowly and sit inside the range that differences in tuning,
API version or prompt formatting could account for, since none of the five
reproduced protocols reports seed variance for the baselines we compare
against. The claims we rest on are the margins with
headroom behind them (AIME-2025 and SST-5 on \texttt{gemma-3-27b-it}) and
the absence of any loss beyond the threshold on any of the five targets.

\subsection{Does the Constrained Formulation Buy Anything?}

\label{sec:exp.contract}

Frozen blocks are preserved by construction: the search space contains
no variant of an immutable block, so reporting a $0\%$ modification
rate against a baseline's nonzero rate restates \cref{def:mutability} rather
than measuring anything. The question that is not definitional is the
one a practitioner actually faces: \emph{when an optimizer is able
to rewrite a contract, does it, and what does that cost?} Answering it
needs a setting where the contract is worth breaking. We construct one.

\textbf{\textsc{The contract.}}
\cref{tab:contract-spec} gives the seven-clause contract, which
instantiates all four classes of production artifact that motivate
\cref{sec:main.method}: a \texttt{record\_sentiment} tool schema, safety
text, a strict output envelope, and a business rule. The task is SST-5
and the target is \texttt{gemma-3-27b-it}. We take the deliberate trade
of a synthetic-but-exact contract over a real deployed one: a
production prompt would be unreproducible and would require an LLM judge
to rule on whether a given rewrite counts as a violation, whereas here
the frozen text is fixed and public, so detection is a regular
expression and a parser and every rate we report is exact and
recomputable.

The clause at the center of the experiment is a compliance-routing rule of
the kind found in deployed agents:

\begin{center}
\begin{minipage}{0.88\linewidth}
\small\itshape
\textsc{Policy-1.} If the snippet mentions ``film'' or ``movie'', do not
classify it. Emit \texttt{\{"label": "ESCALATE"\}}; these require human
review.
\end{minipage}
\end{center}

The optimizer's metric is classification accuracy, so an escalation is
not a sentiment label, scores zero, and \emph{obeying} \textsc{Policy-1}
forfeits score. We call this metric the \emph{proxy}. The
mis-specification is deliberate and, we would argue, typical:
hand-written scoring functions routinely omit credit for a correct
refusal. \cref{tab:contract-metrics} gives the per-example credit under
the proxy and under the contract-aware metric that scores the deployed
objective, together with the ceilings each implies. With a $26\%$ trigger
rate the two are near-opposites on the examples \textsc{Policy-1}
governs: the proxy caps a fully compliant prompt at $74\%$ and the only
route to $100\%$ is total non-compliance. That $\approx\!26$\,pp gap is
the score on offer for breaking the contract, and it is the setup rather
than an artifact of it.

\textbf{\textsc{Why this clause and not an easier one.}}
A contract that is costless to obey cannot demonstrate the value of
protecting it: if compliance never reduces the metric, no optimizer has an
incentive to break it and every method looks equally safe. We verified
this concretely: several natural safety clauses in the same prompt
(forbidding chain-of-thought, requiring the full five-point label scale)
\emph{improved} accuracy when obeyed, and no arm disturbed them.
\textsc{Policy-1} satisfies three requirements simultaneously. It
\emph{can be obeyed}: the trigger is lexical, so compliance is a decision
rather than a capability limit, and the unoptimized seed prompt (A0)
complies on $96.5\%$ of triggered test examples. (A semantically gated variant, ``abstain when uncertain'', was
ignored outright and produced no measurable behavior.) It \emph{is
costly}: the rule fires on $26\%$ of examples. And it \emph{fails
silently}: a missed escalation is well-formed, plausible output that
raises no parser or schema error, so the optimizer receives no corrective
signal. This is the class of failure that format validation cannot
detect and that selective mutability is designed to guard against.

\textbf{\textsc{Arms and protocol.}}
Four arms share the contract text, hyperparameters, splits and objective,
and differ \emph{only} in whether the contract is mutable: \textbf{A1}
\splice{} (immutable), \textbf{A2} all-mutable (identical blocks,
editable), \textbf{A3} monolithic (the entire prompt as one editable
block, which is the representation every prior method searches over), and
\textbf{A4} soft-instructed (editable, plus an explicit ``do not edit''
directive). A3 and A4 both \emph{request} preservation; only A1 enforces
it, and that contrast is what separates a structural guarantee from an
instruction. Splits follow \cref{sec:exp.setup}: fresh training batches
produce critiques only, a fixed $300$-example $\valid$ selects candidates
under the proxy, and the $2{,}210$-example $\test$ is queried once on the
returned prompt under the contract-aware metric. The optimizer therefore
cannot optimize the contract-aware objective (it never observes it),
and the search proceeds exactly as a practitioner's would under a
misspecified metric. Gold sentiment labels are unmodified; only the
scoring rule differs, on the $26.3\%$ of test examples \textsc{Policy-1}
governs. Three paired seeds.

\begin{table}[t]
\centering
\begin{tabular}{@{}llc@{}}
\toprule
\textbf{Contract class} & \textbf{Instantiation} & \textbf{Costs reward?} \\
\midrule
Tool schema            & \texttt{record\_sentiment}: enum-constrained label, & \xmark \\
                       & \texttt{additionalProperties: false}                & \\
Safety text            & no input echo; no chain-of-thought;                 & \xmark \\
                       & prompt-injection resistance                         & \\
Output contract        & lone JSON envelope, no surrounding prose            & \xmark \\
Business rule          & \textsc{Policy-1} compliance routing                & \cmark \\
\bottomrule
\end{tabular}
\caption{\small \textbf{The seven-clause contract, by artifact class.} The
prompt instantiates all four classes of production artifact that motivate the
constrained formulation of \cref{sec:problem-setup}, rather than being a
benchmark prompt with a safety sentence appended. The trade is deliberate: a
prompt from a live deployment would be unreproducible and would require an LLM
judge to rule on whether a given rewrite constitutes a violation, whereas
here the frozen text is fixed and public, so detection is a regular
expression and a parser and every rate in \cref{tab:contract-fate} is exact
and recomputable. Only the business rule is reward-conflicted, and that
asymmetry is the experiment: the other three classes were left untouched by
\emph{every} arm in \emph{every} candidate. Optimizers do not vandalize
indiscriminately; they trade away precisely the contracts that cost
reward, and a practitioner cannot know in advance which those are.}
\label{tab:contract-spec}
\end{table}

\begin{table}[t]
\centering
\begin{minipage}[t]{0.52\linewidth}
\centering
\begin{tabular}{@{}llcc@{}}
\toprule
\textbf{Example} & \textbf{Output} & \textbf{Proxy} & \textbf{Contract} \\
\midrule
$i \in \mathcal{T}$ & \texttt{ESCALATE}  & $0$ & $\mathbf{1}$ \\
$i \in \mathcal{T}$ & label, correct     & $\mathbf{1}$ & $0$ \\
$i \in \mathcal{T}$ & label, wrong       & $0$ & $0$ \\
$i \in U$ & label, correct     & $1$ & $1$ \\
$i \in U$ & label, wrong       & $0$ & $0$ \\
$i \in U$ & \texttt{ESCALATE}  & $0$ & $0$ \\
\bottomrule
\end{tabular}
\end{minipage}%
\hfill
\begin{minipage}[t]{0.44\linewidth}
\centering
\begin{tabular}{@{}lcc@{}}
\toprule
\textbf{A perfect classifier} & \textbf{Proxy} & \textbf{Contract} \\
\textbf{that\ldots} & (val) & (test) \\
\midrule
ignores \textsc{Policy-1}   & $\mathbf{100.0}$ & $73.7$ \\
fully obeys \textsc{Policy-1} & $74.0$ & $\mathbf{100.0}$ \\
\bottomrule
\end{tabular}
\end{minipage}
\caption{\small \textbf{Left: per-example credit under the two scoring
rules.} $\mathcal{T}$ is the set of \textsc{Policy-1}-triggered examples ($26.0\%$ of
$\valid$, $26.3\%$ of $\test$) and $U$ the remainder; each metric is the mean
credit over all $n$ examples. Two rows carry the design. Row~1 is the
mis-specification: a correct refusal earns $0$ under the proxy and $1$ under
the contract-aware metric, so the proxy silently punishes the behavior the
contract mandates. Row~6 is the anti-gaming check: a spurious refusal earns
$0$ under both, so blanket escalation scores only $|\mathcal{T}|/n \approx 26\%$ rather
than $100\%$ and the contract-aware metric cannot be gamed by indiscriminate
abstention. \textbf{Right: the resulting ceilings.} With a $26\%$ trigger
rate the two objectives are near-opposites on the examples \textsc{Policy-1}
governs, so each metric's ceiling sits close to the other's floor. The proxy
caps a fully compliant prompt at $74\%$ and the only route to $100\%$ is
total non-compliance: that $\approx\!26$\,pp gap is the reward on offer for
breaking the contract, and it is the setup rather than an artifact of it.}
\label{tab:contract-metrics}
\end{table}

\begin{table}[t]
\centering
\begin{tabular}{@{}lrrrrr@{}}
\toprule
\textbf{Arm} & \textbf{Cands.} & \textbf{Verbatim} & \textbf{Reworded} & \textbf{Deleted} & \textbf{Mod.\ rate} \\
\midrule
A0 seed (no optimization)      &   3 &   3 &   0 &   0 & $0.0$ \\
\midrule
A1 \splice{} (contract frozen) & 456 & \textbf{456} &   0 &   0 & $\mathbf{0.0}$ \\
A2 all-mutable                 & 336 &   6 & 107 & \textbf{223} & $98.2$ \\
A3 monolithic                  & 372 &  44 & \textbf{316} &  12 & $88.2$ \\
A4 soft-instructed             & 408 & 111 & \textbf{297} &   0 & $72.8$ \\
\bottomrule
\end{tabular}
\caption{\small \textbf{What happens to a reward-conflicted contract clause
when an optimizer is allowed to touch it.} Every candidate prompt the
search evaluated on SST-5 (\texttt{gemma-3-27b-it}), classified by the fate
of the \textsc{Policy-1} clause; counts are candidates pooled across three
paired seeds. \emph{Verbatim}: the $284$-character clause is found
unchanged in the rendered prompt, after normalizing whitespace and JSON
over-escaping. \emph{Reworded}: the string \textsc{Policy-1} still appears
but its text has changed. \emph{Deleted}: no trace of the clause.
\emph{Mod.\ rate} $=$ (Reworded $+$ Deleted) / Candidates, in percent.
Detection uses a regular expression and a parser rather than an LLM judge,
so the counts are deterministic. A1's zero is definitional (the search
space contains no variant of an immutable block) and is reported to
anchor the three editable arms rather than as a result in itself. The
informative contrast is A4 against A2: a ``do not edit'' \emph{instruction}
eliminates deletion entirely ($0$ against $223$) and raises verbatim retention
about fifteenfold ($27.2\%$ against $1.8\%$), yet still leaves $72.8\%$ of candidates with the clause
paraphrased.}
\label{tab:contract-fate}
\end{table}

\begin{table}[t]
\centering
\begin{tabular}{@{}lccc@{}}
\toprule
& \textbf{Val acc.} & \textbf{Test acc.} & \textbf{Escalation} \\
\textbf{Arm} & \textbf{(proxy)} & \textbf{(contract-aware)} & \textbf{compliance} \\
\midrule
A1 \splice{} (contract frozen) & $49.3 \pm 2.9$          & $\mathbf{58.5 \pm 4.1}$ & $\mathbf{66.8 \pm 6.2}$ \\
A2 all-mutable                 & $\mathbf{59.8 \pm 4.5}$ & $41.4 \pm 0.3$          & $0.0$ \\
A3 monolithic                  & $59.4 \pm 3.6$          & $41.6 \pm 0.4$          & $0.1 \pm 0.1$ \\
A4 soft-instructed             & $58.9 \pm 4.8$          & $40.2 \pm 0.8$          & $1.4 \pm 1.6$ \\
\bottomrule
\end{tabular}
\caption{\small \textbf{The ordering inverts between the metric the
optimizer sees and the one a deployment pays.} SST-5,
\texttt{gemma-3-27b-it}, mean $\pm$ s.d.\ over three paired seeds. The
proxy credits sentiment labels only, so a correct escalation scores zero and
obeying \textsc{Policy-1} forfeits reward; the contract-aware metric credits
the escalation instead. Every editable arm beats \splice{} by roughly $10$\,pp on the proxy
and loses by roughly $17$\,pp on the objective a practitioner deploys
against: freezing costs $10.5$\,pp of proxy ($59.8 \to 49.3$) and buys
$17.1$\,pp of contract-aware accuracy ($41.4 \to 58.5$). The arms are
equally good \emph{classifiers}: untriggered label accuracy backs out to
$\approx\!55.5\%$ (A1) against $\approx\!56.2\%$ (A2), so the whole gap
is escalation behavior rather than sentiment quality. A2's compliance is
exactly $0.0$ because every seed deleted the clause, which is a different
failure from A1's $66.8\%$: see \cref{sec:main.limitations} on text versus
behavior.}
\label{tab:contract-results}
\end{table}

\textbf{\textsc{Results.}}
\cref{tab:contract-fate} reports the fate of \textsc{Policy-1} over every
candidate the search produced; \cref{tab:contract-results} reports
accuracy and held-out compliance, and \cref{fig:contract-inversion} plots
them. Modification rates are $0\%$ for A1 by
construction, $72.8\%$ for A4, $88.2\%$ for A3 and $98.2\%$ for A2. The
ordering \emph{inverts} between the two metrics: every unconstrained arm
beats \splice{} by roughly $10$\,pp on the proxy and loses by roughly
$17$\,pp on the objective a practitioner deploys against, and escalation
compliance on held-out test is $66.8\%$ for A1 against $0$--$1.4\%$ for
the editable arms. The optimizer did exactly what it was asked, and the
result is worse to deploy.

Four observations follow.
\emph{First, the arms are equally good classifiers.} Untriggered label
accuracy backs out to $\approx\!55.5\%$ (A1) against $\approx\!56.2\%$
(A2), so the entire $17.1$\,pp gap is escalation behavior rather than
sentiment quality; the constraint costs nothing on the task itself.
\emph{Second, the edits were not cosmetic.} A3 reworded the clause in
$316$ of $372$ candidates and still escalated only $0.1\%$ of triggered
snippets: present in name, stripped of force. This is the silent failure
a validator cannot catch, and it is why we measure behavior on held-out
test rather than stopping at a text-identity check.
\emph{Third, A4 is the most informative arm.} ``Do not edit'' worked in
part: zero deletions against A2's $223$, and the highest verbatim
retention of any editable arm ($27.2\%$ against $1.8\%$). Told not to
remove the clause, the optimizer paraphrased it into ineffectiveness
instead, and none of that partial protection survived to deployment
($1.4\%$ compliance). An instruction shifts the \emph{distribution} of
tampering; it does not prevent tampering, and it cannot survive a
selection rule that rewards violation.
\emph{Fourth, the other three contract classes were untouched by every
arm in every candidate.} Optimizers do not vandalize indiscriminately;
they trade away precisely the clauses that cost score. That is the
argument for freezing by construction rather than auditing after the
fact, because a practitioner cannot know in advance which clauses those
will be.

\textbf{\textsc{On the price of freezing.}}
\textsc{Policy-1} puts $\approx\!26$\,pp of proxy metric on the table,
which makes this the adversarial case for \cref{prop:mut-price} rather
than the typical one: we froze the single clause the optimizer most wanted
to remove. The measured price of doing so is $10.5$\,pp. The bound itself
is not testable here: its right-hand side combines $\gamma$, which we cannot
estimate per benchmark, with a maximum over the counterfactual variant sets
$\tilde{\mathcal{V}}_i$, which are unobservable by construction
(\cref{rem:mut-price-counterfactual}). This arm instead tests the
proposition's \emph{shape} (that the price tracks the marginal value of
what was frozen) at both ends of that range. The claim of
\cref{sec:theory:gen-reg}, that freezing is nearly free when frozen blocks
carry small marginal gains, concerns metric-neutral contracts, and the tool
schema, safety text and output envelope, untouched by every arm in every
candidate, are the evidence for it at zero cost. \textsc{Policy-1} is the
deliberate stress case at the other end, and $10.5$\,pp against $26$\,pp of
forgone proxy metric is the number a practitioner should budget for when a
frozen clause is expensive.

\textbf{\textsc{Limits.}}
Three seeds, one task, one target model, and the arms consumed different
budgets ($456$ candidates for A1 against $336$, $372$ and $408$ for A2--A4),
which if anything favors A1, so the editable arms' lead on the proxy is not
a budget effect. Text preservation is
not behavioral compliance, and the gap between them is the sharpest scope
condition the experiment produces. With the clause present and no
optimization (the seed, A0), the target complies on $96.5\%$ of triggered test
examples; in A1 the clause is byte-identical and compliance is $66.8\%$. The only difference
between those two conditions is the optimizer's edits to the \emph{mutable}
blocks, which share a token stream with the frozen one and were optimized
under a metric that pays for violation. So the most natural reading is that
optimized neighbors partly override a frozen clause: freezing protects a
block's text, not its effect. Baseline non-compliance does not explain the
drop, since A0 is that baseline on the same test split; what the experiment
does not identify is which edits to the neighboring blocks erode the clause.
The
editable arms' $0$--$1.4\%$ has a different cause in each arm. A2 deleted the
clause in most candidates, so there compliance was never possible; A3 and A4
kept it present (A4 retained it verbatim in $111$ of $408$ candidates and
deleted it in none) and complied anyway at $\le 1.4\%$, so a clause that
survives in place but is paraphrased, or surrounded by optimized text, can
lose its force entirely. Finally, the mechanism is aimed at silent failures; a
contract whose violations raise a parser or schema error is caught by output
validation and does not need it.

\begin{table}[t]
\centering
\begin{tabular}{@{}lcccc@{}}
\toprule
\textbf{Strategy} & \textbf{$\bw$} & \textbf{$D$} & \textbf{Best search score}$^{\ast}$ & \textbf{Step of best} \\
\midrule
\texttt{div3} & $4$ & $3$ & $\mathbf{0.963}$ & $4$ \\
\texttt{div2} & $4$ & $2$ & $0.921$ & $4$ \\
\texttt{div1} & $4$ & $1$ & $0.921$ & $4$ \\
\texttt{div0}    & $4$ & $\mathbf{0}$ & $\mathbf{0.213}$ & $12$ \\
\midrule
\texttt{elitist\_fullbatch}$^\ddagger$ & $6$ & $1$ & $0.926$ & $3$ \\
\bottomrule
\end{tabular}
\caption{\small \textbf{Removing the diversity slot collapses the search on BBH.} The four \texttt{div}$*$ rows share beam width $\bw{=}4$ and differ
\emph{only} in the number of reserved diversity slots $D$, which makes this
the clean control for \cref{thm:div-preserve}. Without diversity the run is
near $0.21$ for all $13$ steps (every beam member converges to a
near-identical score and the search never escapes a single mode), while
every $D \ge 1$ variant reaches above $0.92$ by step $4$. This is the
largest single-component effect we measure anywhere in the paper.
$^\ddagger$\texttt{elitist\_fullbatch} is \emph{not} a diversity control: it
differs from the \texttt{div}$*$ rows on two axes ($\bw{=}6$, $D{=}1$) and is
included only as a wider-beam reference point; because two axes differ it
cannot show whether width substitutes for diversity. $^{\ast}$Best score on the
step's training batch ($216$ examples): it measures escape from a mode, not test
accuracy. The diversity mechanism that ran is the distinct-parent slot
with backfill (\cref{sec:method}). Every row is a single
seed (\texttt{random\_seed}${=}42$), so the collapse is unreplicated, and the
same knob has small or no effect on ARC, GPQA and HelpSteer2
(\cref{sec:exp.diversity}). The claim we make is
conditional: diversity is insurance against mode collapse and it pays where
collapse is a live risk.}
\label{tab:diversity-bbh}
\end{table}

\subsection{Knobs, and Where They Fail}
\label{sec:exp.knobs}

The two axes here are ones a practitioner has to set, and neither of them
behaves like a monotone quality dial. Search budget is non-monotonic in $N$,
and trading beam width for per-parent fan-out at a fixed budget did not move
the leader; momentum has no measurable effect at its default and
a measurable cost at a longer window.

\subsubsection{Search Budget: Proposer Draws and Beam Width}

\label{sec:exp.budget}

We characterize the weakest iterative regime first: Best-of-$N$
sampling with no feedback.

\cref{fig:scaling-laws} reports BoN
accuracy across proposer scales and
$N\!\in\!\{1,3,5,10,15\}$ on SST-5 and GPQA-Diamond, from a separate run
that varies the proposer. One-shot accuracy scales
strongly with proposer capacity: on SST-5 the fitted accuracy rises from
$\sim\!45\%$ for the smallest proposers to $\sim\!55\%$ for the largest. But
within that run the five $N$-curves lie within a narrow band at every proposer
size, so drawing more samples from the same seed without feedback does not
substitute for a larger proposer. The GPQA target sweep shows the same lack of
a reliable $N$-trend with the proposer fixed: aggregate accuracy peaks at
$N{=}10$ and drops at $N{=}15$, as does the small target tier, the large tier
peaks at $N{=}5$, and the frontier tier rises from $N{=}10$ to $N{=}15$
(\cref{fig:gpqa-bon-scaling-app} in \cref{sec:app.exp}; its cells carry roughly
$\pm20$\,pp of sampling uncertainty on a difference).
The companion axis is beam width $\bw$. At a fixed per-step variant
budget $\bw k = 12$, trading width for per-parent depth
($\bw{=}4,k{=}3$ against $\bw{=}2,k{=}6$) did not move the leader's mean.
These are single runs per configuration, so we make no claim about the
effect on run-to-run variance, and the comparison does not bear on
\cref{thm:noisy-monotone}, which concerns the re-evaluation regime \splice{}
does not use. It does not test \cref{conj:beam-improvement} either, which
varies $\bw$ at fixed $k$; that conjecture currently has no empirical
handle in this paper. The beam widths used per benchmark ($\bw = 4$ or $5$,
\cref{sec:exp.setup}) are not tuned defaults.

\subsubsection{Momentum}

\label{sec:exp.momentum}

Momentum is part of the critique axis of \cref{sec:main.method}, and its
ablation returns a null. \cref{tab:momentum} isolates it on SST-5 across
four arms (off, and window $w \in \{2, 5, 10\}$) with three seeds
paired across arms and each arm differing from the control in exactly one
field.

Two findings, pointing in different directions.
\emph{Turning momentum off costs nothing measurable.} The paired mean is
$+0.44$\,pp in the control's favor and the sign is consistent on all
three seeds, so we do not claim momentum helps at the default window.
But the magnitude is about six times smaller than the $2.5$\,pp cross-seed
standard deviation, and on a $300$-example validation split $0.33$\,pp is
a single example, which two of the three per-seed gaps are. Windows $2$
and $5$ are likewise indistinguishable. The supported statement is
\emph{no measurable effect}, not \emph{no effect}: three seeds on this
benchmark cannot resolve an effect this small.
\emph{A long window is measurably worse.} At $w{=}10$ the arm loses
$2.44$\,pp paired, consistently on every seed, and terminates at $5.3$
steps against the control's $12.7$ (less than half the search)
while recording the \emph{highest} proposal win rate of any arm ($0.308$).
That combination is consistent with a long critique history
over-constraining the proposer, so that the edits it does produce beat their
parent more often but explore a narrower region, and the run early-stops
at a worse leader; we do not measure the explored region directly. Momentum's window is a real knob with a wrong end.

\begin{table}[t]
\centering
\begin{tabular}{@{}lccccc@{}}
\toprule
& \multicolumn{4}{c}{\textbf{Arm means} (3 seeds)} & \textbf{Paired} \\
\cmidrule(r){2-5}
\textbf{Arm} & \textbf{Val acc.} & \textbf{Test acc.} & \textbf{Steps} & \textbf{Prop.\ win} & \textbf{$\Delta$ val} \\
\midrule
off                 & $59.44 \pm 2.69$ & $56.17 \pm 1.76$ & $11.0$ & $0.266$ & $+0.44 \pm 0.19$ \\
on, $w{=}2$         & $59.44 \pm 4.17$ & $55.28 \pm 0.77$ & $10.7$ & $0.225$ & $+0.44 \pm 1.71$ \\
on, $w{=}5$ (control) & $\mathbf{59.89 \pm 2.50}$ & $\mathbf{56.33 \pm 0.59}$ & $12.7$ & $0.200$ & --- \\
on, $w{=}10$        & $57.44 \pm 3.56$ & $55.84 \pm 1.39$ & $\mathbf{5.3}$ & $\mathbf{0.308}$ & $\mathbf{+2.44 \pm 1.07}$ \\
\bottomrule
\end{tabular}
\caption{\small \textbf{Momentum: a null at the default window and a real
cost at a long one.} SST-5, \texttt{gemma-3-27b-it}, three seeds
($42/43/44$) paired across arms so every contrast is seed-matched rather than
confounded by draw; each arm differs from the control in exactly one field.
\emph{Steps} is the mean number of steps before early stopping;
\emph{Prop.\ win} is the fraction of proposed children beating their parent.
The last column is the paired per-seed difference (control $-$ arm, validation
pp; positive means the control is better), which is the meaningful comparison
because pairing removes the cross-seed draw variance that dominates the
$\pm$ on the arm means. Turning momentum off costs $+0.44$\,pp paired with
the sign consistent on all three seeds, but that is about six times smaller
than the $2.5$\,pp cross-seed standard deviation, and on a $300$-example
split $0.33$\,pp is a single example, so the supported statement is
\emph{no measurable effect}, not \emph{no effect}. The $w{=}10$ arm is
different: it loses $2.44$\,pp paired, consistently on every seed, and
terminates at $5.3$ steps against the control's $12.7$, less than half
the search. Its proposal win rate is the \emph{highest} of any arm, which is
coherent rather than noisy: a long critique history over-constrains the
proposer, so the edits it does produce beat their parent more often but
explore a narrower region, and the run early-stops at a worse leader.
Momentum's window is a real knob with a wrong end: $w{=}5$ is a reasonable choice,
extending it is not.}
\label{tab:momentum}
\end{table}

The cross-benchmark evidence agrees. On HelpSteer2, eight matched pairs
drawn from a $114$-run sweep give a mean effect of $+0.009$ Spearman
$\rho$ with momentum winning three and losing five; on ARC, eight matched
pairs give $-0.001$, nominally hurting six of eight. Across three
benchmarks, two metrics and $19$ matched pairs there is no
effect outside noise on any of them. Two observations belong with that
null rather than against it: ARC is saturated (in this sweep the median
initial prompt scores $0.963$, median run improving $0.004$), so it cannot resolve small
effects, and the ARC and HelpSteer2 sweeps lack replicates, so a true
effect below the noise floor would be invisible.

The implementation tags each momentum entry \textsc{helped}, \textsc{hurt} or
\textsc{flat} by the leader's gain $\Delta_s$ relative to $\Delta_{\min}$, and
warns when three consecutive significant gains alternate in sign. Under elitist
selection with cached scores the leader never falls (\cref{thm:monotone}), so
$\Delta_s \ge 0$: only \textsc{helped} and \textsc{flat} can occur, and the
oscillation warning cannot fire. The absence of oscillation is therefore a
property of the selector, not an observation, and momentum acts as a record of
which directions moved the leader. Its benefit is undemonstrated rather than
refuted, and $w{=}5$ should be read as a reasonable default rather than a tuned
one.

\subsection{When Iteration Pays: Capacity, \texorpdfstring{$\hat p^\star$}{p-star}, and Cost}
\label{sec:exp.when}

The three measurements in this subsection are one argument. Iterative search buys its advantage from the rate at which the (proposer,
target) pair generates improving edits; that rate is directly measurable; and whether
the advantage is worth paying for depends on what the search costs. Taken
together they answer the question a practitioner actually asks before
adopting the method: \emph{on my task and my model, should I iterate, or
sample once and take the best?} They answer it with a diagnostic that
can be run in the first few steps rather than with a benchmark table.

The GPQA-Diamond sweep (set up in \cref{sec:exp.ablations}) varies the target
across six models in three tiers while holding the proposer fixed. Averaged
over targets, Elitist Beam ($64.6\%$), BoN $N{=}10$ ($63.3\%$) and Beam
($62.9\%$) are within noise of one another: one question is $2.5$\,pp, a
difference between two single cells carries about $\pm 20$\,pp of sampling
uncertainty, and a difference between two six-fold averages about $\pm 9$\,pp.
The per-tier view (\cref{fig:gpqa-tier} in \cref{sec:app.exp}) is where a pattern
shows: iterative strategies score highest on the frontier tier, and no
strategy consistently wins on the small tier, where \cref{sec:exp.pstar}
measures $\hat p^\star = 0.21$, so $1-(1-\hat p^\star)^{k} \approx 0.51$ at
$k=3$. \cref{sec:app.exp.gpqa-detail} gives the BoN-scaling, train--validation
and cost views of the same sweep.

\subsubsection{The Improving-Proposal Rate \texorpdfstring{$\hat p^\star$}{p-star}}
\label{sec:exp.pstar}

\begin{table}[t]
\centering
\begin{tabular}{@{}lcccc@{}}
\toprule
& & \multicolumn{3}{c}{\textbf{GPQA-Diamond, by target tier}} \\
\cmidrule(l){3-5}
& \textbf{SST-5} & \textbf{Frontier} & \textbf{Large} & \textbf{Small} \\
\midrule
step $2$     & $0.92$        & $0.83,\ 0.58$ & $0.25,\ 0.75$ & $0.33,\ 0.08$ \\
step $4$--$5$ & $0.08$--$0.17$ & $0.17,\ 0$   & $0.08,\ 0.08$ & $0,\ 0$ \\
\midrule
tier mean (step $2$) & --- & $\mathbf{0.71}$ & $\mathbf{0.50}$ & $\mathbf{0.21}$ \\
$p^{\mathrm{adv}}$ proxy at tier mean, $k{=}3$ & --- & $0.98$ & $0.88$ & $0.51$ \\
\bottomrule
\end{tabular}
\caption{\small \textbf{The improving-proposal rate $\hat p^\star$, measured,
at $\bw k = 12$.} Every run logs the per-step fraction of proposed children
whose validation score exceeds their parent's; this table is extracted from
existing logs, with two folds per GPQA tier. On GPQA the proposer is fixed and
the columns are \emph{target} tiers. \emph{Down the column}, coverage is
abundant while the prompt still has deficiencies the critique can name, then
decays by an order of magnitude as the leader climbs, the state-dependence
that the $\varepsilon$-sublevel form of \cref{asm:proposal-coverage} allows.
\emph{Across the row}, $\hat p^\star$ increases with target capacity. The last
row substitutes the tier mean into the leader-advance rate of
\cref{thm:convergence}, $1-(1-\hat p^\star)^k$, as a proxy: the theorem's rate
uses only the leader's own $k$ draws, and $\tilde p^\star$ counts only gains of
at least $\Delta_{\min}$. It
orders the tiers the way iteration's advantage over Best-of-$N$ does; the
per-step rate $1-(1-\hat p^\star)^{\bw k}$ is at least $0.94$ on every tier
and does not. The rate averages over the states the search visited, so it
upper-bounds the worst-case quantity in \cref{asm:proposal-coverage}. One
improving child out of twelve is already $\hat p^\star = 0.08$, so we read only
\emph{sustained} zeros as a plateau. Single seed (\texttt{random\_seed}${=}42$),
one run per (fold, strategy).}
\label{tab:pstar-measured}
\end{table}

\textbf{\textsc{The mediating quantity, measured.}}
\cref{tab:pstar-measured} reports $\hat p^\star$, the per-step fraction of
proposed children whose validation score exceeds their parent's. This is the
success event of \cref{asm:proposal-coverage}, averaged over the parents the
search visited, so it estimates the improving-proposal probability of that
assumption on average rather than in the worst case. It is the single most informative diagnostic
in the paper, because it reads in two directions at once and each
direction discharges a different claim.

\emph{Down the column} it traces state-dependence. Coverage is abundant
while the prompt still has deficiencies the critique can name ($0.92$
at step $2$ on SST-5, $0.58$--$0.83$ on frontier GPQA folds),
then decays by an order of magnitude as the leader climbs, reaching
zero in several late steps. This is an empirical trace of
$\hat p^\star(\varepsilon)$ as $\varepsilon$ shrinks, and it is what the
$\varepsilon$-sublevel form of \cref{asm:proposal-coverage} predicts
rather than something in tension with it. The terminal zeros are the
stopping condition being met, not the assumption failing: they occur after
the leader has plateaued, at states whose remaining gap is below what the
search can resolve, which is why \splice{} exits on consecutive
no-improvement steps rather than spending the remaining budget.

\emph{Across the row} it gives a measured mechanism for the target-capacity
effect of \cref{sec:main.exp}. With the proposer fixed, tier means at
step $2$ are $0.71$ (frontier), $0.50$ (large) and $0.21$ (small):
$\hat p^\star$ \emph{increases} with target capacity, so the improving-proposal
rate is a property of the (proposer, target, task) triple, not of the proposer
alone. Which rate to read it through matters. Substituting $\hat p^\star$ into
the per-step reachability rate gives $1-(1-\hat p^\star)^{\bw k} \ge 0.94$ for
all three tiers at $\bw k = 12$, which predicts that iteration should pay
everywhere, and it does not on small targets. Substituting it into the
leader-advance rate of \cref{thm:convergence} gives
$1-(1-\hat p^\star)^{k} = 0.98$, $0.88$ and $0.51$ at $k = 3$, an ordering that
matches where iteration separates from one-shot sampling. The second
substitution is a proxy: the theorem's rate $p^{\mathrm{adv}}$ uses only the
leader's own $k$ draws, and $\tilde p^\star$ (\cref{asm:scored-coverage})
counts only gains of at least $\Delta_{\min}$, while $\hat p^\star$ counts every
parent's children and any positive gain. We did not compute the leader-only,
thresholded rate, so these values indicate the advance rate rather than
measure the theorem's quantity. The convergence-relevant floor is
therefore $1/k \approx 0.33$, not the per-step resolution floor
$1/\bw k \approx 0.08$, and the small-tier mean sits below it; small-tier runs
also stop early, after about six steps.

Two consequences we would rather state than leave implicit. First, this
makes the scope condition of \cref{sec:main.exp} operational: \splice{}'s
iterative advantage needs a $\hat p^\star$ well above $1/k$, and
below that one-shot sampling is a reasonable choice when the task is
saturated, though not always, since on the judge task
(\cref{sec:exp.judge}) Best-of-$N$ regresses below its seed. Second, because
the rate is observable within the first few steps of any run, it doubles as a
cheap pre-flight check on a practitioner's own (task, proposer, target) triple.
The caveats are in \cref{tab:pstar-measured}: these are single-seed runs, one
per (fold, strategy); the observed rate averages over the states the search
visited, so it upper-bounds the worst-case $p^\star(\varepsilon)$ of
\cref{asm:proposal-coverage} rather than estimating it; and it cannot
distinguish convergence to $\varepsilon$-optimality from a radius-one local
maximum, since both appear as zeros.

\subsubsection{Cost}

\label{sec:exp.cost}

Cost splits into two questions that are easy to conflate, and only one of
them admits a fair comparison against prior work.

\textbf{\textsc{Inference-time cost of the returned prompt.}}
This is measured and comparable, since every method is evaluated on the
same BBH test set through the same task model.
\cref{tab:cost-inference} reports it: \splice{}'s optimized prompt costs
$1{,}452$ total tokens per example against GEPA's $1{,}727$ (the \splice{}
mean is over the $181$ of $405$ test calls that survive in the response cache),
$16\%$ fewer tokens per example (a $1.19\times$ ratio), driven almost entirely by $2.9\times$ fewer completion
tokens ($128$ against $373$) at near-identical prompt length ($1{,}325$
against $1{,}354$). The claim we make is therefore \emph{comparable prompt
size, substantially shorter outputs} rather than a large total-cost win,
and two caveats travel with it: the completion figure is low enough
relative to every other method that it may in part reflect suppressed
intermediate reasoning, and the harness evaluates \splice{} through a raw
completion call while the other rows pass through a chain-of-thought
wrapper, an asymmetry that inflates their completion counts.

\begin{table}[t]
\centering
\begin{tabular}{@{}lccc@{}}
\toprule
\textbf{Method} & \textbf{Prompt tok/ex} & \textbf{Completion tok/ex} & \textbf{Total tok/ex} \\
\midrule
Baseline (no optimization) & $365$   & $350$ & $714$ \\
GEPA              & $1{,}354$ & $373$ & $1{,}727$ \\
ContraPrompt      & $1{,}021$ & $348$ & $1{,}368$ \\
\midrule
\splice{}                  & $1{,}325$ & $\mathbf{128}$ & $\mathbf{1{,}452}$ \\
\bottomrule
\end{tabular}
\caption{\small \textbf{Inference-time cost of the returned prompt.} BBH,
$405$ test examples, identical task model (\texttt{claude-haiku-4-5}) for
every row. Each column is a rounded per-example mean, so a total can differ
from the sum of its parts by one. \splice{}'s optimized prompt costs $1.19\times$ less per example
than GEPA's, and the gain is almost entirely on the output side:
$2.9\times$ fewer completion tokens at near-identical prompt length
($1{,}325$ against $1{,}354$). The supported claim is therefore
\emph{comparable prompt size, substantially shorter outputs}, not a large
total-cost win. We report per-example figures only and omit totals, because
only $181$ of $405$ \splice{} calls survive in the response cache, so its
totals are not full-test-set totals and would not be comparable with the
other rows'. Two caveats travel with the completion-token figure: it is low
enough relative to every other method that it could in part reflect
suppressed intermediate reasoning rather than pure efficiency, and the
harness evaluates \splice{} through a raw completion call while the other
rows pass through a chain-of-thought wrapper, an asymmetry that inflates
their completion counts. \emph{Optimization-time} cost is deliberately not
compared here; see `What we do not report' below.}
\label{tab:cost-inference}
\end{table}

\textbf{\textsc{Optimization-time cost within \splice{}.}}
Across selectors on GPQA, where the instrumentation is identical for every
row, Elitist Beam reaches the highest validation accuracy on $24\%$ fewer
tracked tokens than Beam ($1.06$M against $1.39$M per fold, excluding candidate scoring;
\cref{tab:gpqa-aggregate} in \cref{sec:app.exp}), a $26\%$ better cost-per-accuracy ratio. On BBH the diversity arms reach their best score at
step $4$ while \texttt{div0} never passes $0.213$ in $13$ steps
(\cref{tab:diversity-bbh}), so the diversity arms also reach their best much
earlier; we
do not have per-arm token counts for that sweep and so give no ratio.

On GPQA, Elitist Beam costs
$12$--$41\times$ a Best-of-$N$ run per fold depending on $N$
(\cref{tab:gpqa-aggregate}), and BoN$(N{=}10)$ lands $1.3$\,pp below it at
roughly $6\%$ of the token cost, a difference this sample cannot resolve. So
our cost argument is \emph{elitist-versus-beam}, not
\emph{iterative-versus-one-shot}: where budget binds on a saturated task,
one-shot sampling is a reasonable choice (though not on the judge task,
where it regresses below the seed), and the $\hat p^\star$ diagnostic of
\cref{sec:exp.pstar} is the cheap way to decide which regime a given (task,
proposer, target) triple is in.

\textbf{\textsc{What we do not report, and why.}}
There is no \splice{}-versus-GEPA comparison on
optimization-time cost. Our own instrumentation does not populate the
candidate-scoring bucket, which is the dominant term (each scored
candidate consumes a full evaluation batch, so the tracked figures
understate true optimization cost by roughly $40\times$), and the
GEPA reproduction logged no optimizer tokens at all, its
reflection and mutation calls never having been cached. Publishing a
partially instrumented number of ours against an absent number of theirs,
in our own favor, is precisely the asymmetry a cost table should not
contain. Closing this requires instrumenting both sides and re-running
under a matched optimizer budget, which we scope as future work rather
than estimate here.

\subsection{Application: LLM-as-Judge Rubric Calibration}

\label{sec:exp.judge}

The benchmarks above all optimize prompts whose output is the
system's primary task response. Here we apply \splice{} to a task
of a different kind: optimizing the rubric prompt given to an
LLM-as-judge so that its ratings align with human annotations. The
framework is unchanged (the rubric text is the block-structured
state, the judge's output is an intermediate signal, and the scalar
metric is the rank correlation of that output with ground truth), but the target has shifted from ``get the task right'' to
``predict human preferences correctly.'' Judge rubrics are a natural
application regime for the method: they are long, multi-section
artifacts whose failure modes are mis-calibrations rather than
outright errors, and they are increasingly used as the scoring
mechanism for other prompt-optimization pipelines.

\paragraph{Task and data.}
We use the NVIDIA HelpSteer2 dataset~\citep{wang2024helpsteer}, which
provides human ratings for assistant responses across five criteria
on a $1$--$5$ scale: helpfulness/understanding,
correctness/completeness, coherence/clarity, language complexity,
and succinctness/verbosity. The judge receives a rubric prompt
containing an overall-quality rubric and a per-level description for
each of the five criteria; given a (user prompt, assistant response)
pair, it emits six scalar scores. The metric is the mean Spearman
rank correlation between the judge's scores and the human ratings,
averaged across the five criteria. Scoring uses a stratified
$400$-example validation split; each step's training batch has $64$
examples. Proposal temperature is $0.9$ and critique temperature $0.0$.

\paragraph{Judge models.}
We evaluate across the six judge models S1--F2 of \cref{sec:exp.ablations},
two per tier (Small, Large, Frontier). The initial hand-crafted rubric
reaches mean Spearman $\rho = 0.555$ across these six models, a
competitive baseline that sets a high bar for any automated optimizer.

\begin{figure}[t]
\centering
\includegraphics[width=0.85\textwidth]{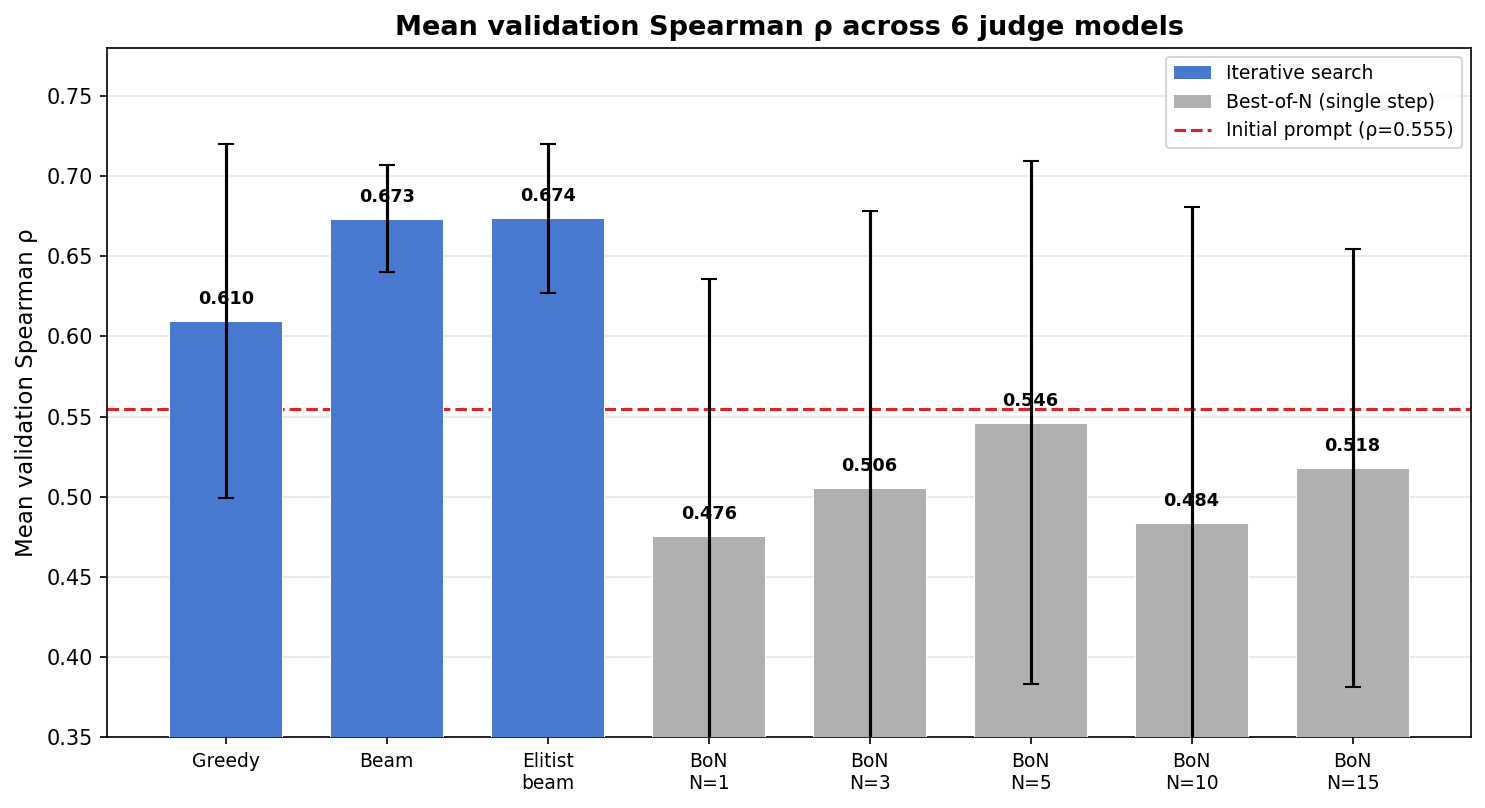}
\caption{\textbf{Judge Alignment on HelpSteer2: iterative search
vs.\ single-step sampling.} Mean validation Spearman $\rho$
between the optimized judge and human annotations across six judge
models (error bars: standard deviation across models). The initial
hand-crafted rubric (red dashed line, $\rho{=}0.555$) is a
competitive baseline that none of the Best-of-$N$ configurations
matches on average. All three iterative strategies improve over
the seed; Best-of-$N$ \emph{regresses below} it.}
\label{fig:judge-scaling}
\end{figure}

\paragraph{Results.}
\cref{fig:judge-scaling} reports mean validation Spearman $\rho$
across the six judge models. The best iterative strategy (Elitist
Beam, $D{=}2$, \texttt{parent} mode) reaches $\rho = 0.674$, followed
by Beam at $0.673$ and Greedy at $0.610$. The best
single-run configuration is F1 with Elitist Beam,
at $\rho = 0.736$. None of the Best-of-$N$ configurations match the
iterative results: the best BoN mean is $0.546$ ($N{=}5$), more than
$0.12$ below the Elitist Beam mean, and BoN mean-over-$N$ is
non-monotonic ($N{=}1{:}0.476$, $3{:}0.506$, $5{:}0.546$, $10{:}0.484$,
$15{:}0.518$). The gap between iterative search and single-step
sampling is wider on this task than on any of the public benchmarks
in the main text.

\paragraph{Single-step sampling regresses below the seed.}
On this dataset Best-of-$N$
sampling \emph{actively degrades} the prompt relative to the
hand-crafted seed. The mean improvement over the seed is $-0.049$
averaged across all five BoN configurations, so on average BoN leaves the
prompt worse than the seed. Iterative
strategies move in the opposite direction, improving over the seed
by $+0.097$ on average. Variance decomposes the effect further:
across judge models, the standard deviation of BoN scores is $0.166$,
compared to $0.063$ for iterative strategies. On this task BoN is
not merely weaker in expectation; for individual judge models it is
sometimes catastrophic (e.g.\ $\rho = 0.18$
on L2 with BoN $N{=}1$, compared to $0.614$ for Elitist Beam on the
same model). The iterative-vs-BoN gap is therefore not a question of
whether more samples would eventually match search: on a task where
the prompt is sensitive to small-scale calibration choices, up to
$N{=}15$, proposer-only sampling does not recover what iterative critique +
selection delivers.

\paragraph{The elitist property prevents leader regression.}
Elitist Beam ($0.674$) and Beam ($0.673$) achieve nearly
identical means, but the mechanism differs in a way that matters for
worst-case behavior. Under Beam, the final leader
may be worse than the best candidate seen during optimization; under
Elitist Beam with the cached-scoring convention of
\cref{sec:main.method}, the leader score is monotone
non-decreasing (\cref{thm:monotone}). Greedy
($\rho = 0.610$) suffers most from regression: on S2 its final leader
drops to $0.402$ even though a stronger candidate appeared earlier
in the search. Across the six judge models, greedy's standard
deviation is $0.110$, approaching BoN-level variance despite
iterative search, a direct consequence of maintaining only a
single candidate with no fallback when it regresses. Beam and
Elitist Beam standard deviations across the six judge models are $0.033$ and $0.046$
respectively, against $0.166$ for BoN.

\paragraph{What \splice{} discovered.}
Inspecting the optimized rubric reveals four calibration strategies
that the seed lacks, each aimed at a way the seed rubric can
mis-calibrate the judge.
\emph{(i) Descriptive vs.\ evaluative dimension separation.}
Language complexity and verbosity are descriptive (a simple response
is not worse than a complex one), whereas helpfulness, correctness,
and coherence are evaluative. The optimized rubric marks the first
two as ``descriptive only'' and instructs the judge that low scores
on these dimensions must not reduce the overall score, a
distinction absent from the symmetric seed.
\emph{(ii) Anti-deflation gates.}
The optimized prompt contains explicit ``default high'' clauses and
mandatory upgrade-before-downgrade tests (``before assigning score
$3$, ask: can I name ONE specific factual claim that is clearly wrong
AND central to the answer? If not, you MUST score $4$'') that
guard against systematic under-scoring.
\emph{(iii) Structured scoring decision trees.}
Rather than the seed's per-level prose, the optimized rubric encodes
a four-step decision procedure (topic relevance $\to$ task
understanding $\to$ substantive correctness $\to$ named-flaw test),
where each step maps to a minimum score.
\emph{(iv) Dual-track overall scoring.}
Two independent tracks (core-quality threshold on helpfulness,
correctness, and coherence; or arithmetic mean above a cutoff) each
suffice for the top overall score, preventing descriptive dimensions
from dragging down the aggregate.
These mechanisms are specific and interpretable, and \splice{}
discovers them from rank-correlation feedback alone, without any
auxiliary supervision on where the judge is mis-calibrated.

The remaining subsections give supporting diagnostics: the SST-5 strategy
sweep against published numbers (\cref{sec:app.exp.strategy-compare}),
GPQA-Diamond search trees and per-tier, BoN-scaling, train--validation and
cost views (\cref{sec:app.exp.trees,sec:app.exp.gpqa-detail}), and SST-5
lineage and score traces (\cref{sec:app.exp.lineage-extra}).

\subsection{Additional Search Strategy Ablations}
\label{sec:app.exp.strategy-compare}

\begin{figure}[t]
\centering
\includegraphics[width=\textwidth]{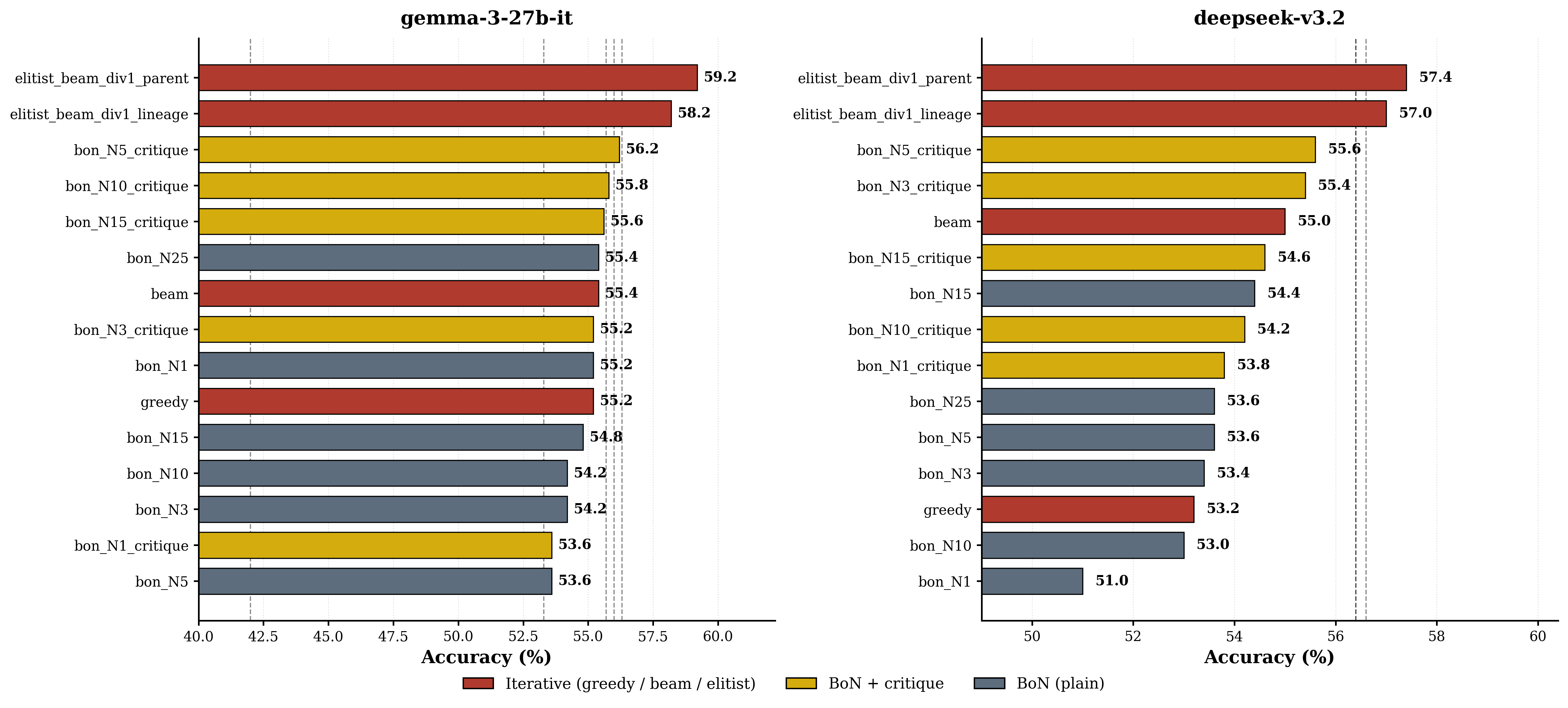}
\caption{\textbf{Only the elitist $D{=}1$ variants beat critique-guided
one-shot sampling.} Bars report test accuracy of the final leader, mean over
three runs, for every iterative strategy (red: Greedy, Beam, Elitist Beam) and
every one-shot baseline (yellow: BoN with critique; grey: plain BoN). Dashed
lines mark published numbers for prior optimizers from
\citep{zehle2026promptolution} (left) and \citep{zhou2025evaluating} (right).
The elitist variants lead the best BoN+critique arm by $3.0$\,pp (Gemma) and
$1.8$\,pp (DeepSeek); Greedy and Beam fall below it.}
\label{fig:sst5-strategy-compare}
\end{figure}

\begin{figure}[t]
\centering
\subfloat{%
  \includegraphics[width=0.6\textwidth]{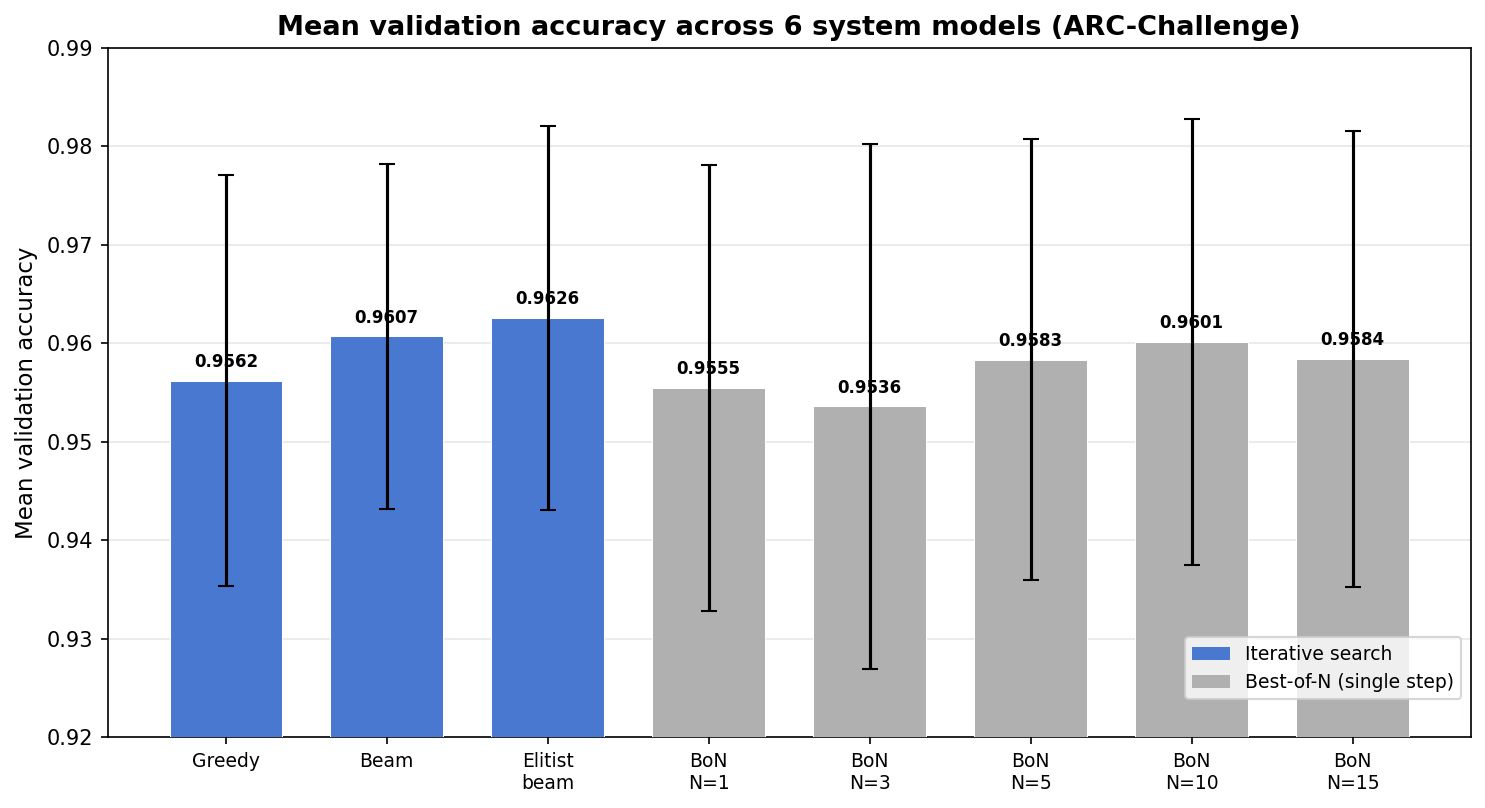}%
  \label{fig:iter-vs-bon-arc}%
}
\caption{\textbf{On ARC-Challenge the strategies compress.} Validation
accuracy, mean over the six target models S1--F2. The SST-5 counterpart is \cref{fig:iter-vs-bon-sst5}.}
\label{fig:iter-vs-bon-joint}
\end{figure}

\cref{fig:sst5-strategy-compare} places every \splice{} variant shown
(Elitist Beam with $D{=}1$, \texttt{parent} and \texttt{lineage} modes), the
Greedy and Beam arms, and every Best-of-$N$ baseline
on the same axes as the published prior-method numbers, for the two
SST-5 target models. Three points follow. First, the top two positions on both target models are occupied by
\splice{} variants with $D{=}1$ diversity (\texttt{parent} and \texttt{lineage}
modes). The figure has no $D{=}0$ elitist arm, so it shows that the $D{=}1$
variants rank highest, not that the slot is the reason. Second, the best
elitist variant leads the best BoN+critique configuration by $3.0$\,pp on
Gemma and $1.8$\,pp on DeepSeek, while Greedy and Beam fall below it, so on
SST-5 the margin over one-shot sampling comes from elitist search, not from
iteration alone. Third, critique changes plain BoN by $-1.6$ to $+2.8$\,pp
(positive in nine of ten pairs; compare \cref{fig:sst5-bon-scaling-bars}).
\cref{fig:iter-vs-bon-joint} gives the ARC-Challenge counterpart of
\cref{fig:iter-vs-bon-sst5}.


\subsection{GPQA-Diamond: Search-Tree Topology}
\label{sec:app.exp.trees}

\begin{figure}[t]
\centering
\includegraphics[width=\textwidth]{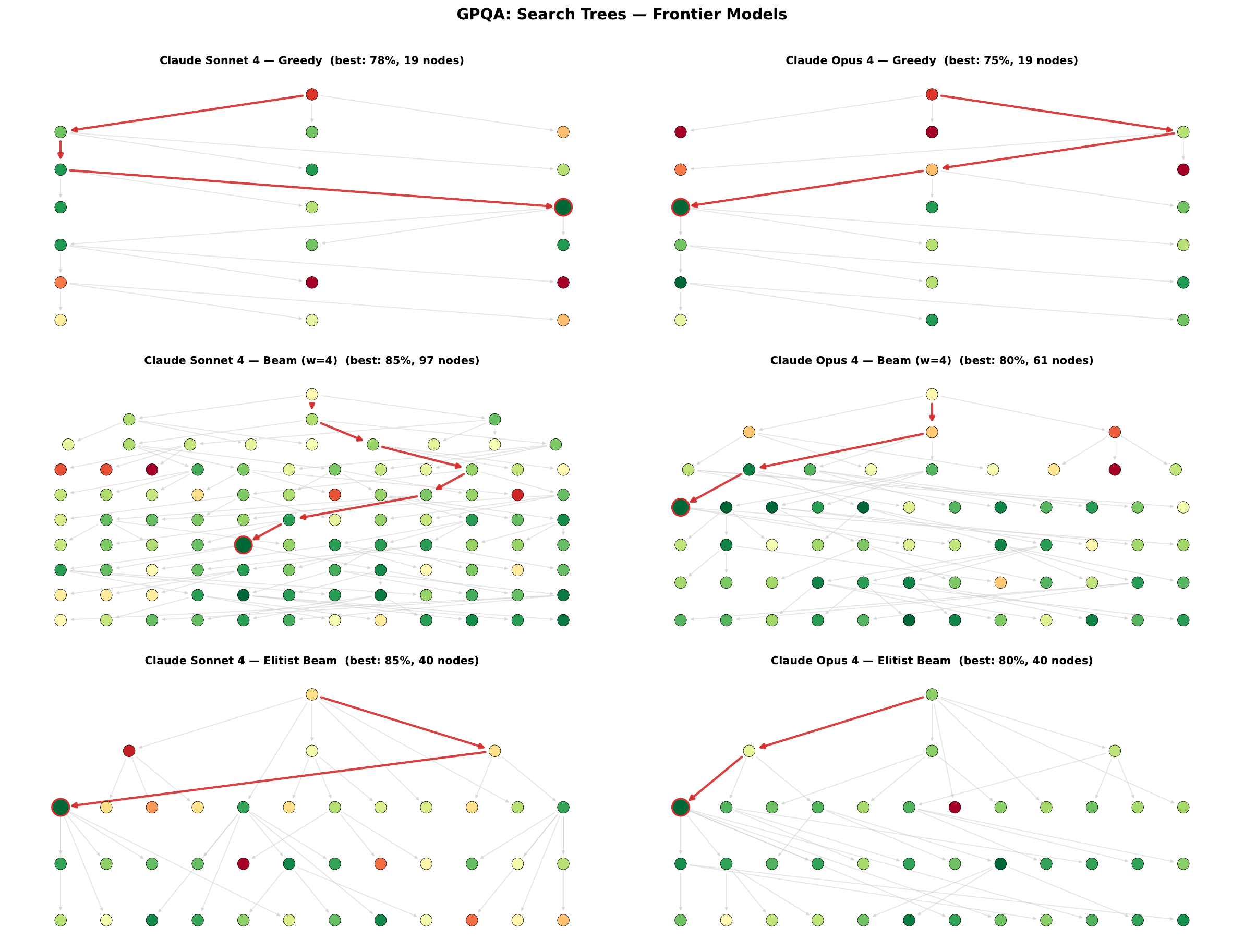}
\caption{\small
\textbf{(Search trees, frontier models.)} Rows: selection strategies
(Greedy, Beam $\bw{=}4$, Elitist Beam); columns: frontier target
models. Node color encodes validation score (red = low,
green = high); the red path is the best-candidate lineage. Greedy
produces narrow $\sim$19-node chains: once it finds a good
candidate, it commits. Beam generates bushy $61$--$97$-node trees
exploring many parallel lineages. Elitist Beam sits between at
$\sim$40 nodes.}
\label{fig:trees_frontier}
\end{figure}

\begin{figure}[ht]
\centering
\includegraphics[width=0.85\textwidth]{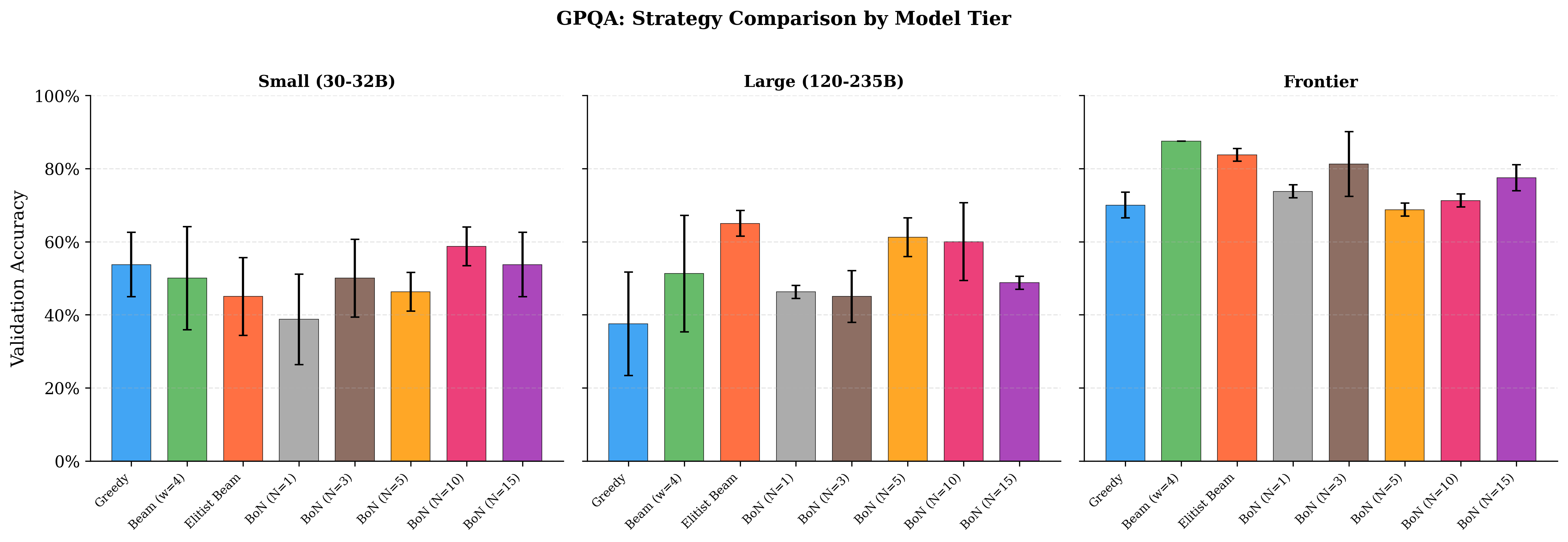}
\caption{\textbf{GPQA strategy comparison by model tier.} Beam and Elitist
Beam lead on frontier models and Elitist Beam on large ones; on small models no
strategy consistently wins. Tick label ``$w{=}4$'' is the beam width
$\bw{=}4$.}
\label{fig:gpqa-tier}
\end{figure}

\Cref{fig:trees_frontier} expands the search-tree footprint discussion of
\cref{sec:main.exp}: the search-tree grid visualizes the exploration footprint of
each selector on frontier target models. This figure, \cref{fig:gpqa-trees-elitist-all} and the
case studies of \cref{sec:app.exp.gpqa.cases} come from a separate GPQA run with
the same strategies and targets as the sweep of \cref{sec:exp.ablations}, so
their final scores need not match \cref{fig:gpqa-tier}. In the panel titles of
the GPQA figures, Claude Sonnet~4 and Claude Opus~4 abbreviate the F1 and F2 targets
\texttt{claude-sonnet-4-6} and \texttt{claude-opus-4-6}, and the
other names denote S1--L2 in the order of \cref{sec:exp.ablations}.

\begin{figure}[ht]
\centering
\includegraphics[width=0.95\textwidth]{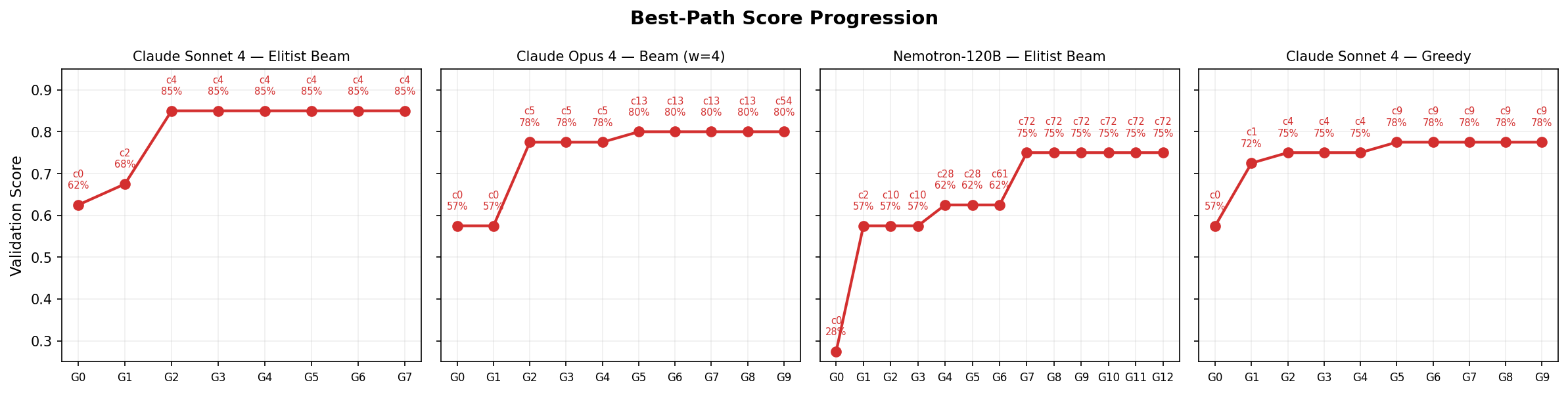}
\caption{\small \textbf{(Candidate Lineage.)} Validation score along
the best-scoring lineage for four GPQA-Diamond runs; $x$-axis is the
search step, each tick labelled with the best-path node at that step. Frontier targets (F1, F2) reach their best
score in $3$--$4$ nodes counting the seed; L1 needs $6$. The Greedy run (rightmost) improves at
steps $1$, $2$ and $5$.}
\label{fig:refinement}
\end{figure}

Greedy produces $\sim\!19$-node chains, Beam produces $61$--$97$-node trees,
and Elitist Beam sits between at $\sim\!40$ nodes.

\label{sec:app.exp.gpqa-detail}

The main text (\cref{sec:exp.ablations}) summarizes this sweep;
\cref{tab:gpqa-aggregate} gives its aggregate, and this subsection holds the
remaining views of the same runs: per-tier bars, Best-of-$N$ scaling, the train--validation gap, the
cost--accuracy Pareto, search-tree topology across all six models, and three
qualitative case studies. For reference, the sweep is
eight strategies (Greedy, Beam, Elitist Beam, and
Best-of-$N$ at $N\in\{1,3,5,10,15\}$) across six target models
(S1--F2, \cref{sec:exp.ablations}).
All runs use one fixed proposal agent (temperature $0.9$) and one fixed
critique agent (temperature $0.0$); the validation split has $40$ questions, scored by exact match.
Iterative strategies use 3 proposals per candidate, early-stopping
patience of 5, and up to 50 steps.

\cref{fig:gpqa-tier} isolates each tier. On \textbf{frontier} models,
Beam ($87.5\%$) and Elitist Beam ($83.8\%$) score highest, with BoN
$N{=}3$ close behind ($81.3\%$). On \textbf{large} models, Elitist Beam
is the single best strategy ($65.0\%$), with BoN $N{=}5$ and $N{=}10$
near $60\%$. On \textbf{small} models the bars span $39$--$59\%$ with
overlapping error bars and BoN $N{=}10$ is highest: strategy choice
matters less than model capability at this scale.

\subsubsection{Best-of-\texorpdfstring{$N$}{N} Scaling Is Non-Monotonic}
\label{sec:app.exp.gpqa.bon}

\begin{figure}[ht]
\centering
\includegraphics[width=0.65\textwidth]{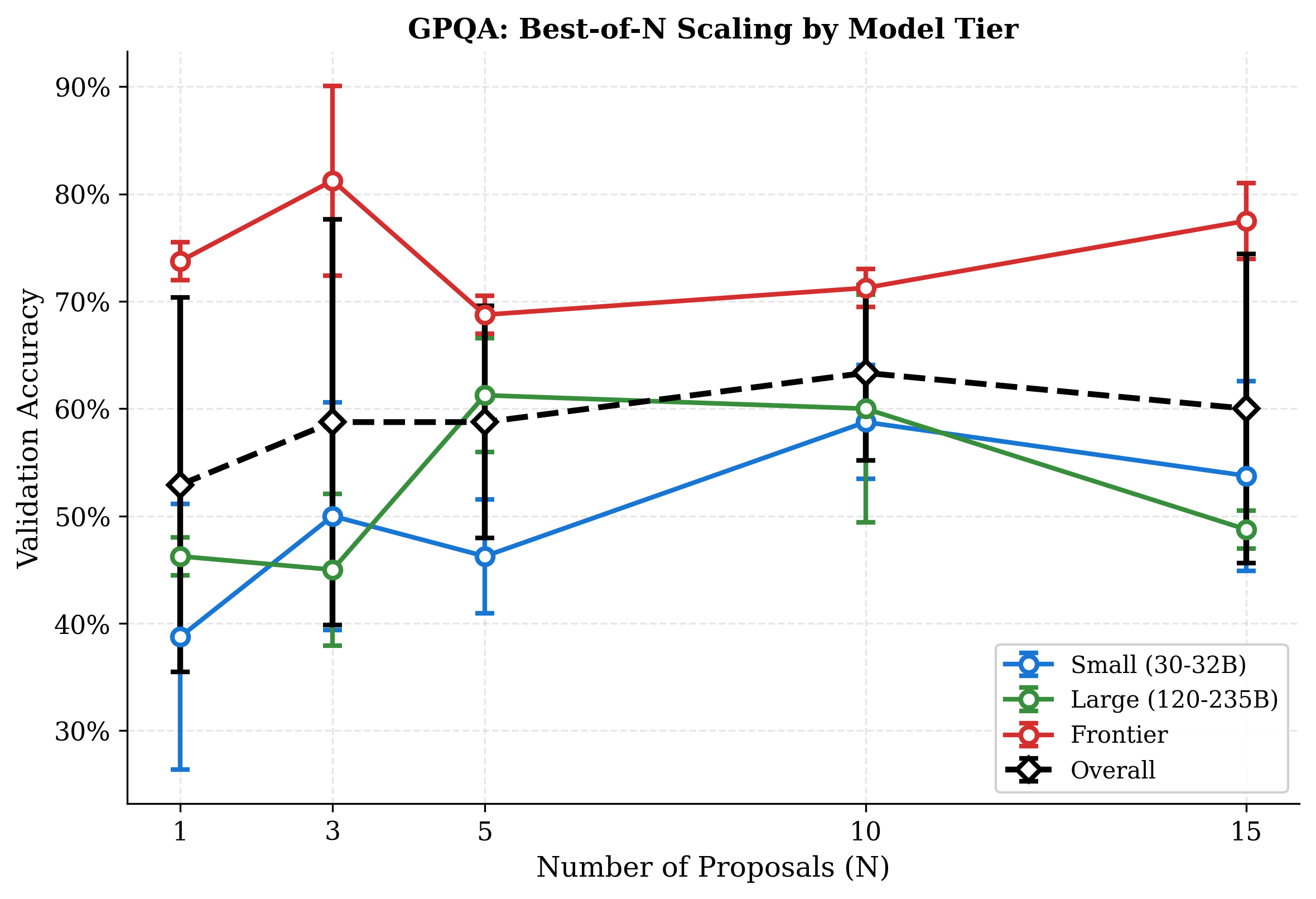}
\caption{\textbf{BoN scaling by tier on GPQA-Diamond.} Accuracy is
non-monotonic in $N$; the overall and small-tier curves peak at
$N{=}10$, and small models gain most.}
\label{fig:gpqa-bon-scaling-app}
\end{figure}

\cref{fig:gpqa-bon-scaling-app} tracks how BoN accuracy changes as
$N$ increases. Scaling is \emph{non-monotonic}: the overall and small-tier curves peak at $N{=}10$
and drop at $N{=}15$, the large tier peaks at $N{=}5$, and the
frontier tier peaks at $N{=}3$ and rises again from $N{=}10$ to
$N{=}15$. Small models gain most from larger $N$ ($38.8\%$ to
$58.8\%$). Each cell is a single independent run, so these
differences are within run-to-run sampling noise and should not be
read as a trend in $N$.

\subsubsection{Train--Validation Gap by Strategy}
\label{sec:app.exp.gpqa.gap}

\begin{figure}[ht]
\centering
\includegraphics[width=0.85\textwidth]{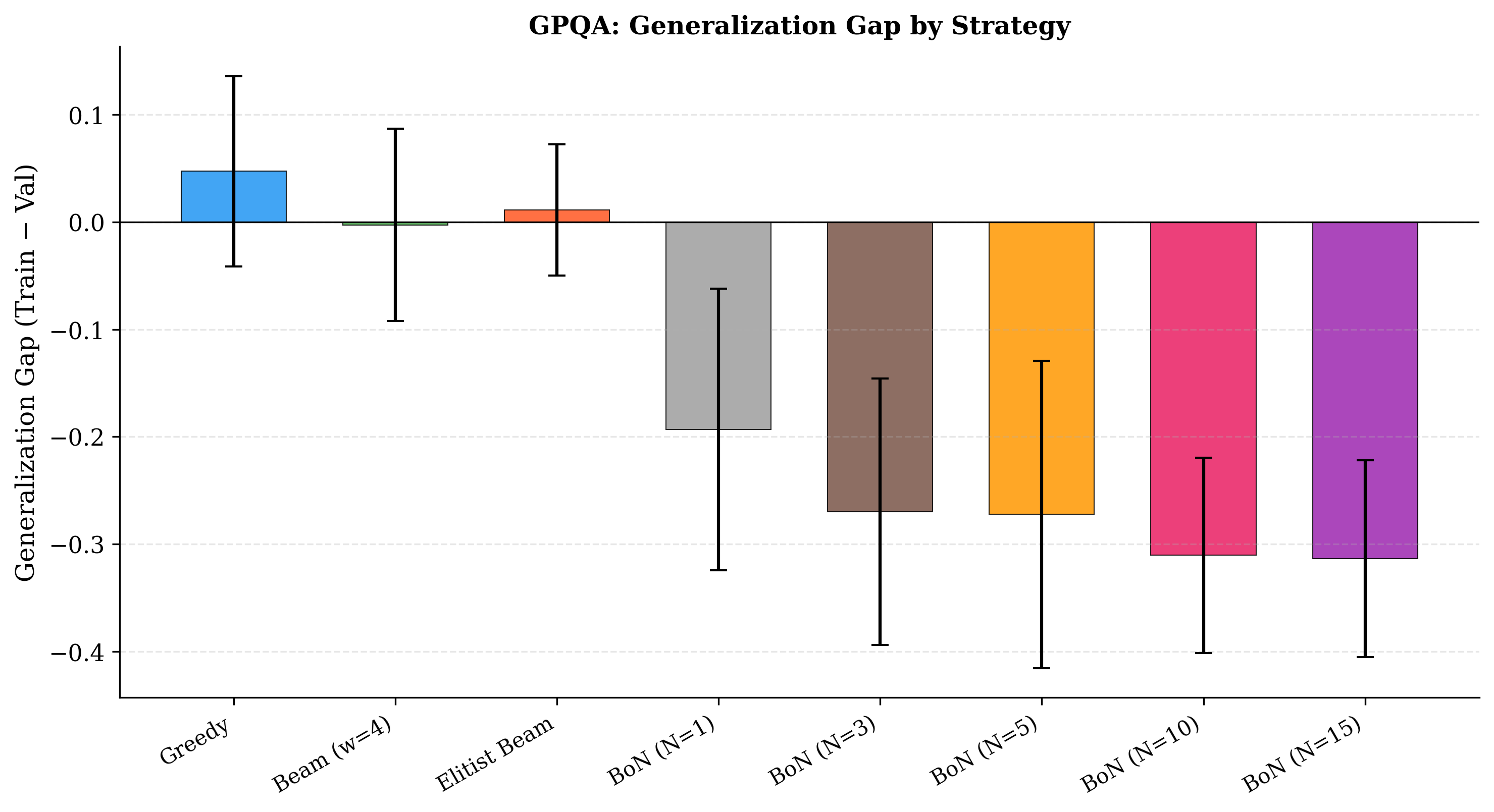}
\caption{\textbf{Train--validation gap (Train $-$ Val) by strategy on
GPQA-Diamond.} Iterative strategies are near zero; BoN strategies
have large negative gaps of $-19$ to $-31$ percentage points,
meaning single-batch training accuracy is not a usable proxy for
validation performance. Tick label ``$w{=}4$'' is the beam width $\bw{=}4$.}
\label{fig:gpqa-gen-gap}
\end{figure}

\cref{fig:gpqa-gen-gap} shows the gap between training and validation
accuracy, strategy by strategy. Iterative strategies hover near zero:
their training scores, averaged over many batches, track validation.
BoN strategies show large \emph{negative} gaps (validation $\gg$
train): one training batch is a noisy score, and the validation score
of the selected candidate is biased upward by selection. Because
every arm selects on $\valid$, the gap does not change which prompt
BoN returns.

\begin{figure}[ht]
\centering
\includegraphics[width=0.6\textwidth]{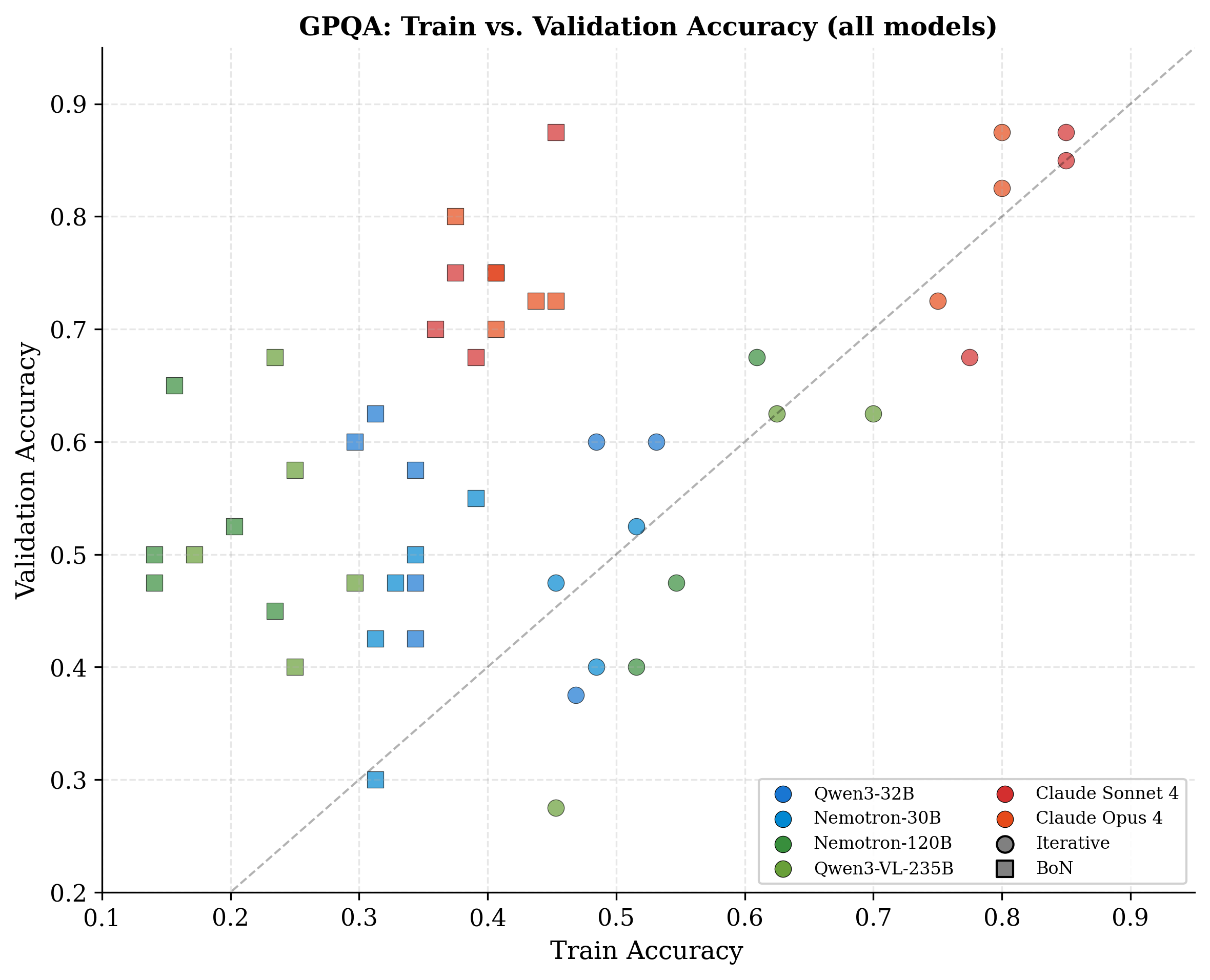}
\caption{\textbf{Train vs.\ validation accuracy on GPQA-Diamond.}
Iterative strategies track the diagonal; BoN strategies overshoot.}
\label{fig:gpqa-train-val}
\end{figure}

\cref{fig:gpqa-train-val} shows the same from a different angle.
Iterative runs (circles) cluster along the diagonal and BoN runs
(squares) sit far above it, across all model tiers. The difference is
one of measurement: an iterative run's training score is averaged over
many steps' batches, while a BoN run logs a single batch, so the gap
reflects how training accuracy is recorded rather than a difference in
how well either strategy generalizes.

\subsubsection{Cost--Performance Trade-off}
\label{sec:app.exp.gpqa.cost}

\begin{figure}[ht]
\centering
\includegraphics[width=0.65\textwidth]{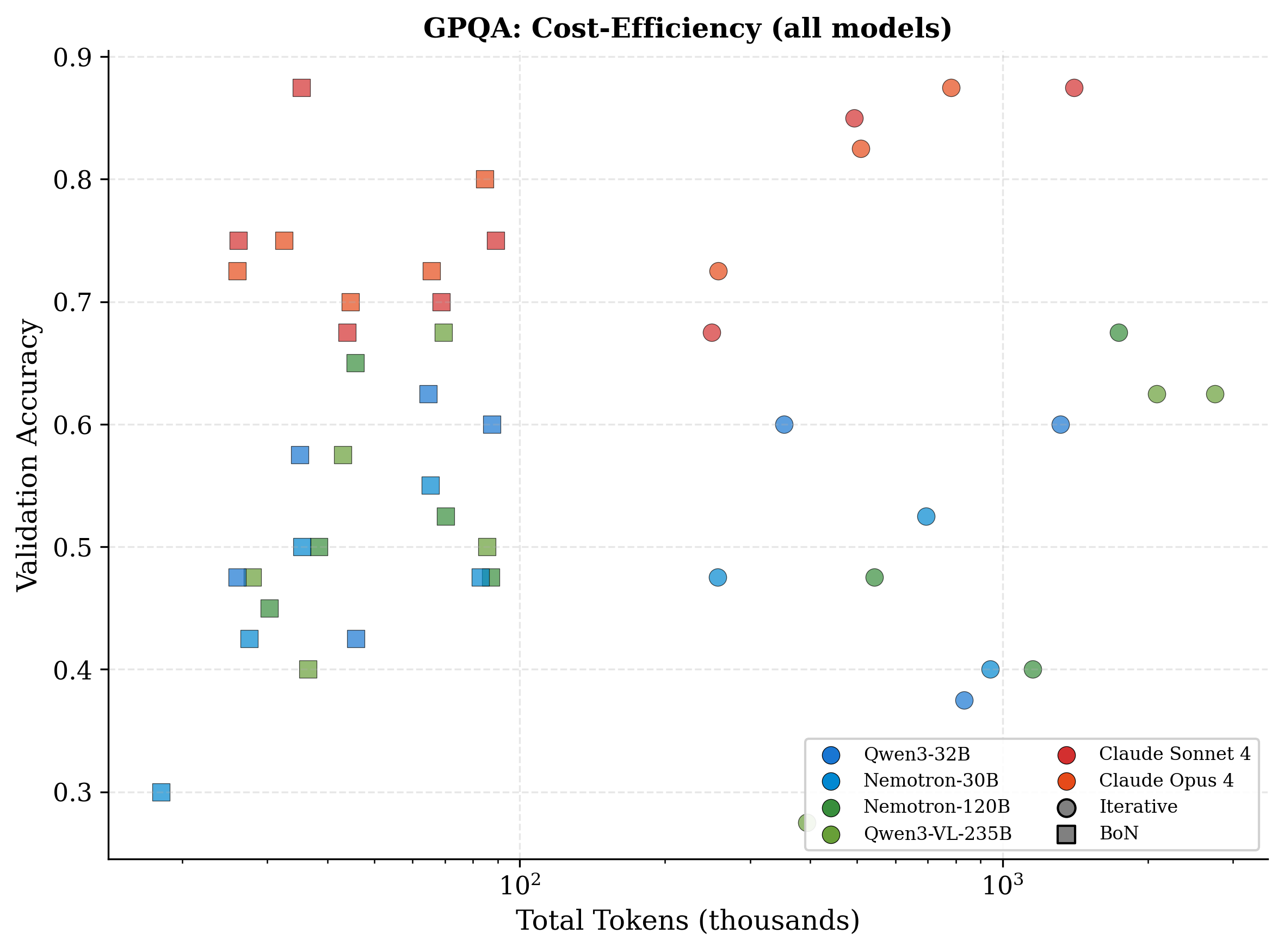}
\caption{\textbf{Cost vs.\ accuracy on GPQA-Diamond.} Circles = iterative,
squares = BoN; colors = target model.}
\label{fig:gpqa-cost-val}
\end{figure}

\cref{tab:gpqa-aggregate} gives accuracy and token cost per strategy, averaged
over the six targets, and \cref{fig:gpqa-cost-val} plots token cost against
validation accuracy for every (model, strategy) run. The runs split by cost
rather than accuracy: BoN runs use about $18$--$90$K tokens and iterative runs
$250$K--$2.7$M, while both span roughly $30$--$88\%$ accuracy, ordered mainly
by target model. For budget-constrained
settings, BoN $N{=}10$ at $67$K tokens offers $63.3\%$ accuracy,
competitive with iterative methods that cost $5$--$21\times$ more;
this is the one regime in which one-shot sampling is a reasonable
default, and it relies on a validation split large enough to
control selection noise, which the $40$-question GPQA split is not.

\begin{table}[ht]
\centering
\caption{\small \textbf{GPQA validation accuracy by search strategy}
(Elitist Beam here uses $D{=}0$, so it differs from Beam only in keeping
parents),
mean $\pm$ s.d.\ \textbf{across the six target models}, \emph{not} across
seeds. Every run is \texttt{random\_seed}${=}42$, one run per (model,
strategy), so the large spreads are substantially the tier effect of
\cref{fig:gpqa-tier} rather than run-to-run variance; no seed replicates
exist for this sweep. Train--val gap is a
diagnostic, not an objective: iterative training scores are averaged over the
steps' training batches, while a BoN run logs a single batch; every arm
selects on $\valid$. Gap is
computed from unrounded means and can differ from the rounded columns by
$0.1$. The sweep has no test split, so the gap says nothing about validation
optimism.}
\label{tab:gpqa-aggregate}
\small
\begin{tabular}{lccccc}
\toprule
\textbf{Strategy} & \textbf{Models} & \textbf{Val (\%)} & \textbf{Train (\%)} & \textbf{Gap} & \textbf{Tok.\ (K)} \\
\midrule
Greedy ($\bw{=}1$)       & 6 & $53.8 \pm 16.4$ & $58.5 \pm 14.3$ & $+4.7$  & 342  \\
Beam ($\bw{=}4$)         & 6 & $62.9 \pm 21.3$ & $62.7 \pm 16.3$ & $-0.3$  & 1{,}394 \\
\textbf{Elitist Beam}  & 6 & $\mathbf{64.6 \pm 18.1}$ & $\mathbf{65.7 \pm 15.3}$ & $+1.2$  & 1{,}060 \\
\midrule
BoN ($N{=}1$)          & 6 & $52.9 \pm 17.4$ & $33.6 \pm 7.4$  & $-19.3$ & 26   \\
BoN ($N{=}3$)          & 6 & $58.8 \pm 18.9$ & $31.8 \pm 11.2$ & $-27.0$ & 34   \\
BoN ($N{=}5$)          & 6 & $58.8 \pm 10.8$ & $31.5 \pm 9.5$  & $-27.2$ & 44   \\
BoN ($N{=}10$)         & 6 & $63.3 \pm 8.2$  & $32.3 \pm 9.1$  & $-31.0$ & 67   \\
BoN ($N{=}15$)         & 6 & $60.0 \pm 14.4$ & $28.7 \pm 10.8$ & $-31.4$ & 86   \\
\bottomrule
\end{tabular}
\end{table}

\subsubsection{Elitist Beam Search Trees Across All Six Models}
\label{sec:app.exp.gpqa.trees-all}

\begin{figure}[ht]
\centering
\includegraphics[width=\textwidth]{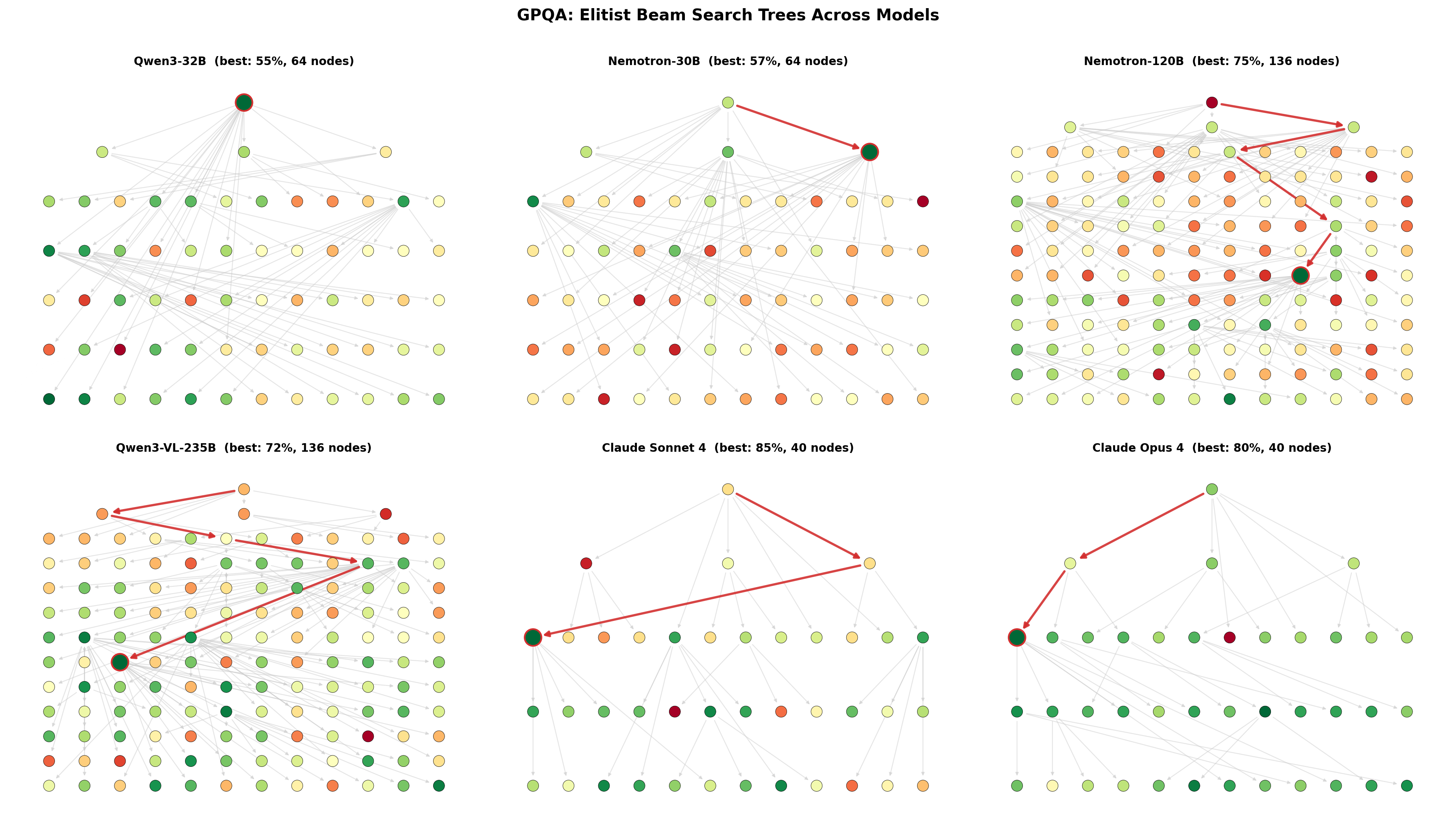}
\caption{\textbf{Elitist Beam search trees across all six target
models on GPQA-Diamond.} Frontier models (bottom middle and right)
converge in $\sim\!40$ nodes, small models (top left and middle) take
$64$, and large models (top right, bottom left) need up to $136$. Node
color encodes validation score (red = low, green = high); the red
path traces the lineage of the best candidate.}
\label{fig:gpqa-trees-elitist-all}
\end{figure}

\cref{fig:trees_frontier} shows the search-tree
footprint across strategies on frontier targets; the complementary
view in \cref{fig:gpqa-trees-elitist-all} fixes the strategy at
Elitist Beam and sweeps the target model across all six target models (three tiers). Frontier models converge in $\sim\!40$ nodes, matching the
frontier column of \cref{fig:trees_frontier}, while large models
need up to $136$ nodes, $3.4\times$ more exploration to find a
good prompt. This mirrors the per-tier accuracy results: harder
optimization landscapes require more search, and the elitist
selector supplies it without sacrificing the monotonicity guarantee
of \cref{thm:monotone}.

\subsubsection{Case Studies: What Iterative Refinement Discovers}
\label{sec:app.exp.gpqa.cases}

Three representative F1 runs illustrate how a prompt is refined along the
best-scoring lineage.

\paragraph{Elitist Beam (62\% \textrightarrow{} 85\%).}
The seed prompt ($c_0$) contains only a bare objective: ``Answer
the following multiple choice question by reasoning through it
step by step.'' In generation~$1$, the optimizer adds a
\texttt{<role>} (``expert scientist and problem solver''), a
structured \texttt{<approach>} with numbered steps, and a few-shot
example, but the score rises only to $68\%$ ($c_2$). The breakthrough comes
in generation~$2$ ($c_4$, $85\%$): the approach is rewritten to
enforce \emph{plain prose} reasoning (``no markdown headers, no
numbered sub-sections, no bullet lists'') and a $2$--$5$ sentence
budget, plus the objective is tightened to require \emph{sufficient}
justification. This formatting constraint prevents the model from
wasting tokens on verbose structure and forces it to reason
concisely, a lesson the optimizer discovered from critique
feedback on failed examples.

\paragraph{Greedy (57\% \textrightarrow{} 78\%).}
The Greedy run shows a different refinement trajectory across four
nodes. Generation~$1$ ($c_1$, $72\%$) introduces a decisive persona
(``prioritize reaching a clear, correct final answer over
exhaustive explanation'') and a concise approach. Generation~$2$
($c_4$, $75\%$) adds a strict word budget (``if your reasoning
would exceed $8$ sentences, stop and state your best answer'') and
option-scanning heuristics. Generation~$3$ ($c_9$, $78\%$) bans all
markdown formatting and imposes a hard $150$-word reasoning cap.
Each refinement targets a specific failure mode: verbose responses
that get truncated before stating an answer, markdown that confuses
answer extraction, and over-analysis that leads to second-guessing.

\paragraph{Beam \texorpdfstring{$\bw{=}4$}{W=4} (52\% \textrightarrow{} 85\%).}
The Beam run traverses seven nodes
($c_0\!\to\!c_2\!\to\!c_9\!\to\!c_{22}\!\to\!c_{33}\!\to\!c_{42}\!\to\!c_{53}$),
the longest best-path lineage among our runs. Generation~$1$
($c_2$, $65\%$) adds a SCAN/FOCUS/REASON structure and a ``no
backtracking'' directive. The critical insight arrives at
generation~$3$ ($c_{22}$, $68\%$): the objective is rewritten to
declare that ``reasoning without a final answer is a failure
regardless of quality,'' and the approach shifts to
answer-first-then-verify. Generation~$4$ ($c_{33}$, $70\%$)
escalates further: ``state your best answer BEFORE verification.''
Generation~$5$ ($c_{42}$, $78\%$) introduces explicit word budgets
and simplifies the approach to three steps. The final node
($c_{53}$, generation~$6$, $85\%$) imposes a hard $250$-word cap and adds a fourth
example. The trajectory reveals a consistent theme: each generation
tightens the contract between reasoning depth and answer
reliability, progressively learning that \emph{completing the
answer matters more than showing all work}.

\paragraph{Recurring refinement themes.}
Across all three runs three patterns emerge. (i) \emph{Format
discipline}: early steps add formatting constraints (plain
text, no markdown, word limits) that prevent answer-extraction
failures. (ii) \emph{Reasoning calibration}: later steps tune
the depth of reasoning, encouraging quick elimination for
conceptual questions and systematic derivation for multi-step
problems. (iii) \emph{Few-shot accumulation}: examples are added
incrementally ($1\!\to\!2\!\to\!3\!\to\!4$ across generations),
each targeting a different question archetype (recall,
calculation, synthesis, spectral analysis). None of these themes
are present in the seed; all are discovered from scalar validation
feedback and textual gradients alone.

\subsection{Candidate Lineage and Score Progression on SST-5}
\label{sec:app.exp.lineage-extra}

\cref{fig:trees_frontier} in \cref{sec:app.exp.trees} shows the
search-tree footprint on GPQA-Diamond. The analogous view for SST-5 is reported here.

\label{sec:app.exp.lineage.sst5}

\begin{figure}[h]
\centering
\includegraphics[width=0.95\textwidth]{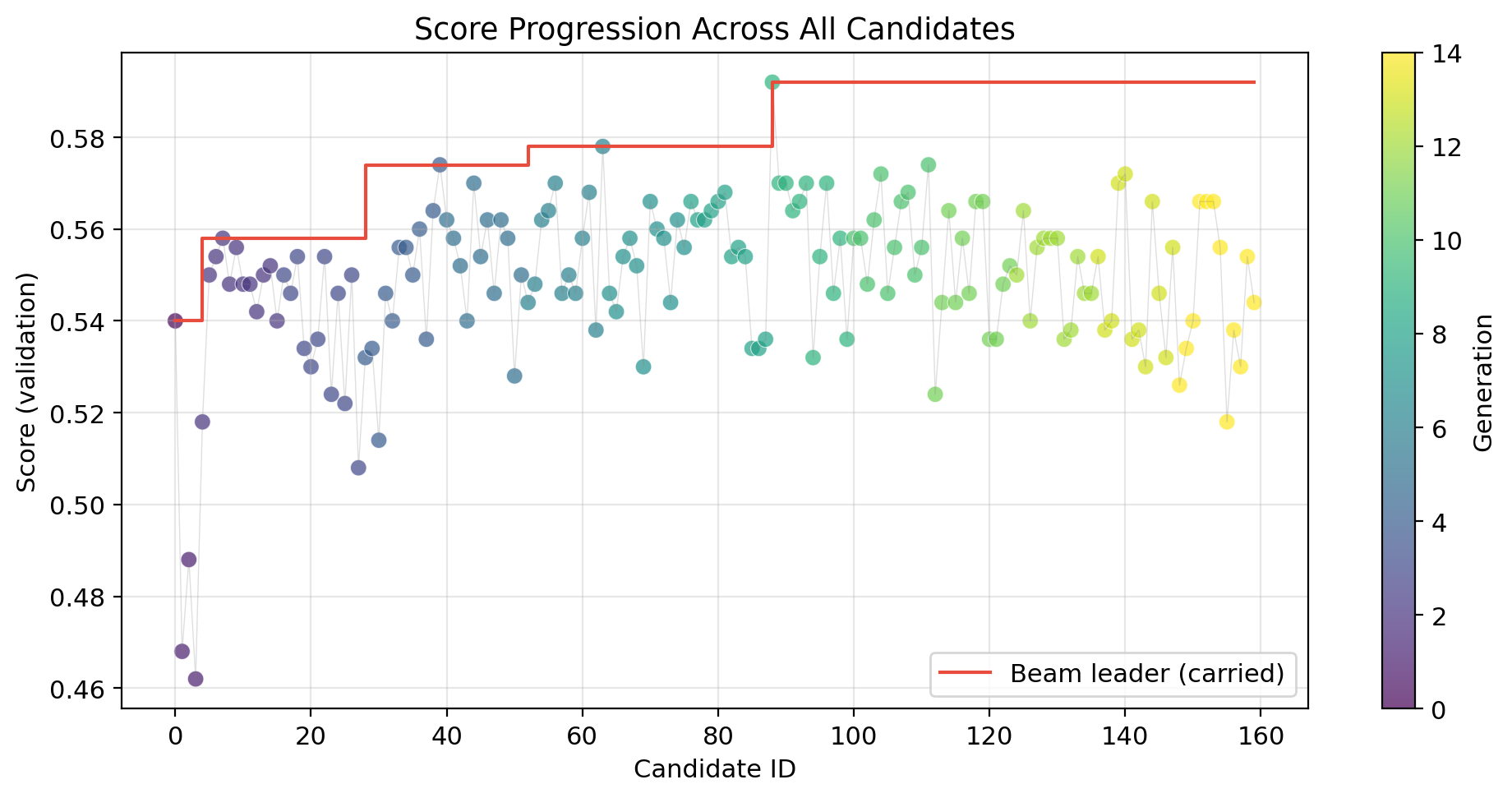}
\caption{\textbf{SST-5: the carried leader never falls, and most of its
gain comes early.} One run. Dots are candidates by ID, colored by
generation; the red line is the beam leader, which rises from $54.0\%$ to
$59.2\%$, about two-thirds of the gain by generation $4$.}
\label{fig:sst5-score-progression}
\end{figure}

\begin{figure}[h]
\centering
\includegraphics[width=0.95\textwidth]{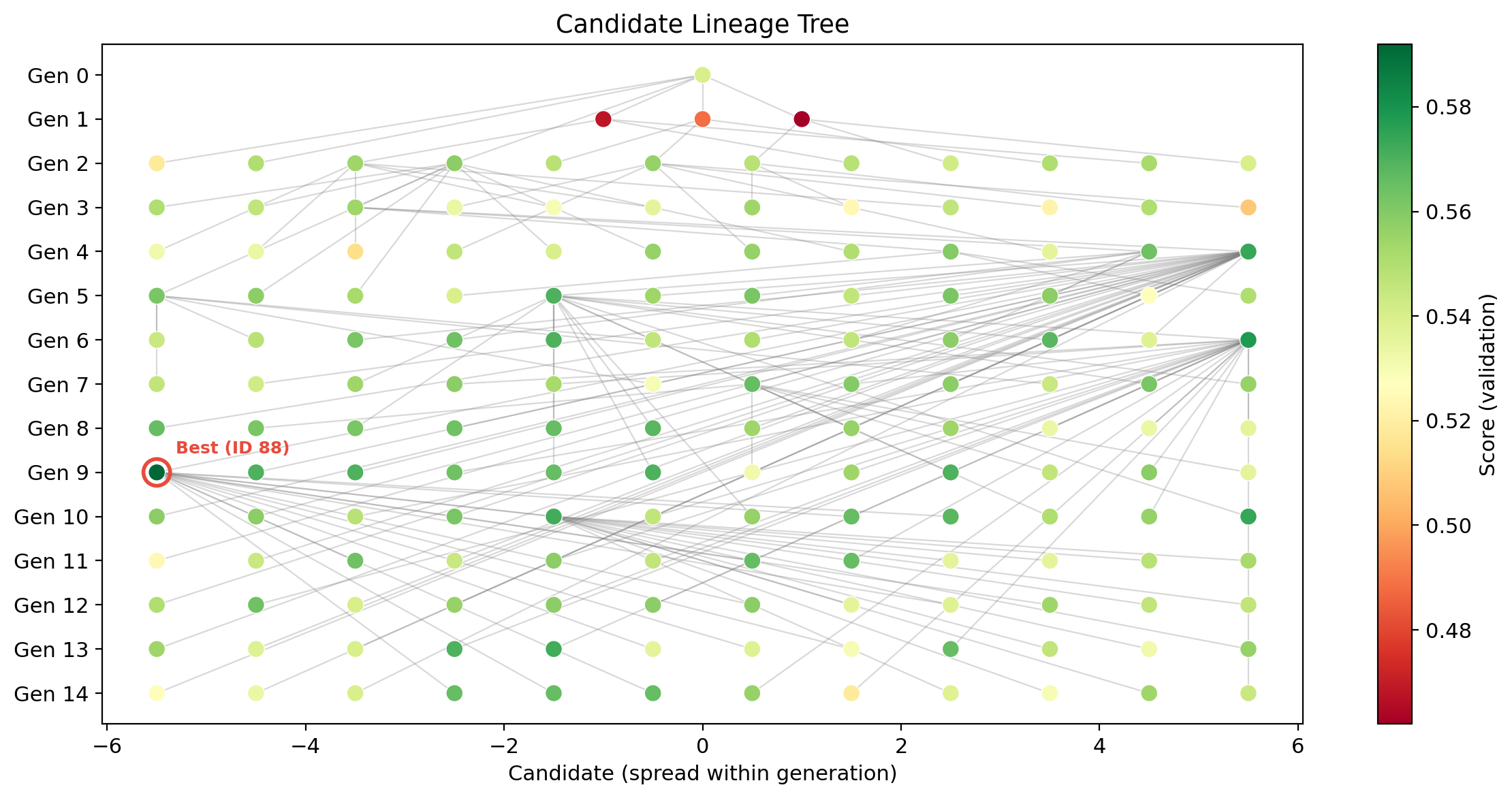}
\caption{\textbf{SST-5: the search fans out early, then concentrates on
one branch.} Candidate lineage graph for one run. Nodes are candidate prompts colored by validation accuracy
(red = low, green = high); edges connect a candidate to the
parent(s) it was proposed from. The ringed node is the final best
candidate. The fan-out pattern at early generations
reflects the Elitist Beam selector filling its diversity slots
(\cref{thm:div-preserve-ext}) before converging on the best-scoring branch.}
\label{fig:sst5-candidate-lineage}
\end{figure}

\cref{fig:sst5-score-progression} traces the best-leader validation
accuracy along the search step index, and
\cref{fig:sst5-candidate-lineage} shows the candidate-lineage graph
on the best SST-5 run. Two patterns appear. \emph{(i) Monotone best-path score.} The leader accuracy is
non-decreasing along the best-scoring lineage, the empirical
signature of \cref{thm:monotone} under the cached-scoring
convention. \emph{(ii) Early but not only early gains.} About two-thirds
of the $5.2$\,pp gain arrives by generation $4$, and the last $+1.4$\,pp at
about generation $9$. This matches the improving-proposal rate being highest early in
the search (\cref{sec:exp.pstar}).

\begin{figure}[h]
\centering
\includegraphics[width=0.7\textwidth]{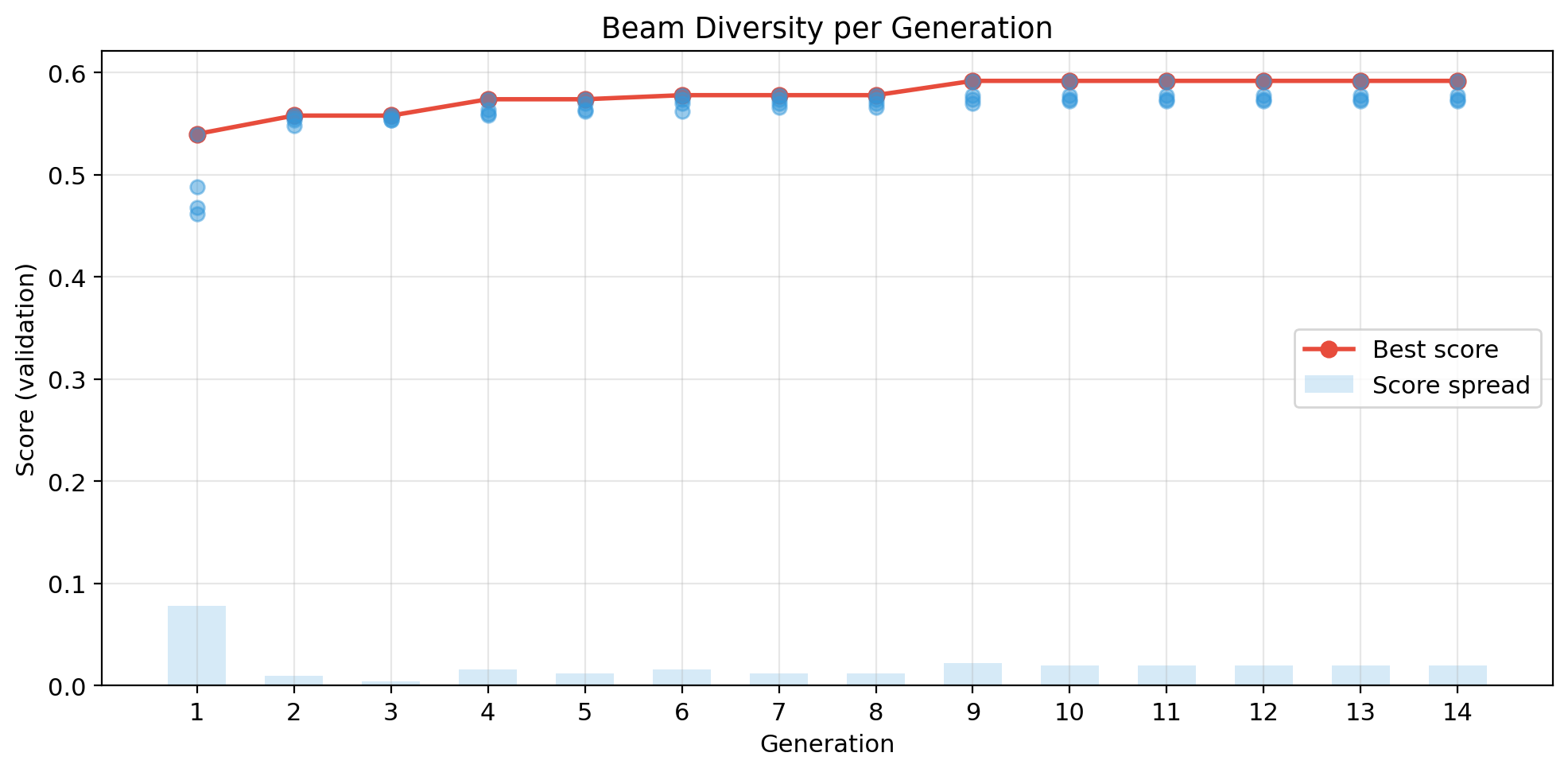}
\caption{\textbf{SST-5: after the first step, beam members score alike.}
Best validation score (line) and the spread of beam members' scores (bars) at
each generation of the Elitist Beam run.}
\label{fig:sst5-beam-diversity}
\end{figure}

\paragraph{Beam diversity and proposer success rate.}
Two further per-step diagnostics on SST-5 track how the selector
and proposer behave inside the search loop.
\cref{fig:sst5-beam-diversity} reports the best validation score and the
spread of beam scores at each generation. After step $1$ the spread stays
small, so the beam's members score alike while the leader climbs.

\begin{figure}[h]
\centering
\includegraphics[width=0.7\textwidth]{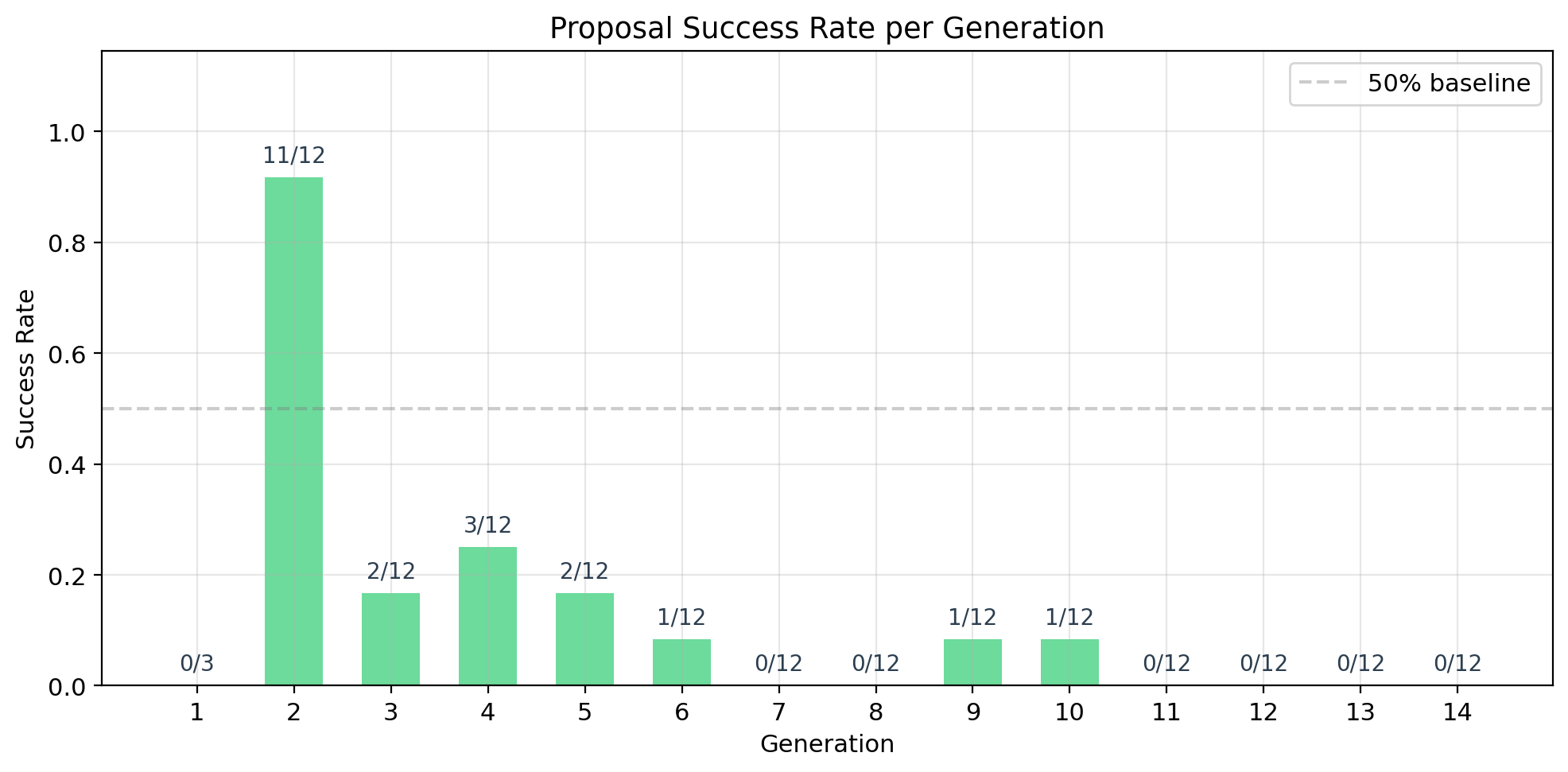}
\caption{\textbf{SST-5: improving proposals are common early and rare
later.} Fraction of proposed children whose validation score exceeds their
parent's, per step: $11/12$ at step $2$, at most $3/12$ from step $3$, and
zero in most steps after step $6$.}
\label{fig:sst5-proposal-success}
\end{figure}

\cref{fig:sst5-proposal-success} reports the complementary quantity
on the proposer side: the fraction of candidate children whose
validation score exceeds their parent's, measured per search step.
Under the deterministic monotonicity of \cref{thm:monotone} the
\emph{leader} is non-decreasing by construction, but the
per-proposal success rate is an empirical quantity: it says how
often a textual-gradient-conditioned proposal actually improves on
its parent. That fraction is itself a per-proposal rate (it is $\hat p^\star$,
a direct estimate of the per-proposal rate, not the per-step reachability probability), and on
SST-5 it is $11/12$ at step $2$, at most $3/12$ from step $3$, and zero in most
steps after step $6$, falling as the leader climbs. This run is not the one tabulated in
\cref{tab:pstar-measured}, whose SST-5 column falls to $0.08$--$0.17$ by steps
$4$--$5$; the two should be read as different runs, not as one rate.

\clearpage
\newpage
\section*{NeurIPS Paper Checklist}

\begin{enumerate}

\item {\bf Claims}
    \item[] Question: Do the main claims made in the abstract and introduction accurately reflect the paper's contributions and scope?
    \item[] Answer: \answerYes{}
    \item[] Justification: The abstract and introduction claim an edit-graph formalization, selective mutability with a price-of-freezing bound, an elitist selector with diversity slots and its guarantees, reachability and convergence results under a stated proposer condition, a length-regularization bias bound, and empirical results: wins on four of seven public benchmarks and matches on three (a margin of at most $0.5$\,pp), plus a contract stress test. These are substantiated in \cref{sec:main.method,sec:theory-main,sec:main.exp} and, in full, \cref{sec:main.theory,sec:main.exp-ext,sec:app.theory}.

\item {\bf Limitations}
    \item[] Question: Does the paper discuss the limitations of the work performed by the authors?
    \item[] Answer: \answerYes{}
    \item[] Justification: \Cref{sec:main.limitations} states the main limitations: the guarantees are conditional on proposal coverage and on a submodularity ratio for which there is no estimator; freezing preserves a contract's text but not its effect (compliance falls from $96.5\%$ to $66.8\%$ in \cref{sec:exp.contract-main}); a misspecified reward is the dominant failure mode. \Cref{sec:main.theory} states what is and is not provable, and \cref{sec:main.exp-ext} reports null and negative ablation results (e.g.\ momentum).

\item {\bf Theory assumptions and proofs}
    \item[] Question: For each theoretical result, does the paper provide the full set of assumptions and a complete (and correct) proof?
    \item[] Answer: \answerYes{}
    \item[] Justification: Every theorem-like statement is given with its assumptions in \cref{sec:theory-main} or \cref{sec:main.theory}, and complete proofs are in \cref{sec:app.theory}. Results that are not proved are labelled as conjectures.

    \item {\bf Experimental result reproducibility}
    \item[] Question: Does the paper fully disclose all the information needed to reproduce the main experimental results of the paper to the extent that it affects the main claims and/or conclusions of the paper (regardless of whether the code and data are provided or not)?
    \item[] Answer: \answerYes{}
    \item[] Justification: \Cref{sec:main.exp} specifies all benchmarks, target models and the published protocols reproduced without modification; \cref{alg:splice} gives the full procedure; \cref{sec:main.exp-ext,sec:app.exp} gives protocols, splits and hyperparameters, and \cref{sec:app.example_prompts} gives optimized prompts.

\item {\bf Open access to data and code}
    \item[] Question: Does the paper provide open access to the data and code, with sufficient instructions to faithfully reproduce the main experimental results, as described in supplemental material?
    \item[] Answer: \answerNo{}
    \item[] Justification: Code is not released with this version. The paper provides full algorithmic pseudocode, all hyperparameters and configuration details sufficient to reimplement the method, and the contract used in the stress test is given verbatim; all benchmarks used are public.

\item {\bf Experimental setting/details}
    \item[] Question: Does the paper specify all the training and test details (e.g., data splits, hyperparameters, how they were chosen, type of optimizer) necessary to understand the results?
    \item[] Answer: \answerYes{}
    \item[] Justification: \Cref{sec:main.exp} specifies benchmarks, target models and protocols; \cref{alg:splice} lists every hyperparameter (beam width $W$, diversity slots $D$, quality-gate ratio $\gateratio$, proposals per candidate $k$, horizon, momentum window, patience); data splits and their roles are defined in \cref{sec:main.method} and per-experiment settings are in \cref{sec:main.exp-ext,sec:app.exp}.

\item {\bf Experiment statistical significance}
    \item[] Question: Does the paper report error bars suitably and correctly defined or other appropriate information about the statistical significance of the experiments?
    \item[] Answer: \answerYes{}
    \item[] Justification: The contract stress test and the momentum ablation report mean $\pm$ s.d.\ over three paired seeds, with paired per-seed differences; the critique ablation reports $95\%$ bootstrap CIs over the target models. Benchmark comparisons use baselines as published, which report no seed variance, so margins of at most $0.5$\,pp are reported as matches rather than wins. Single-run results are identified as such.

\item {\bf Experiments compute resources}
    \item[] Question: For each experiment, does the paper provide sufficient information on the computer resources (type of compute workers, memory, time of execution) needed to reproduce the experiments?
    \item[] Answer: \answerYes{}
    \item[] Justification: All experiments use cloud-hosted LLM APIs rather than local GPU compute. \Cref{alg:splice} determines the number of LLM calls per step, and \cref{sec:main.exp-ext} reports measured inference cost.
    
\item {\bf Code of ethics}
    \item[] Question: Does the research conducted in the paper conform, in every respect, with the NeurIPS Code of Ethics \url{https://neurips.cc/public/EthicsGuidelines}?
    \item[] Answer: \answerYes{}
    \item[] Justification: The research uses publicly available benchmarks (SST-5, GPQA, ARC, BBH, GSM8K, AIME-2025, PUPA, HelpSteer2), does not involve human subjects, and does not raise dual-use concerns. The method optimizes prompts for reasoning and classification tasks and does not generate harmful content.

\item {\bf Broader impacts}
    \item[] Question: Does the paper discuss both potential positive societal impacts and negative societal impacts of the work performed?
    \item[] Answer: \answerYes{}
    \item[] Justification: The paper's motivation is a safety benefit: contract sections such as safety clauses, tool schemas and output formats are made unreachable, so optimization cannot edit them away (\cref{sec:main.intro}). \Cref{sec:main.limitations} states the main risks: \splice{} optimizes the reward it is given, so a misspecified reward is the dominant failure mode, and a frozen clause keeps its text but not necessarily its effect (compliance falls from $96.5\%$ to $66.8\%$), so frozen blocks still need a behavioral audit.
    
\item {\bf Safeguards}
    \item[] Question: Does the paper describe safeguards that have been put in place for responsible release of data or models that have a high risk for misuse (e.g., pre-trained language models, image generators, or scraped datasets)?
    \item[] Answer: \answerNA{}
    \item[] Justification: The paper does not release pre-trained models, scraped datasets, or other high-risk assets. The contribution is an optimization algorithm that operates on existing LLM APIs. The optimized prompts are task-specific text strings without inherent misuse risk beyond what the underlying LLMs already present.

\item {\bf Licenses for existing assets}
    \item[] Question: Are the creators or original owners of assets (e.g., code, data, models), used in the paper, properly credited and are the license and terms of use explicitly mentioned and properly respected?
    \item[] Answer: \answerNo{}
    \item[] Justification: All datasets are properly cited: SST-5~\citep{socher2013recursive}, GPQA~\citep{rein2023gpqa}, GSM8K~\citep{cobbe2021gsm8k}, ARC~\citep{clark2018think}, BBH~\citep{suzgun2023challenging}, AIME-2025~\citep{balunovic2025matharena}, PUPA~\citep{siyan-etal-2025-papillon}, HelpSteer2~\citep{wang2024helpsteer}. All prior methods are cited where techniques are adapted, and the LLM APIs are accessed through their official providers. We use only the public releases of these datasets, under the terms they are distributed with, but the paper does not restate their licenses.

\item {\bf New assets}
    \item[] Question: Are new assets introduced in the paper well documented and is the documentation provided alongside the assets?
    \item[] Answer: \answerNA{}
    \item[] Justification: No new assets (code, datasets, or models) are released with this submission. The algorithmic contribution is fully described via pseudocode (Algorithms 1--3) and formal definitions in the main text and appendix.

\item {\bf Crowdsourcing and research with human subjects}
    \item[] Question: For crowdsourcing experiments and research with human subjects, does the paper include the full text of instructions given to participants and screenshots, if applicable, as well as details about compensation (if any)? 
    \item[] Answer: \answerNA{}
    \item[] Justification: The paper does not involve crowdsourcing or research with human subjects. All evaluations are automated using standard benchmark metrics.

\item {\bf Institutional review board (IRB) approvals or equivalent for research with human subjects}
    \item[] Question: Does the paper describe potential risks incurred by study participants, whether such risks were disclosed to the subjects, and whether Institutional Review Board (IRB) approvals (or an equivalent approval/review based on the requirements of your country or institution) were obtained?
    \item[] Answer: \answerNA{}
    \item[] Justification: The paper does not involve human subjects research. All experiments are conducted using automated evaluation of LLM outputs against benchmark datasets.

\item {\bf Declaration of LLM usage}
    \item[] Question: Does the paper describe the usage of LLMs if it is an important, original, or non-standard component of the core methods in this research? Note that if the LLM is used only for writing, editing, or formatting purposes and does \emph{not} impact the core methodology, scientific rigor, or originality of the research, declaration is not required.
    \item[] Answer: \answerYes{}
    \item[] Justification: LLMs are central to the \splice{} methodology and are fully described. The proposal agent generates improved prompt variants (Algorithm~1, step iv). The critique agent produces dual textual gradients (Algorithm~1, step ii). Multiple system models across capability tiers serve as the target models being optimized (\cref{sec:main.exp}). The paper explicitly describes the role of each LLM component.

\end{enumerate}

\end{document}